\documentclass[12pt]{article}
\usepackage{authblk}
\usepackage{amsmath}
\usepackage{floatpag}
\usepackage{amssymb}
\usepackage{graphicx}
\usepackage{grffile}
\usepackage{caption}
\usepackage{subcaption}
\usepackage{enumerate}
\usepackage{natbib}
\usepackage{amsthm}
\usepackage{url} 
\usepackage{mathtools}
\usepackage{bbm}
\usepackage[normalem]{ulem}
\usepackage{threeparttable}
\usepackage{footnote}
\usepackage{tablefootnote}
\usepackage{adjustbox}
\usepackage{makecell}
\usepackage{booktabs}
\usepackage{multirow}
\usepackage{dsfont}
\usepackage{amsfonts}
\usepackage[ruled,linesnumbered]{algorithm2e}
\usepackage{bm}
\usepackage{tocloft}
\usepackage[title]{appendix}
\usepackage{textcomp}
\usepackage{eufrak}
\usepackage[mathscr]{eucal}
\usepackage{tikz}
\usetikzlibrary{arrows,chains,matrix,positioning,scopes}
\usepackage{hyperref}
\hypersetup{colorlinks=true,linkcolor=blue, citecolor=blue, linktocpage}
\mathtoolsset{showonlyrefs}
\allowdisplaybreaks

\newcommand{\blind}{1}

\usepackage{multibib}
\newcites{app}{References}

\makeatletter
\newcommand*{\Rom}[1]{\expandafter\@slowromancap\romannumeral #1@}
\makeatother
\newcommand*\circled[1]{\tikz[baseline=(char.base)]{
            \node[shape=circle,draw,inner sep=0.05pt] (char) {#1};}}
\newcommand*{\rom}[1]{\romannumeral #1}
\DeclareMathOperator*{\argmax}{arg\,max}
\DeclareMathOperator*{\argmin}{arg\,min}

\newcommand{\wtp}{\widetilde{\tp}}

\newcommand{\tp}{\mathbb{P}}
\newcommand{\tq}{\mathbb{Q}}

\newcommand{\te}{\mathbb{E}}

\newcommand{\btheta}{\bm{\theta}}
\newcommand{\bthetaks}[1]{\bm{\theta}^{(#1)*}}
\newcommand{\bthetak}[1]{\bm{\theta}^{(#1)}}
\newcommand{\gammak}[1]{\gamma^{(#1)}}
\newcommand{\mC}{\mathcal{C}}
\newcommand{\ind}{\perp\!\!\!\!\perp}

\newcommand{\otheta}{\overline{\bm{\theta}}}

\newcommand{\othetaks}[1]{\overline{\bm{\theta}}^{(#1)*}}
\newcommand{\othetak}[1]{\overline{\bm{\theta}}^{(#1)}}

\newcommand{\bepsilon}{\bm{\epsilon}}

\newcommand{\epsilonk}[1]{\epsilon^{(#1)}}

\newcommand{\htheta}{\widehat{\bm{\theta}}}

\newcommand{\hw}{\widehat{w}}

\newcommand{\bbeta}{\bm{\beta}}
\newcommand{\hdelta}{\widehat{\delta}}
\newcommand{\obbeta}{\overline{\bbeta}}

\newcommand{\hbeta}{\widehat{\bm{\beta}}}
\newcommand{\hobeta}{\widehat{\overline{\bm{\beta}}}}
\newcommand{\bgamma}{\bm{\gamma}}
\newcommand{\bmu}{\bm{\mu}}
\newcommand{\hmu}{\widehat{\bm{\mu}}}

\newcommand{\infnorma}[1]{\left\|#1\right\|_{\infty}}

\newcommand{\twonorma}[1]{\left\|#1\right\|_{2}}

\newcommand{\twonorm}[1]{\|#1\|_{2}}

\newcommand{\norma}[1]{\left|#1\right|}
\newcommand{\norm}[1]{|#1|}
\newcommand{\bx}{\bm{x}}
\newcommand{\bv}{\bm{v}}
\newcommand{\bxk}[1]{\bm{x}^{(#1)}}
\newcommand{\yk}[1]{y^{(#1)}}

\newcommand{\by}{\bm{y}}

\newcommand{\bSigma}{\bm{\Sigma}}
\newcommand{\hSigma}{\widehat{\bm{\Sigma}}}
\newcommand{\wk}[1]{w^{(#1)}}
\newcommand{\wks}[1]{w^{(#1)*}}
\newcommand{\bmuk}[1]{\bm{\mu}^{(#1)}}
\newcommand{\bmuks}[1]{\bm{\mu}^{(#1)*}}
\newcommand{\deltak}[1]{\delta^{(#1)}}
\newcommand{\deltaks}[1]{\delta^{(#1)*}}
\newcommand{\bbetak}[1]{\bm{\beta}^{(#1)}}
\newcommand{\bbetaks}[1]{\bm{\beta}^{(#1)*}}
\newcommand{\bSigmak}[1]{\bm{\Sigma}^{(#1)}}
\newcommand{\bSigmaks}[1]{\bm{\Sigma}^{(#1)*}}
\newcommand{\ns}{n_S}
\newcommand{\nsc}{n_{S^c}}
\newcommand{\bz}{\bm{z}}
\newcommand{\bzk}[1]{\bm{z}^{(#1)}}

\newcommand{\bu}{\bm{u}}
\newcommand{\notes}[1]{{\color{red}#1}}

\newcommand{\wepsilon}{\widetilde{\epsilon}}

\newcommand\smallo{
  \mathchoice
    {{\scriptstyle\mathcal{O}}}
    {{\scriptstyle\mathcal{O}}}
    {{\scriptscriptstyle\mathcal{O}}}
    {\scalebox{.7}{$\scriptscriptstyle\mathcal{O}$}}
}

\newtheorem{theorem}{Theorem}
\newtheorem{lemma}{Lemma} 
 
\newtheorem{remark}{Remark}

\newtheorem{assumption}{Assumption}

\def\spacingset#1{\renewcommand{\baselinestretch}%
{#1}\small\normalsize} \spacingset{1}

\begin{document}

\if1\blind
{
  \title{\bf Robust Unsupervised Multi-task and Transfer Learning on Gaussian Mixture Models}
  \date{}

  \author[1]{Ye Tian}
  \affil[1]{Department of Statistics, Columbia University}
  \author[2]{Haolei Weng}
  \affil[2]{Department of Statistics and Probability, Michigan State University}

  \author[3]{Lucy Xia}
  \affil[3]{Department of ISOM, School of Business and Management\\Hong Kong University of Science and Technology}
  \author[4]{Yang Feng}
  \affil[4]{Department of Biostatistics, School of Global Public Health\\New York University}
  \maketitle
} \fi

\if0\blind
{
  \bigskip
  \bigskip
  \bigskip
  \begin{center}
    {\LARGE\bf Robust Unsupervised Multi-task and Transfer Learning on Gaussian Mixture Models}
\end{center}
  \medskip
} \fi

\bigskip

\begin{abstract}
Unsupervised learning has been widely used in many real-world applications. One of the simplest and most important unsupervised learning models is the Gaussian mixture model (GMM). In this work, we study the multi-task learning problem on GMMs, which aims to leverage potentially similar GMM parameter structures among tasks to obtain improved learning performance compared to single-task learning. We propose a multi-task GMM learning procedure based on the EM algorithm that effectively utilizes unknown similarities between related tasks and is robust against a fraction of outlier tasks from arbitrary distributions. The proposed procedure is shown to achieve the minimax optimal rate of convergence for both parameter estimation error and the excess mis-clustering error, in a wide range of regimes. Moreover, we generalize our approach to tackle the problem of transfer learning for GMMs, where similar theoretical results are derived. Additionally, iterative unsupervised multi-task and transfer learning methods may suffer from an initialization alignment problem, and two alignment algorithms are proposed to resolve the issue. Finally, we demonstrate the effectiveness of our methods through simulations and real data examples. To the best of our knowledge, this is the first work studying multi-task and transfer learning on GMMs with theoretical guarantees.
\end{abstract}

\noindent
{\it Keywords: Multi-task learning, transfer learning, unsupervised learning, Gaussian mixture models, robustness, minimax rate, EM algorithm} 
\vspace{1cm}

\spacingset{1.77}
\addtocontents{toc}{\protect\setcounter{tocdepth}{0}}
\section{Introduction}\label{sec: introduction}

\subsection{Gaussian mixture models (GMMs)}\label{subsec: intro gmm}
Unsupervised learning that learns patterns from unlabeled data is a prevalent problem in statistics and machine learning. Clustering is one of the most important problems in unsupervised learning, where the goal is to group the observations based on some similarity metrics. Researchers have developed numerous clustering methods including $k$-means \citep{forgy1965cluster}, $k$-medians \citep{jain1988algorithms}, spectral clustering \citep{ng2001spectral}, and hierarchical clustering \citep{murtagh2012algorithms}, among others. On the other hand, clustering problems have been analyzed from the perspective of the mixture of several probability distributions \citep{scott1971clustering}. The mixture of Gaussian distributions is one of the simplest models in this category and has been widely applied in many real applications \citep{yang1998gaussian, lee2012application}.

In the binary Gaussian mixture models (GMMs) with common covariances, each observation $Z \in \mathbb{R}^p$ comes from the following mixture of two Gaussian distributions:
\begin{align}
	Y &= \begin{cases}
		1, &\text{with probability }1-w,\\
		2, &\text{with probability }w,
	\end{cases}\\
	Z|Y &= r \sim \mathcal{N}(\bmu_r, \bSigma), \quad r = 1, 2,
\end{align}
where $w \in (0,1)$, $\bmu_1 \in \mathbb{R}^p$, $\bmu_2 \in \mathbb{R}^p$ and $\bSigma \succ 0$ are parameters. This is the same setting as the linear discriminant analysis (LDA) problem in classification \citep{hastie2009elements}, except that the label $Y$ is unknown in the clustering problem, while it is observed in the classification case. It has been shown that the Bayes classifier for the LDA problem is
\begin{equation}\label{eq: bayes lda intro section}
	\mathcal{C}(\bz) = \begin{cases}
		1, & \text{if }\bbeta^\top\bz - \delta \leq \log\left(\frac{1-w}{w}\right);\\
		2, & \text{otherwise},
	\end{cases}
\end{equation}
where $\bbeta = \bSigma^{-1}(\bmu_2-\bmu_1) \in \mathbb{R}^p$ and $\delta = \bbeta^\top(\bmu_1 + \bmu_2)/2$. Note that $\bbeta$ is usually referred to as the discriminant coefficient \citep{anderson1958introduction, efron1975efficiency}. Naturally, this classifier is useful in clustering too. In clustering, after learning $w$, $\bmu_1$, $\bmu_2$ and $\bbeta$, we can plug their estimators into \eqref{eq: bayes lda intro section} to group any new observation $Z^{\textup{new}}$. Generally, we define the mis-clustering error rate of any given clustering method $\mC$ as
\begin{equation}
	R(\mC) = \min_{\pi: \{1,2\} \rightarrow \{1, 2\}}\tp(\mC(Z^{\textup{new}}) \neq \pi(Y^{\textup{new}})),
\end{equation}
where $\pi$ is a permutation function, $Y^{\textup{new}}$ is the label of a future observation $Z^{\textup{new}}$, and the probability is taken w.r.t. the joint distribution of $(Z^{\textup{new}}, Y^{\textup{new}})$ based on parameters $w$, $\bmu_1$, $\bmu_2$ and $\bSigma$. Here the error is calculated up to a permutation due to the lack of label information. It is clear that in the ideal case where the parameters are known, $\mC(\cdot)$ in \eqref{eq: bayes lda intro section} achieves the optimal mis-clustering error. Multi-cluster Gaussian mixture models with $R \geq 3$ components can be described in a similar way. To maintain simplicity and highlight key intuitions, we focus on binary GMMs in the main text, while a detailed analysis of the multi-cluster scenario is presented in Section S.2 of the supplementary materials.

There is a large volume of published studies on learning a GMM. The vast majority of approaches can be roughly divided into three categories. The first category is the method of moments, where the parameters are estimated through several moment equations \citep{pearson1894contributions, kalai2010efficiently, hsu2013learning, ge2015learning}. The second category is the spectral method, where the estimation is based on the spectral decomposition \citep{vempala2004spectral, hsu2013learning, jin2017phase}. The last category is the likelihood-based method, which includes the popular expectation-maximization (EM) algorithm as a canonical example. The general form of the EM algorithm was formalized by \cite{dempster1977maximum} in the context of incomplete data, though earlier works \citep{hartley1958maximum, hasselblad1966estimation, baum1970maximization, sundberg1974maximum} have studied EM-style algorithms in various concrete settings. Classical convergence results on the EM algorithm \citep{wu1983convergence, redner1984mixture, meng1994global, mclachlan2007algorithm} guarantee local convergence of the algorithm to fixed points of the sample likelihood. Recent advances in the analysis of the EM algorithm and its variants provide stronger guarantees by establishing geometric convergence rates of the algorithm to the underlying true parameters under mild initialization conditions. See, for example, \cite{dasgupta2000two, wang2014high, xu2016global, balakrishnan2017statistical, yan2017convergence, cai2019chime, kwon2020algorithm, zhao2020statistical} for GMM related works. In this paper, we propose modified versions of the EM algorithm with similarly strong guarantees to learn GMMs, under the new multi-task and transfer learning context.

\subsection{Multi-task learning and transfer learning}\label{subsec: intro mtl tl}
Multi-tasking is an ability that helps people handle more than one task simultaneously. Moreover, the knowledge learned from one task can also be helpful in other tasks. Multi-task learning (MTL) is a learning paradigm inspired by the human learning ability, which aims to learn multiple tasks and improve performance by utilizing the similarity between these tasks \citep{zhang2021survey}. There has been much research on MTL, which can be classified into five categories \citep{zhang2021survey}: feature learning approach \citep{argyriou2008convex, obozinski2006multi}, low-rank approach \citep{ando2005framework}, task clustering approach \citep{thrun1996discovering}, task relation learning approach \citep{evgeniou2004regularized} and decomposition approach \citep{jalali2010dirty}. Most existing works focus on using MTL in supervised learning problems, while the application of MTL in unsupervised learning, such as clustering, has received less attention. \cite{zhang2011multitask} developed an MTL clustering method based on a penalization framework, where the objective function consists of a local loss function and a pairwise task regularization term, both of which are related to the Bregman divergence. In \cite{gu2011learning}, a reproducing kernel Hilbert space (RKHS) was first established, and then a multi-task kernel k-means clustering was applied based on that RKHS. \cite{yang2014multitask} proposed a spectral MTL clustering method with a novel $\ell_{2,p}$-norm, which can also produce a linear regression function to predict labels for out-of-sample data. \cite{zhang2018multi} suggested a new method based on the similarity matrix of samples in each task, which can learn the within-task clustering structure as well as the task relatedness simultaneously. \cite{marfoq2021federated} established a new federated multi-task EM algorithm to learn the mixture of distributions and provided some theory on the convergence guarantee, but the statistical properties of the estimators were not fully understood. \cite{zhang2022distributed} proposed a distributed learning algorithm for GMMs based on transportation divergence when all GMMs are identical. In general, there are very few theoretical results about unsupervised MTL.

Transfer learning (TL) is another learning paradigm similar to multi-task learning but has different objectives. While MTL aims to learn all the tasks well with no priority for any specific task, the goal of TL is to improve the performance on the \textit{target} task using the information from the \textit{source} tasks \citep{zhang2021survey}. According to \cite{pan2009survey}, most of TL approaches can be classified into four categories: instance-based transfer \citep{dai2007boosting}, feature representation transfer \citep{dai2008self}, parameter transfer \citep{lawrence2004learning} and relational-knowledge transfer \citep{mihalkova2007mapping}. Similar to MTL, most of the TL methods focus on supervised learning. Some TL approaches are also developed for the semi-supervised learning setting \citep{chattopadhyay2012multisource, li2013learning}, where only part of the target or source data is labeled. There is much less discussion on the unsupervised TL approaches \footnote{There are different definitions for unsupervised TL. Sometimes, people call the semi-supervised TL an unsupervised TL as well. We follow the definition in \cite{pan2009survey} here.}, which focus on the cases where both target and source data are unlabeled. \cite{dai2008self} developed a co-clustering approach to transfer information from a single source to the target, which relies on the loss of mutual information and requires the features to be discrete. \cite{wang2008transferred} proposed a TL discriminant analysis method, where the target data is allowed to be unlabeled, but some labeled source data is necessary. In \cite{wang2021general}, a TL approach was developed to learn Gaussian mixture models with only one source by weighting the target and source likelihood functions. \cite{zuo2018fuzzy} proposed a TL method based on infinite Gaussian mixture models and active learning, but their approach needs sufficient labeled source data and a few labeled target samples. 

There are some recent studies on TL and MTL under various statistical settings, including high-dimensional linear regression \citep{xu2021learning, li2022transfer, zhang2022class, li2022estimation}, high-dimensional generalized linear models \citep{bastani2021predicting, li2023targeting, tian2023transfer}, functional linear regression \citep{lin2022transfer}, high-dimensional graphical models \citep{li2022transfer}, reinforcement learning \citep{chen2022transferred}, among others. The recent work \cite{duan2023adaptive} developed an adaptive and robust MTL framework with sharp statistical guarantees for a broad class of models. We discuss its connection to our work in Section \ref{sec: mtl}.

\subsection{Our contributions and paper structure}\label{subsec: intro contribution}
Our main contributions in this work can be summarized in the following:
\begin{enumerate}[(i)]
	\item We develop efficient polynomial-time iterative procedures to learn GMMs in both MTL and TL settings. These procedures can be viewed as adaptations of the standard EM algorithm for MTL and TL problems.
	\item The developed procedures come with provable statistical guarantees. Specifically, we derive the upper bounds of their estimation and excess mis-clustering error rates under mild conditions. For MTL, it is shown that when the tasks are close to each other, our method can achieve better upper bounds than those from the single-task learning; when the tasks are substantially different from each other, our method can still obtain competitive convergence rates compared to single-task learning. Similarly for TL, our method can achieve better upper bounds than those from fitting GMM only to target data when the target and sources are similar, and remains competitive otherwise. In addition, the derived upper bounds reveal the robustness of our methods against a fraction of outlier tasks (for MTL) or outlier sources (for TL) from arbitrary distributions. These guarantees certify our procedures as adaptive (to the unknown task relatedness) and robust (to contaminated data) learning approaches.
	\item We derive the minimax lower bounds for parameter estimation and excess mis-clustering errors. In various regimes, the upper bounds from our methods match the lower bounds (up to small order terms), showing that the proposed methods are (nearly) minimax rate optimal.
	\item Our MTL and TL approaches require the initial estimates for different tasks to be ``well-aligned", due to the non-identifiability of GMM. We propose two pre-processing alignment algorithms to provably resolve the alignment problem.  Similar problems arise in many unsupervised MTL settings. However, to our knowledge, there is no formal discussion of the alignment issue in the existing literature on unsupervised MTL \citep{gu2011learning, zhang2011multitask, yang2014multitask, zhang2018multi, dieuleveut2021federated, marfoq2021federated}. Therefore, our rigorous treatment of the alignment problem is an important step forward in this field.
\end{enumerate}

The rest of the paper is organized as follows. In Section \ref{sec: mtl}, we first discuss the multi-task learning problem for binary GMMs, by introducing the problem setting, our method, and the associated theory. The above-mentioned alignment problem is discussed in Section \ref{subsec: alignment mtl}. We present a simulation study and a real-data analysis in Section \ref{sec: numerical} to validate our theoretical findings. Finally, in Section \ref{sec: discussions}, we highlight several interesting extensions that are deferred to the supplementary materials due to space constraints. These include the extension to multi-cluster GMMs, discussions on handling different numbers of clusters across tasks, determination of cluster numbers, and modeling heterogeneous covariance matrices across clusters. Additional numerical results, a complete treatment of the transfer learning setting, and all technical proofs are also provided in the supplementary materials.

We summarize the notations used throughout the paper here for convenience. We use bold capital letters (e.g., $\bSigma$) to denote matrices and bold small letters (e.g., $\bx$, $\by$) to denote vectors. For a matrix $\bm{A} = [a_{ij}]_{p \times q}\in \mathbb{R}^{p \times q}$, its 2-norm or spectral norm is defined as $\twonorm{\bm{A}} = \max_{\bx: \twonorm{\bx}=1}\twonorm{\bm{A}\bx}$. If $q=1$, $\bm{A}$ becomes a $p$-dimensional vector and $\twonorm{\bm{A}}$ equals its Euclidean norm. For symmetric $\bm{A}$, we define $\lambda_{\max}(\bm{A})$ and $\lambda_{\min}(\bm{A})$ as the maximum and minimum eigenvalues of $\bm{A}$, respectively. For two non-zero real sequences $\{a_n\}_{n=1}^{\infty}$ and $\{b_n\}_{n=1}^{\infty}$, we use $a_n \ll b_n$, $b_n \gg a_n$ or $a_n = \smallo(b_n)$ to represent $|a_n/b_n| \rightarrow 0$ as $n \rightarrow \infty$. And $a_n \lesssim b_n$, $b_n \gtrsim a_n$ or $a_n = \mathcal{O}(b_n)$ means $\sup_n|a_n/b_n| < \infty$. For two random variable sequences $\{x_n\}_{n=1}^{\infty}$ and $\{y_n\}_{n=1}^{\infty}$, the notation  $x_n = \mathcal{O}_{\tp}(y_n)$ means that for any $\epsilon > 0$, there exists a positive constant $M$ such that $\sup_n\tp(|x_n/y_n|> M) \leq \epsilon$. For two real numbers $a$ and $b$, $a \vee b$ and $a\wedge b$ represent $\max(a, b)$ and $\min(a, b)$, respectively. For any positive integer $K$, both $1:K$ and $[K]$ stand for the set $\{1, 2, \ldots, K\}$. For any set $S \subseteq [K]$, $|S|$ denotes its cardinality, and $S^c$ denotes its complement. Without further notice, $c$, $C$, $C_1$, $C_2$, $\ldots$ represent some positive constants and can change from line to line. 


\section{Multi-task Learning}\label{sec: mtl}

\subsection{Problem setting}\label{subsec: problem mtl}

Suppose there are $K$ tasks, for which we have $n_k$ observations $\{\bz_i^{(k)}\}_{i=1}^{n_k}$ from the $k$-th task. Suppose there exists an \textit{unknown} subset $S \subseteq 1:K$, such that observations from each task in $S$ independently follow a GMM, while samples from tasks outside $S$ can be arbitrarily distributed. This means,
\begin{align}
	y_i^{(k)} = \begin{cases}
		1, &\text{with probability }1-\wks{k};\\
		2, &\text{with probability }\wks{k};
	\end{cases}\quad\quad   
	\bz_i^{(k)}|y_i^{(k)}=r \sim \mathcal{N}(\bmuks{k}_r, \bSigmaks{k}), \quad \quad r = 1,2,
\end{align}
for all $k \in S$, $i = 1:n_k$, and
\begin{equation}
	\{\bzk{k}_i\}_{i,k\in S^c} \sim \mathbb{Q}_{S},
\end{equation}
where $\mathbb{Q}_S$ is some probability measure on $(\mathbb{R}^p)^{\otimes \nsc}$ and $\nsc = \sum_{k \in S^c}n_k$. In unsupervised learning, we have no access to the true labels $\{\yk{k}_i\}_{i,k}$. To formalize the multi-task learning problem, we first introduce the parameter space for a single GMM:
\begin{align}
	\overline{\Theta} = \{\overline{\btheta} = (w, \bmu_1, \bmu_2, \bSigma): & \twonorm{\bmu_1} \vee \twonorm{\bmu_2} \leq M, w \in (c_w, 1-c_w), \\
	&c_{\bSigma}^{-1} \leq \lambda_{\min}(\bSigma) \leq \lambda_{\max}(\bSigma) \leq c_{\bSigma}\}, \label{eq: parameter space Theta}
\end{align}
where $M$, $c_w \in (0,1/2]$ and $c_{\bSigma}$ are some fixed positive constants. For simplicity, throughout the main text, we assume these constants are fixed. Hence, we have suppressed the dependency on them in the notation $\overline{\Theta}$. The parameter space $\overline{\Theta}$ is a standard formulation. Similar parameter spaces have been considered, for example, in \cite{cai2019chime}.

Our goal of multi-task learning is to leverage the potential similarity shared by different tasks in $S$ to collectively learn them. The tasks outside $S$ can be arbitrarily distributed, and they can potentially be outlier tasks. This motivates us to define a joint parameter space for GMMs in $S$:
\begin{align}
	\overline{\Theta}_{S}(h) 
	= \Big\{\{\overline{\btheta}^{(k)}\}_{k \in S} = \{(\wk{k}, \bmuk{k}_1, \bmuk{k}_2, \bSigmak{k})\}_{k  \in S}: \othetak{k} \in \overline{\Theta}, \inf_{\overline{\bbeta}}\max_{k \in S}\twonorm{\bbetak{k}-\overline{\bbeta}} \leq h\Big\}, \\ \label{eq: parameter space mtl}
\end{align}
where $\bbetak{k} = (\bSigmak{k})^{-1}(\bmuk{k}_2 - \bmuk{k}_1)$ is called the discriminant coefficient in the $k$-th task (recall Section \ref{subsec: intro gmm}). For convenience, we define $\deltak{k} = (\bbetak{k})^\top(\bmuk{k}_1 + \bmuk{k}_2)/2$, which together with $\log((1-\wk{k})/\wk{k})$ is part of the decision boundary. Note that this parameter space is defined only for GMMs of tasks in $S$. To model potentially corrupted or contaminated data, we do not impose any distributional constraints for tasks in $S^c$. Such a modeling framework is reminiscent of Huber's $\epsilon$-contamination model \citep{huber1964robust}. Similar formulations have been adopted in recent multi-task learning research, such as \cite{konstantinov2020sample} and \cite{duan2023adaptive}.

For GMMs in $\overline{\Theta}_S(h)$, we assume that they share similar discriminant coefficients. The similarity is formalized by assuming that all the discriminant coefficients in $S$ are within Euclidean distance $h$ from a ``center". Given that the discriminant coefficient has a major impact on the clustering performance (see the discriminant rule in \eqref{eq: bayes lda intro section}), the parameter space $\overline{\Theta}_S(h)$ is tailored to characterize the task relatedness from the clustering perspective. A similar viewpoint that focuses on modeling the discriminant coefficient has appeared in the study of high-dimensional GMM clustering \citep{cai2019chime} and sparse linear discriminant analysis \citep{cai2011direct, mai2012direct}. With both $S$ and $h$ being unknown in practice, we aim to develop a multi-task learning procedure that is robust to outlier tasks in $S^c$, and achieves improved performance for tasks in $S$ (compared to the single-task learning), in terms of discriminant coefficient estimation and clustering, whenever $h$ is small.

The parameter space does not require the mean vectors $\{\bmuk{k}_1, \bmuk{k}_2\}_{k \in S}$ or the covariance matrices $\{\bSigmak{k}\}_{k \in S}$ to be similar, although they are not free parameters due to the constraint on $\{\bbetak{k}\}_{k \in S}$. And the mixture proportions $\{\wk{k}\}_{k \in S}$ do not need to be similar either. We thus avoid imposing restrictive conditions on those parameters. On the other hand, it implies that estimation of the mixture proportions, mean vectors, and covariance matrices in multi-task learning may not be generally improvable over that in single-task learning. This is verified by the theoretical results in Section \ref{subsec: theory mtl}. While the current treatment in the paper does not consider similarity structure among $\{\bmuk{k}_1, \bmuk{k}_2\}_{k \in S}, \{\bSigmak{k}\}_{k \in S}$ or $\{\wk{k}\}_{k \in S}$, our methods and theory can be readily adapted to handle such scenarios, if desired.

There are two main reasons why this MTL problem can be challenging. First, commonly used strategies like data pooling are fragile with respect to outlier tasks and can lead to arbitrarily inaccurate outcomes in the presence of even a small number of outliers. Also, since the distribution of data from outlier tasks can be adversarial to the learner, the idea of outlier task detection in the recent literature \citep{li2021transfer, tian2023transfer} may not be applicable. Second, to address the nonconvexity of the likelihood, we propose to explore the similarity among tasks via a generalization of the EM algorithm. However, a clear theoretical understanding of such an iterative procedure requires a delicate analysis of the whole iterative process. In particular, as in the analysis of EM algorithms \citep{cai2019chime, kwon2020algorithm}, the estimates of similar discriminant vectors $\{\bbetaks{k}\}_{k \in S}$ and other potentially dissimilar parameters are entangled in the iterations. It is highly non-trivial to separate the impact of estimating $\{\bbetaks{k}\}_{k \in S}$ and other parameters to derive the desired statistical error rates. We address this challenge through a localization technique by carefully shrinking the analysis radius of estimators as the iteration proceeds.

\subsection{Method}\label{subsec: method mtl}
We aim to tackle the problem of GMM estimation under the context of multi-task learning. The EM algorithm is commonly used to address the non-convexity of the log-likelihood function arising from the latent labels. In the standard EM algorithm, we ``classify" the observations (update the posterior) in E-step and update the parameter estimations in M-step \citep{redner1984mixture}. For multi-task and transfer learning problems, the penalization framework is very popular, where we solve an optimization problem based on a new objective function. This objective function consists of a local loss function and a penalty term, forcing the estimators of similar tasks to be close to each other. For examples, see \cite{zhang2011multitask, zhang2015smart, bastani2021predicting, xu2021learning, li2021transfer, duan2023adaptive, lin2022transfer, li2023targeting, tian2023transfer}. Thus motivated, our method seeks a combination of the EM algorithm and the penalization framework.

In particular, we adapt the penalization framework of \cite{duan2023adaptive} and modify the updating formulas in the M-step accordingly. The proposed procedure is summarized in Algorithm \ref{algo: multitask}. For simplicity, in Algorithm \ref{algo: multitask} we have used the notation
\begin{equation}
	\gamma_{\btheta}(\bz) = \frac{w\exp(\bbeta^\top\bz - \delta)}{1-w+w\exp(\bbeta^\top\bz - \delta)}, \quad  \text{for } \btheta = (w, \bbeta, \delta).
\end{equation}
Note that $\gamma_{\btheta}(\bz)$ is the posterior probability $\tp(Y=2|Z=\bm{z})$ given the observation $\bz$. The estimated posterior probability is calculated in every E-step, given the updated parameter estimates.

\begin{algorithm}[!h]
\linespread{1.48}\selectfont
\caption{MTL-GMM}
\label{algo: multitask}
\KwIn{Initialization $\{(\hw^{(k)[0]}, \hbeta^{(k)[0]}, \hmu^{(k)[0]}_1, \hmu^{(k)[0]}_2)\}_{k=1}^K$, maximum number of iteration rounds $T$, initial penalty parameter $\lambda^{[0]}$, tuning parameters $C_{\lambda} > 0$ and $\kappa \in (0, 1)$}
$\htheta^{(k)[0]} = (\hw^{(k)[0]}, \hbeta^{(k)[0]}, \hdelta^{(k)[0]})$ and $\hdelta^{(k)[0]} = \frac{1}{2}(\hbeta^{(k)[0]})^\top(\hmu^{(k)[0]}_1 + \hmu^{(k)[0]}_2)$ for $k = 1:K$\\
\For{$t = 1$ \KwTo $T$}{
	$\lambda^{[t]} = \kappa \lambda^{[t-1]} + C_{\lambda}\sqrt{p+\log K}$ \tcp*[f]{Penalty parameter update}\\
	\For(\tcp*[f]{Local update for each task}){$k = 1$ \KwTo $K$}{ 
		$\hw^{(k)[t]} = \frac{1}{n_k}\sum_{i=1}^{n_k}\gamma_{\htheta^{(k)[t-1]}}(\bz_i^{(k)})$\\ 
		$\hmu^{(k)[t]}_1 = \frac{\sum_{i=1}^{n_k}[1-\gamma_{\htheta^{(k)[t-1]}}(\bz_i^{(k)})]\bzk{k}_i}{n_k(1-\hw^{(k)[t]})}$, $\hmu^{(k)[t]}_2 = \frac{\sum_{i=1}^{n_k}\gamma_{\htheta^{(k)[t-1]}}(\bz_i^{(k)})\bzk{k}_i}{n_k\hw^{(k)[t]}}$ \\
		$\hSigma^{(k)[t]} = \frac{1}{n_k}\sum_{i=1}^{n_k}\left\{[1-\gamma_{\htheta^{(k)[t-1]}}(\bz_i^{(k)})]\cdot (\bzk{k}_i - \hmu^{(k)[t]}_1)(\bzk{k}_i - \hmu^{(k)[t]}_1)^\top \right.$ $\left. \hspace{3.3cm} + \gamma_{\htheta^{(k)[t-1]}}(\bz_i^{(k)}) \cdot (\bzk{k}_i - \hmu^{(k)[t]}_2)(\bzk{k}_i - \hmu^{(k)[t]}_2)^\top\right\}$
	}
	$\{\hbeta^{(k)[t]}\}_{k=1}^K$, $\obbeta^{[t]}= \argmin\limits_{\bbetak{1}, \ldots, \bbetak{K}, \overline{\bbeta}}\bigg\{\sum_{k=1}^K n_k\Big[\frac{1}{2}(\bbetak{k})^\top\hSigma^{(k)[t]}\bbetak{k} - (\bbetak{k})^\top(\hmu_2^{(k)[t]}-\hmu_1^{(k)[t]})\Big] + \sum_{k=1}^K\sqrt{n_k}\lambda^{[t]}\cdot \twonorm{\bbetak{k}-\overline{\bbeta}}\bigg\}$\tcp*[f]{Aggregation to learn $\{\hbeta^{(k)[t]}\}_{k=1}^K$} \\
	\For(\tcp*[f]{Local update for each task}){$k = 1$ \KwTo $K$}{
		$\hdelta^{(k)[t]} = \frac{1}{2}(\hbeta^{(k)[t]})^\top(\hmu^{(k)[t]}_1 + \hmu^{(k)[t]}_2)$\\
		Let $\htheta^{(k)[t]} = (\hw^{(k)[t]}, \hbeta^{(k)[t]}, \hdelta^{(k)[t]})$
	}
	
}
\KwOut{$\{(\htheta^{(k)[T]}, \hmu^{(k)[T]}_1, \hmu^{(k)[T]}_2, \hSigma^{(k)[T]})\}_{k=1}^K$ with $\htheta^{(k)[T]} = (\hw^{(k)[T]}, \hbeta^{(k)[T]}, \hdelta^{(k)[T]})$, and $\obbeta^{[T]}$}
\end{algorithm}

Recall that the parameter space $\overline{\Theta}_S(h)$ introduced in \eqref{eq: parameter space mtl} does not encode similarity for the mixture proportions $\{\wks{k}\}_{k \in S}$, mean vectors $\{\bmuks{k}_1, \bmuks{k}_2\}_{k \in S}$, or covariance matrices $\{\bSigmaks{k}\}_{k \in S}$. Hence, their updates in Steps 5-7 are kept the same as those in the standard EM algorithm. Regarding the update for discriminant coefficients in Step 9, the quadratic loss function is motivated by the direct estimation of the discriminant coefficient in high-dimensional GMM \citep{cai2019chime} and high-dimensional LDA literature \citep{cai2011direct, witten2011penalized, fan2012road, mai2012direct, mai2019multiclass}. The penalty term in Step 9 penalizes the contrasts of $\bbetak{k}$'s to exploit the similarity structure among tasks. The ``center" parameter $\overline{\bbeta}$ in the penalization induces robustness against outlier tasks. We refer to \cite{duan2023adaptive} for a systematic treatment of this penalization framework. It is straightforward to verify that when the tuning parameters $\{\lambda^{[t]}\}_{t=1}^T$ are set to zero, Algorithm \ref{algo: multitask} reduces to the standard EM algorithm performed separately on the $K$ tasks. That is, for each $k = 1:K$, given the parameter estimate from the previous step $\htheta^{(k)[t-1]} = (\hw^{(k)[t-1]}, \hbeta^{(k)[t-1]}, \hdelta^{(k)[t-1]})$, we update $\hw^{(k)[t]}$, $\hmu^{(k)[t]}_1$, $\hmu^{(k)[t]}_2$, $\hSigma^{(k)[t]}$, and $\hdelta^{(k)[t]}$ as in Algorithm \ref{algo: multitask}, and update $\hbeta^{(k)[t]}$ via
\begin{equation}
	\hbeta^{(k)[t]} = (\hSigma^{(k)[t]})^{-1}(\hmu^{(k)[t]}_2 - \hmu^{(k)[t]}_1).
\end{equation}
For the maximum number of iteration rounds, $T$, our theory will show that $T \gtrsim \log (\max_{k=1:K}n_k)$ is sufficient to reach the desired statistical error rates. In practice, we can terminate the iteration when the change of estimates within two successive rounds falls below some pre-set small tolerance level. We discuss the initialization in detail in Sections \ref{subsec: theory mtl} and \ref{subsec: alignment mtl}.

Lastly, it is worth noting that the aggregation in Step 9 is central to Algorithm \ref{algo: multitask}. The joint optimization over the individual parameters $\{\bbetak{k}\}_{k=1}^K$ and a global parameter $\obbeta$ using an $\ell_2$-penalty provides sufficient flexibility for multi-task learning. As we will show later, with appropriate choices of the penalty parameter $\lambda^{[t]}$, Step 9 enables adaptive information sharing: when the $\bbetaks{k}$'s are sufficiently similar, the penalty reduces estimation variance and encourages the estimators $\{\bbetak{k}\}_{k=1}^K$ to be close to one another; when the $\bbetaks{k}$'s differ significantly, the strongly convex squared loss dominates the penalty term, ensuring that the estimators remain close to their corresponding single-task solutions. Moreover, as discussed in \cite{she2011outlier} and \cite{donoho2016high}, this form of penalization is connected to robust empirical risk minimization. Specifically, the equivalent loss function (i.e., the objective function in Step 9 after profiling out $\{\bbetak{k}\}_{k=1}^K$) for estimating the global parameter $\obbeta$ can be shown to have some robustness properties, providing robustness against a small proportion of outlier tasks.

\subsection{Theory}\label{subsec: theory mtl}
In this section, we develop statistical theories for our proposed procedure MTL-GMM (see Algorithm \ref{algo: multitask}). As mentioned in Section \ref{subsec: problem mtl}, we are interested in the performance of both parameter estimation and clustering, although the latter is the main focus and motivation. First, we impose conditions in the following assumption set.

\begin{assumption}\label{asmp: upper bound multitask est error}	
Denote $\Delta^{(k)} = \sqrt{(\bmuks{k}_1-\bmuks{k}_2)^\top(\bSigmaks{k})^{-1}(\bmuks{k}_1-\bmuks{k}_2)}$ for $k \in S$. The quantity $\Delta^{(k)}$ is the Mahalanobis distance between $\bmuks{k}_1$ and $\bmuks{k}_2$ with covariance matrix $\bSigmaks{k}$, and can be viewed as the signal-to-noise ratio (SNR) in the $k$-th task \citep{anderson1958introduction}. Suppose the following conditions hold: 
\begin{enumerate}[(i)]
		\item $\ns= \sum_{k \in S}n_k \geq C_1|S|\max_{k = 1:K}n_k$ with a constant $C_1 \in (0, 1]$;
		\item $\min_{k \in S}n_k \geq C_2(p+\log K)$ with some constant $C_2 > 0$;
		\item Either of the following two conditions holds with some constant $C_3 > 0$:
		\begin{enumerate}
			\item $\max_{k \in S}\big(\twonorm{\hbeta^{(k)[0]} - \bbetaks{k}} \vee \twonorm{\hmu^{(k)[0]}_1 - \bmuks{k}_1}\vee \twonorm{\hmu^{(k)[0]}_2 - \bmuks{k}_2}\big) \leq C_3\min_{k \in S}\Delta^{(k)}$, $\max_{k \in S}\norm{\hw^{(k)[0]}-\wks{k}} \leq c_w/2$;
			\item $\max_{k \in S}\big(\twonorm{\hbeta^{(k)[0]} + \bbetaks{k}} \vee \twonorm{\hmu^{(k)[0]}_1 - \bmuks{k}_2}\vee \twonorm{\hmu^{(k)[0]}_2 - \bmuks{k}_1} \big)  \leq C_3\min_{k \in S}\Delta^{(k)}$, $\max_{k \in S}\norm{1-\hw^{(k)[0]}-\wks{k}} \leq c_w/2$.
		\end{enumerate}
		\item $\min_{k \in S}\Delta^{(k)} \geq C_4 > 0$ with some constant $C_4 > 0$;
	\end{enumerate}
\end{assumption}

\begin{remark}\label{rmk: asmp1}
	These are common and mild conditions related to the sample size, initialization, and signal-to-noise ratio of GMMs. Condition (\rom{1}) requires the maximum sample size of all tasks not to be much larger than the average sample size of tasks in $S$. Similar conditions can be found in \cite{duan2023adaptive}. Condition (\rom{2}) is the requirement of the sample size of tasks in $S$. The usual condition for low-dimensional single-task learning is $n_k \gtrsim p$ \citep{cai2019chime}. The additional $\log K$ term arises from the simultaneous control of performance on all tasks in $S$, where $S$ can be as large as $1:K$. Condition (\rom{3}) requires that the initialization should not be too far away from the truth, which is commonly assumed in either the analysis of EM algorithm \citep{redner1984mixture, balakrishnan2017statistical, cai2019chime} or other iterative algorithms like the local estimation used in semi-parametric models \citep{carroll1997generalized, li2008variable} and adaptive Lasso \citep{zou2006adaptive}. The two possible forms considered in this condition are due to the fact that binary GMM is only identifiable up to label permutation. Condition (\rom{4}) requires that the signal strength of GMM (in terms of Mahalanobis distance) is strong enough, which is usually assumed in the literature about the likelihood-based methods of GMMs \citep{dasgupta2000two, azizyan2013minimax, balakrishnan2017statistical, cai2019chime}. 
\end{remark}

We first establish the rate of convergence for the estimation. Recalling the parameter space $\overline{\Theta}_S(h)$ in \eqref{eq: parameter space mtl}, let us denote the true parameter by 
\[
\{\othetaks{k}\}_{k \in S} = \{(\wks{k}, \bmuks{k}_1, \bmuks{k}_2, \bSigmaks{k})\}_{k  \in S} \in \overline{\Theta}_{S}(h).
\]
To better present the results for parameters related to the optimal discriminant rule \eqref{eq: bayes lda intro section}, we further denote 
\[
\bthetaks{k} = (\wks{k}, \bbetaks{k}, \deltaks{k}), \quad \forall k \in S,
\]
where $\bbetaks{k}=(\bSigmaks{k})^{-1}(\bmuks{k}_2-\bmuks{k}_1), \deltaks{k}=\frac{1}{2}(\bbetaks{k})^\top(\bmuks{k}_1+\bmuks{k}_2)$. Note that $\bthetaks{k}$ is a function of $\othetaks{k}$. For the estimators returned by MTL-GMM (see Algorithm \ref{algo: multitask}), we are particularly interested in the following two error metrics:
\begin{align}
& d(\htheta^{(k)[T]}, \bthetaks{k}):= \min\{\norm{\hw^{(k)[T]}-\wks{k}} \vee \twonorm{\hbeta^{(k)[T]}-\bbetaks{k}} \vee \norm{ \hdelta^{(k)[T]}-\deltaks{k}}, \\ 
&\hspace{4cm} \norm{1-\hw^{(k)[T]}-\wks{k}} \vee \twonorm{\hbeta^{(k)[T]}+\bbetaks{k}} \vee \norm{ \hdelta^{(k)[T]}+\deltaks{k}}\},  \label{errm:1} \\
& \Big(\min_{\pi: [2] \rightarrow [2]}\max_{r=1:2}\twonorm{\hmu^{(k)[T]}_r - \bmuks{k}_{\pi(r)}}\Big) \vee \twonorm{\hSigma^{(k)[T]}-\bSigmaks{k}}, \label{errm:2}
\end{align}
where $\pi: [2] \rightarrow [2]$ is a permutation on $\{1,2\}$. Again, we take the minimum above because binary GMM is identifiable up to label permutation. The first error metric $d(\htheta^{(k)[T]}, \bthetaks{k})$ involves the error for discriminant coefficients and is closely related to the clustering performance. It reveals how well our method utilizes similarity structure in multi-task learning. The second error metric is about the mean vectors and covariance matrix. As discussed in Section \ref{subsec: problem mtl}, we shall not expect it to be improved compared to that in single-task learning, as these parameters are not necessarily similar.

We are ready to present upper bounds for the estimation error of MTL-GMM. We recall that $\overline{\Theta}_{S}(h)$ and $\mathbb{Q}_S$ are the parameter space and probability measure that we use in Section \ref{subsec: problem mtl} to describe the data distributions for tasks in $S$ and $S^c$, respectively.

\begin{theorem}(Upper bounds of the estimation error of GMM parameters for MTL-GMM)\label{thm: upper bound multitask est error}
	Suppose Assumption \ref{asmp: upper bound multitask est error} holds for some $S$ with $|S| \geq s$ and $\epsilon \coloneqq \frac{K-s}{K} < 1/3$. Let $\lambda^{[0]} \geq C_1\max_{k=1:K}\sqrt{n_k}$, $C_{\lambda} \geq C_1$ and $\kappa > C_2$ with some constants $C_1 > 0, C_2 \in (0, 1)$ \footnote{$C_1$ and $C_2$ depend on the constants $M$, $c_w$, and $c_{\bSigma}$ etc.}. Then there exists a constant $C_3 > 0$, such that for any $\{\othetaks{k}\}_{k \in S} = \{(\wks{k}, \bmuks{k}_1, \bmuks{k}_2, \bSigmaks{k})\}_{k  \in S} \in \overline{\Theta}_{S}(h)$ and any probability measure $\mathbb{Q}_S$ on $(\mathbb{R}^p)^{\otimes n_{S^c}}$, with probability $1-C_3K^{-1}$, the following hold for all $k \in S$:
	\begin{align}
		d(\htheta^{(k)[T]}, \bthetaks{k}) &\lesssim \sqrt{\frac{p}{\ns}}+ \sqrt{\frac{\log K}{n_k}} + h\wedge \sqrt{\frac{p+\log K}{n_k}} + \epsilon \sqrt{\frac{p + \log K}{\max_{k=1:K}n_k}}+ T^2(\kappa')^\top, \quad
	\end{align}
	\begin{equation}
		\Big(\min_{\pi: [2] \rightarrow [2]}\max_{r=1:2}\twonorm{\hmu^{(k)[T]}_r - \bmuks{k}_{\pi(r)}}\Big)  \vee \twonorm{\hSigma^{(k)[T]} - \bSigmaks{k}} \lesssim \sqrt{\frac{p + \log K}{n_k}} +  T^2(\kappa')^\top,
	\end{equation}
	where $\kappa' \in (0, 1)$ is some constant and $\ns= \sum_{k \in S}n_k$. When $T \geq C\log(\max_{k=1:K} n_k)$ with a large constant $C > 0$, the last term on the right-hand side will be dominated by other terms in both inequalities.
\end{theorem}

The upper bound of $\big(\min_{\pi: [2] \rightarrow [2]}\max_{r=1:2}\twonorm{\hmu^{(k)[T]}_r - \bmuks{k}_{\pi(r)}}\big) \vee \twonorm{\hSigma^{(k)[T]}-\bSigmaks{k}}$ contains two parts. The first part is comparable to the single-task learning rate \citep{cai2019chime} (up to a $\sqrt{\log K}$ term due to the simultaneous control over all tasks in $S$), and the second part characterizes the geometric convergence of iterates. As expected, since $\bmuks{k}_1$, $\bmuks{k}_2$, $\bSigmaks{k}$ in $S$ are not necessarily similar, an improved error rate over single-task learning is generally impossible. The upper bound for $d(\htheta^{(k)[T]}, \bthetaks{k})$ is directly related to the clustering performance of our method. Thus, we will provide a detailed discussion of it after presenting the clustering result in the next theorem.

As introduced in Section \ref{subsec: intro gmm}, using the estimate $\htheta^{(k)[T]} = (\hw^{(k)[T]}, \hbeta^{(k)[T]}, \hdelta^{(k)[T]})$ from Algorithm \ref{algo: multitask}, we can construct a classifier for task $k$ as
\begin{equation}\label{eq: clustering method ck}
	\widehat{\mathcal{C}}^{(k)[T]}(\bz) = \begin{cases}
		1, & \text{if } (\hbeta^{(k)[T]})^\top\bz - \hdelta^{(k)[T]} \leq \log\left(\frac{1-\hw^{(k)[T]}}{\hw^{(k)[T]}}\right);\\
		2, & \text{otherwise}.
	\end{cases}
\end{equation}
Recall that for a clustering method $\mC: \mathbb{R}^p \rightarrow \{1,2\}$, its mis-clustering error rate under GMM with parameter $\otheta =(w, \bmu_1, \bmu_2, \bSigma)$ is
\begin{equation}\label{eq: clustering method ck error}
	R_{\otheta}(\mC) = \min_{\pi: [2] \rightarrow [2]}\tp_{\otheta}(\mC(Z^{\textup{new}}) \neq \pi(Y^{\textup{new}})),
\end{equation}
where $Z^{\textup{new}} \sim (1-w)\mathcal{N}(\bmu_1, \bSigma) + w\mathcal{N}(\bmu_2, \bSigma)$ is a future observation associated with the label $Y^{\textup{new}}$, independent from $\mC$; the probability $\tp_{\otheta}$ is w.r.t. $(Z^{\textup{new}}, Y^{\textup{new}})$, and the minimum is taken over two permutation functions on $\{1, 2\}$. Denote $\mC_{\otheta}$ as the Bayes classifier that minimizes $R_{\otheta}(\mC)$. In the following theorem, we obtain the upper bound of the excess mis-clustering error of $\widehat{\mathcal{C}}^{(k)[T]}$ for $k \in S$.

\begin{theorem}(Upper bound of the excess mis-clustering error for MTL-GMM)\label{thm: upper bound multitask classification error}
Suppose the same conditions in Theorem \ref{thm: upper bound multitask est error} hold. Then there exists a constant $C_1 > 0$ such that for any $\{\othetaks{k}\}_{k \in S} \in \overline{\Theta}_{S}(h)$ and any probability measure $\mathbb{Q}_S$ on $(\mathbb{R}^p)^{\otimes n_{S^c}}$, with probability $1-C_1K^{-1}$, the following holds for all $k \in S$:
\begin{align}
	R_{\othetaks{k}}(\widehat{\mathcal{C}}^{(k)[T]}) - R_{\othetaks{k}}(\mC_{\othetaks{k}}) &\lesssim d^2(\htheta^{(k)[T]}, \bthetaks{k})\\
	&\lesssim \underbrace{\frac{p}{\ns}}_{\rm (\Rom{1})} + \underbrace{\frac{\log K}{n_k}}_{\rm (\Rom{2})}  + \underbrace{h^2 \wedge \frac{p+\log K}{n_k}}_{\rm (\Rom{3})} + \underbrace{\epsilon^2 \frac{p + \log K}{\max_{k=1:K}n_k}}_{\rm (\Rom{4})} + \underbrace{T^4(\kappa')^{2T}}_{\rm (\Rom{5})},
\end{align}
with some $\kappa' \in (0, 1)$. When $T \geq C\log (\max_{k=1:K} n_k)$ with a large constant $C > 0$, the last term on the right-hand side will be dominated by the second term.
\end{theorem}

The upper bounds of $d(\htheta^{(k)[T]}, \bthetaks{k})$ in Theorem \ref{thm: upper bound multitask est error} and $R_{\othetaks{k}}(\widehat{\mathcal{C}}^{(k)[T]}) - R_{\othetaks{k}}(\mC_{\othetaks{k}})$ in Theorem \ref{thm: upper bound multitask classification error} consist of five parts with one-to-one correspondence. It is sufficient to discuss the bound of $R_{\othetaks{k}}(\widehat{\mathcal{C}}^{(k)[T]}) - R_{\othetaks{k}}(\mC_{\othetaks{k}})$. Part (\Rom{1}) represents the ``oracle rate", which can be achieved when all tasks in $S$ are the same. This is the best rate to achieve possibly. Part (\Rom{2}) is a dimension-free error caused by estimating scalar parameters $\deltaks{k}$ and $\wks{k}$ that appears in the optimal discriminant rule. Part (\Rom{3}) includes $h$ that measures the degree of similarity among the tasks in $S$. When these tasks are very similar, $h$ will be small, contributing a small term to the upper bound. Nicely, even when $h$ is large, the term becomes $\frac{p+\log K}{n_k}$, and it is still comparable to the minimax error rate of single-task learning $\mathcal{O}_{\tp}(p/n_k)$ (e.g., Theorems 4.1 and 4.2 in \cite{cai2019chime}). We have the extra $\log K$ term here due to the simultaneous control over all tasks in $S$. Part (\Rom{4}) quantifies the influence from the outlier tasks in $S^c$. When there are more outlier tasks, $\epsilon$ increases, and the bound becomes worse. On the other hand, as long as $\epsilon$ is small enough to make this term dominated by any other part, the error rate induced by outlier tasks becomes negligible. Given that data from outlier tasks can be arbitrarily contaminated, we can conclude that our method is robust against a fraction of outlier tasks from arbitrary sources.  The term in Part (\Rom{5}) decreases geometrically in the iteration number $T$, implying that the iterates in Algorithm \ref{algo: multitask} converge geometrically to a ball of radius determined by the errors from Parts (\Rom{1})-(\Rom{4}).

After explaining each part of the upper bound, we now compare it with the convergence rate $\mathcal{O}_{\tp}(\frac{p+\log K}{n_k})$ (including $\log K$ here since we consider all the tasks simultaneously) in the single-task learning and reveal how our method performs. With a quick inspection, we can conclude the following:
\begin{itemize}
\item The rate of the upper bound is never larger than $\frac{p+\log K}{n_k}$. So, in terms of convergence rate, our method MTL-GMM performs at least as well as single-task learning, regardless of the similarity level $h$ and outlier task fraction $\epsilon$. 
\item When $\ns \gg n_k$ (large total sample size for tasks in $S$), $p$ increases with $n_k$ (diverging dimension), $h \ll \sqrt{\frac{p+\log K}{n_k}}$ (sufficient similarity between tasks in $S$), and $\epsilon \ll \sqrt{(\max_{k=1:K}n_k)/n_k}$  (small fraction of outlier tasks), MTL-GMM attains a faster excess mis-clustering error rate and improves over single-task learning. 
\end{itemize}

The preceding discussions on the upper bounds have demonstrated the superiority of our method. But can we do better? To further evaluate the upper bounds of our method, we next derive complementary minimax lower bounds for both estimation error and excess mis-clustering error. We will show that our method is (nearly) minimax rate optimal in a broad range of regimes.

\begin{theorem}(Lower bounds of the estimation error of GMM parameters in multi-task learning)\label{thm: lower bound multitask est error}
	Suppose $\epsilon = \frac{K-s}{K} < 1/3$. Suppose there exists a subset $S$ with $|S| \geq s$ such that $\min_{k \in S} n_k \geq C_1(p + \log K)$ and $\min_{k \in S}\Delta^{(k)} \geq C_2$, where $C_1, C_2 > 0$ are some constants. Then 
	\begin{align}
		\inf_{\{\htheta^{(k)}\}_{k=1}^K} \sup_{S: |S| \geq s}\sup_{\substack{\{\othetaks{k}\}_{k \in S} \in \overline{\Theta}_S(h) \\ \mathbb{Q}_S}} &\tp\Bigg(\bigcup_{k \in S}\Bigg\{d(\htheta^{(k)}, \bthetaks{k}) \gtrsim \sqrt{\frac{p}{\ns}} + \sqrt{\frac{\log K}{n_k}} \\
		&\quad\quad + h \wedge \sqrt{\frac{p+\log K}{n_k}} + \frac{\epsilon}{\sqrt{\max_{k=1:K}n_k}}\Bigg\}\Bigg) \geq \frac{1}{10},	
	\end{align}
	
	\begin{align}
		\inf_{\substack{\{\hmu^{(k)}_1, \hmu^{(k)}_2\}_{k=1}^K \\\{\hSigma^{(k)}\}_{k=1}^K}} \sup_{S: |S| \geq s}\sup_{\substack{\{\othetaks{k}\}_{k \in S} \in \overline{\Theta}_S(h) \\ \mathbb{Q}_S}} \tp&\Bigg(\bigcup_{k \in S}\Bigg\{\Big(\min_{\pi: [2] \rightarrow [2]}\max_{r=1:2}\twonorm{\hmu^{(k)}_r - \bmuks{k}_{\pi(r)}}\Big) \vee \twonorm{\hSigma^{(k)} - \bSigmaks{k}}
		\\
		&\quad\quad\quad \gtrsim \sqrt{\frac{p + \log K}{n_k}}\Bigg\}\Bigg) \geq \frac{1}{10}.
	\end{align}
\end{theorem}

\begin{theorem}(Lower bound of the excess mis-clustering error in multi-task learning)\label{thm: lower bound multitask classification error}
	Suppose the same conditions in Theorem \ref{thm: lower bound multitask est error} hold. Then
	\begin{align}
		\inf_{\{\widehat{\mathcal{C}}^{(k)}\}_{k=1}^K} \sup_{S: |S| \geq s}\sup_{\substack{\{\othetaks{k}\}_{k \in S} \in \overline{\Theta}_S(h) \\ \mathbb{Q}_S}} &\tp\Bigg(\bigcup_{k \in S}\Bigg\{R_{\othetaks{k}}(\widehat{\mathcal{C}}^{(k)}) - R_{\othetaks{k}}(\mC_{\othetaks{k}}) \gtrsim \frac{p}{\ns} + \frac{\log K}{n_k}   \\
		&\quad\quad\quad\quad +  h^2 \wedge \frac{p+\log K}{n_k} + \frac{\epsilon^2}{\max_{k=1:K}n_k} \Bigg\}\Bigg) \geq \frac{1}{10}.
	\end{align}
\end{theorem}
Comparing the upper and lower bounds in Theorems \ref{thm: upper bound multitask est error}-\ref{thm: lower bound multitask classification error}, we make several remarks:
\begin{itemize}
	\item Regarding the estimation of mean vectors $\{\bmuks{k}_1, \bmuks{k}_2\}_{k \in S}$ and covariance matrices $\{\bSigmaks{k}\}_{k \in S}$, the upper and lower bounds match, hence our method is minimax rate optimal.
	\item For the estimation error $d(\htheta^{(k)}, \bthetaks{k})$ and excess mis-clustering error $R_{\othetaks{k}}(\widehat{\mathcal{C}}^{(k)[T]}) - R_{\othetaks{k}}(\mC_{\othetaks{k}})$ with $k \in S$, the first three terms in the upper and lower bounds match. Only the term involving $\epsilon$ in the lower bound differs from that in the upper bound by a factor $\sqrt{p+\log K}$ or $p+\log K$. As a result, in the classical low-dimensional regime where $p$ is bounded, the upper and lower bounds match (up to a logarithmic factor). Therefore, our method is (nearly) minimax rate optimal for estimating $\{\bthetaks{k}\}_{k \in S}$ and clustering in such a classical regime. 
	\item When the dimension $p$ diverges, there might exist a non-negligible gap between the upper and lower bounds for $d(\htheta^{(k)}, \bthetaks{k})$ and $R_{\othetaks{k}}(\widehat{\mathcal{C}}^{(k)[T]}) - R_{\othetaks{k}}(\mC_{\othetaks{k}})$ with $k \in S$. Nevertheless, this only occurs when the fraction of outlier task $\epsilon$ is above the threshold $\sqrt{\frac{\max_{k=1:K}n_k}{p+\log K}\big(h^2 \vee \frac{\log K}{n_k} \vee \frac{p}{n_S}\big)}$. Below the threshold, our method remains minimax rate optimal even though $p$ is unbounded. 
	\item Does the gap, when it exists, arise from the upper bound or the lower bound? It is the upper bound that sometimes becomes not sharp. As can be seen from the proof of Theorem \ref{thm: upper bound multitask est error}, the term $\epsilon \sqrt{\frac{p + \log K}{\max_{k = 1:K}n_k}}$ is due to the estimation of those ``center" parameters in Algorithm \ref{algo: multitask}. Recent advances in robust statistics \citep{chen2018robust} have shown that estimators based on statistical depth functions such as Tukey's depth function \citep{tukey1975mathematics} can achieve optimal minimax rate under Huber's $\epsilon$-contamination model for location and covariance estimation. It might be possible to utilize depth functions to estimate ``center" parameters in our problem and kill the factor $\sqrt{p}$ in the upper bound. We leave the rigorous development of optimal robustness as an interesting future research topic. On the other hand, such statistical improvement may come with expensive computation, as depth function-based estimation typically requires solving a challenging non-convex optimization problem.
\end{itemize}

\subsection{Initialization and cluster alignment}\label{subsec: alignment mtl}
As specified by Condition (\rom{3}) in Assumption \ref{asmp: upper bound multitask est error}, our proposed learning procedure requires that for each task in $S$, initial values of the GMM parameter estimates lie within a distance of SNR-order from the ground truth. This condition can be satisfied by various estimators, such as the method of moments proposed in \cite{ge2015learning} and the robust initialization procedures developed in \cite{jana2024provable}. In practice, a natural initialization method is to run the standard EM algorithm or other common clustering methods like $k$-means on each task and use the corresponding estimate as the initial values. We adopted the standard EM algorithm in our numerical experiments, and the numerical results in Section \ref{sec: numerical} and supplements showed that this practical initialization works quite well. However, in the context of multi-task learning, Condition (\rom{3}) further requires a correct alignment of those good initializations from each task, owing to the non-identifiability of GMMs. We discuss in detail the alignment issue in Section \ref{subsubsection: alignment issue} and propose two algorithms to resolve this issue in Section \ref{subsubsection: two alignment algs}.

\subsubsection{The alignment issue}\label{subsubsection: alignment issue}
Recall that Section \ref{subsec: problem mtl} introduces the binary GMM with parameters $(\wks{k}, \bmuks{k}_1, \bmuks{k}_2, \bSigmaks{k})$ for each task $k \in S$. Because the two sets of parameter values $\{(w, \bu, \bv, \bSigma), (1-w, \bv, \bu, \bSigma)\}$ for $(\wks{k}, \bmuks{k}_1, \bmuks{k}_2, \bSigmaks{k})$ index the same distribution, a good initialization close to the truth is up to a permutation of the two cluster labels. The permutations in the initialization of different tasks could be different. Therefore, in light of the joint parameter space $\overline{\Theta}_S(h)$ defined in \eqref{eq: parameter space mtl} and Condition (\rom{3}) in Assumption \ref{asmp: upper bound multitask est error}, for given initializations from different tasks, we may need to permute their cluster labels to feed the well-aligned initialization into Algorithm \ref{algo: multitask}.

We further elaborate on the alignment issue using Algorithm 1. The penalization in Step 9 aims to push the estimators $\hbeta^{(k)[t]}$'s with different $k$ towards each other, which is expected to improve the performance thanks to the similarity among underlying true parameters $\{\bbetaks{k}\}_{k \in S}$. However, due to the potential permutation of two cluster labels, the vanilla single-task initializations (without alignment) cannot guarantee that the estimators $\{\hbeta^{(k)[t]}\}_{k \in S}$ at each iteration are all estimating the corresponding $\bbetaks{k}$'s (some may estimate $-\bbetaks{k}$'s). 

\begin{figure}[!h]
	\centering
	\includegraphics[width=0.95\linewidth]{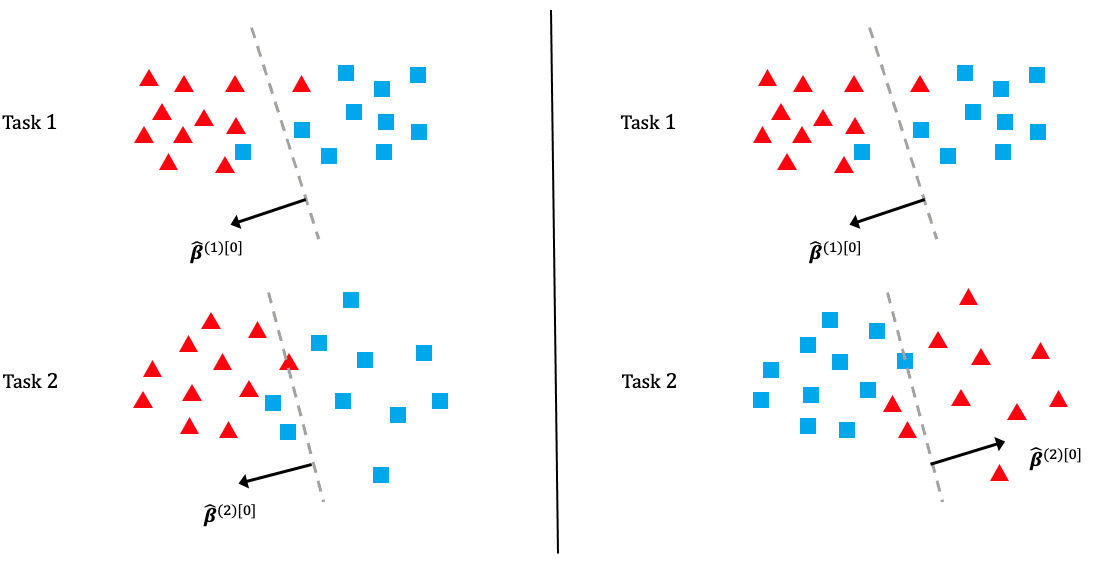}
	\caption{Examples of well-aligned (left) and badly-aligned (right) initializations.}
	\label{fig: alignment}
\end{figure}

Figure \ref{fig: alignment} illustrates the alignment issue in the case of two tasks. The left-hand-side situation is ideal where $\hbeta^{(1)[0]}$, $\hbeta^{(2)[0]}$ are estimates of $\bbetaks{1}$, $\bbetaks{2}$ (which are similar). The right-hand-side situation is problematic because $\hbeta^{(1)[0]}$, $\hbeta^{(2)[0]}$ are estimates of $\bbetaks{1}$, $-\bbetaks{2}$ (which are not similar). Therefore, after obtaining the initializations from each task, it is necessary to align their cluster labels to ensure that estimators of similar parameters are correctly put together in the penalization framework in Algorithm \ref{algo: multitask}. We formalize the problem and provide two solutions in the next subsection.

\subsubsection{Two alignment algorithms}\label{subsubsection: two alignment algs}

Suppose $\{\hbeta^{(k)[0]}\}_{k=1}^K$ are the initial estimates of discriminant coefficients with potentially bad alignment for $k\in S$. Note that a good initialization and alignment is not required (in fact, it is not even well defined) for the outlier tasks in $S^c$, because they can be from arbitrary distributions. However, since $S$ is unknown, we will have to address the alignment issue for tasks in $S$ based on initial estimates from all the tasks. For binary GMMs, each alignment of $\{\hbeta^{(k)[0]}\}_{k=1}^K$ can be represented by a $K$-dimensional Rademacher vector $\bm{r} \in \{\pm 1\}^K$. Define the ideal alignment as $r^*_k = \argmin_{r_k = \pm 1} \twonorm{r_k\hbeta^{(k)[0]} - \bbetaks{k}}, k \in S$. The goal is to recover the well-aligned initializers $\{r_k^*\hbeta^{(k)[0]}\}_{k \in S}$ from the initial estimates $\{\hbeta^{(k)[0]}\}_{k=1}^K$ (equivalently, to recover $\{r^*_k\}_{k\in S}$), which can then be fed into Algorithm \ref{algo: multitask}. Once $\{\hbeta^{(k)[0]}\}_{k\in S}$ are well aligned, other initial estimates in Algorithm \ref{algo: multitask} will be automatically well aligned.

In the following, we will introduce two alignment algorithms. The first one is the ``exhaustive search" method (Algorithm \ref{algo: exhaustive alignment}), where we search among all possible alignments to find the best one. The second one is the ``greedy search" method (Algorithm 3 in Section S.1 of the supplements), where we flip the sign of $\hbeta^{(k)[0]}$ in a greedy way to recover $\{r_k^*\hbeta^{(k)[0]}\}_{k \in S}$. Both methods are proved to recover $\{r_k^*\}_{k \in S}$ under mild conditions. The conditions required by the ``exhaustive search" method are slightly weaker than those required by the ``greedy search" method. As for computational complexity, the latter enjoys a linear time complexity $\mathcal{O}(K)$, while the former suffers from an exponential time complexity $\mathcal{O}(2^{K})$ due to optimization over all possible $2^K$ alignments. Due to space constraints, we will introduce only the ``exhaustive search" method, while the details of the ``greedy search" algorithm are provided in Section S.1 of the supplementary materials.

To this end, for a given alignment $\bm{r} = \{r_k\}_{k=1}^K \in \{\pm 1\}^K$ with the correspondingly aligned estimates $\{r_k\hbeta^{(k)[0]}\}_{k=1}^K$, define its alignment score as
\begin{equation}
	\text{score}(\bm{r}) = \sum_{1 \leq k_1 \neq k_2 \leq K}\twonorm{r_{k_1}\hbeta^{(k_1)[0]} - r_{k_2}\hbeta^{(k_2)[0]}}.
\end{equation}
The intuition is that as long as the initializations $\{\hbeta^{(k)[0]}\}_{k \in S}$ are close to the ground truth, a smaller score indicates less difference among $\{r_k\hbeta^{(k)[0]}\}_{k\in S}$, which implies a better alignment. The score can be thus used to evaluate the quality of an alignment. Note that the score is defined in a symmetric way, that is, $\text{score}(\bm{r}) = \text{score}(-\bm{r})$. The exhaustive search algorithm is presented in Algorithm \ref{algo: exhaustive alignment}, where scores of all alignments are calculated, and the alignment that minimizes the score is output. Since the score is symmetric, there are at least two alignments with the minimum score. The algorithm can arbitrarily choose and output one of them. 

\begin{algorithm}[!h]
\linespread{1.4}\selectfont
\caption{Exhaustive search for the alignment}
\label{algo: exhaustive alignment}
\KwIn{Initialization $\{\hbeta^{(k)[0]}\}_{k=1}^K$}
$\widehat{\bm{r}} \leftarrow \argmin_{\bm{r} \in \{\pm 1\}^K}\text{score}(\bm{r})$\\
\KwOut{$\widehat{\bm{r}}$}
\end{algorithm}

The following theorem reveals that the exhaustive search algorithm can successfully find the ideal alignment under mild conditions.
\begin{theorem}[Alignment correctness for Algorithm \ref{algo: exhaustive alignment}]\label{thm: brute force alignment}
	Assume that
	\begin{enumerate}[(i)]
		\item $\epsilon < \frac{1}{3}$;
		\item $\min_{k \in S}\twonorm{\bbetaks{k}} \geq \frac{4(1-\epsilon)}{1-3\epsilon}h + \frac{2(2-\epsilon)}{1-3\epsilon}\max_{k \in S}\big(\twonorm{\hbeta^{(k)[0]} - \bbetaks{k}} \wedge \twonorm{\hbeta^{(k)[0]} + \bbetaks{k}}\big)$,
	\end{enumerate}
	where $\epsilon = \frac{K-|S|}{K}$ is the outlier task proportion introduced in Theorem \ref{thm: upper bound multitask est error}, and $h$ is the similarity level of discriminant coefficient in \eqref{eq: parameter space mtl}. Then the output of Algorithm \ref{algo: exhaustive alignment} satisfies
	\begin{equation}
		\widehat{r}_k = r_k^* \text{ for all } k \in S
		\quad \text{ or }\quad 
		\widehat{r}_k = - r_k^* \text{ for all } k \in S
	\end{equation}
\end{theorem}

\begin{remark}\label{rmk: exhaustive alignment}
	The conditions imposed in Theorem \ref{thm: brute force alignment} are \textbf{no stronger} than conditions required by Theorem \ref{thm: upper bound multitask est error}. First of all, Condition (\rom{1}) is also required in Theorem \ref{thm: upper bound multitask est error}. Moreover, from the definition of $h$ in \eqref{eq: parameter space mtl}, it is bounded by a constant. This, together with Conditions (\rom{3}) and (\rom{4}) in Assumption \ref{asmp: upper bound multitask est error} implies Condition (\rom{2}) in Theorem \ref{thm: brute force alignment}.
\end{remark}

\begin{remark}
	With Theorem \ref{thm: brute force alignment}, we can relax the original Condition (\rom{3}) in Assumption \ref{asmp: upper bound multitask est error} to
	the following condition: 
	
	For all $k \in S$, either of the following two conditions holds with a sufficiently small constant $C_3$:
		\begin{enumerate}[(a)]
			\item $\twonorm{\hbeta^{(k)[0]} - \bbetaks{k}} \vee \twonorm{\hmu^{(k)[0]}_1 - \bmuks{k}_1}\vee \twonorm{\hmu^{(k)[0]}_2 - \bmuks{k}_2} \leq C_3\min_{k \in S}\Delta^{(k)}$, $\norm{\hw^{(k)[0]}-\wks{k}} \leq c_w/2$;
			\item $\twonorm{\hbeta^{(k)[0]} + \bbetaks{k}} \vee \twonorm{\hmu^{(k)[0]}_1 - \bmuks{k}_2}\vee \twonorm{\hmu^{(k)[0]}_2 - \bmuks{k}_1} \leq C_3\min_{k \in S}\Delta^{(k)}$, $\norm{1-\hw^{(k)[0]}-\wks{k}} \leq c_w/2$.
		\end{enumerate}
   In the relaxed version, the initialization for each task only needs to be good up to an \textbf{arbitrary} permutation, while in the original version, the initialization for each task needs to be good under \textbf{the same permutation}. 
\end{remark}

In contrast with supervised MTL, the alignment issue commonly exists in unsupervised MTL. It generally occurs when aggregating information (up to latent label permutation) across different tasks. Alignment pre-processing is thus necessary and important. However, to our knowledge, there is no formal discussion regarding alignment in the existing literature of unsupervised MTL \citep{gu2011learning, zhang2011multitask, yang2014multitask, zhang2018multi, dieuleveut2021federated, marfoq2021federated}. Our treatment of alignment is an important step forward in this field. Our algorithms can potentially be extended to other unsupervised MTL scenarios, and we will leave it for future studies.

Lastly, we emphasize that our alignment algorithms, including Algorithm \ref{algo: exhaustive alignment} and the greedy search algorithm described in Section S.1, do not require any specific initialization procedures. For instance, as Condition (\rom{2}) in Theorem \ref{thm: brute force alignment} shows, Algorithm~\ref{algo: exhaustive alignment} performs well given a reasonably good initialization, and better initializations allow for weaker assumptions on the SNR. As discussed at the beginning of Section \ref{subsec: alignment mtl}, several estimators in the literature satisfy this requirement. In practice, we prefer simple choices such as those obtained from $k$-means or the standard single-task EM algorithm.

\vspace{-0.66cm}

\section{Numerical Experiments}\label{sec: numerical}

In Section \ref{subsec: simulation binary}, we present a simulation study for binary GMMs, followed by a real-data analysis on the Human Activity Recognition dataset for a binary clustering problem in Section \ref{subsec: har main binary}. Due to space limitation, additional simulation studies in various settings, experiments on tuning parameter selection, a multi-cluster analysis of the Human Activity Recognition dataset, and a real-data study on handwritten digits clustering are provided in Section S.5 of the supplementary materials.

As mentioned earlier, we also explored the transfer learning versions of our algorithms in the supplementary material. Given their close similarity in framework to the multi-task versions, we expect them to exhibit comparable performance and therefore do not include them here for comparison.

\subsection{Simulations}\label{subsec: simulation binary}
In this section, we present a simulation study of our multi-task learning procedure MTL-GMM, i.e.,  Algorithm \ref{algo: multitask}. The tuning parameter $\kappa \in (0,1)$ is set as $1/3$, and the value of $C_{\lambda}$ is determined by a 10-fold cross-validation based on the log-likelihood of the final fitted model. The candidates of $C_{\lambda}$ are chosen in a data-driven way, which is described in detail in Section S.5.1.7 of the supplements. All the experiments in this section are implemented in R. Function \texttt{Mcluster} in R package \texttt{mclust} is called to fit a single GMM. We also conducted two additional simulation studies. Due to space constraints, we included these in Section S.5 of the supplementary materials.

We consider a binary GMM setting. There are $K = 10$ tasks, each of which has a sample size of $n_k=100$ and a dimension of $p=15$. When $k \in S$, we generate each $\wks{k}$ from $\text{Unif}(0.1, 0.9)$ and $\bmuks{k}_1$ from $(2, 2, \bm{0}_{p-2})^\top + h/2\cdot (\bSigmaks{k})^{-1}\bm{u}$, where $\bu \sim \text{Unif}(\{\bu \in \mathbb{R}^p: \twonorm{\bm{u}}=1\})$, $\bSigmaks{k} = (0.2^{|i-j|})_{p\times p}$, and let $\bmuks{k}_2 = -\bmuks{k}_1$. When $k \notin S$, the distributions still follow GMM, but we generate each $\wks{k}$ from $\text{Unif}(0.2,0.4)$ and $\bmuks{k}_1$ from $\text{Unif}(\{\bu \in \mathbb{R}^p: \twonorm{\bm{u}}=5\})$, and let $\bmuks{k}_2 = -\bmuks{k}_1$, $\bSigmaks{k} = (0.5^{|i-j|})_{p\times p}$. In this setup, it is clear that $h$ quantifies the similarity among tasks in $S$, and tasks in $S^c$ have very distinct distributions and can be viewed as outlier tasks. For a given $\epsilon\in [0,1)$, the outlier task index set $S^c$ in each replication is uniformly sampled from all subsets of $1:K$ with cardinality $K\epsilon$. We consider two cases:
\begin{enumerate}[(i)]
	\item No outlier tasks ($\epsilon = 0$), and $h$ changes from 0 to 10 with increment 1;
	\item 2 outlier tasks ($\epsilon = 0.2$), and $h$ changes from 0 to 10 with increment 1;
\end{enumerate}

\begin{figure}[!t]
	\centering
	\includegraphics[width=\linewidth]{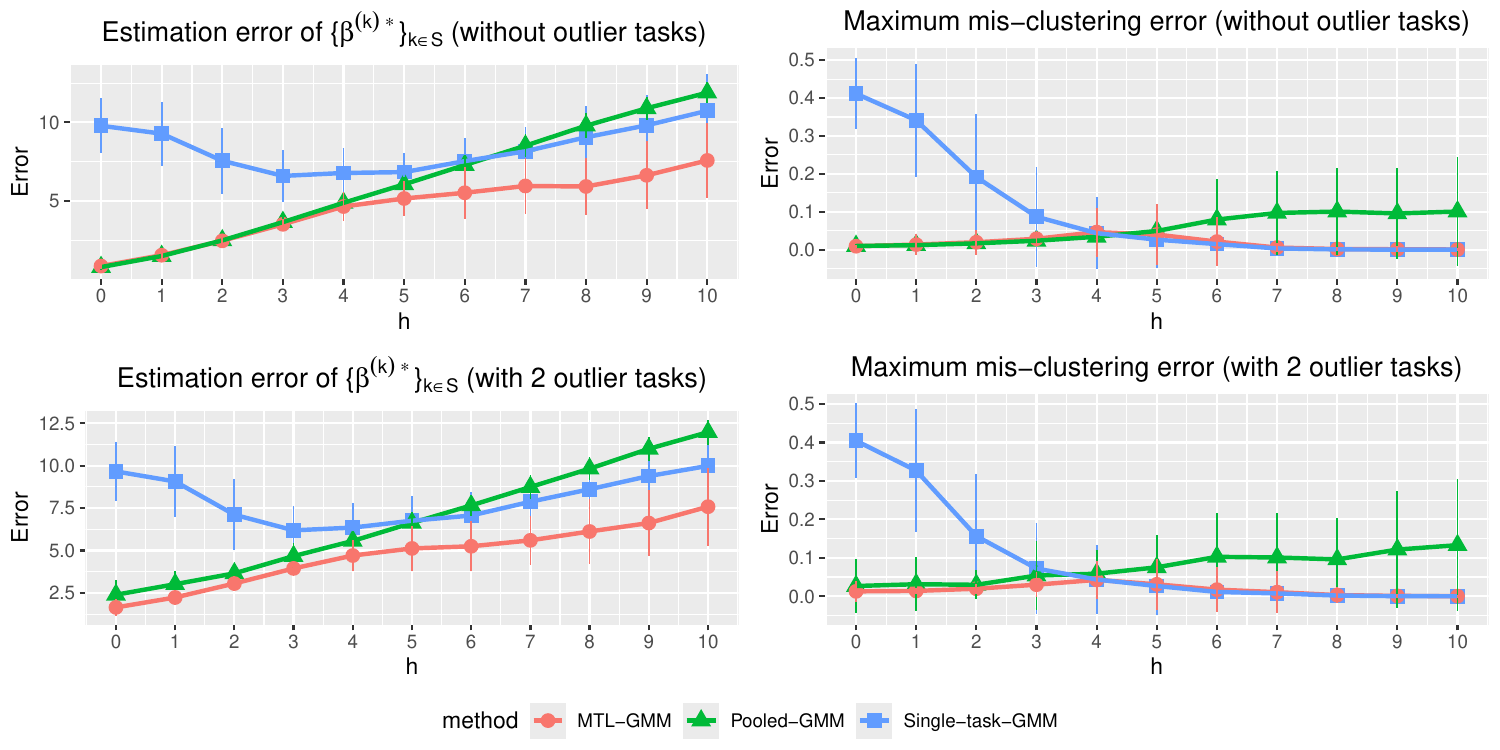}
	\caption{The performance of different methods in Simulation 1 under different outlier proportions. The upper panel shows the performance without outlier tasks ($\epsilon = 0$), and the lower panel shows the performance with two outlier tasks ($\epsilon = 0.2$). $h$ changes from 0 to 10 with increment 1. Estimation error of $\{\bbetaks{k}\}_{k \in S}$ stands for $\max_{k \in S}(\twonorm{\hbeta^{(k)[T]} - \bbetaks{k}} \wedge \twonorm{\hbeta^{(k)[T]} + \bbetaks{k}})$ and maximum mis-clustering error represents the maximum empirical mis-clustering error rate calculated on the test set of tasks in $S$. }
	\label{fig: simulation 1}
\end{figure}

We fit Single-task-GMM on each separate task, Pooled-GMM on the merged data of all tasks, and our MTL-GMM in Algorithm \ref{algo: multitask} coupled with the exhaustive search for the alignment in Algorithm \ref{algo: exhaustive alignment}. The performances of all three methods are evaluated by the estimation error of $\wks{k}$, $\bmuks{k}_1$, $\bmuks{k}_2$, $\bbetaks{k}$, $\deltaks{k}$, and $\bSigmaks{k}$, as well as the empirical mis-clustering error calculated on a test data set of size 500, for tasks in $S$. Due to page limit, we only present the estimation error of $\bbetaks{k}$ and the mis-clustering error here, and leave the others to Section S.5.1.1 of the supplements. These two errors are the maximum errors over tasks in $S$. For each setting, the simulation is replicated 200 times, and the average of the maximum errors and the standard deviation are reported in Figure \ref{fig: simulation 1}. 

When there are no outlier tasks, it can be seen that MTL-GMM and Pooled-GMM are competitive when $h$ is small (i.e., the tasks are similar), and they outperform Single-task-GMM. As $h$ increases (i.e., tasks become more heterogenous), MTL-GMM starts to outperform Pooled-GMM by a large margin. Moreover, MTL-GMM is significantly better than Single-task-GMM in terms of both estimation and mis-clustering errors over a wide range of $h$. These comparisons demonstrate that MTL-GMM not only effectively utilizes the unknown similarity structure among tasks, but also adapts to it. When the outlier tasks exist, even when $h$ is very small, MTL-GMM still performs better than Pooled-GMM, showing the robustness of MTL-GMM against a fraction of outlier tasks.

\subsection{A Real-data Study}\label{subsec: har main binary}
Human Activity Recognition (HAR) Using Smartphones Data Set contains the data collected from 30 volunteers when they performed six activities (walking, walking upstairs, walking downstairs, sitting, standing, and laying) wearing a smartphone \citepapp{anguita2013public}. Each observation has 561 time and frequency domain variables. Each volunteer can be viewed as a task, and the sample size of each task varies from 281 to 409. The original data set is available at UCI Machine Learning Repository: \url{https://archive.ics.uci.edu/ml/datasets/human+activity+recognition+using+smartphones}.

Here, we first focus on two activities, standing and laying, and perform clustering without the label information, to test our method in the binary case. This is a binary MTL clustering problem with 30 tasks. The sample size of each task varies from 95 to 179. For each task, in each replication, we use 90\% of the samples as training data and hold 10\% of the samples as test data.

\begin{table}[!ht]
\centering
\renewcommand{\arraystretch}{0.65}
\begin{tabular}{@{} cccc @{}} 
 \toprule 
Method   & Single-task  & Pooled & MTL \\
 \midrule 
Maximum error  & 0.49 (0.02)     & 0.38 (0.12)     & 0.36 (0.09) \\
Average error & 0.28 (0.02)     & 0.15 (0.17)     & 0.03 (0.01) \\
 \bottomrule 
\end{tabular}
\caption{Maximum and average mis-clustering errors and standard deviations (numbers in the parentheses) in the HAR data set.}
\label{table: har binary}
\end{table}

\begin{figure}[!h]
	\centering
	\includegraphics[width=\linewidth]{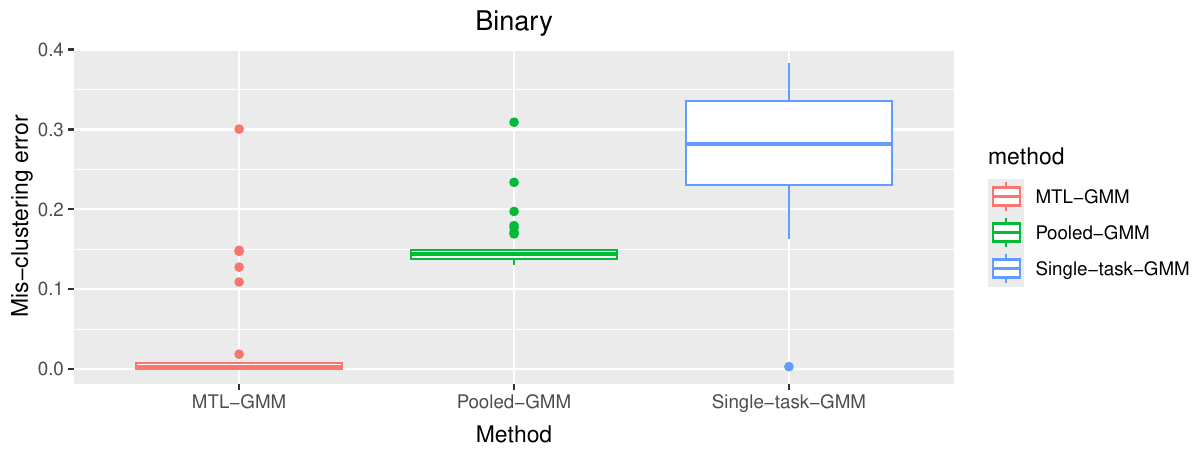}
	\caption{Box plots of mis-clustering errors of 30 tasks for each method for HAR data set.}
	\label{fig: har box plot binary}
\end{figure}

We first run a principal component analysis (PCA) on the training data of each task and project both the training and test data onto the first 15 principal components. PCA has often been used for dimension reduction in pre-processing the HAR data set \citepapp{walse2016pca, aljarrah2019human, duan2023adaptive}. We fit Single-task-GMM on each task separately, Pooled-GMM on merged data from 30 tasks, and our MTL-GMM with the greedy label swapping alignment algorithm. The performance of the three methods is evaluated by the mis-clustering error rate on the test data of all 30 tasks. The maximum and average mis-clustering errors among the 30 tasks are calculated in each replication. The mean and standard deviation of these two errors over 200 replications are reported in Table \ref{table: har binary}. To better display the clustering performance on each task, we further generate the box plot of mis-clustering errors of 30 tasks (averaged over 200 replications) for each method in Figure \ref{fig: har box plot binary}. It is clear that MTL-GMM outperforms both Pooled-GMM and Single-task-GMM. Note that MTL-GMM requires only the discriminant coefficients to be similar across tasks in order to improve clustering performance over Single-task-GMM, whereas Pooled-GMM relies on stronger assumptions -- namely, similarity in both the mean vectors and the covariance matrices. In the current context, individual differences in height, weight, and body shape can affect the observed data patterns, even when the same gestures are performed. This practical heterogeneity makes the assumptions of Pooled-GMM less realistic and, consequently, favors the use of MTL-GMM in this real-data study.

Due to space constraints, additional results from this real-data study focusing on the multi-component GMM extension and an additional application to handwritten digit recognition are provided in Section S.5.2 of the supplementary materials. We refer interested readers to that section for further details.

\section{Discussions}\label{sec: discussions}


We briefly comment on several extensions that are developed in the supplementary materials. First, Section S.2 generalizes the algorithms and theory from binary to multi-component GMMs with $R \geq 3$ clusters. Section S.3.1 considers settings where tasks have different numbers of clusters; we discuss both a direct extension that aggregates only over clusters shared across tasks and a representation-learning approach in which task-specific component means lie in a common low-dimensional subspace. Section S.3.2 discusses how to select the number of clusters when it is unknown and the impact of mis-specification. Section S.3.3 discusses the possibility of relaxing the common-covariance assumption within each task, leading to quadratic decision boundaries and suggesting that cross-task similarity should be imposed jointly on the linear and quadratic terms of the classifier. Finally, Section S.4 develops the transfer learning counterpart of our method, including the algorithms, theoretical guarantees, and label-alignment procedures, where information from multiple source tasks is robustly aggregated to improve performance on a target task. We refer interested readers to these sections for further details.

\section*{Acknowledgment}
We thank the co-Editor Dylan Small, the AE, and the anonymous referees for their insightful comments, which greatly improved the scope and quality of the paper.  \if1\blind
{Haolei Weng was partially supported by NSF grant DMS-2210505. Lucy Xia was partially supported by RGC GRF grant 16308625. Yang Feng was partially supported by NSF Grant DMS-2324489 and NIH Grant 1R21AG074205-01.  All experiments were conducted on Ginsburg HPC Cluster of Columbia University. The authors report there are no competing interests to declare.}\fi

\spacingset{0.8} 
{\small
\setlength{\bibsep}{8pt}
\bibliography{reference.bib}}
\bibliographystyle{apalike}
\spacingset{1.2}

\newpage
\setcounter{page}{1}
\setcounter{section}{0}
\renewcommand{\thesection}{S.\arabic{section}}
\renewcommand{\theequation}{\thesection.\arabic{equation}}
\renewcommand{\thefigure}{S.\arabic{figure}}
\renewcommand{\thetable}{S.\arabic{table}}
\begin{center}
{\large\bf Supplementary Materials of ``Robust Unsupervised Multi-task and Transfer Learning on Gaussian Mixture Models"}
\end{center}
\renewcommand{\cftsecnumwidth}{30pt}
\renewcommand{\cftsubsecnumwidth}{35pt}
\renewcommand{\cftsubsubsecnumwidth}{50pt}
\tableofcontents
\addtocontents{toc}{\protect\setcounter{tocdepth}{5}}

\section{A Greedy Alignment Algorithm for Binary GMMs}\label{sec: alignment binary appendix}
We continue the discussion of alignment algorithms in Section \ref{subsec: alignment mtl} by introducing the second alignment algorithm, the ``greedy search" method, summarized in Algorithm \ref{algo: greedy alignment}. The main idea is to flip the sign of the discriminant coefficient estimates (equivalently, swap the two cluster labels) from $K$ tasks in a sequential fashion to check whether the alignment score decreases or not. If yes, we keep the alignment after the flip and proceed with the next task. Otherwise, we keep the alignment before the flip and proceed with the next task. A surprising fact of Algorithm \ref{algo: greedy alignment} is that it is sufficient to iterate this procedure for all tasks just \textbf{once} to recover the ideal alignment, making the algorithm computationally efficient. 

\begin{algorithm}[!h]
\caption{Greedy search for the alignment}
\label{algo: greedy alignment}
\KwIn{Initialization $\{\hbeta^{(k)[0]}\}_{k=1}^K$}
$\widehat{\bm{r}} = (1, \ldots, 1) \in \{\pm 1\}^K$\\
	\For{$k=1$ \KwTo $K$}{
		$\widetilde{\bm{r}} \leftarrow \text{flip the sign of } \widehat{r}_k$ in $\widehat{\bm{r}}$\\
		\If{$\textup{score}(\widehat{\bm{r}}) > \textup{score}(\widetilde{\bm{r}})$}{
			$\widehat{\bm{r}} \leftarrow \widetilde{\bm{r}}$\\
		}
	}
\KwOut{$\widehat{\bm{r}}$}
\end{algorithm}

To help state the theory of the greedy search algorithm, we define the ``mismatch proportion" of $\{\hbeta^{(k)[0]}\}_{k \in S}$ as
\begin{equation}
	p_a = \min\{\#\{k\in S: r_k^* = 1\}, \#\{k\in S: r_k^* = -1\}\}/|S| 
\end{equation}
Intuitively, $p_a$ represents the level of mismatch between the initial alignment and the ideal
one. It's straightforward to verify that $p_a \in [0, 1/2]$; $p_a = 0$ means the initial alignment equals the ideal one, while $p_a = 1/2$ (or $\frac{|S|-1}{2|S|}$ when $|S|$ is odd) represents the ``worst" alignment, where almost half of the tasks are badly-aligned. The smaller $p_a$ is, the better alignment $\{\hbeta^{(k)[0]}\}_{k \in S}$ is. Note that we only care about the alignment of tasks in $S$.

The following theorem shows that the greedy search algorithm can succeed in finding the ideal alignment under mild conditions.

\begin{theorem}[Alignment correctness for Algorithm \ref{algo: greedy alignment}]\label{thm: greedy alignment}
	Assume that
	\begin{enumerate}[(i)]
		\item $\epsilon < \frac{1}{2}$;
		\item $\min_{k \in S}\twonorm{\bbetaks{k}} \geq \frac{2(1-\epsilon)}{2(1-\epsilon)(1-p_a)-1}h + \frac{2-\epsilon}{2(1-\epsilon)(1-p_a)-1}\max_{k \in S}\big(\twonorm{\hbeta^{(k)[0]} - \bbetaks{k}} \wedge \twonorm{\hbeta^{(k)[0]} + \bbetaks{k}}\big)$;
		\item $p_a < \frac{1-2\epsilon}{2(1-\epsilon)}$,
	\end{enumerate}
	where $\epsilon$ and $h$ are the same as in Theorem \ref{thm: brute force alignment}. Then the output of Algorithm \ref{algo: greedy alignment} satisfies
	\begin{equation}
		r_k = r_k^* \text{ for all } k \in S
		\quad \text{ or }\quad 
		r_k = -r_k^* \text{ for all } k \in S
	\end{equation}
\end{theorem}

\begin{remark}
	Conditions (\rom{1}) and (\rom{2}) required by Theorem \ref{thm: greedy alignment} are similar to the requirements in Theorem \ref{thm: brute force alignment}, which have been shown to be \textbf{no stronger} than conditions in Assumption \ref{asmp: upper bound multitask est error} and Theorem \ref{thm: upper bound multitask est error} (See Remark \ref{rmk: exhaustive alignment}). However, Condition (\rom{3}) is an additional requirement for the success of the greedy label-swapping algorithm. The intuition is that in the exhaustive search algorithm, we compare the scores of all alignments and only need to ensure the ideal alignment can defeat the badly-aligned ones in terms of the alignment score. In contrast, the success of the greedy search algorithm relies on the correct move at each step. We need to guarantee that the ``better" alignment after the swap (which may still be badly aligned) can outperform the ``worse" one before the swap. This is more difficult to satisfy. Hence, more conditions are needed for the success of Algorithm \ref{algo: greedy alignment}. Condition (\rom{3}) is one such condition to provide a reasonably good initial alignment to start the greedy search process. More details of the analysis can be found in the proofs of Theorems \ref{thm: brute force alignment} and \ref{thm: greedy alignment} in the supplements.
\end{remark}

\begin{remark}\label{rmk: greedy search binary}
	In practice, Condition (\rom{3}) can fail to hold with a non-zero probability. One solution is to start with random alignments, run the greedy search algorithm multiple times, and use the alignment that appears most frequently in the output. Nevertheless, this will increase the computational burden. In our numerical studies, Algorithm \ref{algo: greedy alignment} without multiple random alignments works well.
\end{remark}

One appealing feature of the two alignment algorithms is that they are robust against a fraction of outlier tasks from arbitrary distributions. According to the definition of the alignment score, this may appear impossible at first glance because the score depends on the estimators from all tasks. However, it turns out that the impact of outliers when comparing the scores in Algorithm \ref{algo: exhaustive alignment} and \ref{algo: greedy alignment} can be bounded by parameters and constants that are unrelated to outlier tasks via the triangle inequality of Euclidean norms. The key idea is that the alignment of outlier tasks in $S^c$ does not matter in Theorems \ref{thm: brute force alignment} and \ref{thm: greedy alignment}. More details can be found in the proof of Theorems \ref{thm: brute force alignment} and \ref{thm: greedy alignment} in the supplementary materials.

\section{Extension to Multi-cluster GMMs}\label{sec: mtl multi}
In the main text, we have discussed the MTL problem for binary GMMs. In this section, we extend our methods and theory to Gaussian mixtures with $R$ clusters ($R \geq 3$).

 We first generalize the problem setting introduced in Sections \ref{subsec: intro gmm} and \ref{subsec: problem mtl}. There are $K$ tasks where we have $n_k$ observations $\{\bz_i^{(k)}\}_{i=1}^{n_k}$ from the $k$-th task. An \emph{unknown} subset $S\subseteq 1: K$ denotes tasks whose samples follow multi-cluster GMMs, and $S^c$ refers to outlier tasks that can have arbitrary distributions. Specifically, for all $k \in S, i = 1:n_k$,
\begin{align}
	\yk{k}_i = r  \text{ \,\,with probability }\wks{k}_r, \quad \quad
	\bz_i^{(k)}|y_i^{(k)}=r \sim \mathcal{N}(\bmuks{k}_r, \bSigmaks{k}), \quad \quad r = 1:R,
\end{align}
with $\sum_{r=1}^R \wks{k}_r = 1$, and
\begin{equation}
	\{\bzk{k}_i\}_{i,k\in S^c} \sim \mathbb{Q}_{S},
\end{equation}
where $\mathbb{Q}_S$ is some probability measure on $(\mathbb{R}^p)^{\otimes \nsc}$ and $\nsc = \sum_{k \in S^c}n_k$. We focus on the following joint parameter space
\begin{equation}
	\overline{\Theta}_{S}(h) = \Big\{\{\overline{\btheta}^{(k)}\}_{k \in S} = \{(\{\wk{k}_r\}_{r=1}^R, \{\bmuk{k}_r\}_{r=1}^R, \bSigmak{k})\}_{k  \in S}: \othetak{k} \in \overline{\Theta}, \max_{r=2:R}\inf_{\bbeta}\max_{k \in S}\twonorm{\bbetak{k}_r-\bbeta} \leq h\Big\}, \label{eq: parameter space mtl multi-component}
\end{equation}
where $\bbetak{k}_r = (\bSigmak{k})^{-1}(\bmuk{k}_r - \bmuk{k}_1)$ is the $r$-th discriminant coefficient in the $k$-th task, and $\overline{\Theta}$ is the parameter space for a single multi-cluster GMM,
\begin{align}
	\overline{\Theta} = \bigg\{\overline{\btheta} = (\{w_r\}_{r=1}^R, \{\bmu_r\}_{r=1}^R, \bSigma): & \max_{r=1:R}\twonorm{\bmu_r} \leq M, c_w \leq \min_{r=1:R}w_r \leq \max_{r=1:R}w_r \leq  1-c_w, \\
	&\sum_{r=1}^R w_r = 1, c_{\bSigma}^{-1} \leq \lambda_{\min}(\bSigma) \leq \lambda_{\max}(\bSigma) \leq c_{\bSigma}\bigg\}. \label{multic:single}
\end{align}
Note that \eqref{eq: parameter space mtl multi-component} and \eqref{multic:single} are natural generalizations of \eqref{eq: parameter space mtl} and \eqref{eq: parameter space Theta}, respectively.

Under a multi-cluster GMM with parameter $\otheta =(\{w_r\}_{r=1}^R, \{\bmu_r\}_{r=1}^R, \bSigma)$, compared with \eqref{eq: bayes lda intro section}, the optimal discriminant rule now becomes 
\begin{align}
\label{optimal:lda:multi}
\mC_{\otheta}(\bz) = \argmax_{r=1:R} \bigg\{\bigg(\bz - \frac{\bmu_1 + \bmu_r}{2}\bigg)^\top \bbeta_r + \log\bigg(\frac{w_r}{w_1}\bigg)\bigg\},
\end{align}
where $\bbeta_r = \bSigma^{-1}(\bmu_r - \bmu_1)$. Once we have the parameter estimators, we plug them into the above rule to obtain the plug-in clustering method. Recall that for a clustering method $\mC: \mathbb{R}^p \rightarrow [R]$, its mis-clustering error is
\begin{equation}\label{eq: clustering method ck error multi-cluster}
	R_{\otheta}(\mC) = \min_{\pi: [R] \rightarrow [R]}\tp_{\otheta}(\mC(Z^{\textup{new}}) \neq \pi(Y^{\textup{new}})).
\end{equation}
Here, $Z^{\textup{new}} \sim \sum_{r=1}^R w_r\mathcal{N}(\bmu_r, \bSigma)$ is an independent future observation associated with the label $Y^{\textup{new}}$, the probability $\tp_{\otheta}$ is w.r.t. $(Z^{\textup{new}}, Y^{\textup{new}})$, and the minimum is taken over $R!$ permutation functions on $[R]$. Since \eqref{optimal:lda:multi} is the optimal clustering method that minimizes $R_{\otheta}(\mC)$, the excess mis-clustering error for a given clustering $\mC$ is $R_{\otheta}(\mC)-R_{\otheta}(\mC_{\otheta})$. The rest of the section aims to extend the EM-stylized multi-task learning procedure and the two alignment algorithms in Section \ref{sec: mtl} to the general multi-cluster GMM setting, and provide similar statistical guarantees in terms of estimation and excess mis-clustering errors. For simplicity, throughout this section, we assume the number of clusters $R$ to be bounded and known. We leave the case of diverging $R$ as a future work.

Since both the EM algorithm and the penalization framework work beyond binary GMM, our methodological idea described in Section \ref{subsec: method mtl} can be directly adapted to extend Algorithm \ref{algo: multitask} to the multi-cluster case. We summarize the general procedure in Algorithm \ref{algo: multitask multi-component}. Like in Algorithm \ref{algo: multitask}, we have adopted the following notation for posterior probability in Algorithm \ref{algo: multitask multi-component},
\begin{equation}
	\gammak{r}_{\btheta}(\bz) = \frac{w_r\exp(\bbeta_r^\top\bz - \delta_r)}{\sum_{r=1}^R w_r\exp(\bbeta_r^\top\bz - \delta_r)}, \quad {\rm for~}\btheta=(\{w_r\}_{r=2}^R, \{\bbeta_r\}_{r=2}^R, \{\delta_r\}_{r=2}^R),
\end{equation}
where $w_1 = 1-\sum_{r=2}^R w_r$, $\bbeta_1 \coloneqq \bm{0}$, and $\delta_1 \coloneqq 0$. Specifically, $\gammak{r}_{\btheta}(\bz)$ is the posterior probability $ \tp(Y=r |Z=\bm{z}) $ given the observation $\bm{z}$, when the true parameter of a multi-cluster GMM  $(\{w^*_r\}_{r=1}^R, \{\bmu^*_r\}_{r=1}^R, \bSigma^*)$ satisfies $w_r=w_r^*, \bbeta_r=(\bSigma^*)^{-1}(\bmu^*_r-\bmu^*_1), \delta_r=\frac{1}{2}\bbeta_r^\top(\bmu^*_1+\bmu^*_r)$, for $r=1:R$. 

\begin{algorithm}[!]
\caption{MTL-GMM (Multi-cluster)}
\label{algo: multitask multi-component}
\KwIn{Initialization $\{(\{\hw^{(k)[0]}_r\}_{r=1}^R, \{\hbeta^{(k)[0]}_r\}_{r=2}^R, \{\hmu^{(k)[0]}_r\}_{r=1}^R)\}_{k=1}^K$, maximum number of iteration rounds $T$, initial penalty parameter $\lambda^{[0]}$, tuning parameters $C_{\lambda} > 0$, $\kappa \in (0, 1)$}
$\htheta^{(k)[0]} = (\{\hw^{(k)[0]}_r\}_{r=1}^R, \{\hbeta^{(k)[0]}_r\}_{r=2}^R, \{\hdelta^{(k)[0]}_r\}_{r=2}^R)$, where $\hdelta^{(k)[0]}_r = \frac{1}{2}(\hbeta^{(k)[0]}_r)^\top(\hmu^{(k)[0]}_1 + \hmu^{(k)[0]}_r)$, for $k = 1:K$\\
\For{$t = 1$ \KwTo $T$}{
	$\lambda^{[t]} = \kappa \lambda^{[t-1]} + C_{\lambda}\sqrt{p+\log K}$ \tcp*[f]{Update the penalty parameter}\\
	\For(\tcp*[f]{Local update for each task}){$k = 1$ \KwTo $K$}{ 
	\For{$r = 1$ \KwTo $R$}{
		$\hw^{(k)[t]}_r = \frac{1}{n_k}\sum_{i=1}^{n_k}\gammak{r}_{\htheta^{(k)[t-1]}}(\bz_i^{(k)})$\\ 
		$\hmu^{(k)[t]}_r = \frac{\sum_{i=1}^{n_k}\gammak{r}_{\htheta^{(k)[t-1]}}(\bz_i^{(k)})\bzk{k}_i}{n_k\hw^{(k)[t]}_r}$ \\
	}
	$\hSigma^{(k)[t]} = \frac{1}{n_k}\sum_{i=1}^{n_k}\sum_{r=1}^R\gammak{r}_{\htheta^{(k)[t-1]}}(\bz_i^{(k)}) \cdot (\bzk{k}_i - \hmu^{(k)[t]}_r)(\bzk{k}_i - \hmu^{(k)[t]}_r)^\top$
	}
	
	\For{$r = 2$ \KwTo $R$}{
		$\{\hbeta^{(k)[t]}_r\}_{k=1}^K$, $\overline{\bbeta}_r^{[t]} = \argmin\limits_{\bbetak{1}, \ldots, \bbetak{K}, \overline{\bbeta}}\bigg\{\sum_{k=1}^K n_k\Big[\frac{1}{2}(\bbetak{k})^\top\hSigma^{(k)[t]}\bbetak{k} - (\bbetak{k})^\top(\hmu_r^{(k)[t]}-\hmu_1^{(k)[t]})\Big] + \sum_{k=1}^K\sqrt{n_k}\lambda^{[t]}\cdot \twonorm{\bbetak{k}-\overline{\bbeta}}\bigg\}$\tcp*[f]{Aggregation} \\
	}
	
	\For(\tcp*[f]{Local update for each task}){$k = 1$ \KwTo $K$}{
		\For{$r = 2$ \KwTo $R$}{
			$\hdelta^{(k)[t]}_r = \frac{1}{2}(\hbeta^{(k)[t]}_r)^\top(\hmu^{(k)[t]}_1 + \hmu^{(k)[t]}_r)$
		}
		Let $\htheta^{(k)[t]} = (\{\hw^{(k)[t]}_r\}_{r=1}^R, \{\hbeta^{(k)[t]}_r\}_{r=2}^R, \{\hdelta^{(k)[t]}\}_{r=2}^R)$
	}
	
}
\KwOut{$\{(\htheta^{(k)[T]}, \{\hmu^{(k)[T]}_r\}_{r=1}^R, \hSigma^{(k)[T]})\}_{k=1}^K$ with $\htheta^{(k)[T]} = (\{\hw^{(k)[T]}_r\}_{r=1}^R, \{\hbeta^{(k)[T]}_r\}_{r=2}^R, \allowbreak \{\hdelta^{(k)[T]}\}_{r=2}^R)$}
\end{algorithm}

Having the estimates $(\{\hw^{(k)[T]}_r\}_{r=1}^R, \{\hbeta^{(k)[T]}_r\}_{r=2}^R, \allowbreak \{\hmu^{(k)[T]}_r\}_{r=1}^R)$ from Algorithm \ref{algo: multitask multi-component}, we can plug them into \eqref{optimal:lda:multi} to construct the clustering method, denoted by $\widehat{\mathcal{C}}^{(k)[T]}(\bz)$. Equivalently, 
\begin{equation}\label{eq: clustering method ck multi-cluster}
	\widehat{\mathcal{C}}^{(k)[T]}(\bz) =  \argmax_{r=1:R} \gammak{r}_{\htheta^{(k)[T]}}(\bz).
\end{equation}

\subsection{Theory}
We need the following assumption before stating the theory.

\begin{assumption}\label{asmp: upper bound multitask est error multi-cluster}	
Denote $\Delta^{(k)}_{rj} = \sqrt{(\bmuks{k}_r-\bmuks{k}_j)^\top(\bSigmaks{k})^{-1}(\bmuks{k}_r-\bmuks{k}_j)}$ for $k \in S$. Suppose the following conditions hold: 
\begin{enumerate}[(i)]
		\item $\ns= \sum_{k \in S}n_k \geq C_1|S|\max_{k = 1:K}n_k$ with a constant $C_1 > 0$;
		\item $\min_{k \in S}n_k \geq C_2(p+\log K)$ with some constant $C_2$;
		\item There exists a permutation $\pi: [R] \rightarrow [R]$ such that 
			\begin{enumerate}
				\item $\max_{k \in S}\big\{\big[\max_{r=2:R}\twonorm{\hbeta^{(k)[0]}_r - (\bSigmaks{k})^{-1}(\bmuks{k}_{\pi(r)} - \bmuks{k}_{\pi(1)})}\big] \vee \big(\max_{r=1:R}\twonorm{\hmu^{(k)[0]}_r - \bmuks{k}_{\pi(r)}}\big)\big\} \leq  C_3\min_{k \in S}\min_{r\neq j}\Delta^{(k)}_{rj}$, with some constant $C_3$;
				\item $\max_{k \in S}\max_{r=2:R}\norm{\hw^{(k)[0]}_r-\wks{k}_{\pi(r)}}\leq c_w/2$.
			\end{enumerate}
		\item $\min_{k \in S}\min_{r\neq j}\Delta^{(k)}_{rj} \geq C_4 > 0$ with some constant $C_4$;
	\end{enumerate}
\end{assumption}

\begin{remark}
\label{rem:assump2}
The above set of conditions are analogues of those in Assumption \ref{asmp: upper bound multitask est error}. We refer to Remark \ref{rmk: asmp1} for a detailed explanation of each condition.
\end{remark}

We first present the result for parameter estimation. We adopt similar error metrics as the ones in \eqref{errm:1} and \eqref{errm:2}. Specifically, denote the true parameter by $\{\othetaks{k}\}_{k \in S} = \{(\{\wks{k}_r\}_{r=2}^R, \{\bmuks{k}_r\}_{r=1}^R,  \bSigmaks{k})\}_{k  \in S} $ which belongs to the parameter space $\overline{\Theta}_{S}(h)$ in \eqref{eq: parameter space mtl multi-component}. For each $k\in S$, define the functional $\bthetaks{k}= (\{\wks{k}_r\}_{r=2}^R, \{\bbetaks{k}_r\}_{r=2}^R, \{\deltaks{k}_r\}_{r=2}^R)$, where $\bbetaks{k}_r=(\bSigmaks{k})^{-1}(\bmuks{k}_r-\bmuks{k}_1), \deltaks{k}_r=\frac{1}{2}(\bbetaks{k})^\top(\bmuks{k}_1+\bmuks{k}_r)$. For the estimators returned by Algorithm \ref{algo: multitask multi-component}, we are interested in the error metrics \footnote{Similar to the binary case, the minimum is taken due to the non-identifiability in multi-cluster GMMs.}:
\begin{align}
&d(\htheta^{(k)[T]}, \bthetaks{k})= \min_{\pi: [R]\rightarrow [R]}\max_{r=2:R} \Big\{\norm{\hw^{(k)[T]}_r-\wks{k}_{\pi(r)}}\vee \twonorm{\hbeta^{(k)[T]}_r -(\bSigmaks{k})^{-1}(\bmuks{k}_{\pi(r)} - \bmuks{k}_{\pi(1)})} \\
&\hspace{6cm} \vee \norm{\hdelta^{(k)[T]}_r - (\bmuks{k}_{\pi(r)} + \bmuks{k}_{\pi(1)})^\top(\bSigmaks{k})^{-1}(\bmuks{k}_{\pi(r)} - \bmuks{k}_{\pi(1)})/2}\Big\}, \\
&\Big(\min_{\pi: [R] \rightarrow [R]}\max_{r=1:R}\twonorm{\hmu^{(k)[T]}_r - \bmuks{k}_{\pi(r)}}\Big) \vee \twonorm{\hSigma^{(k)[T]} - \bSigmaks{k}}.
\end{align}

\begin{theorem}(Upper bounds of the estimation error of GMM parameters for multi-cluster MTL-GMM)\label{thm: upper bound multitask est error multi-cluster}
	Suppose Assumption \ref{asmp: upper bound multitask est error multi-cluster} holds, $|S| \geq s$, and $\epsilon = \frac{K-s}{K}<1/3$. Let $\lambda^{[0]} \geq C_1\max_{k=1:K}\sqrt{n_k}$, $C_{\lambda} \geq C_1  $ and $\kappa > C_2$ with some constants $C_1 > 0, C_2 \in (0, 1)$ \footnote{$C_1$ and $C_2$ depend on the constants $M$, $c_w$, and $c_{\bSigma}$ etc.} Then there exists a constant $C_3 > 0$, such that for any $\{\othetaks{k}\}_{k \in S} \in \overline{\Theta}_{S}(h)$ and any probability measure $\mathbb{Q}_S$ on $(\mathbb{R}^p)^{\otimes n_{S^c}}$, with probability $1-C_3K^{-1}$, the following hold for all $k \in S$:
	\begin{align}
		d(\htheta^{(k)[T]}, \bthetaks{k}) &\lesssim \sqrt{\frac{p}{\ns}} + \sqrt{\frac{\log K}{n_k}}+ h \wedge \sqrt{\frac{p+\log K}{n_k}} + \epsilon \sqrt{\frac{p + \log K}{\max_{k=1:K}n_k}}+  T^2(\kappa')^{T},
	\end{align}
	\begin{align}
		\left(\min_{\pi: [R] \rightarrow [R]}\max_{r=1:R}\twonorm{\hmu^{(k)[T]}_r - \bmuks{k}_{\pi(r)}}\right) \vee \twonorm{\hSigma^{(k)[T]} - \bSigmaks{k}} \lesssim \sqrt{\frac{p + \log K}{n_k}} +  T^2(\kappa')^{T}, 
	\end{align}	
	where $\kappa' \in (0, 1)$ is some constant and $\ns = \sum_{k \in S}n_k$. When $T \geq C\log(\max_{k=1:K} n_k)$ with some large constant $C > 0$, the last term on the right-hand side will be dominated by other terms in both inequalities.
\end{theorem}

Recall the clustering method $\widehat{\mathcal{C}}^{(k)[T]}$ defined in \eqref{eq: clustering method ck multi-cluster}. The next theorem obtains the upper bound of the excess mis-clustering error of $\widehat{\mathcal{C}}^{(k)[T]}$ for $k \in S$.

\begin{theorem}(Upper bound of the excess mis-clustering error for multi-cluster MTL-GMM)\label{thm: upper bound multitask classification error multi-cluster}
Suppose the same conditions in Theorem \ref{thm: upper bound multitask est error multi-cluster} hold. Then there exists a constant $C_1 > 0$ such that for any $\{\othetaks{k}\}_{k \in S} \in \overline{\Theta}_{S}(h)$ and any probability measure $\mathbb{Q}_S$ on $(\mathbb{R}^p)^{\otimes n_{S^c}}$, with probability at least $1-C_1K^{-1}$, the following holds for all $k \in S$: 
\begin{align}
	&R_{\othetaks{k}}(\widehat{\mathcal{C}}^{(k)[T]}) - R_{\othetaks{k}}(\mC_{\othetaks{k}}) \lesssim d^2(\htheta^{(k)[T]}, \bthetaks{k}) \cdot \log d^{-1}(\htheta^{(k)[T]}, \bthetaks{k})\\
&\lesssim \bigg[\frac{p}{\ns} + \frac{\log K}{n_k} + h^2\wedge \frac{p+\log K}{n_k}+ \epsilon^2\frac{p + \log K}{\max_{k=1:K}n_k} + T^4(\kappa')^{2T}\bigg] \cdot \log\left(\frac{\ns}{p} \wedge \frac{n_k}{\log K}\right),
\end{align}
where $\kappa' \in (0, 1)$ is some constant. When $T \geq C\log (\max_{k=1:K}n_k)$ with a large constant $C > 0$, the term involving $T$ on the right-hand side will be dominated by other terms.
\end{theorem}

Comparing the upper bounds in Theorems \ref{thm: upper bound multitask est error multi-cluster} and \ref{thm: upper bound multitask classification error multi-cluster} with those in Theorems \ref{thm: upper bound multitask est error} and \ref{thm: upper bound multitask classification error}, the only difference is an extra logarithmic term $\log\big(\frac{\ns}{p} \wedge \frac{n_k}{\log K}\big)$ in Theorem \ref{thm: upper bound multitask classification error multi-cluster}, which we believe is a proof artifact. Similar logarithmic terms appear in other multi-cluster GMM literatures as well, see for example, \citeapp{yan2017convergence} and \citeapp{zhao2020statistical}. To understand the upper bounds in Theorems \ref{thm: upper bound multitask est error multi-cluster} and \ref{thm: upper bound multitask classification error multi-cluster}, we can follow the discussion after Theorems \ref{thm: upper bound multitask est error} and \ref{thm: upper bound multitask classification error}. We do not repeat it here.

The following lower bounds together with the derived upper bounds will show that our method is (nearly) minimax optimal in a wide range of regimes.

\begin{theorem}(Lower bounds of the estimation error of GMM parameters in multi-task learning)\label{thm: lower bound multitask est error multi-cluster}
	Suppose $\epsilon = \frac{K-s}{K} < 1/3$. When there exists a subset $S$ with $|S| \geq s$ such that $\min_{k \in S} n_k \geq C_1(p + \log K)$ and $\min_{k \in S}\min_{r,j}\Delta_{rj}^{(k)} \geq C_2$, where $C_1, C_2 > 0$ are some constants, we have
	\begin{align}
		\inf_{\{\htheta^{(k)}\}_{k=1}^K} \sup_{S: |S| \geq s}\sup_{\substack{\{\othetaks{k}\}_{k \in S} \in \overline{\Theta}_S(h) \\ \mathbb{Q}_S}} &\tp\Bigg(\bigcup_{k \in S}\Bigg\{d(\htheta^{(k)}, \bthetaks{k}) \gtrsim \sqrt{\frac{p}{\ns}}+ \sqrt{\frac{\log K}{n_k}} \\ 
		&\quad\quad  + h \wedge \sqrt{\frac{p+\log K}{n_k}} +\frac{\epsilon}{\sqrt{\max_{k=1:K}n_k}} \Bigg\}\Bigg) \geq \frac{1}{10},	
	\end{align}
	\begin{align}
		\inf_{\substack{\{\hmu^{(k)}_r\}_{k=1:K,r=1:R} \\\{\hSigma^{(k)}\}_{k=1}^K}} \sup_{S: |S| \geq s}\sup_{\substack{\{\othetaks{k}\}_{k \in S} \in \overline{\Theta}_S(h) \\ \mathbb{Q}_S}} \tp\Bigg(\bigcup_{k \in S}\Bigg\{&\left(\min_{\pi: [R] \rightarrow [R]}\max_{r=1:R}\twonorm{\hmu^{(k)}_r - \bmuks{k}_{\pi(r)}}\right) \vee \\
		&  \twonorm{\hSigma^{(k)} - \bSigmaks{k}} \gtrsim \sqrt{\frac{p + \log K}{n_k}}\Bigg\}\Bigg) \geq \frac{1}{10}.	
	\end{align}
\end{theorem}

\begin{theorem}(Lower bound of the excess mis-clustering error in multi-task learning)\label{thm: lower bound multitask classification error multi-cluster}
	Suppose the same conditions in Theorem \ref{thm: lower bound multitask est error multi-cluster} hold. Then 
	\begin{align}
		\inf_{\{\widehat{\mathcal{C}}^{(k)}\}_{k=1}^K} \sup_{S: |S| \geq s}\sup_{\substack{\{\othetaks{k}\}_{k \in S} \in \overline{\Theta}_S(h) \\ \mathbb{Q}_S}} &\tp\Bigg(\bigcup_{k \in S}\Bigg\{R_{\othetaks{k}}(\widehat{\mathcal{C}}^{(k)}) - R_{\othetaks{k}}(\mC_{\othetaks{k}}) \gtrsim  \frac{p}{\ns} + \frac{\log K}{n_k}    \\
		&\quad\quad\quad + h^2\wedge \frac{p+\log K}{n_k} + \frac{\epsilon^2}{\max_{k=1:K}n_k}\Bigg\}\Bigg) \geq \frac{1}{10}.
	\end{align}
\end{theorem}

The lower bounds in Theorems \ref{thm: lower bound multitask est error multi-cluster} and \ref{thm: lower bound multitask classification error multi-cluster} are the same as those in Theorems \ref{thm: lower bound multitask est error} and \ref{thm: lower bound multitask classification error}. Therefore, the remarks on the comparison of upper and lower bounds presented after Theorem \ref{thm: lower bound multitask classification error} carry over to the multi-cluster setting (up to the logarithmic term from Theorem \ref{thm: upper bound multitask classification error multi-cluster}). We do not repeat the details here.

\subsection{Alignment}\label{subsec: alignment multi-cluster mtl}
Similar to the binary case, we have the alignment issues in multi-cluster case as well. In this section, we propose two alignment algorithms as the extension to the Algorithms \ref{algo: exhaustive alignment} and \ref{algo: greedy alignment}. 

In the multi-cluster case, the alignment of each task can be represented as a permutation of $[R]$. Consider a series of permutations $\bm{\pi} = \{\pi_k\}_{k=1}^K$, where each $\pi_k$ is a permutation function on $[R]$. Define a score of $\bm{\pi}$ as
\begin{equation}
	\textup{score}(\bm{\pi}) = \sum_{r=2}^R\sum_{k\neq k'}\twonorma{(\hSigma^{(k)[0]})^{-1}\big(\hmu^{(k)[0]}_{\pi_k(r)} - \hmu^{(k)[0]}_{\pi_k(1)}) - (\hSigma^{(k')[0]})^{-1}(\hmu^{(k')[0]}_{\pi_{k'}(r)} - \hmu^{(k')[0]}_{\pi_{k'}(1)}\big)}.
\end{equation}
We want to recover the correct alignment $\pi_k^* = \argmin\limits_{\pi_k: [R] \rightarrow [R]} \sum_{r=2}^R\twonorma{(\hSigma^{(k)[0]})^{-1}\big(\hmu^{(k)[0]}_{\pi_k(r)} - \hmu^{(k)[0]}_{\pi_k(1)}) - \bbetaks{k}_r}$. We propose an exhaustive search algorithm, which is summarized in Algorithm \ref{algo: exhaustive alignment multi component}.

\begin{algorithm}[!h]
\caption{Exhaustive search for the alignment in multi-cluster GMMs}
\label{algo: exhaustive alignment multi component}
\KwIn{Initialization $\{\{\hmu^{(k)[0]}_r\}_{r=1}^R, \hSigma^{(k)[0]}\}_{k=1}^K$}
$\widehat{\bm{\pi}} = \{\widehat{\pi}_k\}_{k=1}^K \leftarrow \argmin_{\bm{\pi}}\text{score}(\bm{\pi})$\\
\KwOut{$\widehat{\bm{\pi}}= \{\widehat{\pi}_k\}_{k=1}^K$}
\end{algorithm}

The following theorem shows that under certain conditions, the output from Algorithm \ref{algo: exhaustive alignment multi component} recovers the correct alignment up to a permutation.

\begin{theorem}[Alignment correctness for Algorithm \ref{algo: exhaustive alignment multi component}]\label{thm: brute force alignment multi component}
	Assume that
	\begin{enumerate}[(i)]
		\item $\max_{k \in S}\max_{r\neq j}\Delta^{(k)}_{rj}/\min_{k \in S}\min_{r\neq j}\Delta^{(k)}_{rj} \leq D$ with $D \geq 1$.
		\item $\epsilon < \frac{1}{24Dc_{\bSigma} + 1}$;
		\item $\min_{k \in S}\min_{r\neq j}\Delta^{(k)}_{rj} \geq \left[\frac{4(1-\epsilon)c_{\bSigma}^{1/2}}{1-(24Dc_{\bSigma} + 1)\epsilon}h + \frac{(4+20\epsilon)c_{\bSigma}^{1/2}}{1-(24Dc_{\bSigma} + 1)\epsilon}\xi\right] \vee \left[\frac{13(1-\epsilon)c_{\bSigma}^{1/2}}{1-(9Dc_{\bSigma} + 1)\epsilon}h + \frac{(13-4\epsilon)c_{\bSigma}^{1/2}}{1-(9Dc_{\bSigma} + 1)\epsilon}\xi\right]$,
	\end{enumerate}
	where $\epsilon = \frac{K-s}{K}$ is the outlier task proportion introduced in Theorem \ref{thm: upper bound multitask est error multi-cluster}, $h$ is degree of discriminant coefficient similarity defined in \eqref{eq: parameter space mtl multi-component}, and
	\begin{align}
		\xi &= \max_{k \in S}\min_{\pi:[R] \rightarrow [R]}\max_{r=1:R}\twonorma{(\hSigma^{(k)[0]})^{-1}\big(\hmu^{(k)[0]}_{\pi(r)} - \hmu^{(k)[0]}_{\pi(1)}\big) - \bbetaks{k}_r} \\
		&= \max_{k \in S}\min_{\pi:[R] \rightarrow [R]}\max_{r=1:R}\twonorma{(\hSigma^{(k)[0]})^{-1}\big(\hmu^{(k)[0]}_{\pi(r)} - \hmu^{(k)[0]}_{\pi(1)}\big) - (\bSigmaks{k})^{-1}\big(\bmuks{k}_r - \bmuks{k}_1\big)}.
	\end{align}
	Then there exists a permutation $\iota: [R] \rightarrow [R]$, such that the output of Algorithm \ref{algo: exhaustive alignment multi component} satisfies
	\begin{equation}
		\widehat{\pi}_k = \iota \circ \pi_k^*,
	\end{equation}
	for all $k \in S$.
\end{theorem}

The biggest issue of Algorithm \ref{algo: exhaustive alignment multi component} is the computational time. It is easy to see that the time complexity of it is $\mathcal{O}((R!)^K\cdot RK^2)$, because it needs to search over all permutations. This is not practically feasible when $R$ and $K$ are large. Therefore, we propose the following greedy search algorithm to reduce the computational cost, which is summarized in Algorithm \ref{algo: greedy alignment multi component}. Note that its main idea is similar to Algorithm \ref{algo: greedy alignment} for the binary GMM, but the procedure is different. We define the score of alignments $\{\pi_{k'}\}_{k'=1}^k$ of tasks $1$-$k'$ as
\begin{align}
	&\textup{score}(\{\pi_{k'}\}_{k'=1}^k| \{\{\hmu^{(k')[0]}_r\}_{r=1}^R\}_{k'=1}^{k}, \{\hSigma^{(k')[0]}\}_{k'=1}^k) \\
	&= \sum_{r=2}^R\sum_{\tilde{k}, k' \leq k}\twonorma{(\hSigma^{(\tilde{k})[0]})^{-1}\big(\hmu^{(\tilde{k})[0]}_{\pi_{\tilde{k}}(r)} - \hmu^{(\tilde{k})[0]}_{\pi_{\tilde{k}}(1)}) - (\hSigma^{(k')[0]})^{-1}(\hmu^{(k')[0]}_{\pi_{k'}(r)} - \hmu^{(k')[0]}_{\pi_{k'}(1)}\big)}. \label{eq: score multi-component mtl alignment}
\end{align}

\begin{algorithm}[!h]
\caption{Greedy search for the alignment in multi-cluster GMMs}
\label{algo: greedy alignment multi component}
\KwIn{Initialization $\{\{\hmu^{(k)[0]}_r\}_{r=1}^R, \hSigma^{(k)[0]}\}_{k=1}^K$}
\For{$k = 1$ \KwTo $K$}{
	With $\{\widehat{\pi}_{k'}\}_{k'=1}^{k-1}$ fixed ($\emptyset$ when $k=1$), set $\widehat{\pi}_k = \argmin\limits_{\pi:[R]\rightarrow [R]}\textup{score}(\{\widehat{\pi}_{k'}\}_{k'=1}^{k-1} \cup \pi| \{\{\hmu^{(k')[0]}_r\}_{r=1}^R\}_{k'=1}^{k}, \{\hSigma^{(k')[0]}\}_{k'=1}^k)$\\
}
\KwOut{$\widehat{\bm{\pi}}= \{\widehat{\pi}_k\}_{k=1}^K$}
\end{algorithm}

The subsequent theorem demonstrates that, with slightly stronger assumptions than those required by Algorithm \ref{algo: exhaustive alignment multi component}, the greedy search algorithm can recover the correct alignment up to a permutation with high probability. Importantly, this approach significantly alleviates the computational cost from  $\mathcal{O}((R!)^K\cdot RK^2)$ to $\mathcal{O}(R!K\cdot RK^2)$.

\begin{theorem}\label{thm: greedy alignment multi component}
Assume there are no outlier tasks in the first $K_0$ tasks, and
\begin{enumerate}[(i)]
	\item $\max_{k \in S}\max_{r\neq j}\Delta^{(k)}_{rj}/\min_{k \in S}\min_{r\neq j}\Delta^{(k)}_{rj} \leq D$ with $D \geq 1$.
	\item $\epsilon < \frac{1}{2Dc_{\bSigma} + 1}$;
	\item $K_0 > 2Dc_{\bSigma}K\epsilon$;
	\item $\min_{k \in S}\min_{r\neq j}\Delta^{(k)}_{rj} \geq \left[\frac{4K_0c_{\bSigma}^{1/2}}{K_0 - Dc_{\bSigma}K\epsilon}h + \frac{(4K_0 + K\epsilon)c_{\bSigma}^{1/2}}{K_0 - Dc_{\bSigma}K\epsilon}\xi\right] \vee \left[\frac{2K_0c_{\bSigma}^{1/2}}{K_0 - 2Dc_{\bSigma}K\epsilon}h + \frac{(2K_0 + 2K\epsilon)c_{\bSigma}^{1/2}}{K_0 - 2Dc_{\bSigma}K\epsilon}\xi\right]$,
\end{enumerate}
	where $\epsilon = \frac{K-s}{K}$ is the outlier task proportion and $c_{\bSigma}$ appears in the condition that $c_{\bSigma}^{-1} \leq \min_{k \in S}\lambda_{\min}(\bSigmaks{k}) \leq \max_{k \in S}\lambda_{\max}(\bSigmaks{k}) \leq c_{\bSigma}$. Then there exists a permutation $\iota: [R] \rightarrow [R]$, such that the output of Algorithm \ref{algo: greedy alignment multi component} satisfies
	\begin{equation}
		\widehat{\pi}_k = \iota \circ \pi_k^*,
	\end{equation}
	for all $k \in S$.
\end{theorem}

\begin{remark}
	Conditions (\rom{2})-(\rom{4}) are similar to the conditions in Theorem \ref{thm: greedy alignment}. The inclusion of Condition (\rom{1}) aims to facilitate the analysis in the proof, and we conjecture that the obtained results persist even if this condition is omitted.
\end{remark}

When $R$ is very large, the computational burden becomes prohibitive, rendering even the $\mathcal{O}(R!K\cdot RK^2)$ time complexity impractical. Addressing this computational challenge requires the development of more efficient alignment algorithms, a pursuit that we defer to future investigations. In addition, one caveat of the greedy search algorithm is that we need to know $K_0$ non-outlier tasks a priori, which may not be unrealistic in practice. In our empirical examinations, we enhance the algorithm's performance by introducing a random shuffle of the $K$ tasks in each iteration. Specifically, we execute Algorithm \ref{algo: greedy alignment multi component} for 200 times, yielding 200 alignment candidates. The final alignment is then determined by selecting the configuration that attains the minimum score among the candidates.

\section{Other Extensions}\label{sec: other extensions}
\subsection{Extensions to different cluster numbers}\label{subsec: different cluster numbers}
Previously in Section \ref{sec: mtl multi}, we assumed that all non-outlier tasks share the same number of clusters, $R$. In this subsection, we explore potential extensions to situations in which different tasks have varying cluster numbers.

Throughout this subsection, we assume that the data for the $k$-th task are generated from a multi-component GMM with $R^{(k)}$ clusters, where $R^{(k)}$ is an integer greater than 1. Other notations, such as $\wks{k}_r, \bmuks{k}_r, \bSigmaks{k}, \bbetaks{k}_r, \deltaks{k}_r$, remains consistent with previous usage.

\subsubsection{A direct extension of MTL-GMM and alignment algorithms}\label{subsubsec: direct extension diff R}
\begin{figure}[!h]
	\centering
	\includegraphics[width=\textwidth]{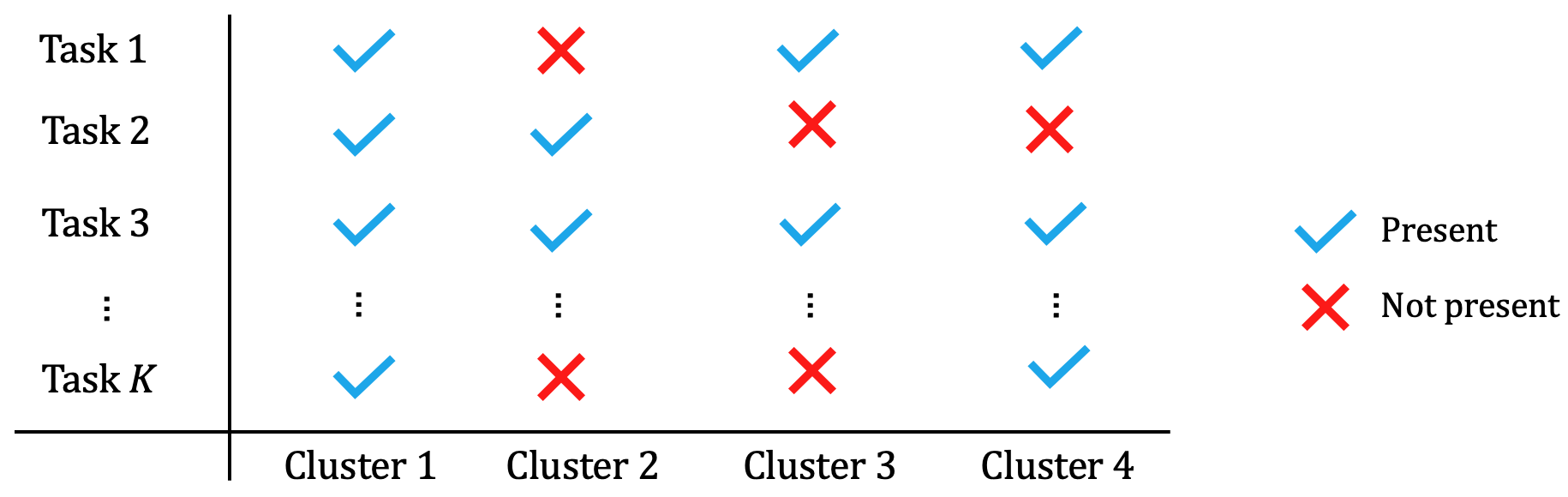}
	\caption{An illustration of multi-component GMM with varying numbers of clusters across tasks.}
	\label{fig: diff-R-schametic}
\end{figure}

We begin by discussing a direct extension of the MTL-GMM model and corresponding alignment algorithms in the multi-task learning (MTL) framework described earlier. Consider the scenario in which the number of clusters, $R^{(k)}$, differs across tasks, yet Gaussian components belonging to the same cluster remain similar according to the earlier $\ell_2$-similarity conditions imposed on $\{\bbetaks{k}_r\}_{k \in S}$. This can be viewed as a missing-data scenario, wherein certain clusters are not observed in some tasks. For simplicity, we assume a reference cluster is observed in every task. This assumption is nearly necessary for leveraging cross-task similarity in $\{\bbetaks{k}_r\}_{k \in S}$, since such similarity cannot be meaningfully defined without a common reference class. This setting is illustrated in Figure \ref{fig: diff-R-schametic}. 

It is straightforward to see that the local update for each task in Algorithm \ref{algo: multitask multi-component} remains applicable even when $R^{(k)}$ varies among tasks. The primary challenge arises during the aggregation steps 11-12. Nevertheless, if the initialization is ``well-aligned", as depicted in Figure \ref{fig: diff-R-schametic}, we can conduct aggregation separately for each cluster, including only tasks in which that cluster is observed. Hence, the key consideration here is alignment.

Analogous to the scenario where the number of clusters is uniform across tasks, we can adopt alignment procedures similar to those introduced in Section \ref{subsec: alignment multi-cluster mtl}. For simplicity and computational efficiency, we only discuss specifically how the greedy search algorithm (Algorithm \ref{algo: greedy alignment multi component}) can be generalized to accommodate the current setting.

Suppose we have initializations $(\{\hmu^{(k)[0]}_r\}_{r=1}^{R^{(k)}}, \hSigma^{(k)[0]})$ from task $k$ for all $k \in [K]$. Define $R_{\max} = \max_{k \in S}R^{(k)}$ and let an alignment $\pi: [R_{\max}] \rightarrow [R^{(k)}] \cup \{\clubsuit\}$ be a permutation function for task $k$. The definition of alignment here slightly differs from the earlier setting due to varying numbers of clusters. Specifically, under the alignment function $\pi$, each cluster $r \in [R_{\max}]$ is either uniquely mapped onto an element in $[R^{(k)}]$ or assigned the special symbol $\clubsuit$, indicating that the corresponding cluster is absent in task $k$ (see Figure \ref{fig: diff-R-schametic}). Since we assume the reference cluster is always observed, we enforce the constraint $\pi(1) \neq \clubsuit$.

Under this framework, we modify the scoring function in \eqref{eq: score multi-component mtl alignment} for an alignment plan $\{\pi_{k'}\}_{k'=1}^k$ of tasks $1$ through $k$ as follows:
\begin{align}
	&\textup{score}(\{\pi_{k'}\}_{k'=1}^k| \{\{\hmu^{(k')[0]}_r\}_{r=1}^{R^{(k')}}\}_{k'=1}^{k}, \{\hSigma^{(k')[0]}\}_{k'=1}^k) \\
	&= \sum_{r=2}^{R_{\max}} \notes{\sum_{\substack{\tilde{k}, k' \leq k \\ \pi_{\tilde{k}}(r), \pi_{k'}(r) \neq \clubsuit}}}\twonorma{(\hSigma^{(\tilde{k})[0]})^{-1}\big(\hmu^{(\tilde{k})[0]}_{\pi_{\tilde{k}}(r)} - \hmu^{(\tilde{k})[0]}_{\pi_{\tilde{k}}(1)}) - (\hSigma^{(k')[0]})^{-1}(\hmu^{(k')[0]}_{\pi_{k'}(r)} - \hmu^{(k')[0]}_{\pi_{k'}(1)}\big)} \notes{\times \frac{\binom{k}{2}}{\binom{\#\{k' \leq k: \pi_{k'}(r)\neq \clubsuit\}}{2}}}. \\\label{eq: score multi-component mtl alignment diff R}
\end{align}
Compared to the original definition \eqref{eq: score multi-component mtl alignment}, this updated definition differs mainly in the inner summation, which now only considers pairs of tasks where the corresponding clusters are present, and the introduction of a scaling factor. This scaling factor mitigates bias toward alignments involving fewer tasks. Without it, alignments that map a cluster to less frequent clusters might artificially achieve higher scores due to fewer terms in the inner summation.  The minimization in Step 2 of Algorithm \ref{algo: greedy alignment multi component} is now over all alignment functions $\pi: [R_{\max}] \rightarrow [R^{(k)}] \cup \{\clubsuit\}$ with $\pi(1) \neq \clubsuit$. While in the original version of the algorithm (when all $R^{(k)}$ are equal), the greedy alignment can begin from the second task, in this extended setting we must also optimize over the reference class $\pi_1(1)$ in the first task. This is because the choice of reference class is inconsequential when all tasks have the same number of clusters, but becomes essential when $R^{(k)}$ varies across tasks. After replacing \eqref{eq: score multi-component mtl alignment} with \eqref{eq: score multi-component mtl alignment diff R} in Algorithm \ref{algo: greedy alignment multi component} along with the aforementioned modifications, the other algorithmic steps remain unchanged, resulting in the greedy search alignment algorithm for the scenario of different cluster numbers.

As indicated in Remark \ref{rmk: greedy search binary}, the alignment obtained by the greedy search may depend on the task ordering in practice. To enhance robustness, one might perform the greedy alignment algorithm multiple times with randomly shuffled task orders and subsequently select the alignment with the lowest score. However, this approach increases computational demands. In our numerical studies, the algorithm performs sufficiently well even without multiple random repetitions.

As we will discuss further in Section \ref{subsubsec: simulation diff R}, this proposed extension proves effective in practice. Nevertheless, the theoretical characterization of this approach is very complex and is left for future investigation.

\subsubsection{An extension based on representation learning}\label{subsubsec: extension repre diff R}
In addition to the straightforward extension described in Section \ref{subsubsec: direct extension diff R}, another potential framework can be developed from multi-task representation learning methods (e.g., \citealpapp{du2021few, tripuraneni2021provable}). We briefly outline this alternative framework below. \footnote{We thank one reviewer for suggesting this potential extension.}

Adopting the same notation from the previous subsection, suppose the $k$-th task comprises $R^{(k)}$ Gaussian clusters. Departing from the $\ell_2$-similarity notion emphasized earlier, we assume a shared low-rank structure among different tasks. Specifically, assume there exist a $p \times d$ orthonormal matrix $\bm{A}^*$ (i.e. $(\bm{A}^*)^\top \bm{A}^* = \bm{I}_{d\times d}$) and a $d$-dimensional vector $\bthetaks{k}_r$ such that
\begin{equation}\label{eq: low-rank setting}
	\bmuks{k}_r = \bm{A}^*\bthetaks{k}_r, \quad  r = 1:R^{(k)}, \quad  k \in S,
\end{equation} 
where $d \leq p$. Equivalently, this assumption indicates that the mean vectors $\{\bmuks{k}_r\}_{k \in S, r \in [R^{(k)}]}$ live in the column space of $\bm{A}^*$, a $d$-dimensional subspace. When $d$ is substantially smaller than $p$, exploiting such shared low-dimensional structure can considerably enhance estimation accuracy.

The biggest advantage of the framework \eqref{eq: low-rank setting} is that it circumvents the alignment requirement inherent in the $\ell_2$-similarity-based framework. Given that this significantly deviates from our primary framework, we do not provide a full exploration here. Instead, we briefly illustrate it as an alternative to handle varying numbers of clusters across tasks.

Various strategies exist in the literature to estimate this kind of low-rank model in the supervised learning context, including direct empirical risk minimization \citepapp{du2021few, tripuraneni2021provable}, alternative gradient descent \citepapp{thekumparampil2021statistically}, spectral methods \citepapp{tian2025learning}, and method of moments \citepapp{tripuraneni2021provable}. For simplicity, we only discuss how the spectral method proposed by \citeapp{tian2025learning} might be adapted to the current GMM setup. Moreover, we assume $S = [K]$, meaning there are no outlier tasks.

The spectral-based representation learning method consists of three primary steps:
\begin{enumerate}
\item Obtain single-task initial estimates of $\bmuks{k}_r$ for all clusters and tasks. Collect these estimates into a large matrix $\bm{M}$ by stacking each initial estimate as columns.
\item Conduct the singular value decomposition (SVD) on the matrix $\bm{M}$, and select the left singular vectors corresponding to the largest $d$ singular values to form an estimator $\widehat{\bm{A}}$ of the subspace matrix $\bm{A}^*$.
\item Project each observation $\bzk{k}_i$ onto the low-dimensional subspace by replacing it with the projected vector $\widehat{\bm{A}}^\top \bzk{k}_i$, and subsequently fit a GMM separately for each task within this lower-dimensional space.
\end{enumerate}
Note that the initial estimators for $\bmuks{k}_r$ and the subsequent fitting of the GMM in the projected subspace can be performed using any appropriate method. For simplicity, in the numerical experiments presented in Section \ref{subsubsec: simulation diff R}, we use the EM algorithm for both steps.

Lastly, we highlight that the exact low-rank assumption in \eqref{eq: low-rank setting} can be restrictive in practice. \citeapp{tian2025learning} suggests several possible extensions to relax the strict assumption of sharing an identical representation matrix $\bm{A}^*$ and to incorporate the presence of outlier tasks in supervised settings. Extending these relaxations and robustness considerations to unsupervised settings may involve significant challenges and warrants future research efforts.

\subsection{Determination of cluster numbers}
Throughout this paper, we have assumed the number of clusters for each task to be known a priori, which may be too idealistic in practice. Although we do not investigate scenarios with unknown numbers of clusters, we recognize this as a critical problem in clustering analysis, particularly in the context where cluster numbers vary across tasks. Therefore, we briefly discuss this issue below.

Determining the optimal number of clusters is a classical problem in clustering, with extensive literature available, especially within the model-based clustering framework considered here. Commonly used approaches include likelihood-based cross-validation \citepapp{smyth2000model}, BIC-based model selection \citepapp{fraley1998many, fraley2002model}, the gap statistic inspired by the heuristic elbow method \citepapp{tibshirani2001estimating}, and the Calinski-Harabasz index \citepapp{calinski1974dendrite}, among others. Most of these approaches rely on selecting the number of clusters according to specific model selection criteria. In theory, any method that applies to classical single-task clustering can also be employed here individually on each task's dataset. Once the number of clusters per task is determined, our multi-task algorithms can then be directly applied. From a theoretical perspective, if the chosen criterion for determining the cluster number satisfies certain assumptions and consequently identifies the correct number of clusters with high probability, then our theoretical analyses presented previously can be extended by conditioning on this high-probability event.

In cases where one believes the number of clusters is consistent across all tasks, it may be advantageous to propose a method that jointly selects the cluster number using data from all tasks simultaneously. A straightforward method could involve evaluating model-selection criteria separately on each task and subsequently aggregating the criterion values, such as averaging or using the median, to determine a single unified cluster number.

\begin{figure}[!h]
	\centering
	\includegraphics[width=\textwidth]{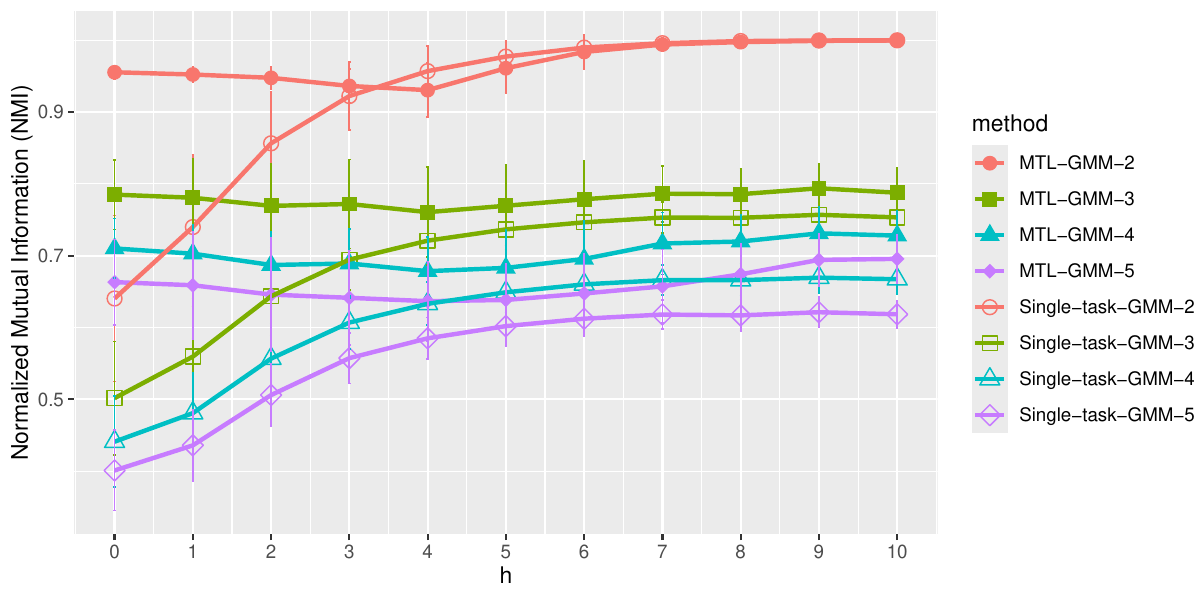}
	\caption{Comparison of NMI across different methods in Simulation 1 (no outliers), under varying numbers of clusters specified in the algorithms. The numbers in the method names (e.g., Single-task-GMM-2, MTL-GMM-3) indicate the specified cluster number. In Simulation 1, the true number of clusters is 2.}
	\label{fig: misspecified_em}
\end{figure}

Another natural question is: what happens if the number of clusters is misspecified? In general, when the number of clusters used in our algorithms is smaller than the true number, theoretical guarantees no longer hold. On the other hand, if the number of clusters is overestimated, we conjecture that an excess mis-clustering error bound similar to ours can still be derived, although the convergence rate may differ.  This is partially supported by existing theoretical studies on the behavior of EM algorithms under over-specified cluster numbers in certain simplified GMM settings (e.g., \citealpapp{dwivedi2018theoretical, dwivedi2020singularity, zhou2025global}). Moreover, we expect that the MTL-GMM algorithm can still outperform the single-task EM algorithm even when both use the same misspecified number of clusters. To verify this, we conduct an experiment using the same setting as in Simulation 1 in Section \ref{subsec: simulation binary}, varying the number of clusters used in both MTL-GMM and single-task EM. Specifically, we set the cluster number to $2$, $3$, $4$, and $5$ and use Normalized Mutual Information (NMI) as the clustering accuracy metric \footnote{We do not use estimation or mis-clustering error here, as these are difficult to define when the predicted and true numbers of clusters differ. For two random variables $Y_1$ and $Y_2$, $\textup{NMI}(Y_1; Y_2) \coloneqq \frac{I(Y_1; Y_2)}{\sqrt{H(Y_1)H(Y_2)}}$, where $I(Y_1; Y_2)$ is the mutual information and $H(Y_1)$, $H(Y_2)$ are the corresponding entropies.}. The results are presented in Figure \ref{fig: misspecified_em}. As expected, over-specifying the number of clusters generally reduces clustering accuracy. Nevertheless, MTL-GMM continues to outperform Single-task-GMM across all settings, indicating its advantage even under model misspecification. An additional direction worth exploring is to start with a sufficiently large number of clusters and then prune unnecessary components to reduce the model complexity, as suggested in \citeapp{dasgupta2000two}.

Given the extensive material already presented in this paper, we refrain from providing additional detailed discussions or theoretical investigations on this topic. However, we hope this brief outline is helpful and encourages further exploration in future studies.

\subsection{Heterogeneous covariance matrices across clusters}
Throughout this paper, we have assumed that different Gaussian clusters share identical covariance matrices in GMMs. Here, we briefly discuss a potential extension to the case where clusters have heterogeneous covariance matrices. For simplicity, we consider a binary GMM setup where $\bzk{k}_i \overset{\textup{i.i.d.}}{\sim} (1-\wks{k})N(\bmuks{k}_1, \bSigmaks{k}_1) + \wks{k}N(\bmuks{k}_2, \bSigmaks{k}_2)$. Unlike the homogeneous case \eqref{eq: bayes lda intro section}, where $\bSigmaks{k}_1 = \bSigmaks{k}_2$ and the Bayes classifier is linear, the Bayes decision boundary in this setting becomes quadratic, resulting in the classifier
\begin{equation}
	\mathcal{C}^{(k)}(\bz) = \begin{cases}
		1, & \text{if }\frac{1}{2}\bz^\top\bm{\Omega}^{(k)*}\bz + (\bbetaks{k})^\top\bz \leq \log\Big(\frac{1-\wks{k}}{\wks{k}}\Big) + \frac{1}{2}\log\Big(\frac{|\bSigmaks{k}_2|}{|\bSigmaks{k}_1|}\Big)\\
		& \hspace{5cm}+ \frac{1}{2}(\bmuks{k}_2)^\top(\bSigmaks{k}_2)^{-1}\bmuks{k}_2 - \frac{1}{2}(\bmuks{k}_1)^\top(\bSigmaks{k}_1)^{-1}\bmuks{k}_1;\\
		2, & \text{otherwise},
	\end{cases}
\end{equation}
where $\bm{\Omega}^{(k)*} = (\bSigmaks{k}_1)^{-1} - (\bSigmaks{k}_2)^{-1}$ and $\bbetaks{k} = (\bSigmaks{k}_2)^{-1}\bmuks{k}_2 - (\bSigmaks{k}_1)^{-1}\bmuks{k}_1$.

To reduce the mis-clustering error in this heterogeneous setting, we would need to impose similarity assumptions on both $\bm{\Omega}^{(k)*}$ and $\bbetaks{k}$. In the homogeneous case, we use the Euclidean distance as a similarity metric for $\bbetaks{k}$, since the excess mis-clustering error can be explicitly written in terms of $\bbetaks{k}$ and its estimator, and further upper bounded by the $\ell_2$-estimation error of $\bbetaks{k}$. In contrast, the heterogeneous case introduces more complexity, as the mis-clustering error does not admit a closed-form expression \citepapp{li2015sparse}. To the best of our knowledge, existing work in both supervised quadratic discriminant analysis (QDA) and unsupervised clustering contexts only provides loose upper bounds, often using the $\ell_{\infty}$-error on the vectorized forms of $\bm{\Omega}^{(k)*}$ and $\bbetaks{k}$ (e.g., \citealpapp{fan2015quadro, jiang2018direct}).

If tighter control of the excess mis-clustering error in terms of the estimation errors of $\bm{\Omega}^{(k)*}$ and $\bbetaks{k}$ can be established, then a multi-task EM algorithm similar to Algorithm \ref{algo: multitask} could potentially be developed. This would involve aggregating information across tasks through penalization, with the specific penalty form depending on the type of estimation error used in the error bound. This presents the main challenge in extending our methods and theoretical guarantees to the heterogeneous case. Other challenges, such as generalizing the alignment algorithms, may be more straightforward. Given the current paper's comprehensive coverage, we leave this important direction for future work.

\section{Transfer Learning}\label{sec: tl}

\subsection{Problem setting}\label{subsec: problem tl}
In the main text and Section \ref{sec: mtl multi}, we discussed GMMs under the context of \textit{multi-task learning}, where the goal is to learn all tasks jointly by utilizing the potential similarities shared by different tasks. In this section, we will study binary GMMs in the \textit{transfer learning} context where the focus is on the improvement of learning in one target task through the transfer of knowledge from related source tasks. Multi-cluster results can be obtained similarly as in the MTL case, and we omit the details given the extensive length of the paper.

Suppose that there are $(K + 1)$ tasks in total, where the first task is called the target and the $K$ remaining ones are called $K$ sources. As in multi-task learning, we assume that there exists an unknown subset $S \subseteq 1:K$, such that samples from sources in $S$ follow an independent GMM, while samples from sources outside $S$ can be arbitrarily distributed. This means,
\begin{align}
	y_i^{(k)} = \begin{cases}
		1, &\text{with probability }1-\wks{k};\\
		2, &\text{with probability }\wks{k};
	\end{cases}
\end{align}
\begin{equation}
	\bz_i^{(k)}|y_i^{(k)}=j \sim \mathcal{N}(\bmuks{k}_j, \bSigmaks{k}), \,\,j = 1,2,
\end{equation}
for all $k \in S$, $i = 1:n_k$, and
\begin{equation}
	\{\bzk{k}_i\}_{i,k\in S^c} \sim \mathbb{Q}_{S},
\end{equation}
where $\mathbb{Q}_S$ is some probability measure on $(\mathbb{R}^p)^{\otimes \nsc}$ and $\nsc = \sum_{k \in S^c}n_k$. 

For the target task, we observe sample $\{\bz_i^{(0)}\}_{i=1}^{n_0}$ independently sampled from the following GMM:
\begin{align}
	y_i^{(0)} = \begin{cases}
		1, &\text{with probability }1-\wks{0};\\
		2, &\text{with probability }\wks{0};
	\end{cases}
\end{align}
\begin{equation}
	\bz_i^{(0)}|y_i^{(0)}=j \sim \mathcal{N}(\bmuks{0}_j, \bSigmaks{0}), \,\,j = 1,2.
\end{equation}

The objective of transfer learning is to use source data to help improve GMM learning in the \textit{target} task. As for multi-task learning, we measure the learning performance by both parameter estimation error and the excess mis-clustering error, but only on the \textit{target} GMM. Toward this end, we define the joint parameter space for GMM parameters of the target and sources in $S$:
\begin{equation}
	\overline{\Theta}_{S}'(h) = \Big\{\{\overline{\btheta}^{(k)}\}_{k  \in \{0\} \cup S} = \{(\wk{k}, \bmuk{k}_1, \bmuk{k}_2, \bSigmak{k})\}_{k \in \{0\} \cup S}: \othetak{k} \in \overline{\Theta}, \max_{k  \in S}\twonorm{\bbetak{k}-\bbetak{0}} \leq h\Big\},  \label{eq: parameter space tl}
\end{equation}
where $\overline{\Theta}$ is the single GMM parameter space introduced in \eqref{eq: parameter space Theta}, and $\bbetak{k} = (\bSigmak{k})^{-1}(\bmuk{k}_2 - \bmuk{k}_1)$, $k \in \{0\} \cup S$. Comparing $\overline{\Theta}_{S}'(h)$ with the parameter space $\overline{\Theta}_{S}(h)$ from multi-task learning in \eqref{eq: parameter space mtl}, here the target discriminant coefficient $\bbetak{0}$ serves as the ``center" of discriminant coefficients of sources in $S$. The quantity $h$ characterizes the closeness between sources in $S$ and the target.

\subsection{Method}\label{subsec: method tl}
Like the MTL-GMM procedure developed in Section \ref{subsec: method mtl}, we combine the EM algorithm and the penalization framework to develop a variant of the EM algorithm for transfer learning. The key idea is to first apply MTL-GMM to all the sources to obtain estimates of discriminant coefficient ``center" as good summary statistics of the $K$ source data sets, and then shrink the target discriminant coefficient towards those center estimates in the EM iterations to explore the relatedness between sources and the target. See Section 3.3 of \citeapp{duan2023adaptive} for more general discussions on this idea. Our proposed transfer learning procedure TL-GMM is summarized in Algorithm \ref{algo: transfer}.

While the steps of TL-GMM look very similar to those of MTL-GMM, there exist two major differences between them. First, for each optimization problem in TL-GMM, the first part of the objective function only involves the target data $\{\bzk{0}_i\}_{i=1}^{n_0}$, while in MTL-GMM, it is a weighted average of all tasks. Second, in TL-GMM, the penalty is imposed on the distance between a discriminant coefficient estimator and a given center estimator produced by MTL-GMM from the source data. In contrast, the center is estimated simultaneously with other parameters through the penalization in MTL-GMM. In light of existing transfer learning approaches in the literature, TL-GMM can be considered as the ``debiasing" step described in \citeapp{li2021transfer} and \citeapp{tian2023transfer}, which corrects potential bias of the center estimate using the target data. 

The tuning parameters $\{\lambda^{[t]}_0\}_{t=1}^{T_0}$ in Algorithm \ref{algo: transfer} control the amount of knowledge to be transferred from sources. Setting tuning parameters large enough pushes parameter estimates for the target task to be exactly equal to the center learned from sources while letting them be zero makes TL-GMM reduce to the standard EM algorithm on the target data.

\begin{algorithm}[!h]
\caption{TL-GMM}
\label{algo: transfer}
\KwIn{Initialization $\htheta^{(0)[0]} = (\hw^{(0)[0]}, \hbeta^{(0)[0]}, \hdelta^{(0)[0]})$, output $\obbeta^{[T]}$ from Algorithm \ref{algo: multitask}, maximum number of iteration rounds $T_0$, initial penalty parameter $\lambda^{[0]}_0$, tuning parameters $C_{\lambda_0} > 0$ and $\kappa_0 \in (0, 1)$}
\For{$t = 1$ \KwTo $T_0$}{
		$\lambda^{[t]}_0 = \kappa_0 \lambda^{[t-1]}_0 + C_{\lambda_0}\sqrt{p+\log K}$ \tcp*[f]{Update the penalty parameter}\\
	$\hw^{(0)[t]} = \frac{1}{n_0}\sum_{i=1}^{n_0}\gamma_{\htheta^{(0)[t-1]}}(\bz_i^{(0)})$\\ 
	$\hmu^{(0)[t]}_1 = \frac{\sum_{i=1}^{n_0}[1-\gamma_{\htheta^{(0)[t-1]}}(\bz_i^{(0)})]\bzk{0}_i}{n_0(1-\hw^{(0)[t]})}$, $\hmu^{(0)[t]}_2 = \frac{\sum_{i=1}^{n_0}\gamma_{\htheta^{(0)[t-1]}}(\bz_i^{(0)})\bzk{0}_i}{n_0\hw^{(0)[t]}}$ \\
	$\hSigma^{(0)[t]} = \frac{1}{n_0}\sum_{i=1}^{n_0}\left\{[1-\gamma_{\htheta^{(0)[t-1]}}(\bz_i^{(0)})]\cdot (\bzk{0}_i - \hmu^{(0)[t]}_1)(\bzk{0}_i - \hmu^{(0)[t]}_1)^\top \right.$ $\left. \hspace{3.3cm} + \gamma_{\htheta^{(0)[t-1]}}(\bz_i^{(0)}) \cdot (\bzk{0}_i - \hmu^{(0)[t]}_2)(\bzk{0}_i - \hmu^{(0)[t]}_2)^\top\right\}$\\
	$\hbeta^{(0)[t]} = \argmin\limits_{\bbetak{0}}\left\{\left[\frac{1}{2}(\bbetak{0})^\top\hSigma^{(0)[t]}\bbetak{0} - (\bbetak{0})^\top(\hmu_2^{(0)[t]}-\hmu_1^{(0)[t]})\right] + \frac{\lambda^{[t]}_0}{\sqrt{n_0}}\twonorm{\bbetak{0}-\obbeta^{[T]}}\right\}$ \\
	$\hdelta^{(0)[t]} = \frac{1}{2}(\hbeta^{(0)[t]})^\top(\hmu^{(0)[t]}_1 + \hmu^{(0)[t]}_2)$\\
		Let $\htheta^{(0)[t]} = (\hw^{(0)[t]}, \hbeta^{(0)[t]}, \hdelta^{(0)[t]})$
}
\KwOut{$(\htheta^{(0)[T_0]}, \hmu^{(0)[T_0]}_1, \hmu^{(0)[T_0]}_2, \hSigma^{(0)[T_0]})$ with $\htheta^{(0)[T_0]} = (\hw^{(0)[T_0]}, \hbeta^{(0)[T_0]}, \hdelta^{(0)[T_0]})$}
\end{algorithm}

\subsection{Theory}\label{subsec: theory tl}
In this section, we will establish the upper and lower bounds for the GMM parameter estimation error and the excess mis-clustering error on the \textit{target} task. First, we impose the following assumption set.

\begin{assumption}\label{asmp: upper bound transfer est error}
	Denote $\Delta^{(0)} = \sqrt{(\bmuks{0}_1 - \bmuks{0}_2)^\top(\bSigmaks{0})^{-1}(\bmuks{0}_1 - \bmuks{0}_2)}$. Assume the following conditions hold: 
	\begin{enumerate}[(i)]
		\item $C_1\left[\frac{\log K}{p} + \epsilon^2\big(1+\frac{\log K}{p}\big)\right]\leq \frac{\max_{k \in S}n_k}{n_0} \leq C_2\big(1+\frac{\log K}{p}\big)$ with constants $C_1$ and $C_2$, where $\epsilon = \frac{K-s}{K}$.
		\item $n_0 \geq C_3p$ with some constant $C_3$;
		\item Either of the following two conditions holds with some constant $C_4$:
			\begin{enumerate}
				\item $\twonorm{\hbeta^{(0)[0]} - \bbetaks{0}} \vee \norm{\hdelta^{(0)[0]} - \deltaks{0}}\leq C_4\Delta^{(0)}$, $\norm{\hw^{(0)[0]}-\wks{0}}\leq c_w/2$;
				\item $\twonorm{\hbeta^{(0)[0]} + \bbetaks{0}} \vee \norm{\hdelta^{(0)[0]} + \deltaks{0}} \leq C_4\Delta^{(0)}$, $\norm{1-\hw^{(0)[0]}-\wks{0}}\leq c_w/2$.
			\end{enumerate}
		\item $\Delta^{(0)} \geq C_5 > 0$ with some constant $C_5$;
	\end{enumerate}
\end{assumption}

\begin{remark}
	Condition (\rom{1}) requires the target sample size not to be much smaller than the maximum source sample size, which appears due to technical reasons in the proof. Conditions (\rom{2})-(\rom{4}) can be seen as the counterpart of Conditions (\rom{2})-(\rom{4}) in Assumption \ref{asmp: upper bound multitask est error} for the target GMM.
\end{remark}

We are in the position to present the upper bounds of the estimation error of GMM parameters for TL-GMM.

\begin{theorem}(Upper bounds of the estimation error of GMM parameters for TL-GMM)\label{thm: upper bound transfer est error}
	Suppose the conditions in Theorem \ref{thm: upper bound multitask est error} and Assumption \ref{asmp: upper bound transfer est error} hold. Let $\lambda^{[0]}_0 \geq C_1\max_{k=1:K}\sqrt{n_k}$, $C_{\lambda_0} \geq C_1  $, $\kappa_0 > C_2$ with some specific constants $C_1 > 0, C_2 \in (0, 1)$. Then there exists a constant $C_3 > 0$, such that for any $\{\othetaks{k}\}_{k \in \{0\} \cup S} \in \overline{\Theta}_{S}'(h)$ and any probability measure $\mathbb{Q}_S$ on $(\mathbb{R}^p)^{\otimes n_{S^c}}$, we have
	\begin{align}
		d(\htheta^{(0)[T_0]}, \bthetaks{0}) &\lesssim \sqrt{\frac{p}{\ns + n_0}} + \sqrt{\frac{1}{n_0}}+ h \wedge \sqrt{\frac{p}{n_0}} + \bigg(\epsilon\sqrt{\frac{p+\log K}{\max_{k=1:K}n_k}}\bigg)\wedge \sqrt{\frac{p}{n_0}} \\
		&\quad\quad\quad\quad +\sqrt{\frac{\log K}{\max_{k =1:K}n_k}}  + T_0(\kappa_0')^{T_0},
	\end{align}
	\begin{equation}
		\min_{\pi: [2] \rightarrow [2]}\max_{r=1:2}\twonorm{\hmu^{(0)[T_0]}_{\pi(r)} - \bmuks{0}_r} \vee \twonorm{\hSigma^{(0)[T_0]} - \bSigmaks{0}} \lesssim \sqrt{\frac{p}{n_0}} + T_0(\kappa_0')^{T_0},
	\end{equation}
	with probability at least $1-C_3K^{-1}$, where $\kappa'_0 \in (0, 1)$ and $\ns = \sum_{k \in S}n_k$. When $T_0 \geq C\log n_0$ with a large constant $C > 0$, in both inequalities, the last term on the right-hand side will be dominated by other terms.
\end{theorem}

Next, we present the upper bound of the excess mis-clustering error on the target task for TL-GMM. Having the estimator $\htheta^{(0)[T_0]}$ and the truth $\bthetaks{0}$, the clustering method $\widehat{\mC}^{(0)[T_0]}$  and its mis-clustering error $R_{\othetaks{0}}(\widehat{\mC}^{(0)[T_0]})$ are defined in the same way as in \eqref{eq: clustering method ck} and \eqref{eq: clustering method ck error}.

\begin{theorem}(Upper bound of the target excess mis-clustering error for TL-GMM)\label{thm: upper bound transfer classification error}
Suppose the same conditions in Theorem \ref{thm: upper bound transfer est error} hold.  Then there exists a constant $C_1 > 0$ such that for any $\{\othetaks{k}\}_{k \in \{0\} \cup S} \in \overline{\Theta}_S'(h)$ and any probability measure $\mathbb{Q}_S$ on $(\mathbb{R}^p)^{\otimes n_{S^c}}$, with probability at least $1-C_1K^{-1}$ the following holds:
\begin{align}
	R_{\othetaks{0}}(\widehat{\mathcal{C}}^{(0)[T_0]}) - R_{\othetaks{0}}(\mC_{\othetaks{0}}) &\lesssim \underbrace{\frac{p}{\ns + n_0}}_{\rm (\Rom{1})} + \underbrace{\frac{1}{n_0}}_{\rm (\Rom{2})} + \underbrace{h^2 \wedge \frac{p}{n_0}}_{\rm (\Rom{3})} + \underbrace{\epsilon^2\frac{p+\log K}{\max_{k=1:K}n_k}\wedge \frac{p}{n_0}}_{\rm (\Rom{4})} \\
		&\quad\quad\quad\quad + \underbrace{\frac{\log K}{\max_{k =1:K}n_k}}_{\rm (\Rom{5})} + \underbrace{T_0^2(\kappa_0')^{2T_0}}_{\rm (\Rom{6})},
\end{align}
with some constant $\kappa'_0 \in (0,1)$. When $T_0  \geq C\log n_0$ with some large constant $C > 0$, the last term in the upper bound will be dominated by the second term.
\end{theorem}

Similar to the upper bounds of $d(\htheta^{(k)[T]}, \bthetaks{k})$ and $R_{\othetaks{k}}(\widehat{\mathcal{C}}^{(k)[T]}) - R_{\othetaks{k}}(\mC_{\othetaks{k}})$ in Theorems \ref{thm: upper bound multitask est error} and \ref{thm: upper bound multitask classification error}, the upper bounds for $d(\htheta^{(0)[T_0]}, \bthetaks{0})$ and $R_{\othetaks{0}}(\widehat{\mathcal{C}}^{(0)[T_0]}) - R_{\othetaks{0}}(\mC_{\othetaks{0}})$ consist of multiple parts with one-to-one correspondence. We take the bound of $R_{\othetaks{0}}(\widehat{\mathcal{C}}^{(0)[T_0]}) - R_{\othetaks{0}}(\mC_{\othetaks{0}})$ in Theorem \ref{thm: upper bound transfer classification error} as an example. Part (\Rom{1}) is the oracle rate $\mathcal{O}_{\tp}\big(\frac{p}{\ns+n_0}\big)$. Part (\Rom{2}) is the error caused by estimating scalar parameters $\deltaks{0}$ and $\wks{0}$  in the decision boundary, which thus do not depend on dimension $p$. Part (\Rom{3}) quantifies the contribution of related sources to the target task learning. The more related the sources in $S$ to the target (i.e. the smaller $h$ is), the smaller Part (\Rom{3}) becomes. Part (\Rom{4}) captures the impact of outlier sources on the estimation error. As $\epsilon$ increases (i.e. the proportion of outlier sources increases), Part (\Rom{4}) first increases and then flats out. It never exceeds the minimax rate $\mathcal{O}_{\tp}(p/n_0)$ of the single task learning on \textit{target} task \citepapp{balakrishnan2017statistical, cai2019chime}. Therefore, our method is robust against a fraction of outlier sources with arbitrary contaminated data. Part (\Rom{5}) is an extra term caused by estimating the center in MTL-GMM, which by Assumption \ref{asmp: upper bound transfer est error}.(\rom{1}) is smaller than the single-task learning rate $\mathcal{O}_{\tp}(p/n_0)$. Part (\Rom{6}) decreases geometrically in the iteration number $T_0$ of Algorithm \ref{algo: transfer}, which becomes negligible by setting the iteration numbers $T_0$ large enough.

 Consider the general scenario $T_0 \gtrsim \log n_0$. Then the upper bound of excess mis-clustering error rate $R_{\othetaks{0}}(\widehat{\mathcal{C}}^{(0)[T_0]}) - R_{\othetaks{0}}(\mC_{\othetaks{0}})$ is guaranteed to be no worse than the optimal single-task learning rate $\mathcal{O}_{\tp}(p/n_0)$. More importantly, in the general regime where $\epsilon \ll \sqrt{\frac{p\max_{k=1:K}n_k}{(p+\log K)n_0}}$ (small number of outlier sources), $h \ll \sqrt{p/n_0}$ (enough similarity between sources and target), $\ns \gg n_0$ (large total source sample size), and $\max_{k \in S}n_k/n_0 \gg \log K/p$ (large maximum source sample size), TL-GMM improves the GMM learning on the target task by achieving a better estimation error rate. As for the upper bound of $\min_{\pi: [2] \rightarrow [2]}\max_{r=1:2}\twonorm{\hmu^{(0)[T_0]}_{\pi(r)} - \bmuks{0}_r}$ and $\twonorm{\hSigma^{(0)[T_0]} - \bSigmaks{0}}$, when $T_0 \gtrsim \log n_0$, it has the single-task learning rate $\mathcal{O}_{\tp}(\sqrt{p/n_0})$. This is expected since the mean vectors and covariance matrices from sources are not necessarily similar to the one from target in the parameter space $\overline{\Theta}_{S}'(h)$.

The following result of minimax lower bounds shows that the upper bounds in Theorems \ref{thm: upper bound transfer est error} and \ref{thm: upper bound transfer classification error} are optimal in a broad range of regimes.

\begin{theorem}(Lower bounds of the estimation error of GMM parameters in transfer learning)\label{thm: lower bound transfer est error}
	Suppose $\epsilon = \frac{K-s}{K} < 1/3$. Suppose there exists a subset $S$ with $|S| \geq s$ such that $\min_{k \in S} n_k \geq C_1(p + \log K)$, $n_0 \geq C_1p$ and $\min_{k \in \{0\}\cup S}\Delta^{(k)} \geq C_2$ with some constants $C_1, C_2 > 0$. Then we have
	\begin{align}
		\inf_{\htheta^{(0)}} \sup_{S: |S| \geq s}\sup_{\substack{\{\othetaks{k}\}_{k \in \{0\}\cup S} \in \overline{\Theta}_{S}'(h) \\ \mathbb{Q}_S}} &\tp\Bigg(d(\htheta^{(0)}, \bthetaks{0}) \gtrsim \sqrt{\frac{p}{\ns+n_0}} + \sqrt{\frac{1}{n_0}} + h\wedge \sqrt{\frac{p}{n_0}} \\
		&\hspace{4cm} + \frac{\epsilon}{\sqrt{\max_{k =1:K}n_k}}\wedge \sqrt{\frac{p}{n_0}}\Bigg) \geq \frac{1}{10},
	\end{align}
	\begin{equation}
		\inf_{\substack{\hmu^{(0)}_1, \hmu^{(0)}_2 \\\hSigma^{(0)}}} \sup_{S: |S| \geq s}\sup_{\substack{\{\othetaks{k}\}_{k \in \{0\}\cup S} \in \overline{\Theta}_{S}'(h) \\ \mathbb{Q}_S}} \tp\Bigg(\min_{\pi: [2] \rightarrow [2]}\max_{r=1:2}\twonorm{\hmu^{(0)}_{\pi(r)} - \bmuks{0}_r} \vee \twonorm{\hSigma^{(0)} - \bSigmaks{0}} \gtrsim  \sqrt{\frac{p}{n_0}}\Bigg) \geq \frac{1}{10}.	
	\end{equation}
\end{theorem}

\begin{theorem}(Lower bound of the target excess mis-clustering error in transfer learning)\label{thm: lower bound transfer classification error}
	Suppose the same conditions in Theorem \ref{thm: lower bound transfer est error} hold. Then we have
	\begin{align}
		\inf_{\widehat{\mathcal{C}}^{(0)}} \sup_{S: |S| \geq s}\sup_{\substack{\{\othetaks{k}\}_{k \in \{0\}\cup S} \in \overline{\Theta}_{S}'(h) \\ \mathbb{Q}_S}} &\tp\Bigg(R_{\othetaks{0}}(\widehat{\mathcal{C}}^{(0)}) - R_{\othetaks{0}}(\mC_{\othetaks{0}}) \gtrsim \frac{p}{\ns + n_0} + \frac{1}{n_0}+ h^2 \wedge \frac{p}{n_0} \\
		&\hspace{5cm} + \frac{\epsilon^2}{\max_{k=1:K}n_k} \wedge \frac{1}{n_0} \Bigg) \geq \frac{1}{10}.
	\end{align}
\end{theorem}

Comparing the upper and lower bounds in Theorems \ref{thm: upper bound transfer est error}- \ref{thm: lower bound transfer classification error}, several remarks are in order:
\begin{itemize}
	\item With $T_0 \gtrsim \log n_0$, our estimators $\hmu^{(0)[T_0]}_1$, $\hmu^{(0)[T_0]}_2$, $\hSigma^{(0)[T_0]}$ achieve the minimax optimal rate for estimating the mean vectors $\bmuks{0}_1$, $\bmuks{0}_2$ and the covariance matrix $\bSigmaks{0}$.
	\item Regarding the target excess mis-clustering error, with the choices $T_0 \gtrsim \log n_0$,  Part (\Rom{6}) in the upper bound becomes negligible. We thus compare the other five terms in the upper bound with the corresponding terms in the lower bound.
		\begin{enumerate}
			\item Part (\Rom{4}) in the upper bound differs from the one in the lower bound by a factor $p$ (up to $\log K$). Hence the gap can arise when the dimension $p$ diverges. The reason is similar to the one in a multi-task learning setting and using statistical depth function based ``center" estimates might be able to close the gap. We refer to the paragraph after Theorem \ref{thm: lower bound multitask classification error} for more details.
			\item Part (\Rom{5}) in the upper bound does not appear in the lower bound. This term is due to the center estimate from the upper bound in MTL-GMM. When $\max_{k \in S}n_k/n_0 \gtrsim \log K$, this term is dominated by Part (\Rom{2}).
			\item The other three terms from the upper bound match with the ones in the lower bound.
		\end{enumerate}
	\item Based on the above comparisons, we can conclude that under the mild condition $\max_{k \in S}n_k/n_0 \gtrsim \log K$, our method is minimax rate optimal for the estimation of $\bthetaks{0}$ in the classical low-dimensional regime $p = O(1)$. Even when $p$ is unbounded, the gap between the upper and lower bounds appears only when the fourth or fifth term is the dominating term in the upper bound. Like the discussions after Theorem \ref{thm: lower bound multitask classification error}, similar restricted regimes where our method might become sub-optimal can be derived.
\end{itemize}

\subsection{Label alignment}\label{subsec: alignment tl}
As in multi-task learning, the alignment issue exists in transfer learning as well. Referring to the parameter space $\overline{\Theta}'(h)$ and the conditions of initialization in Assumptions \ref{asmp: upper bound multitask est error} and \ref{asmp: upper bound transfer est error}, the success of Algorithm \ref{algo: transfer} requires correct alignments in two places. First, the center estimate $\overline{\bbeta}^{[T]}$ used as input of Algorithm \ref{algo: transfer} are obtained from Algorithm \ref{algo: multitask} which involves the alignment of initial estimates for sources. This alignment problem can be readily solved by Algorithm \ref{algo: exhaustive alignment} or \ref{algo: greedy alignment}. Second, the initialization of the target problem $\hbeta^{(0)[0]}$ needs to be correctly aligned with the aforementioned center estimates. This is easy to address using the alignment score described in Section \ref{subsubsection: two alignment algs} as there are only two different alignment options. We summarize the steps in Algorithm \ref{algo: alignment for TL}.

Like Algorithms \ref{algo: exhaustive alignment} and \ref{algo: greedy alignment}, Algorithm \ref{algo: alignment for TL} is able to find the correct alignments under mild conditions. Suppose $\{\hbeta^{(k)[0]}\}_{k = 0}^K$ are the initialization values with potentially wrong alignment. Define the correct alignment as $\bm{r}^* = (r_0^*, r^*_1, \ldots, r^*_K)$ with $r^*_k = \argmin_{r_k = \pm 1} \twonorm{r_k\hbeta^{(k)[0]} - \bbetaks{k}}$. For any $\bm{r} = \{r_k\}_{k=0}^{K} \in \{\pm 1\}^{K+1}$ which is a permutation order of $\{\hbeta^{(k)[0]}\}_{k=0}^K$ and its corresponding alignment $\{r_k\hbeta^{(k)[0]}\}_{k=0}^K$, define its alignment score as
\begin{equation}
	\text{score}(\bm{r}) = \sum_{0 \leq k_1 \neq k_2 \leq K}\twonorm{r_{k_1}\hbeta^{(k_1)[0]} - r_{k_2}\hbeta^{(k_2)[0]}}.
\end{equation}

\begin{algorithm}[!h]
\caption{Alignment for transfer learning}
\label{algo: alignment for TL}
\KwIn{Initialization $\{(\hbeta^{(k)[0]})\}_{k=0}^K$, and $\widehat{\bm{r}}$ from Algorithm \ref{algo: exhaustive alignment} or \ref{algo: greedy alignment}}
\uIf{$\textup{score}((-1, \widehat{\bm{r}})) > \textup{score}((1, \widehat{\bm{r}}))$}{
	$\widehat{\bm{r}}' = (1, \widehat{\bm{r}})$\\
}
\Else{$\widehat{\bm{r}}' = (-1, \widehat{\bm{r}})$}
\KwOut{$\widehat{\bm{r}}'$}
\end{algorithm}

As expected, under the conditions from Algorithms \ref{algo: exhaustive alignment} or \ref{algo: greedy alignment} for sources together with some similar conditions on the target, Algorithm \ref{algo: alignment for TL} will output the ideal alignment $\widehat{\bm{r}}'$ (equivalently, the good initialization $\widehat{r}_0'\hbeta^{(0)[0]}$ for Algorithm \ref{algo: transfer}). 

\begin{theorem}[Alignment correctness for Algorithm \ref{algo: alignment for TL}]\label{thm: tl alignment}
	Assume that
	\begin{enumerate}[(i)]
		\item $\epsilon < \frac{1}{2}$;
		\item $\twonorm{\bbetaks{0}} > \frac{2(1-\epsilon)}{1-2\epsilon}h + \frac{2-\epsilon}{1-2\epsilon}\max_{k \in \{0\}\cup S}\big(\twonorm{\hbeta^{(k)[0]} - \bbetaks{k}} \wedge \twonorm{\hbeta^{(k)[0]} + \bbetaks{k}}\big)$,
	\end{enumerate}
	where $\epsilon = \frac{K-s}{K}$ is the outlier source task proportion, and $h$ is the degree of discriminant coefficient relatedness defined in \eqref{eq: parameter space tl}.
	
	For $\widehat{\bm{r}}$ in Algorithm \ref{algo: alignment for TL}: if it is from Algorithm \ref{algo: exhaustive alignment}, assume the conditions of Theorem \ref{thm: brute force alignment} hold; if it is from Algorithm \ref{algo: greedy alignment}, assume the conditions of Theorem \ref{thm: greedy alignment} hold. Then the output of Algorithm \ref{algo: alignment for TL} satisfies
	\begin{equation}
		\widehat{\bm{r}}'_k = r_k^* \text{ for all } k \in \{0\} \cup S
		\quad \text{ or }\quad 
		\widehat{\bm{r}}'_k = -r_k^* \text{ for all } k \in \{0\} \cup S.
	\end{equation}
\end{theorem}

\section{Additional Numerical Studies}\label{sec: numerical supp}
In this section, we present results from additional numerical studies, including supplementary results from the simulation study and the real-data study in Section \ref{sec: numerical} of the main text. Additionally, we provide results from two new MTL simulations, one TL simulation, explorations of different penalty parameters, and another real-data study.

Before presenting the results, we would like to clarify the relationship between MTL-GMM and TL-GMM and explain why we do not include TL-GMM and Target-GMM in the multi-task learning (MTL) simulations. MTL-GMM and TL-GMM are two algorithmic variants derived from the same underlying framework, tailored to different learning contexts. In the multi-task setting, TL-GMM can be applied to each task individually, and we expect its performance to closely mirror that of MTL-GMM due to the strong similarity in their aggregation steps. For instance, Step 9 of MTL-GMM (Algorithm 1) can be interpreted as a two-step procedure: a global estimator $\hobeta$ is first computed using all tasks, and then a task-specific estimator $\hbeta^{(k)}$ is obtained by solving $\argmin_{\bbeta}\{n_k[\frac{1}{2}\bbeta^\top \hSigma^{(k)[t]}\bbeta - \bbeta^\top (\hmu^{(k)[t]}_2 - \hmu^{(k)[t]}_1) + \sqrt{n_k}\lambda^{[t]}\twonorm{\bbeta - \hobeta}\}$, where the second step is structurally similar to Step 6 of TL-GMM (Algorithm 7). 

TL-GMM may be more appropriate in scenarios where we aim to generalize to a new task without retraining the entire system. It also avoids transferring any data or intermediate estimators from the target to the sources, which can be beneficial for privacy considerations. Another situation where TL-GMM may be preferred is when the source tasks are highly similar but the target task deviates significantly from them. In such cases, the parameter tuning process in MTL-GMM -- driven by the model fitting performance across all tasks -- can be biased toward the majority (source) tasks, potentially hurting performance on the target. In contrast, TL-GMM tunes its parameters specifically for the target task. We explore one such setting in Section \ref{subsubsec: simulation tl}.

Moreover, Single-task GMM and Target-GMM are essentially the same algorithm applied in different contexts. Both of them refer to fitting GMMs independently using the EM algorithm on each task's individual dataset. As a result, they should yield identical performance.

Given the close similarity between MTL-GMM and TL-GMM, and the equivalence of Single-task GMM and Target-GMM, we chose not to include TL-GMM and Target-GMM in the MTL numerical comparisons.

\subsection{Simulations}

\subsubsection{Simulation 1 of MTL}\label{subsubsec: simulation mtl_1 app}
\begin{figure}[!h]
	\centering
	\includegraphics[width=\linewidth]{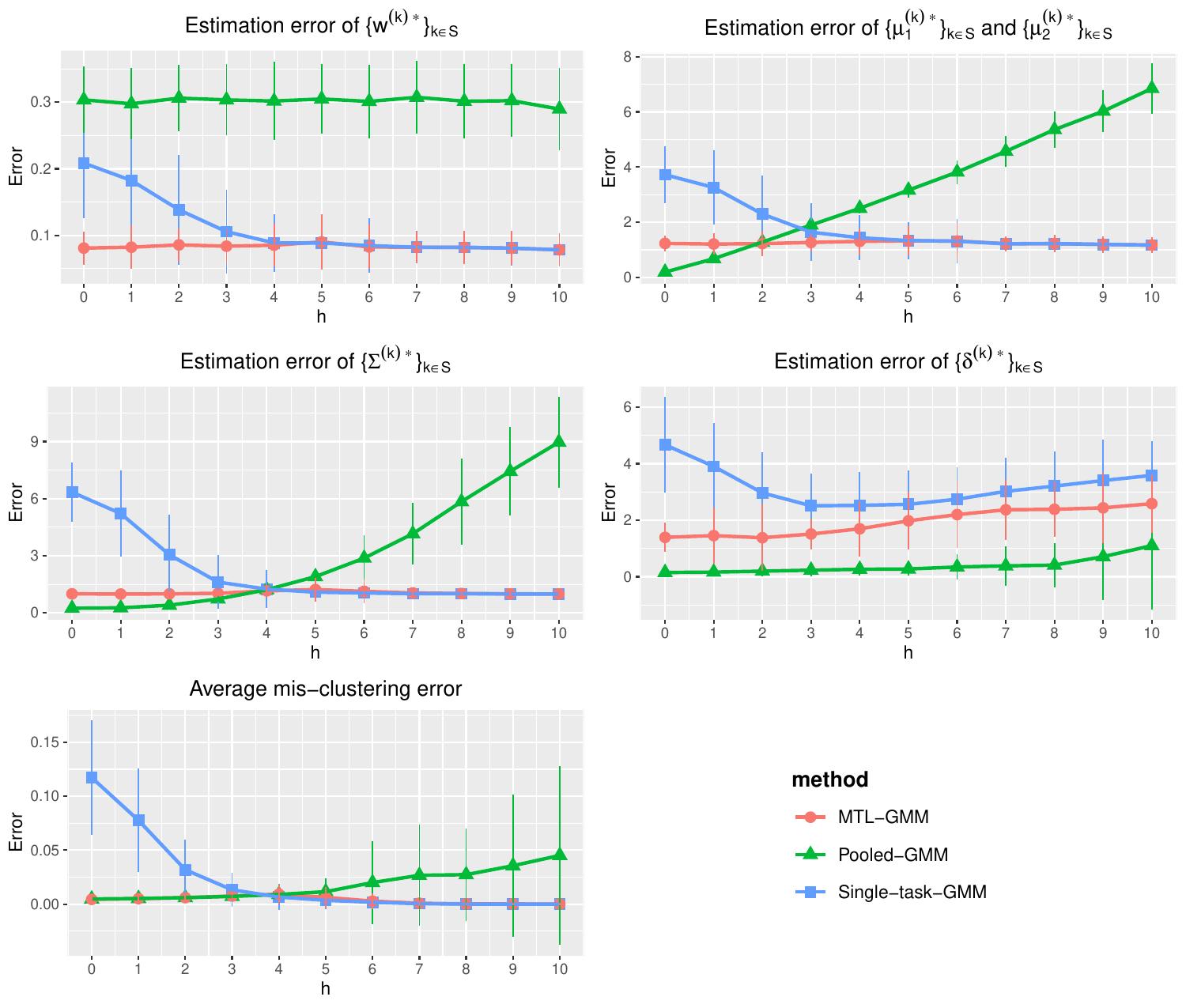}
	\caption{The performance of different methods in Simulation 1.(\rom{1}) of multi-task learning, with no outlier tasks ($\epsilon = 0$), and $h$ changing from 0 to 10 with increment 1. Estimation error of $\{\wks{k}\}_{k \in S}$ stands for $\max_{k \in S}(\norm{\hw^{(k)[T]} - \wks{k}} \wedge \norm{1-\hw^{(k)[T]} - \wks{k}})$. Estimation error of $\{\bmuks{k}_1\}_{k \in S}$ and $\{\bmuks{k}_2\}_{k \in S}$ stands for $\max_{k \in S}\min_{\pi:[2] \rightarrow [2]}(\twonorm{\hmu^{(k)[T]}_1 - \bmuks{k}_{\pi(1)}} \vee \twonorm{\hmu^{(k)[T]}_2 - \bmuks{k}_{\pi(2)}})$. Estimation error of $\{\bSigmaks{k}\}_{k \in S}$ stands for $\max_{k \in S}\twonorm{\hSigma^{(k)[T]} - \bSigmaks{k}}$. Estimation error of $\{\deltaks{k}\}_{k \in S}$ stands for $\max_{k \in S}(\norm{\hdelta^{(k)[T]} - \deltaks{k}} \wedge \norm{\hdelta^{(k)[T]} + \deltaks{k}})$. Average mis-clustering error represents the average empirical mis-clustering error rate calculated on the test set of tasks in $S$.}
	\label{fig: simulation 1.(i) app}
\end{figure}

\begin{figure}[!h]
	\centering
	\includegraphics[width=\linewidth]{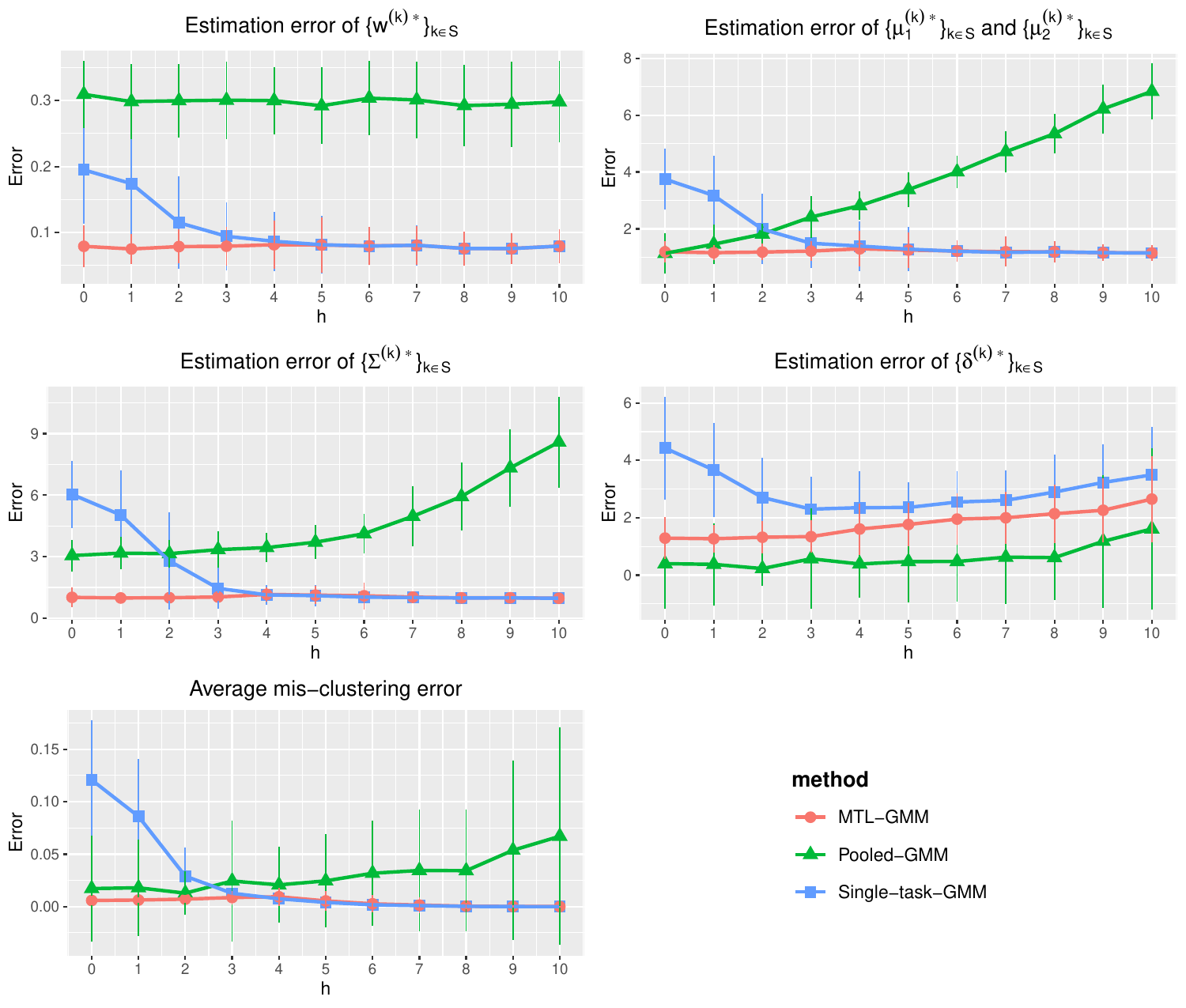}
	\caption{The performance of different methods in Simulation 1.(\rom{2}) of multi-task learning, with 2 outlier tasks ($\epsilon = 0.2$), and $h$ changing from 0 to 10 with increment 1. The meaning of each subfigure's title is the same as in Figure \ref{fig: simulation 1.(i) app}.}
	\label{fig: simulation 1.(ii) app}
\end{figure}

In this subsection, we provide additional performance evaluations for the three methods (MTL-GMM, Pooled-GMM, and Single-task-GMM) in the simulation presented in the main text (referred to as Simulation 1). The results are displayed in Figures \ref{fig: simulation 1.(i) app} and \ref{fig: simulation 1.(ii) app}.

Referring to Figure \ref{fig: simulation 1.(i) app} for the case without outlier tasks, MTL-GMM outperforms Pooled-GMM in estimating $\wks{k}$ all the time. This makes sense because Pooled-GMM does not take the heterogeneity of $\wks{k}$'s into account. For the estimation of other parameters (except $\deltaks{k}$ \footnote{Actually it is not surprising to see Pooled-GMM estimates $\bmuks{k}_1$, $\bmuks{k}_2$, $\deltaks{k}$, and $\bSigmaks{k}$ better than MTL-GMM when $h$ is small in this example. The reason is that these parameters are similar to each other (although MTL-GMM does not rely on this similarity) which makes pooling the data a good approach.}) and clustering, MTL-GMM and Pooled-GMM are competitive when $h$ is small (i.e. the tasks are similar). As $h$ increases (i.e. tasks become more heterogenous), MTL-GMM starts to outperform Pooled-GMM by a large margin. Moreover, MTL-GMM is significantly better than Single-task-GMM in terms of both estimation and mis-clustering errors over a wide range of $h$. They only become comparable when $h$ is very large. These comparisons demonstrate that MTL-GMM not only effectively utilizes the unknown similarity structure among tasks, but also adapts to it.

The results for the case with two outlier tasks are shown in Figure \ref{fig: simulation 1.(ii) app}. It is clear that the comparison between MTL-GMM and Single-task-GMM is similar to the one in Figure \ref{fig: simulation 1.(i) app}. What is new here is that even when $h$ is very small, MTL-GMM still performs much better than Pooled-GMM, showing the robustness of MTL-GMM against a fraction of outlier tasks. Note that in this simulation, $\deltaks{k} = 0$ for all $k \in [K]$, which might explain the phenomenon where Pooled-GMM outperforms MTL-GMM in estimating $\deltaks{k}$'s.

\subsubsection{Simulation 2}\label{subsubsec: simulation mtl 3}
The second simulation is a multi-cluster example, which is built based on Simulation 1. Consider a multi-task learning problem with $K = 10$ tasks, where each task has sample size $n_k=100$ and dimension $p=15$, and follows a GMM with $R = 4$ clusters. For all $k \in [K]$, we generate $(\wks{k}_1, \ldots, \wks{k}_R)$ independently from $\text{Dirichlet}(\bm{\alpha})$ with $\bm{\alpha} = 5\cdot \bm{1}_R$. When $k \in S$, we generate $\bmuks{k}_r$ from $(2\cdot \bm{0}_{2r-2}, 2, 2, \bm{0}_{p-2r})^\top + h/2\cdot (\bSigmaks{k})^{-1}\bm{u}$, where $\bu \sim \text{Unif}(\{\bu \in \mathbb{R}^p: \twonorm{\bm{u}}=1\})$, $\bSigmaks{k} = (0.2^{|i-j|})_{p\times p}$. When $k \notin S$, we generate each $\wks{k}$ from the same Dirichlet distribution and set $\bSigmaks{k} = (0.5^{|i-j|})_{p\times p}$ and $\bmuks{k}_r$ from $\text{Unif}(\{\bu \in \mathbb{R}^p: \twonorm{\bm{u}}=0.5\})$ for $r = 1:R$. For a given $\epsilon \in [0,1)$, in each replication the outlier task index set $S^c$ is uniformly sampled from all subsets of $1:K$ with cardinality $K\epsilon$. We consider two cases:
\begin{enumerate}[(i)]
	\item No outlier tasks ($\epsilon = 0$), and $h$ changes from 0 to 10 with increment 1;
	\item 2 outlier tasks ($\epsilon = 0.2$), and $h$ changes from 0 to 10 with increment 1.
\end{enumerate}

\begin{figure}[!t]
	\centering
	\includegraphics[width=\linewidth]{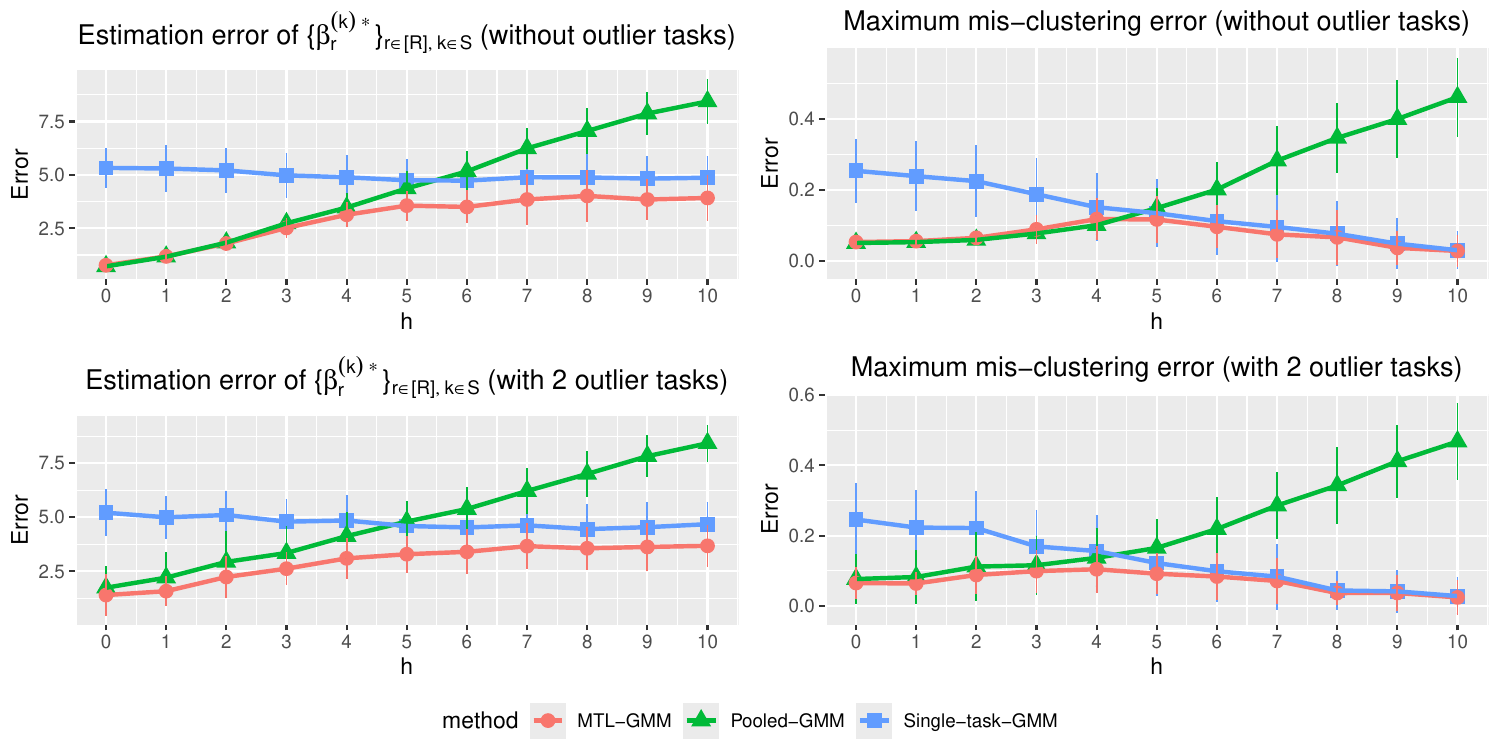}
	\caption{The performance of different methods in Simulation 2 under different outlier proportions. The upper panel shows the performance without outlier tasks ($\epsilon = 0$), and the lower panel shows the performance with 2 outlier tasks ($\epsilon = 0.2$). $h$ changes from 0 to 10 with increment 1. Estimation error of $\{\bbetaks{k}_r\}_{r \in [R], k \in S}$ stands for $\max_{k \in S}\min_{\pi:[R]\rightarrow [R]}\max_{r=1:R}\twonorm{\hbeta^{(k)[T]}_r - (\bSigmaks{k})^{-1}(\bmuks{k}_{\pi(r)}-\bmuks{k}_{\pi(1)})}$ and maximum mis-clustering error represents the maximum empirical mis-clustering error rate calculated on the test set of tasks in $S$.}
	\label{fig: simulation 3}
\end{figure}

Algorithm \ref{algo: multitask multi-component} is run with the alignment Algorithm \ref{algo: greedy alignment multi component}. The results of it and other benchmarks are reported in Figures \ref{fig: simulation 3}. The main message is the same as in Simulation 1: Pooled-GMM is sensitive to outlier tasks and suffers from negative transfer when $h$ is large, while MTL-GMM is robust to outliers and can adapt to the unknown similarity level $h$. Note that in this example, $\{\bmuks{k}_r\}_{k \in S}$ are similar, and $\{\bSigmaks{k}\}_{k \in S}$ are the same, therefore running the EM algorithm by pooling all the data when $h$ is small without outliers may be more effective than our MTL algorithm. This could explain why MTL-GMM performs slightly worse than Pooled-GMM in terms of maximum mis-clustering error when $h$ is small and $\epsilon = 0$.

\begin{figure}[!h]
	\centering
	\includegraphics[width=\linewidth]{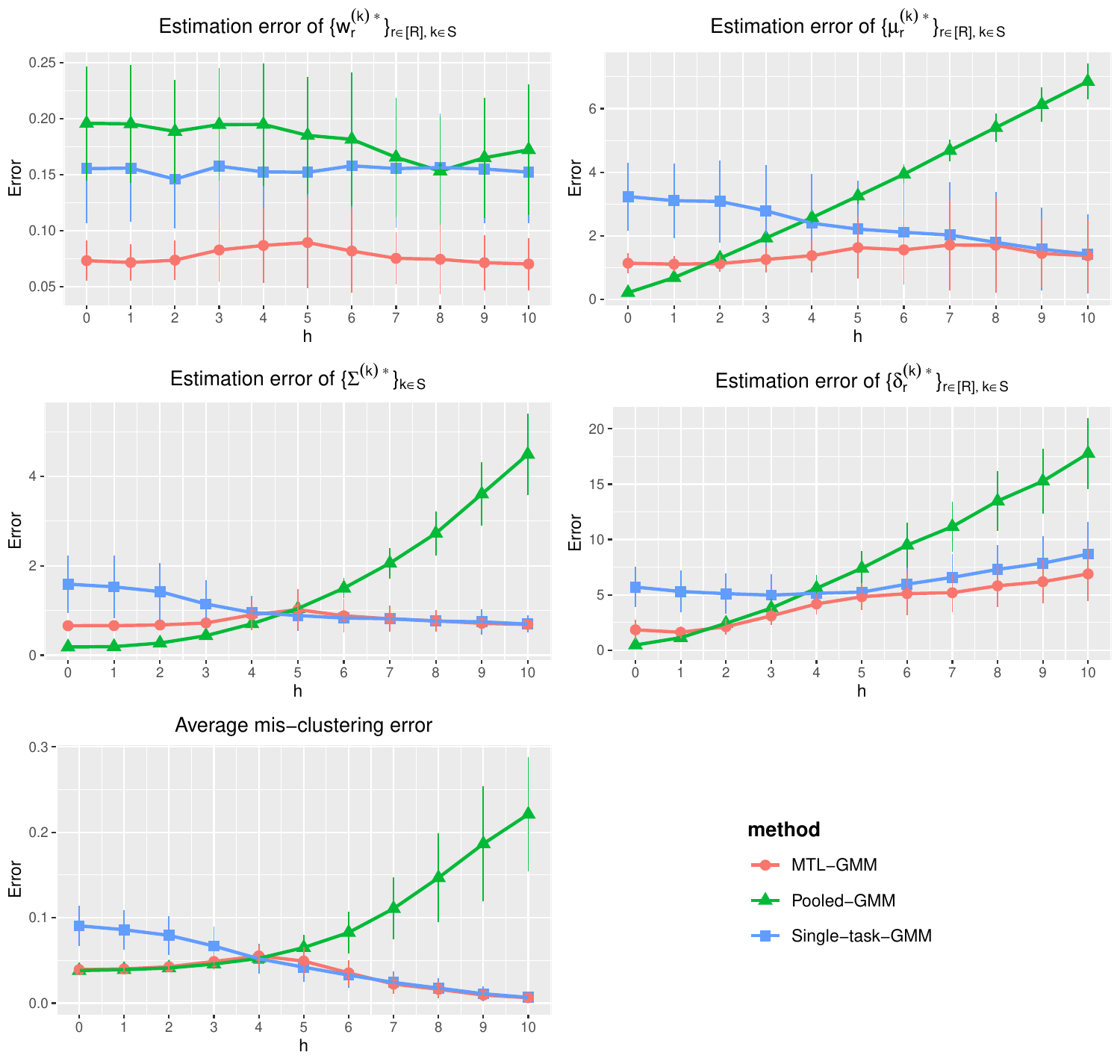}
	\caption{The performance of different methods in Simulation 2.(\rom{1}) of multi-task learning, with no outlier tasks ($\epsilon = 0$), and $h$ changing from 0 to 10 with increment 1. Estimation error of $\{\wks{k}_r\}_{r \in [R], k \in S}$ stands for $\max_{k \in S}\min_{\pi:[R] \rightarrow [R]}\max_{r \in [R]}\norm{\hw^{(k)[T]}_r - \wks{k}_{\pi(r)}}$. Estimation error of $\{\bmuks{k}_r\}_{r \in [R], k \in S}$ stands for $\max_{k \in S}\min_{\pi:[R] \rightarrow [R]}\max_{r \in [R]}\twonorm{\hmu^{(k)[T]}_r - \bmuks{k}_{\pi(r)}}$. Estimation error of $\{\bSigmaks{k}\}_{k \in S}$ stands for $\max_{k \in S}\twonorm{\hSigma^{(k)[T]} - \bSigmaks{k}}$. Estimation error of $\{\deltaks{k}_r\}_{r \in [R], k \in S}$ stands for $\max_{k \in S}\min_{\pi:[R] \rightarrow [R]}\max_{r \in [R]}\norm{\hdelta^{(k)[T]}_r - (\bmuks{k}_{\pi(r)} + \bmuks{k}_{\pi(1)})^\top(\bSigmaks{k})^{-1}(\bmuks{k}_{\pi(r)} - \bmuks{k}_{\pi(1)})/2}$. Average mis-clustering error represents the average empirical mis-clustering error rate calculated on the test set of tasks in $S$.}
	\label{fig: simulation 3.(i) app}
\end{figure}

\begin{figure}[!h]
	\centering
	\includegraphics[width=\linewidth]{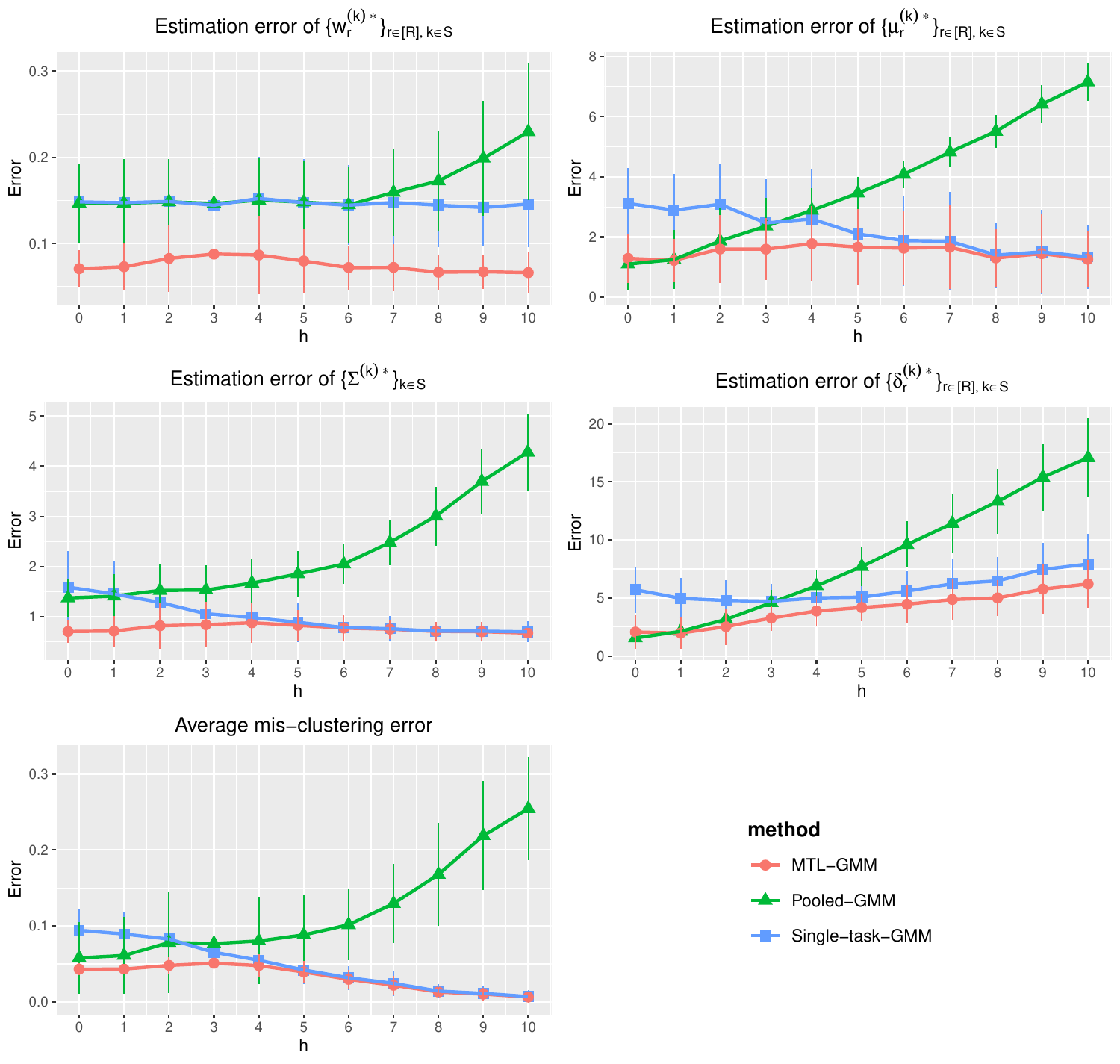}
	\caption{The performance of different methods in Simulation 2.(\rom{2}) of multi-task learning, with 2 outlier tasks ($\epsilon = 0.2$), and $h$ changing from 0 to 10 with increment 1. The meaning of each subfigure's title is the same as in Figure \ref{fig: simulation 3.(i) app}.}
	\label{fig: simulation 3.(ii) app}
\end{figure}

We also provide additional performance evaluations for the three methods in Simulation 2. The results are presented in Figures \ref{fig: simulation 3.(i) app} and \ref{fig: simulation 3.(ii) app}. The main takeaway is the same as in the previous simulation example: Pooled-GMM is sensitive to outlier tasks and suffers from negative transfer when $h$ is large, while MTL-GMM is robust to outliers and can adapt to the unknown similarity level $h$. The results verify the theoretical findings in the multi-cluster case.

\subsubsection{Simulation 3 of MTL}\label{subsubsec: simulation mtl 2}
In the third simulation of MTL, we consider a different similarity structure among tasks in $S$ and a different type of outlier tasks. For a multi-task learning problem with $K = 10$ tasks, set the sample size of each task equal to 100. Let $\bbetaks{1} = (2.5, 0,0,0,0)$, $\bSigmaks{1} = (0.5^{|i-j|})_{5 \times 5}$, and $1 \in S$, i.e., the first task is not an outlier task. We generate each $\wks{k}$ from $\text{Unif}(0.1, 0.9)$ for all $k \in S$. For $k \in S \backslash \{1\}$, we generate $\bSigmaks{k}$  as
\begin{equation}\label{eq: Sigma generation}
	\bSigmaks{k} = \begin{cases}
		(0.5^{|i-j|})_{5 \times 5}, &\text{with probability }1/2, \\
		(a^{|i-j|})_{5 \times 5}, &\text{with probability }1/2,
	\end{cases}
\end{equation}
and set $\bbetaks{k} = (\bSigmaks{k})^{-1}\bSigmaks{1}\bbetaks{1}$. Here, the value of $a$ is determined by $\max\{a \in [0.5, 1): \twonorm{\bbetaks{k}-\bbetaks{1}} \leq h\}$ for a given $h$. Let $\bmuks{k}_2 = \bSigmaks{k}\bbetaks{k}$ and $\bmuks{k}_1 = \bm{0}$, $\forall k \in S$. In this generation process, $\bmuks{k}_2 = \bmuks{1}_2 = \bSigmaks{1}\bbetaks{1} = (5/2, 5/4, 5/8, 5/16, 5/32)^\top$ for all $k \in S$. The covariance matrix of tasks in $S$ can differ. When $k \notin S$, we generate the data of task $k$ from two clusters with probability $1-\wks{k}$ and $\wks{k}$, where $\wks{k} \sim \text{Unif}(0.1,0.9)$. Samples from the second cluster follow $N(\bmuks{k}_2, \bSigmaks{k})$, with $\bSigmaks{k}$ coming from \eqref{eq: Sigma generation}, $\bbetaks{k} = (-2.5, -2.5, -2.5, -2.5, -2.5)^\top$, and $\bmuks{k}_2= \bSigmaks{k}\bbetaks{k}$. For each sample from the first cluster, each component is independently generated from a $t$-distribution with degrees of freedom $4$. In each replication, for given $\epsilon$, the outlier task index set $S^c$ is uniformly sampled from all subsets of $2:K$ with cardinality $K\epsilon$ (since task 1 has been fixed in $S$). We consider two cases:
\begin{enumerate}[(i)]
	\item No outlier tasks ($\epsilon = 0$), and $h$ changes from 0 to 10 with increment 1;
	\item 2 outlier tasks ($\epsilon = 0.2$), and $h$ changes from 0 to 10 with increment 1.
\end{enumerate}

\begin{figure}[!h]
	\centering
	\includegraphics[width=\linewidth]{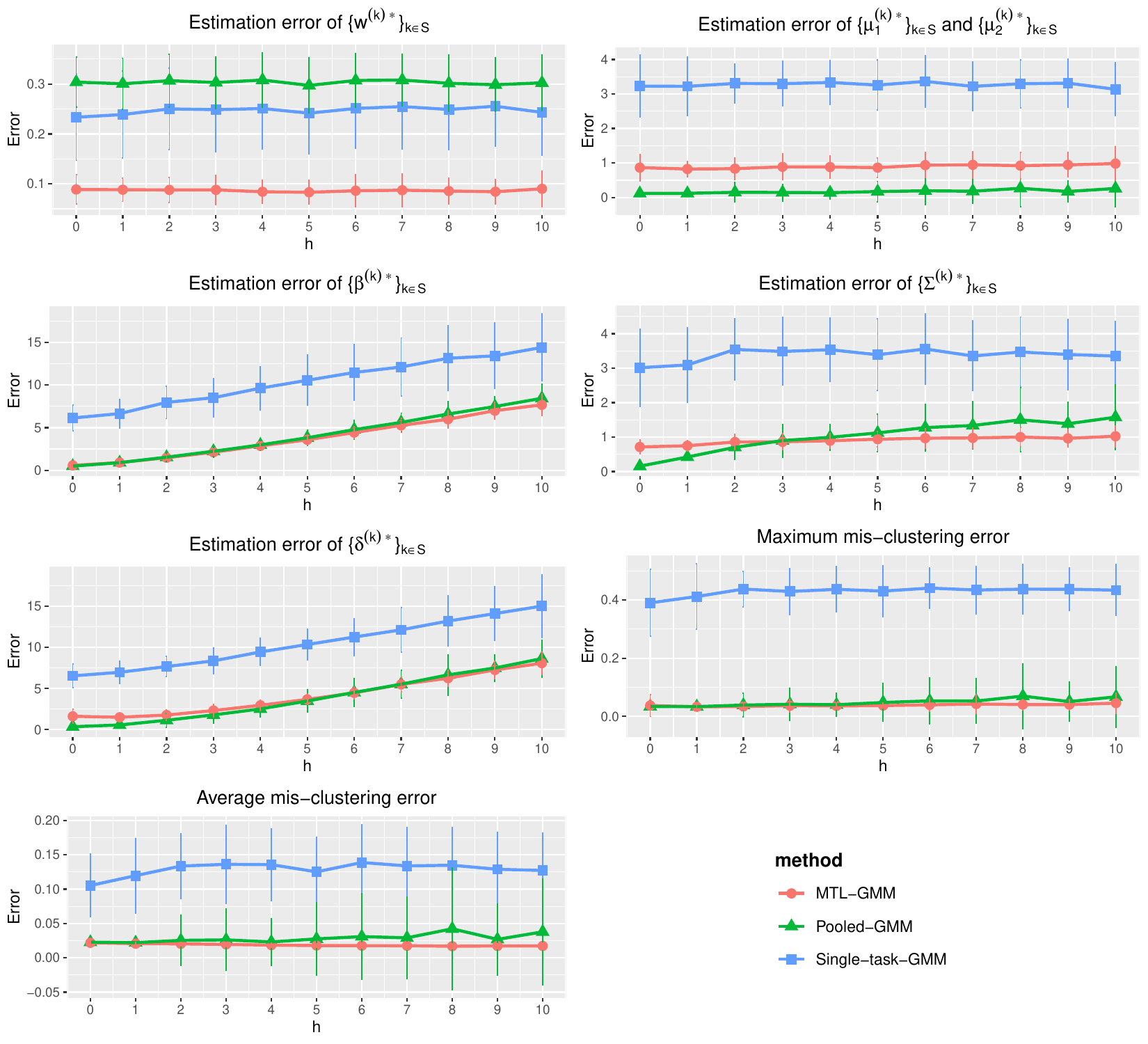}
	\caption{The performance of different methods in Simulation 3.(\rom{1}) of multi-task learning, with no outlier tasks ($\epsilon = 0$), and $h$ changing from 0 to 10 with increment 1. Estimation error of $\{\bbetaks{k}\}_{k \in S}$ stands for $\max_{k \in S}(\twonorm{\hbeta^{(k)[T]} - \bbetaks{k}} \wedge \twonorm{\hbeta^{(k)[T]} + \bbetaks{k}})$. Maximum mis-clustering error represents maximum empirical mis-clustering error rate calculated on test set of tasks in $S$. The meaning of other subfigures' titles is the same as in Figure \ref{fig: simulation 1.(i) app}.}
	\label{fig: simulation 2.(i)}
\end{figure}

\begin{figure}[!h]
	\centering
	\includegraphics[width=\linewidth]{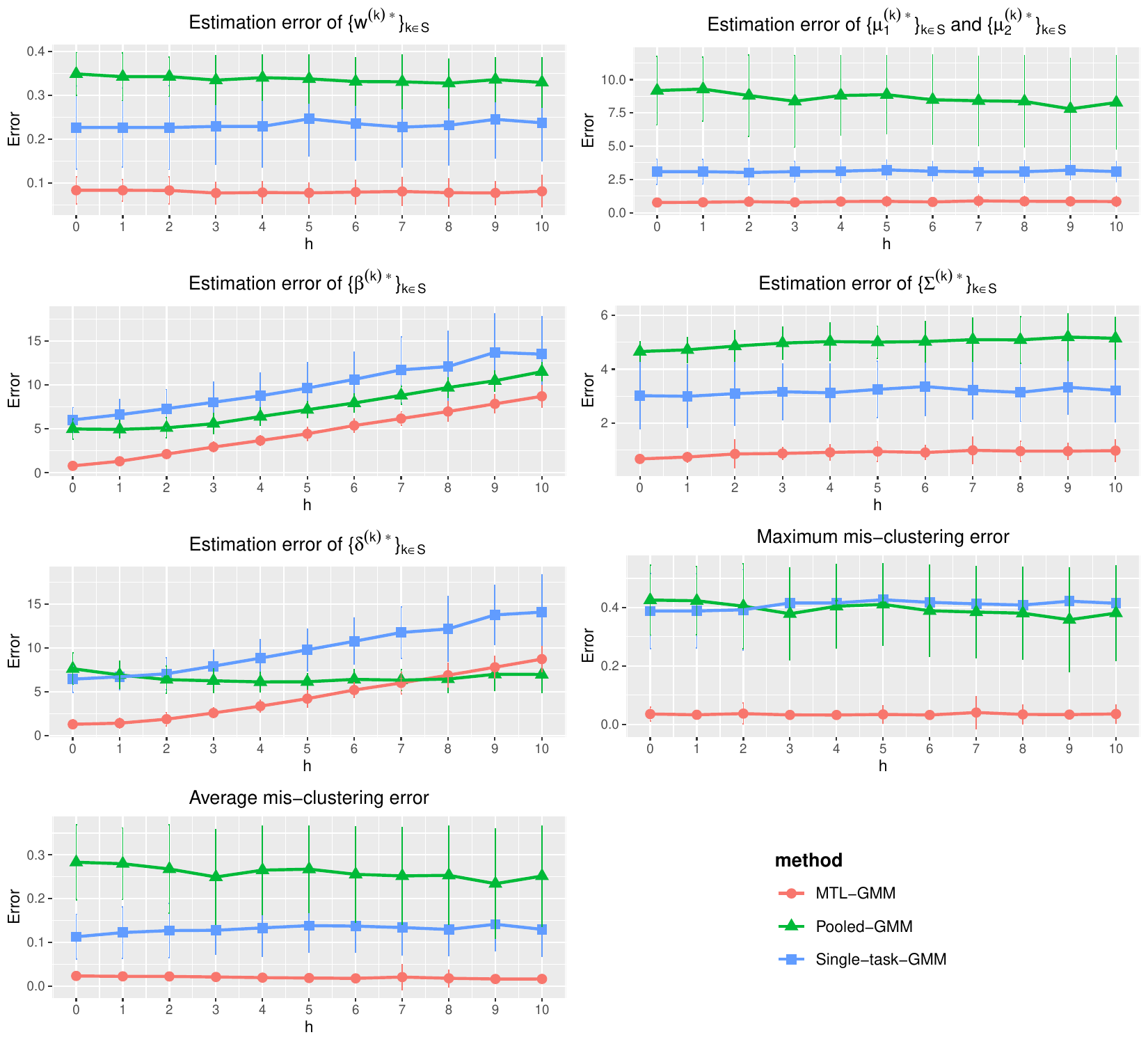}
	\caption{The performance of different methods in Simulation 3.(\rom{2}) of multi-task learning, with 2 outlier tasks ($\epsilon = 0.2$), and $h$ changing from 0 to 10 with increment 1. The meaning of each subfigure's title is the same as in Figure \ref{fig: simulation 2.(i)}.}
	\label{fig: simulation 2.(ii)}
\end{figure}

We implement the same three methods as in Simulation 1 and the results are reported in Figures \ref{fig: simulation 2.(i)} and \ref{fig: simulation 2.(ii)}. When there are no outlier tasks, both MTL-GMM and Pooled-GMM significantly outperform Single-task-GMM. Note that in this simulation, $\bmuks{k}_1 = \bmuks{k'}_1$ and $\bmuks{k}_2 = \bmuks{k'}_2$ for all $k \neq k' \in [K]$, which might explain the phenomenon where Pooled-GMM outperforms MTL-GMM in estimating $\bmuks{k}_1$ and $\bmuks{k}_2$'s. When there are two outlier tasks, Figure \ref{fig: simulation 2.(ii)} shows that Pooled-GMM performs much worse than Single-task-GMM on most of the estimation errors of GMM parameters as well as the mis-clustering error rate. In contrast, MTL-GMM greatly improves the performance of Single-task-GMM, showing the advantage of MTL-GMM when dealing with outlier tasks and heterogeneous covariance matrices. 

\subsubsection{Contamination proportion $\epsilon$}\label{subsubsec: varying epsilon}
In this subsection, we investigate how varying the contamination proportion $\epsilon$ affects the performance of MTL-GMM. By comparing MTL-GMM with Pooled-GMM and Single-task-GMM, we aim to illustrate the robustness of MTL-GMM and demonstrate that this robustness does not significantly compromise performance.

\begin{figure}[!h]
	\centering
	\includegraphics[width=\linewidth]{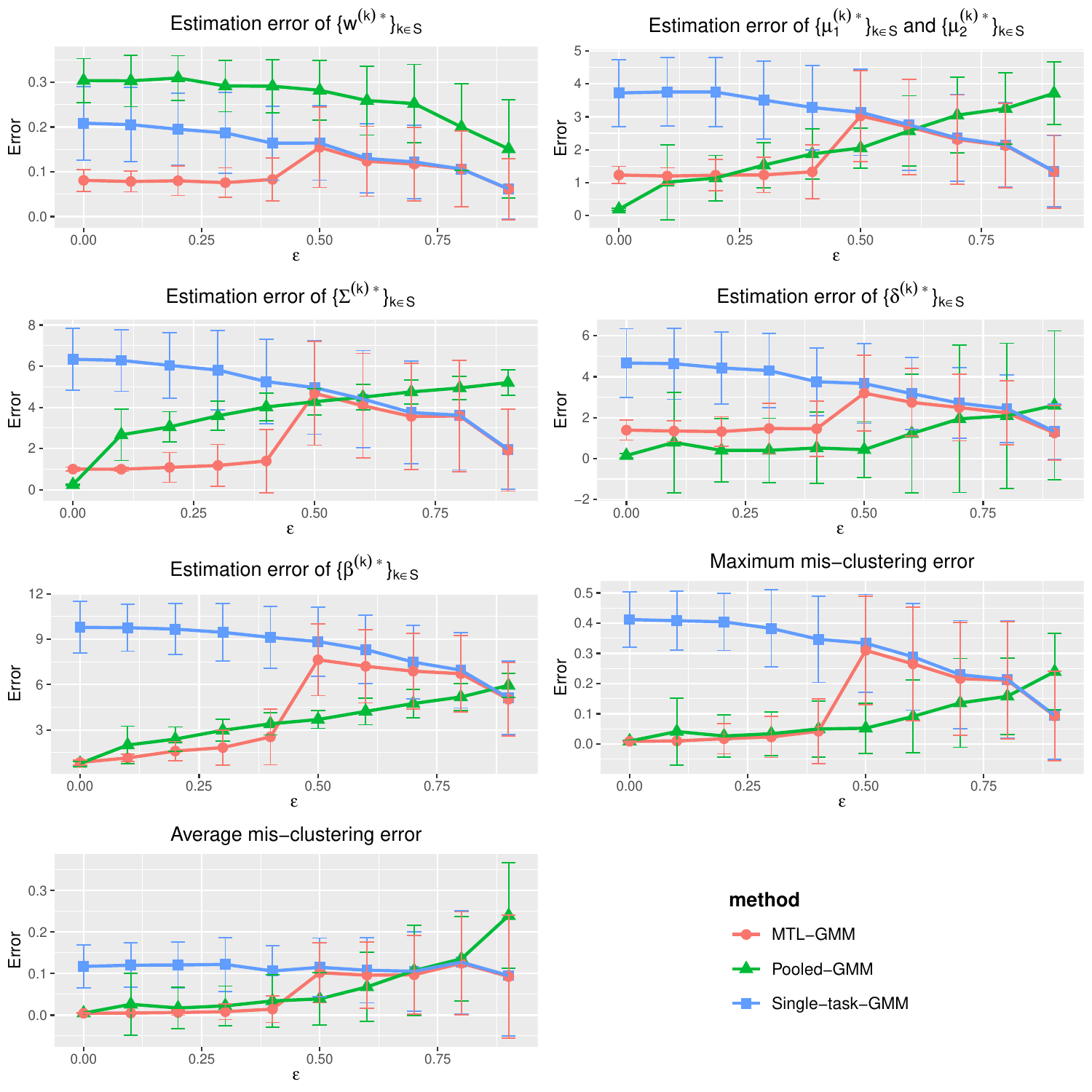}
	\caption{The performance of different methods in Simulation 1 of multi-task learning, with $h = 0$, and $\epsilon$ changing from 0 to 0.9 with increment 0.1. The meaning of each subfigure's title is the same as in Figure \ref{fig: simulation 2.(i)}.}
	\label{fig: mtl_1_epsilon}
\end{figure}

We first consider the setting identical to Simulation 1 in Section \ref{subsec: simulation binary}, using the same data generation mechanism. We set $h = 0$ and vary $\epsilon$ from $0$ to $0.9$ in increments of $0.1$. The results are summarized in Figure \ref{fig: mtl_1_epsilon}. As $\epsilon$ increases, the performance of MTL-GMM deteriorates but consistently remains superior to both Single-task-GMM and Pooled-GMM. Furthermore, when $\epsilon \leq 0.4$, increasing $\epsilon$ has a minimal impact on the performance of MTL-GMM, highlighting its robustness. As $\epsilon$ approaches $1$, MTL-GMM's performance becomes closer to that of Single-task-GMM, effectively avoiding negative transfer.

In general, no algorithm maintains robustness when $\epsilon \geq 0.5$. Nevertheless, due to the aggregation in Step 9 and the strategic integration of local and global updates, MTL-GMM is guaranteed to perform no worse than Single-task-GMM even when the contamination proportion $\epsilon \geq 0.5$. This reflects an important distinction between classical robust statistics and multi-task learning: while in classical robust statistics, the performance of methods can be arbitrarily bad when contamination exceeds $0.5$, whereas in multi-task learning, the baseline performance of single-task learning on uncontaminated tasks remains achievable, as one can always resort to single-task learning individually.

\begin{figure}[!h]
	\centering
	\includegraphics[width=\linewidth]{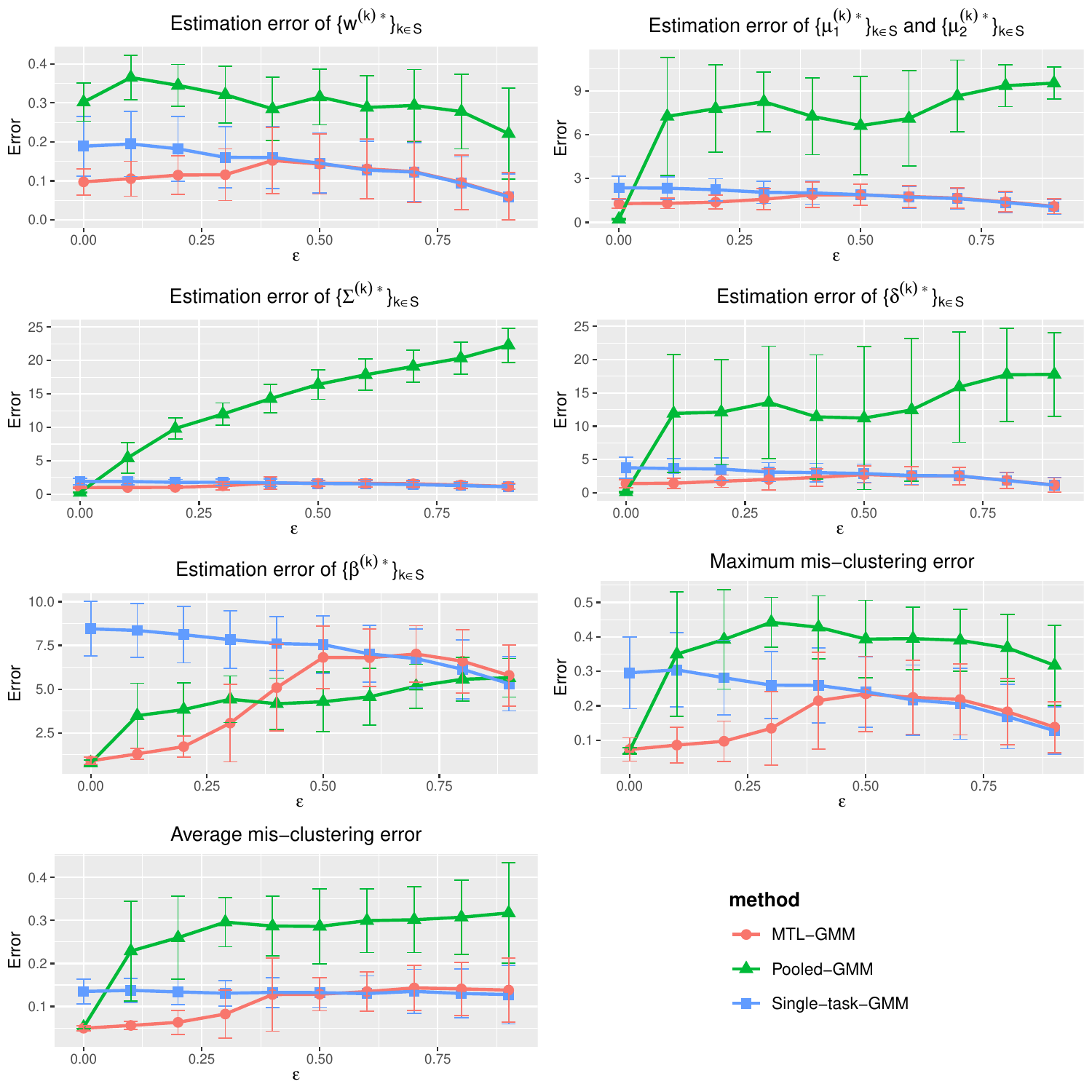}
	\caption{The performance of different methods in a new Simulation 4 described in Section \ref{subsubsec: varying epsilon}, with $\epsilon$ changing from 0 to 0.9 with increment 0.1. The meaning of each subfigure's title is the same as in Figure \ref{fig: simulation 2.(i)}.}
	\label{fig: mtl_epsilon}
\end{figure}

In the previous setting, the GMM means for uncontaminated tasks were sparse, concentrating on a few coordinates, while contaminated tasks had GMM means generated from a sphere. This scenario had limited negative effects on Pooled-GMM when the contamination proportion $\epsilon$ was small. However, this is generally not the case. To further illustrate the robustness of MTL-GMM, we now consider a variant of Simulation 1. Specifically, for tasks with indices $k \in S$, we set $\bmuks{k}_1 = -\bmuks{k}_2 = 2\times \mathds{1}_p/\sqrt{p}$. For contaminated tasks ($k \notin S$), the data still follow a GMM, but we generate $\bmuks{k}_1$ from a uniform distribution on the sphere, i.e., $\text{Unif}({\bu \in \mathbb{R}^p: \twonorm{\bm{u}}=10})$, and set $\bmuks{k}_2 = -\bmuks{k}_1$. We also set the covariance matrix as $\bSigmaks{k} = (0.5^{|i-j|})_{p\times p}$. The remaining settings are identical to Simulation 1. For convenience, we denoted it as Simulation 4. The results are presented in Figure \ref{fig: mtl_epsilon}.

We observe that introducing even a single outlier task ($\epsilon = 0.1$) severely degrades the performance of Pooled-GMM. In contrast, MTL-GMM exhibits substantially more stable performance, consistently outperforming Single-task-GMM when $\epsilon$ is small. As in previous simulation results, when $\epsilon$ approaches $1$, MTL-GMM performance converges to that of Single-task-GMM.

\subsubsection{Varying numbers of clusters}\label{subsubsec: simulation diff R}
In this subsection, we present simulation results to illustrate the extensions discussed in Section \ref{subsec: different cluster numbers}, focusing on scenarios where the number of clusters differs across tasks.

\begin{figure}[!h]
	\centering
	\includegraphics[width=\linewidth]{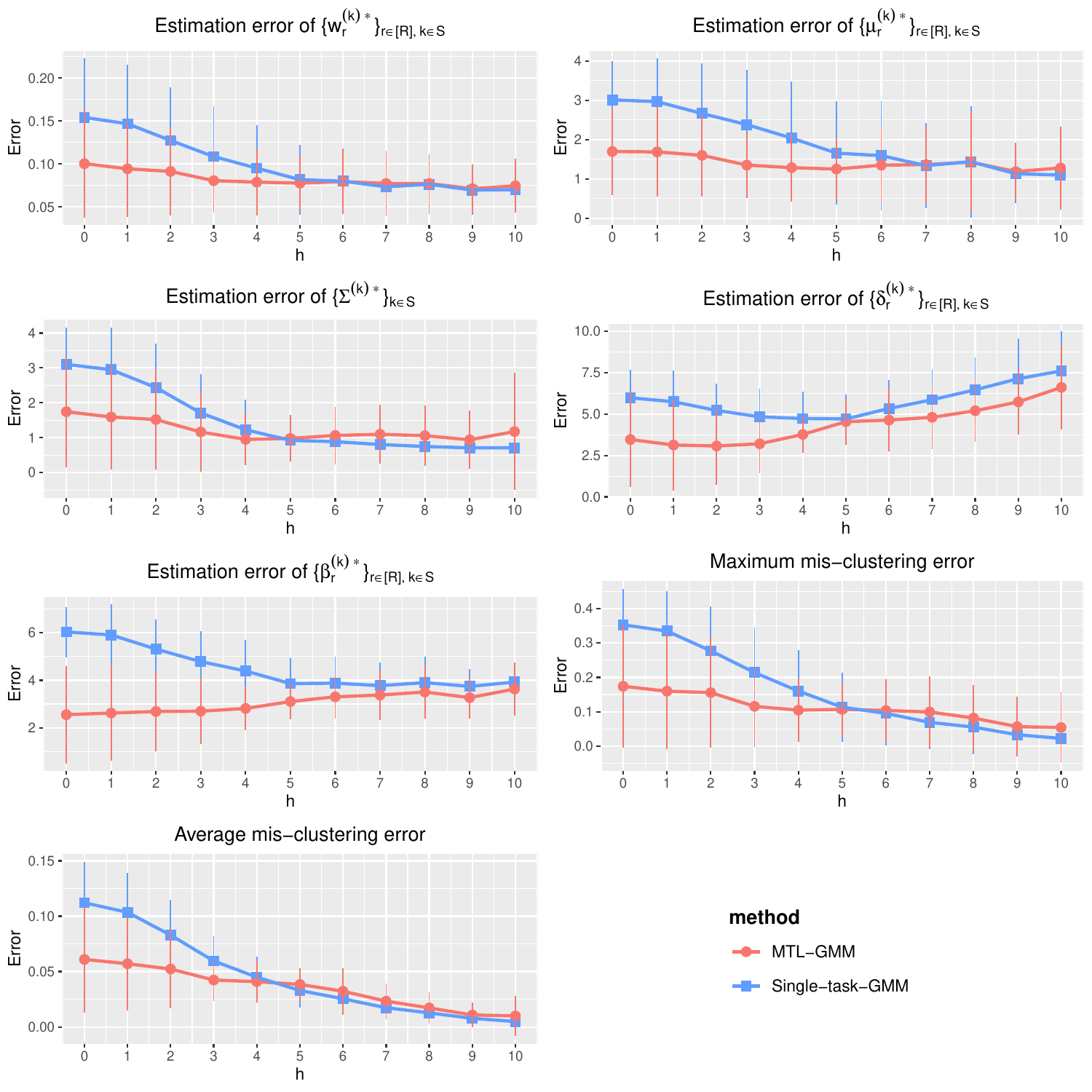}
	\caption{The performance of different methods in Simulation 5 of multi-task learning (where the cluster numbers vary), with no outlier tasks ($\epsilon = 0$), and $h$ changing from 0 to 10 with increment 1. The meaning of each subfigure's title is the same as in Figure \ref{fig: simulation 2.(i)}.}
	\label{fig: mtl_diff_R_0}
\end{figure}

\begin{figure}[!h]
	\centering
	\includegraphics[width=\linewidth]{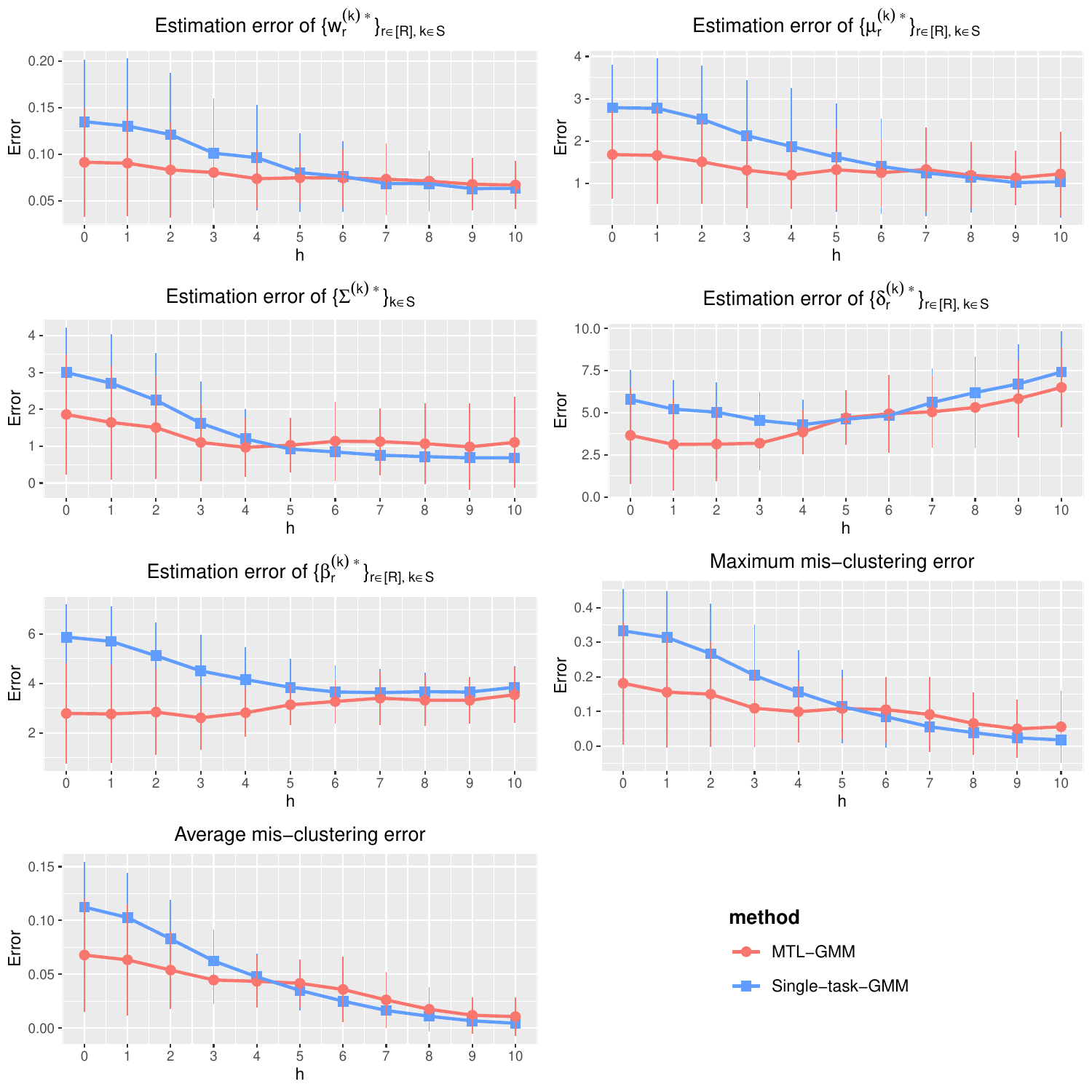}
	\caption{The performance of different methods in Simulation 5 of multi-task learning  (where the cluster numbers vary), with 2 outlier tasks ($\epsilon = 0.2$), and $h$ changing from 0 to 10 with increment 1. The meaning of each subfigure's title is the same as in Figure \ref{fig: simulation 2.(i)}.}
	\label{fig: mtl_diff_R_2}
\end{figure}

Our first simulation setup is related to the direct extension proposed in Section \ref{subsubsec: direct extension diff R} and builds upon Simulation 2 from Section \ref{subsubsec: simulation mtl 3}. Specifically, for each task $k$, the number of clusters $R^{(k)}$ is independently sampled from ${2, 3, 4}$ with probabilities $0.2$, $0.3$, and $0.5$, respectively. Subsequently, the Gaussian mixture model (GMM) data are generated following the same procedures as in Simulation 2, with all other settings kept identical. We refer to this new setting as Simulation 5. The results under scenarios with no outlier tasks ($\epsilon = 0$) and with two outlier tasks ($\epsilon = 0.2$) are shown in Figures \ref{fig: mtl_diff_R_0} and \ref{fig: mtl_diff_R_2}, respectively. Note that we exclude the pooled-data EM algorithm (i.e., Pooled-GMM from previous simulations) here because it inherently requires the number of clusters to be uniform across all tasks, making it inapplicable in this scenario. Applying it in this setting would potentially result in assigning clusters beyond the actual underlying number of clusters present in some tasks.

From the simulation outcomes, we observe that the multi-task GMM (MTL-GMM) performs similarly to its counterpart under uniform cluster numbers. Specifically, MTL-GMM achieves superior performance compared to Single-task-GMM when $h$ is small, and becomes comparable to Single-task-GMM when $h$ is large.

The second simulation we present is related to the representation-learning framework discussed in Section \ref{subsubsec: extension repre diff R}. Specifically, we generate the number of clusters $R^{(k)}$ for each task $k$ from $\textup{Unif}(\{2,3,4,5\})$. A common representation matrix $\bm{A}^* \in \mathbb{R}^{p \times d}$ is constructed as the left singular matrix corresponding to the top $d$ singular values of a $p\times p$ random matrix whose entries are independently drawn from the standard Gaussian distribution. Next, we independently sample $\{\bthetaks{k}_r\}_{r=1}^{R^{(k)}}$ from $3\times \textup{Unif}(\mathcal{S}^{d-1})$ and $\bmuks{k}_r = \bm{A}^*\bthetaks{k}_r$. The mixture proportions $(\wks{k}_1, \ldots, \wks{k}_{R^{(k)}})$ are generated independently from $\text{Dirichlet}(\bm{\alpha})$ with $\bm{\alpha} = 5\times  \bm{1}_{R^{(k)}}$. For each task, the covariance of GMM is set to be $\bSigmaks{k} = (0.2^{|i-j|})_{p \times p}$. We consider $K$ tasks with $n$ observations in each task. For convenience, we denote this setting as Simulation 6. Various configurations of $(K, p, d, n)$ are considered, and the results comparing the Single-task-GMM with the representation learning-based MTL-GMM are summarized in Table \ref{table: mtl repre diff R result}. These results are based on 200 replications.

Our multi-task clustering algorithm effectively leverages the low-rank structure and outperforms single-task learning, particularly when the intrinsic dimension $d$ is significantly smaller than the full dimension $p$.

\begin{table}[!h]
	\begin{tabular}{c|cc|cc}
	\toprule
	\multirow{2}{*}{$(K, p, d, n)$} & \multicolumn{2}{c|}{Maximum mis-clustering error} & \multicolumn{2}{c}{Average mis-clustering error} \\
	& Single-task-GMM & MTL-GMM & Single-task-GMM & MTL-GMM\\
	\hline
		(20, 20, 3, 200) & 0.562 (0.064) & 0.437 (0.067) & 0.259 (0.041) & 0.162 (0.031) \\ 
(20, 30, 5, 200) & 0.739 (0.02) & 0.531 (0.054) & 0.554 (0.032) & 0.277 (0.039) \\ 
(30, 20, 3, 200) & 0.591 (0.06) & 0.463 (0.059) & 0.262 (0.032) & 0.165 (0.025) \\ 
(30, 20, 5, 200) & 0.522 (0.065) & 0.398 (0.069) & 0.201 (0.031) & 0.119 (0.022) \\
(30, 30, 5, 200) & 0.747 (0.02) & 0.528 (0.058) & 0.549 (0.026) & 0.245 (0.036) \\ 
(50, 20, 5, 200) & 0.551 (0.056) & 0.427 (0.059) & 0.199 (0.025) & 0.117 (0.016) \\ 
(50, 30, 5, 300) & 0.751 (0.016) & 0.495 (0.052) & 0.516 (0.025) & 0.191 (0.026) \\ 
(100, 50, 5, 300) & 0.764 (0.012) & 0.515 (0.046) & 0.555 (0.014) & 0.182 (0.018) \\ 
(100, 50, 10, 300) & 0.762 (0.013) & 0.625 (0.041) & 0.556 (0.015) & 0.293 (0.024) \\ 
	\bottomrule
	\end{tabular}
	\caption{Performance comparison between the Single-task-GMM and the representation learning-based MTL-GMM described in Section \ref{subsubsec: extension repre diff R} for Simulation 6 under different $(K, p, d, n)$ configurations. The maximum and average mis-clustering errors (with standard deviations in parentheses) are calculated on independent test datasets, and their definitions match those provided in Figure \ref{fig: simulation 2.(i)}.}
	\label{table: mtl repre diff R result}
\end{table}

\subsubsection{Simulation of TL}\label{subsubsec: simulation tl}
Consider a transfer learning problem with $K = 10$ source data sets, where all sources are from the same GMM. The setting is modified based on Simulation 1 of MTL. The source and target sample sizes are equal to 100. For each of the source and target task, $\wks{k} \sim \textup{Unif}(0.1, 0.9)$ and $\bmuks{k}_1 = -\bmuks{k}_2 = (2, 2, \bm{0}_{p-2})^\top + h/2\cdot (\bSigmaks{k})^{-1}\bm{u}$, where $p = 15$, $\bSigmaks{k} = (0.2^{|i-j|})_{5\times 5}$, and $\bu \sim \text{Unif}(\{\bu \in \mathbb{R}^p: \twonorm{\bm{u}}=1\})$. Note that $\bu$ is generated independently for the source and target, but the same $\bu$ is used across all source tasks. As a result, there is no heterogeneity among the sources. We consider the case that $h$ changes from 0 to 10 with increment 1.

\begin{figure}[!hp]
	\centering
	\includegraphics[width=\linewidth]{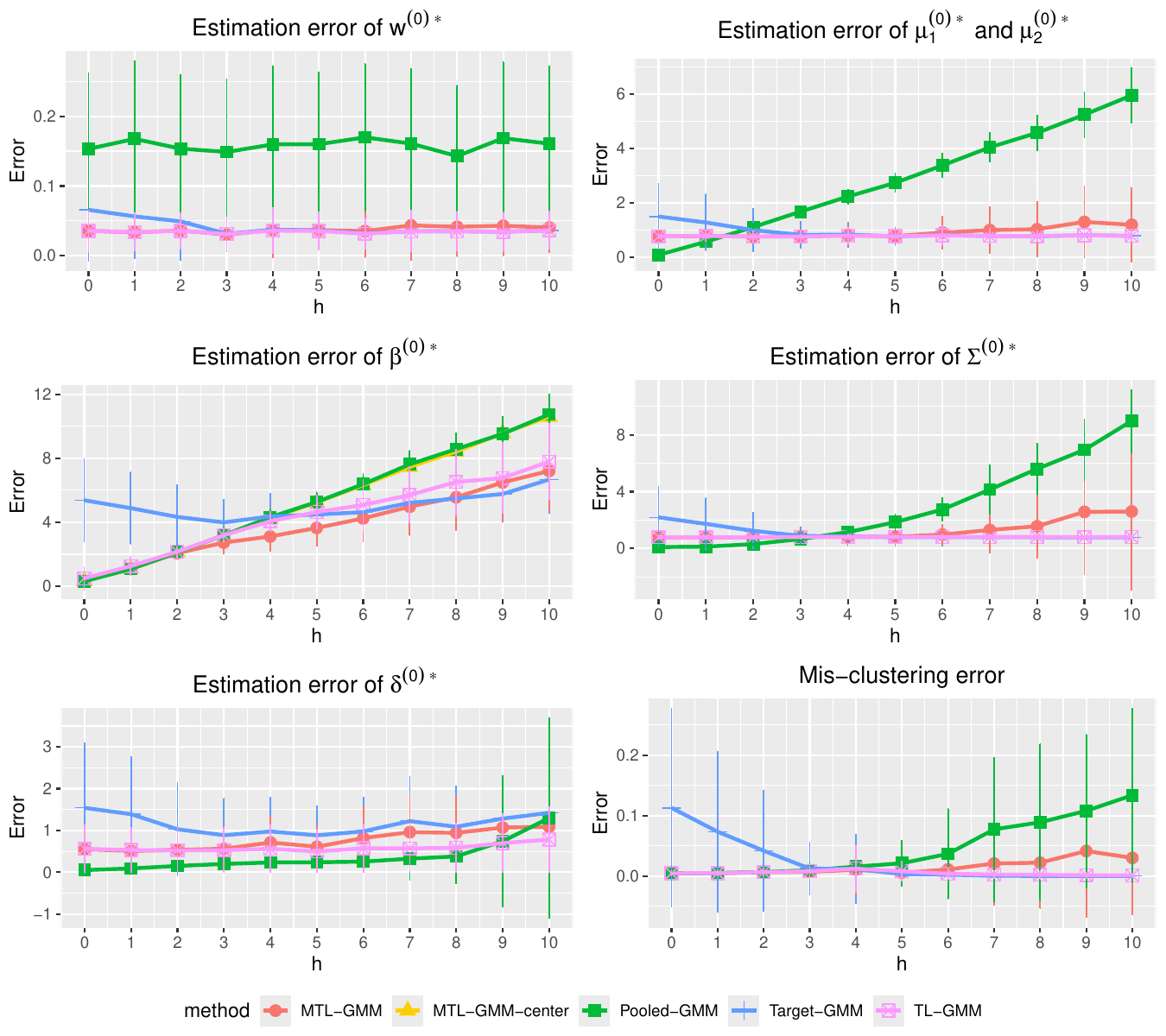}
	\caption{The performance of different methods in the simulation of transfer learning, with no outlier tasks ($\epsilon = 0$) and $h$ changing from 0 to 10 with increment 1. Estimation error of $\wks{0}$ stands for $\norm{\hw^{(0)[T_0]} - \wks{0}} \wedge \norm{1-\hw^{(0)[T_0]} - \wks{0}}$. Estimation error of $\bmuks{0}_1$ and $\bmuks{0}_2$ stands for $\min_{\pi:[2] \rightarrow [2]}\max_{r \in [2]}\twonorm{\hmu^{(0)[T_0]}_r - \bmuks{0}_{\pi(r)}}$. Estimation error of $\bbetaks{0}$ stands for $\twonorm{\hbeta^{(0)[T_0]} - \bbetaks{0}} \wedge \twonorm{\hbeta^{(0)[T_0]} + \bbetaks{0}}$. Estimation error of $\bSigmaks{k}$ stands for $\twonorm{\hSigma^{(0)[T_0]} - \bSigmaks{0}}$. Estimation error of $\deltaks{0}_r$ stands for $\norm{\hdelta^{(0)[T_0]} - \deltaks{0}} \wedge \norm{\hdelta^{(0)[T_0]} + \deltaks{0}}$. Mis-clustering error represents the empirical mis-clustering error rate calculated on the test set of the target data.}
	\label{fig: simulation tl_1_0}
\end{figure}

We compare five different methods, including Target-GMM fitted on target data only, MTL-GMM fitted on all the data, MTL-GMM-center which fits MTL-GMM on source data and outputs the estimated ``center" $\overline{\bbeta}^{[T]}$ as the target estimate \footnote{MTL-GMM-center only appears in the comparison of estimation error of $\bbetaks{k}$'s.}, Pooled-GMM which fits a merged GMM on all the data, and our TL-GMM. The performance is evaluated by the estimation errors of $\wks{0}$, $\bmuks{0}_1$, $\bmuks{0}_2$, $\bbetaks{0}$, $\deltaks{0}$, and $\bSigmaks{0}$ as well as the mis-clustering error rate calculated on an independent test target data of size 500. Results are presented in Figure \ref{fig: simulation tl_1_0}.

Figure \ref{fig: simulation tl_1_0} shows that when $h$ is small, the performances of MTL-GMM, MTL-GMM-center, Pooled-GMM, and TL-GMM are comparable, and all of them are much better than Target-GMM. This is expected, because the sources are very similar to the target and can be easily used to improve the target task learning. As $h$ keeps increasing, the target and sources become increasingly different. This is the phase where the knowledge of sources needs to be carefully transferred for the possible learning improvement on the target task. As is clear from Figure \ref{fig: simulation tl_1_0}, MTL-GMM, MTL-GMM-center, and Pooled-GMM do not handle heterogeneous resources well, thus outperformed by Target-GMM. By contrast, TL-GMM remains effective in transferring source knowledge to improve over Target-GMM; when $h$ is very large so that sources are not useful anymore, TL-GMM is robust enough to still have competitive performance compared to Target-GMM.

Recall that we mentioned at the beginning of Section \ref{sec: numerical supp} that MTL-GMM and TL-GMM are two variants of the same framework designed for different contexts, and thus we generally expect them to behave similarly. Why, then, do we observe a performance gap when $h$ is large? The key reason is that MTL-GMM likely treats the target as an outlier during the tuning of parameters in Step 9. Specifically, the tuning parameter is selected based on a cross-validation procedure that evaluates the average (or trimmed average) performance across all tasks. In this particular setup, since there is no heterogeneity among the source tasks, the tuning parameter is primarily optimized to perform well on the source data, not the target. In contrast, TL-GMM selects its tuning parameter using only the target data, which better adapts it to that task. This discrepancy in tuning procedures explains the observed performance difference. That said, this is a deliberately constructed simulation designed to highlight a scenario where MTL-GMM and TL-GMM can diverge. In most practical settings, their performance should be similar. Moreover, due to its computational efficiency, we recommend using MTL-GMM when the goal is multi-task learning.
 
\begin{figure}[!h]
	\centering
	\includegraphics[width=\textwidth]{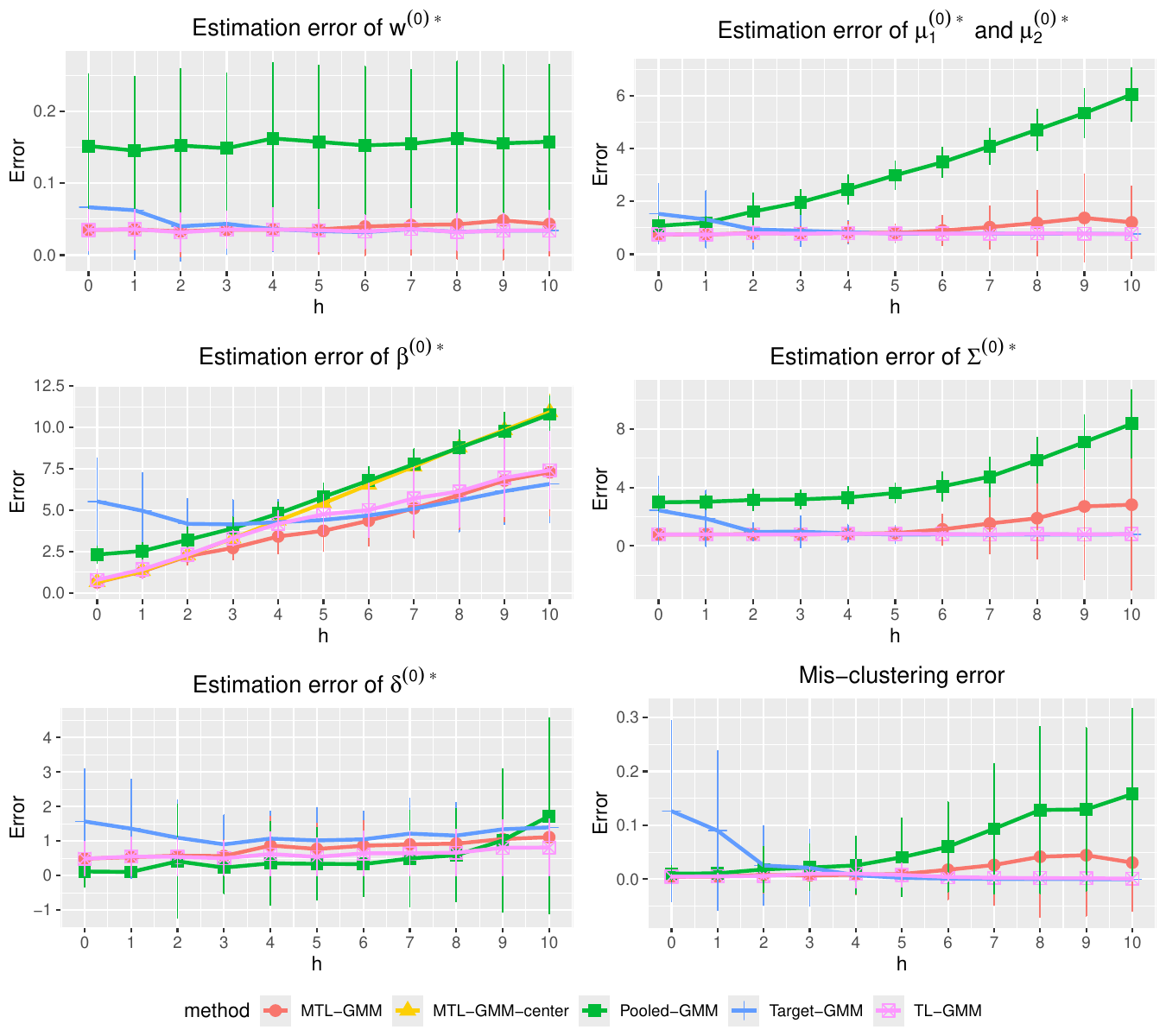}
	\caption{The performance of different methods in the simulation of transfer learning, with 2 outlier tasks ($\epsilon = 0.2$) and $h$ changing from 0 to 10 with increment 1. The meaning of each subfigure's title is the same as in Figure \ref{fig: simulation tl_1_0}.}
	\label{fig: simulation tl_1_2}
\end{figure}

Figure \ref{fig: simulation tl_1_2} shows the results when there are two outlier tasks ($\epsilon = 0.2$). It can be seen that TL-GMM is robust to outliers.

\subsubsection{Tuning parameters $C_{\lambda}$ and $C_{\lambda_0}$ in Algorithms \ref{algo: multitask}, \ref{algo: multitask multi-component}, and \ref{algo: transfer}}\label{subsubsec: simulation mtl 1 tuning}
The candidates of $C_{\lambda}$ and $C_{\lambda_0}$ values used in the 10-fold cross-validation are chosen through a data-driven way. For $C_{\lambda}$ in Algorithm \ref{algo: multitask}, we first determine the smallest $C_{\lambda}$ value which makes all $\bbetak{k}$ estimators identical, which is denoted as $C_{\max}$. Then the $C_{\lambda}$ candidates are set to be the sequence from $C_{\max}/50$ and $2C_{\max}$ with equal logarithm distance. For $C_{\lambda_0}$ in Algorithm \ref{algo: transfer}, we first determine the smallest $C_{\lambda_0}$ value which makes the $\bbetak{0}$ estimator equal to $\obbeta^{[T]}$, which is denoted as $C_{\max}'$. Then the $C_{\lambda_0}$ candidates are set to be the sequence from $C_{\max}'/50$ and $2C_{\max}'$ with equal logarithm distance.

\begin{figure}[!h]
	\centering
	\includegraphics[width=\linewidth]{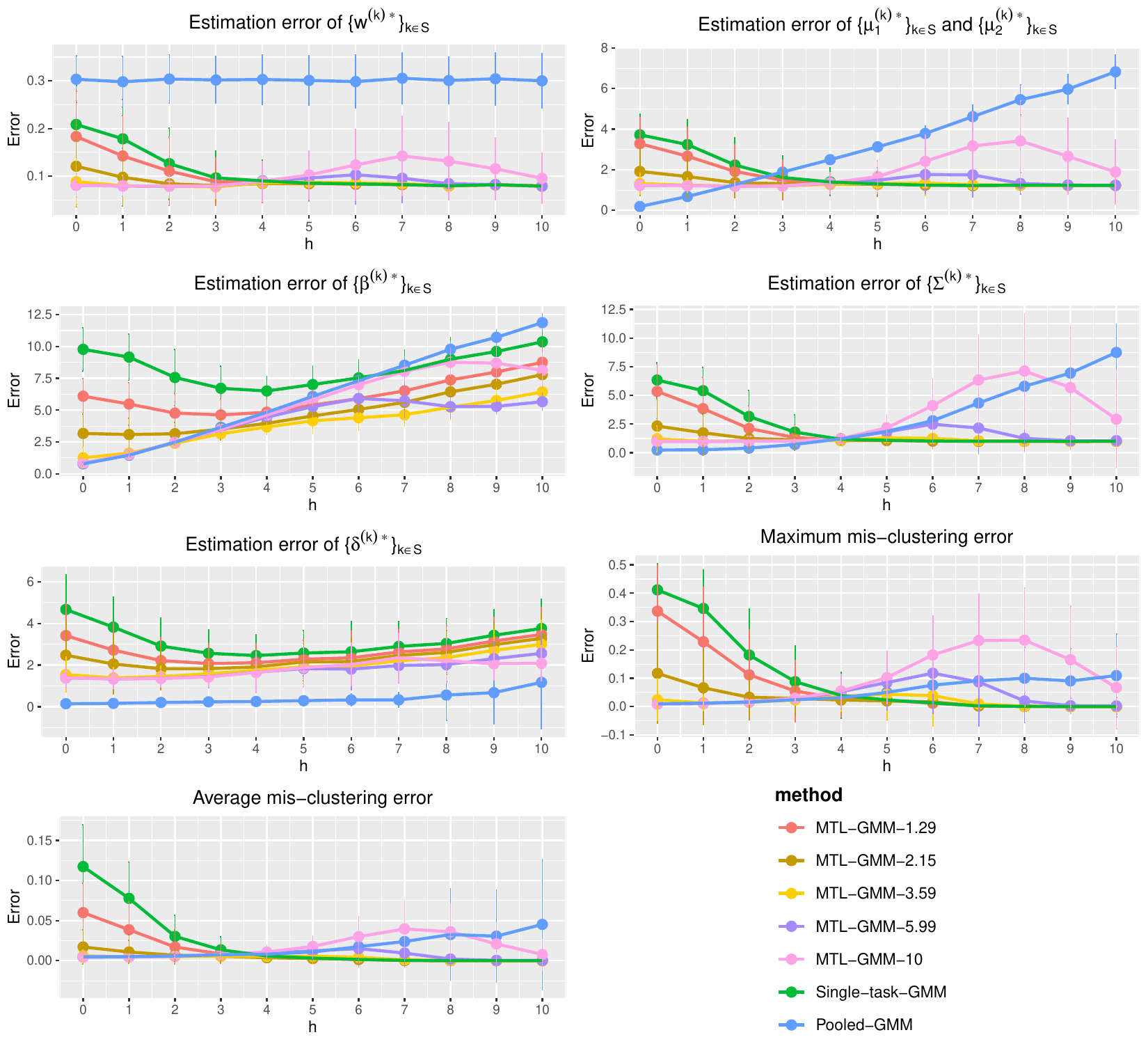}
	\caption{The performance of different methods in Simulation 1 of multi-task learning, with no outlier tasks ($\epsilon = 0$) and $h$ changing from 0 to 10 with increment 1. The meaning of each subfigure's title is the same as in Figures \ref{fig: simulation 1} and \ref{fig: simulation 1.(i) app}.}
	\label{fig: simulation mtl_1_tuning}
\end{figure}

We also run MTL-GMM with different $C_{\lambda}$ values in Simulation 1 to test the impact of the penalty parameter. The results are presented in Figure \ref{fig: simulation mtl_1_tuning}. The values 1.29, 2.15, 3.59, 5.99, and 10 are the last 5 elements in sequence from 0.1 to 10 with equal logarithm distance. It can be seen that with small $C_{\lambda}$ values like 1.29 and 2.15, the performance of MTL-GMM is similar to that of Single-task-GMM, although MTL-GMM-2.15 improves Single-task-GMM a lot when $h$ is small. With large $C_{\lambda}$ values like 5.99 and 10, MTL-GMM performs similarly to Pooled-GMM when $h$ is small while suffering from negative transfer when $h$ is large. However, as $h$ continues to increase, the performance of MTL-GMM with large $C_{\lambda}$ values starts to improve and finally becomes similar to Single-task-GMM. This phenomenon is in accordance with the theory, as the theory predicts that MTL-GMM achieves the same rate as Single-task-GMM for large $h$. The negative transfer effect of MTL-GMM with large $C_{\lambda}$ could be caused by large unknown constants in the upper bound. Comparing Figure \ref{fig: simulation mtl_1_tuning} with figures in Sections \ref{sec: numerical} and \ref{subsubsec: simulation mtl_1 app}, we can see that cross-validation enhances the performance of MTL-GMM.

\subsubsection{Tuning parameter $\kappa$ and $\kappa_0$ in Algorithms \ref{algo: multitask}, \ref{algo: multitask multi-component}, and \ref{algo: transfer}}\label{subsubsec: simulation mtl 1 tuning kappa}
\begin{figure}[!h]
	\centering
	\includegraphics[width=\linewidth]{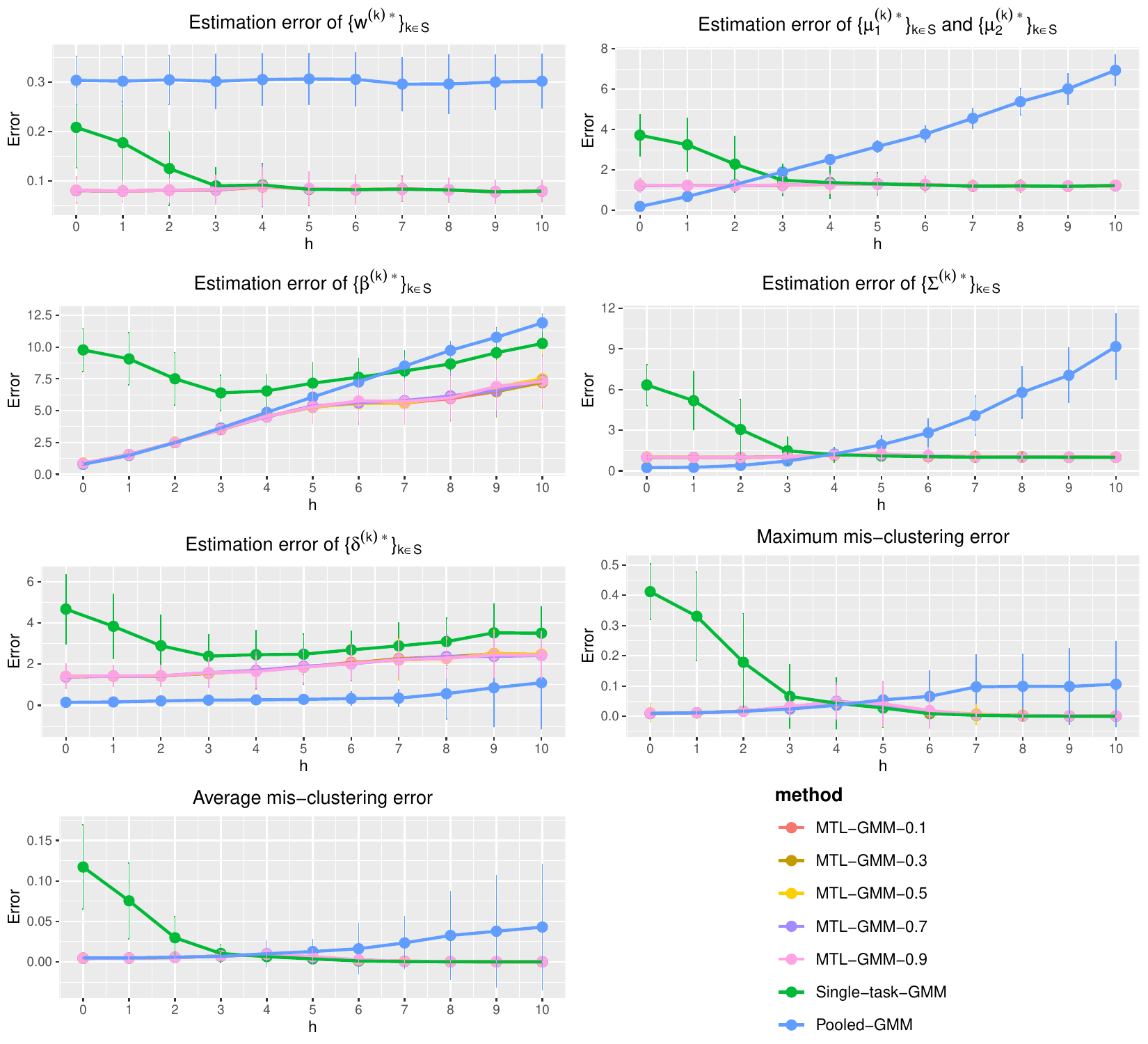}
	\caption{The performance of different methods in Simulation 1 of multi-task learning, with no outlier tasks ($\epsilon = 0$) and $h$ changing from 0 to 10 with increment 1. The meaning of each subfigure's title is the same as in Figures \ref{fig: simulation 1} and \ref{fig: simulation 1.(i) app}.}
	\label{fig: simulation mtl_1_tuning_kappa}
\end{figure}

We set $\kappa = \kappa_0 = 1/3$ in Algorithms \ref{algo: multitask}, \ref{algo: multitask multi-component}, and \ref{algo: transfer}. We run MTL-GMM with different $\kappa$ values in Simulation 1 to test the impact of $\kappa$ on the performance. The results are presented in Figure \ref{fig: simulation mtl_1_tuning_kappa}. We tried $\kappa = 0.1, 0.3, 0.5, 0.7, 0.9$ in Algorithms \ref{algo: multitask}. It can be seen that the lines representing MTL-GMM with different $\kappa$ values highly overlap with each other, which shows that the performance of MTL-GMM is very robust to the choice of $\kappa$. In practice, we take $\kappa = 1/3$ for convenience. 

\subsection{Real-data analysis}\label{subsec: real data mtl}

\subsubsection{Human activity recognition}
This subsection presents additional results for the HAR data set analyzed in Section \ref{subsec: har main binary}, focusing on the multi-cluster scenario. To ensure completeness and facilitate a clearer comparison of algorithm performance between binary and multi-cluster settings, we also include the binary GMM results from Section \ref{subsec: har main binary}.

Human Activity Recognition (HAR) Using Smartphones Data Set contains the data collected from 30 volunteers when they performed six activities (walking, walking upstairs, walking downstairs, sitting, standing, and laying) wearing a smartphone \citepapp{anguita2013public}. Each observation has 561 time and frequency domain variables. Each volunteer can be viewed as a task, and the sample size of each task varies from 281 to 409. The original data set is available at UCI Machine Learning Repository: \url{https://archive.ics.uci.edu/ml/datasets/human+activity+recognition+using+smartphones}.

Here, we first focus on two activities, standing and laying, and perform clustering without the label information, to test our method in the binary case. This is a binary MTL clustering problem with 30 tasks. The sample size of each task varies from 95 to 179. For each task, in each replication, we use 90\% of the samples as training data and hold 10\% of the samples as test data.

\begin{figure}[!h]
	\centering
	\includegraphics[width=\linewidth]{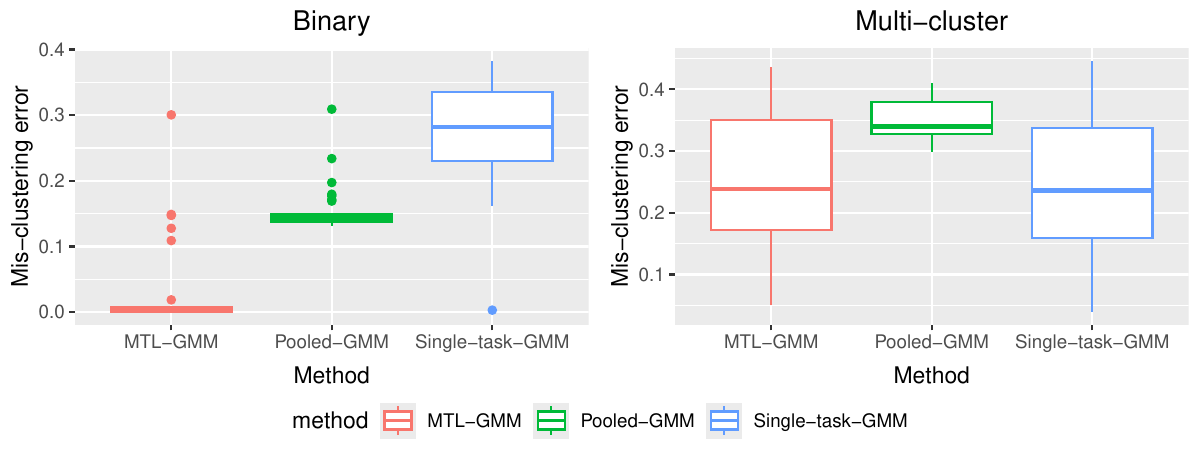}
	\caption{Box plots of mis-clustering errors of 30 tasks for each method for HAR data sets. (Left: binary case; Right: multi-cluster case)}
	\label{fig: har box plot}
\end{figure}

We first run a principal component analysis (PCA) on the training data of each task and project both the training and test data onto the first 15 principal components. PCA has often been used for dimension reduction in pre-processing the HAR data set \citepapp{zeng2014convolutional, walse2016pca, aljarrah2019human, duan2023adaptive}. We fit Single-task-GMM on each task separately, Pooled-GMM on merged data from 30 tasks, and our MTL-GMM with the greedy label swapping alignment algorithm. The performance of the three methods is evaluated by the mis-clustering error rate on the test data of all 30 tasks. The maximum and average mis-clustering errors among the 30 tasks are calculated in each replication. The mean and standard deviation of these two errors over 200 replications are reported on the left side of Table \ref{table: har}. To better display the clustering performance on each task, we further generate the box plot of mis-clustering errors of 30 tasks (averaged over 200 replications) for each method in the left plot of Figure \ref{fig: har box plot}. It is clear that MTL-GMM outperforms both Pooled-GMM and Single-task-GMM.

\begin{table}[!ht]
\centering
\begin{tabular}{@{} ccccccc @{}} 
 \toprule 
  & \multicolumn{3}{c}{Binary} & \multicolumn{3}{c@{}}{Multi-cluster}\\
 \cmidrule(lr){2-4} \cmidrule(l){5-7}
Method   & Single-task  & Pooled & MTL  & Single-task& Pooled & MTL\\
 \midrule 
Max. error  & 0.49 (0.02)     & 0.38 (0.12)     & 0.36 (0.09) & 0.51 (0.04)     & 0.50 (0.04)     & 0.51 (0.05)\\
Avg. error & 0.28 (0.02)     & 0.15 (0.17)     & 0.03 (0.01) & 0.25 (0.01)     & 0.35 (0.03)     & 0.25 (0.01)\\
 \bottomrule 
\end{tabular}
\caption{Maximum and average mis-clustering errors and standard deviations (numbers in the parentheses) in binary and multi-cluster HAR data sets.}
\label{table: har}
\end{table}

Next, we consider all six activities and compare the performance of the three approaches using the same sample-splitting strategy, to test our method in a multi-cluster scenario. Now the sample size of each task varies from 281 to 409. The maximum and average mis-clustering error rates and standard deviations over 200 replications are reported on the right side of Table \ref{table: har}. We can see that Pooled-GMM might suffer from negative transfer with a worse performance than the other two methods, while MTL-GMM and Single-task-GMM have similar performances. The right plot in Figure \ref{fig: har box plot} reveals the same comparison results.

In summary, the HAR data set exhibits different levels of similarity in binary and multi-cluster cases: tasks in the binary data are sufficiently similar so that Pooled-GMM achieves a large margin of improvement over Single-task GMM, while tasks in the multi-cluster data become much more heterogeneous, resulting in the degraded performance of Pooled-GMM compared to Single-task GMM. Nevertheless, our method MTL-GMM performs either competitively or better than the best of the two, regardless of the similarity level. These results lend further support to the effectiveness of our method.

\subsubsection{Pen-based recognition of handwritten digits (PRHD)}
The Pen-based Recognition of Handwritten Digits (PRHD) data set contains 250 samples from each of the 44 writers. Each of these writers was asked to write digits 0-9 on a pressure-sensitive tablet with an integrated LCD display and a cordless stylus. The $x$ and $y$ tablet coordinates and pressure level values of the pen were recorded. After some transformations, each observation has 16 features. The data set and more information about it are available at UCI Machine Learning Repository: \url{https://archive.ics.uci.edu/dataset/81/pen+based+recognition+of+handwritten+digits}.

\begin{figure}[!h]
	\centering
	\includegraphics[width=\linewidth]{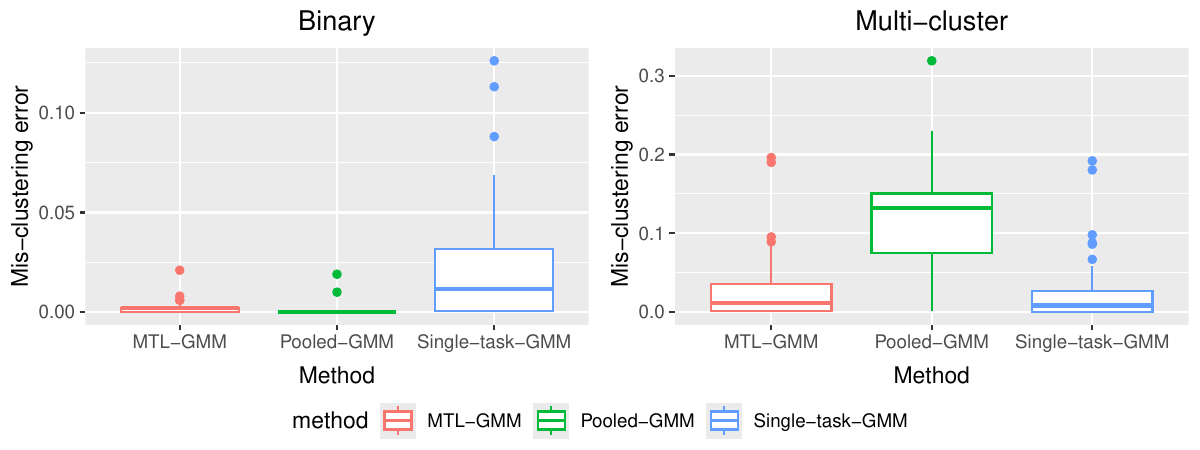}
	\caption{Box plots of mis-clustering errors of 44 tasks for each method for PRHD data set. (Left: binary case; Right: multi-cluster case)}
	\label{fig: pen box plot}
\end{figure}

Similar to the previous real-data example, we first focus on a binary clustering problem by clustering observations of digits 8 and 9. The number of observations varies between 47 and 48 among the 44 tasks, showing that this is a more balanced data set with a smaller sample size (per dimension) than the HAR data. For each task, in each replication, we use 90\% of the samples as training data and hold 10\% of the samples as test data. The maximum and average mis-clustering error rates and standard deviations over 200 replications are reported on the left side of  Table \ref{table: pen}, and the box plots of mis-clustering errors of 44 tasks are shown in Figure \ref{fig: pen box plot}. We can see that Pooled-GMM and MTL-GMM perform similarly and are much better than Single-task-GMM.

\begin{table}[!ht]
\centering
\begin{tabular}{@{} ccccccc @{}} 
 \toprule 
  & \multicolumn{3}{c}{Binary} & \multicolumn{3}{c@{}}{Multi-cluster}\\
 \cmidrule(lr){2-4} \cmidrule(l){5-7}
Method   & Single-task  & Pooled & MTL  & Single-task& Pooled & MTL\\
 \midrule 
Max. error  & 0.32 (0.10)     & 0.03 (0.07)     & 0.03 (0.09) & 0.26 (0.07)     & 0.37 (0.06)     & 0.28 (0.08)\\
Avg. error & 0.03 (0.01)     & 0.00 (0.00)     & 0.00 (0.02) & 0.03 (0.01)     & 0.12 (0.01)     & 0.03 (0.01)\\
 \bottomrule 
\end{tabular}
\caption{Maximum and average mis-clustering errors and standard deviations (numbers in the parentheses) in binary and multi-cluster PRHD data sets.}
\label{table: pen}
\end{table}

Next, we consider the observations of digits 5-9, i.e. a 5-class clustering problem. The maximum and average mis-clustering error rates and standard deviations over 200 replications are reported on the right side of Table \ref{table: pen}, and the box plots of mis-clustering errors of 44 tasks are shown in Figure \ref{fig: pen box plot}. In this multi-cluster case, MTL-GMM and Single-task-GMM have similar performance which is better than that of Pooled-GMM. Like in the first real-data example, our method MTL-GMM adapts to the unknown similarity and is competitive with the best of the other two methods.

\section{Proofs}

\subsection{Proof of Theorem \ref{thm: upper bound multitask est error}}
Define the contraction basin of one GMM as 
\begin{align}
	B_{\text{con}}(\bthetaks{k}) = \bigg\{\btheta = \{w, \bbeta, \delta\}&: w_r\in[c_w/2, 1-c_w/2], \twonorm{\bbeta - \bbetaks{k}} \leq C_b\Delta, \delta = \frac{1}{2}\bbeta^\top(\bmu_1 + \bmu_2)\\
	&\quad\, \max_{r=1:2}\twonorm{\bmu_r - \bmu_r^*} \leq C_b\Delta\bigg\},
\end{align}
for which we may shorthand as $B_{\text{con}}$ in the following.
And given the index set $S$, two joint contraction basins are defined as 
\begin{align}
	B_{\text{con}}^{J, 1}(\{\bthetaks{k}\}_{k \in S}) &= \left\{\{\bthetak{k}\}_{k \in S} = \{(\wk{k}, \bbetak{k}, \deltak{k})\}_{k \in S}: \bthetak{k} \in B_{\text{con}}(\bthetaks{k})\right\}, \\
	B_{\text{con}}^{J, 2}(\{\bthetaks{k}\}_{k \in S}) &= \left\{\{\bthetak{k}\}_{k \in S} = \{(\wk{k}, \bbetak{k}, \deltak{k})\}_{k \in S}: \bthetak{k} \in B_{\text{con}}(\bthetaks{k}), \bbetak{k} \equiv \obbeta \text{ for all } k\right\}.
\end{align}
For simplicity, at some places, we will write them as $B_{\text{con}}^{J, 1}$ and $B_{\text{con}}^{J, 2}$, respectively.

For $\btheta = (w, \bbeta, \delta)$ and $\btheta' = (w', \bbeta', \delta')$, define
\begin{equation}
	d(\btheta, \btheta') = \norm{w-w'}\vee \twonorm{\bbeta - \bbeta'} \vee \norm{\delta - \delta'}.
\end{equation}
And denote the minimum SNR $\Delta = \min_{k \in S}\Delta^{(k)}$.

\subsubsection{Lemmas}
For GMM $\bz \sim  (1-w^*)\mathcal{N}(\bmu_1^*, \bSigma^*) + w^*\mathcal{N}(\bmu_2^*, \bSigma^*)$ and any $\btheta = (w, \bbeta, \delta)$, define
\begin{align}
	\gamma_{\btheta}(\bz) = \frac{w\exp\{\bbeta^\top\bz - \delta\}}{1-w+w\exp\{\bbeta^\top\bz - \delta_r\}},& \quad w(\btheta) = \te [\gamma_{\btheta}(\bz)], \\
	\bmu_1(\btheta) = \frac{\te [(1-\gamma_{\btheta}(\bz))\bz]}{\te [1-\gamma_{\btheta}(\bz)]},& \quad  \bmu_2(\btheta) = \frac{\te [\gamma_{\btheta}(\bz)\bz]}{\te [\gamma_{\btheta}(\bz)]}.
\end{align}

\begin{lemma}[Contraction of binary GMMs, a special case of Lemma \ref{lem: contraction lemma multi-cluster} when $R = 2$]\label{lem: contraction}
	When $C_b \leq cc_{\bSigma}^{-1/2}$ with a small constant $c > 0$ and $\Delta \geq C\log(c_{\bSigma}Mc_w^{-1})$ with a large constant $C > 0$, there exist positive constants $C '> 0$ and $C'' > 0$, for any $\btheta \in B_{\textup{con}}(\bthetaks{k})$, 
	\begin{equation}
		\norm{w_r(\btheta) - w_r^*} \leq C'exp\{-C''\Delta^2\}\cdot d(\btheta, \btheta^*), \quad \twonorm{\bmu_r(\btheta) - \bmu_r^*} \leq C'\exp\{-C''\Delta^2\}\cdot d(\btheta, \btheta^*),
	\end{equation}
	where $C'\Delta\exp\{-C''\Delta^2\} \leq \kappa_0 <1$ with a constant $\kappa_0$.
\end{lemma}


\begin{lemma}\label{lem: U1}
	When $h \leq C_b\Delta$,  $B_{\text{con}}^{J, 2}(\{\bthetaks{k}\}_{k \in S}) \neq \emptyset$.
\end{lemma}

\begin{lemma}[Theorem 3 in \citealpapp{maurer2021concentration}]\label{lem: maurer concentration}
	Let $f: \mathcal{X}^n \rightarrow \mathbb{R}$ and $X = (X_1, \ldots, X_n)$ be a vector of independent random variables with values in a space $\mathcal{X}$. Then for any $t > 0$ we have
	\begin{equation}
		\tp(f(X) - \te f(X) > t) \leq \exp\left\{-\frac{t^2}{32e\infnorma{\sum_{i=1}^n \|f_i(X)\|_{\psi_2}^2}}\right\},
	\end{equation}
	where $f_i(X)$ as a random function of $x$ is defined to be $(f_i(X))(x) \coloneqq f(x_1, \ldots, x_{i-1}, X_{i}, x_{i+1}, \ldots, X_n) - \te_{X_i}[f(x_1, \ldots, x_{i-1}, X_{i}, x_{i+1}, \ldots, X_n)]$, the sub-Gaussian norm $\|Z\|_{\psi_2} \coloneqq \sup_{d\geq 1}\{\|Z\|_d/\sqrt{d}\}$, and $\|Z\|_d = (\te |Z|^d)^{1/d}$.
\end{lemma}

\begin{lemma}\label{lem: concentration w}
	Suppose Assumption \ref{asmp: upper bound multitask est error} holds.
	\begin{enumerate}[(i)]
		\item With probability at least $1-C'K^{-2}$,
			\begin{equation}
				\sup_{\substack{\bthetak{k} \in B_{\text{con}} \\ \twonorm{\bbetak{k} - \bbetaks{k}} \leq \xi^{(k)}}} \norma{\frac{1}{n_k}\sum_{i=1}^{n_k}\gamma_{\bthetak{k}}(\bzk{k}_i) - \te[\gamma_{\bthetak{k}}(\bzk{k})]} \lesssim \xi^{(k)}\sqrt{\frac{p}{n_k}} + \sqrt{\frac{\log K}{n_k}},
			\end{equation}
			for all $k \in S$.
		\item With probability at least $1-C'K^{-2}e^{-C''p}$,
			\begin{equation}
				\sup_{\{\bthetak{k}\}_{k \in S} \in B_{\text{con}}^{J,2}}\sup_{\norm{\widetilde{w}_k} \leq 1} \frac{1}{\ns}\norma{\sum_{k \in S}\widetilde{w}_k\sum_{i=1}^{n_k}\Big[\gamma_{\bthetak{k}}(\bzk{k}_i) - \te[\gamma_{\bthetak{k}}(\bzk{k})]\Big]} \lesssim \sqrt{\frac{p+K}{\ns}}.
			\end{equation}
	\end{enumerate}
\end{lemma}

\begin{lemma}\label{lem: concentration delta}
	Suppose Assumption \ref{asmp: upper bound multitask est error} holds.  
	\begin{enumerate}[(i)]
		\item With probability at least $1-C'(K^{-2}+K^{-2}e^{-C''p})$,
			\begin{align}
				&\sup_{\substack{\bthetak{k} \in B_{\text{con}} \\ \twonorm{\bbetak{k} - \bbetaks{k}} \leq \xi^{(k)}}} \norma{\frac{1}{n_k}\sum_{i=1}^{n_k}\big[1-\gamma_{\bthetak{k}}(\bzk{k}_i)\big](\bzk{k}_i)^\top\bbetaks{k} - \te\big[[1-\gamma_{\bthetak{k}}(\bzk{k})](\bzk{k})^\top\bbetaks{k}\big]} \\
				&\quad\lesssim \xi^{(k)}\sqrt{\frac{p}{n_k}} + \sqrt{\frac{\log K}{n_k}},
			\end{align}
			for all $k \in S$.
		\item With probability at least $1-C'K^{-2}e^{-C''p}$,
			\begin{align}
				&\sup_{\{\bthetak{k}\}_{k \in S} \in B_{\text{con}}^{J,2}}\sup_{\norm{\widetilde{w}_k} \leq 1} \frac{1}{\ns}\norma{\sum_{k \in S}\widetilde{w}_k\sum_{i=1}^{n_k}\Big[\big[1-\gamma_{\bthetak{k}}(\bzk{k}_i)\big](\bzk{k}_i)^\top\bbetaks{k} - \te\big[[1-\gamma_{\bthetak{k}}(\bzk{k})](\bzk{k})^\top\bbetaks{k}\big]\Big]} \\
				&\quad \lesssim \sqrt{\frac{p+K}{\ns}}.
			\end{align}
	\end{enumerate}
\end{lemma}

\begin{lemma}\label{lem: concentration beta}
	Suppose Assumption \ref{asmp: upper bound multitask est error} holds.  
	\begin{enumerate}[(i)]
		\item With probability at least $1-C'(K^{-2}+K^{-2}e^{-C''p})$,
			\begin{equation}
				\sup_{\bthetak{k} \in B_{\text{con}}} \twonorma{\frac{1}{n_k}\sum_{i=1}^{n_k}\gamma_{\bthetak{k}}(\bzk{k}_i)\bzk{k}_i - \te[\gamma_{\bthetak{k}}(\bzk{k})\bzk{k}]} \lesssim \sqrt{\frac{p+\log K}{n_k}} ,
			\end{equation}
			for all $k \in S$.
		\item With probability at least $1-C'K^{-2}e^{-C''p}$,
			\begin{equation}
				\sup_{\{\bthetak{k}\}_{k \in S} \in B_{\text{con}}^{J,2}}\sup_{\norm{\widetilde{w}_k} \leq 1} \frac{1}{\ns}\twonorma{\sum_{k \in S}\widetilde{w}_k\sum_{i=1}^{n_k}\Big[\gamma_{\bthetak{k}}(\bzk{k}_i)\bzk{k}_i - \te[\gamma_{\bthetak{k}}(\bzk{k})\bzk{k}]\Big]} \lesssim \sqrt{\frac{p+K}{\ns}}.
			\end{equation}
		\item With probability at least $1-C'K^{-2}e^{-C''p}$,
			\begin{equation}
				\sup_{\{\bthetak{k}\}_{k \in S} \in B_{\text{con}}^{J,2}}\sup_{\norm{\widetilde{w}_k} \leq 1} \frac{1}{\ns}\twonorma{\sum_{k \in S}\widetilde{w}_k\sum_{i=1}^{n_k}\Big[\gamma_{\bthetak{k}}(\bzk{k}_i) - \te[\gamma_{\bthetak{k}}(\bzk{k})]\Big]\bmuks{k}_1} \lesssim \sqrt{\frac{p+K}{\ns}}.
			\end{equation}
	\end{enumerate}
\end{lemma}

\begin{lemma}\label{lem: concentration beta 2}
	Suppose Assumption \ref{asmp: upper bound multitask est error} holds. 
	\begin{enumerate}[(i)]
		\item With probability at least $1-C'(K^{-2}+K^{-1}e^{-C''p})$,
			\begin{equation}
				\twonorma{\frac{1}{n_k}\sum_{i=1}^{n_k}\big[\bzk{k}_i(\bzk{k}_i)^\top - \te [\bzk{k}_i(\bzk{k}_i)^\top]\big]\bbetaks{k}} \lesssim \sqrt{\frac{p+\log K}{n_k}},
			\end{equation}
			for all $k \in S$.
		\item With probability at least $1-C'K^{-2}e^{-C''p}$,
			\begin{equation}
				\twonorma{\frac{1}{\ns}\sum_{k \in S}\sum_{i=1}^{n_k}\big[\bzk{k}_i(\bzk{k}_i)^\top - \te [\bzk{k}_i(\bzk{k}_i)^\top]\big]\bbetaks{k}} \lesssim \sqrt{\frac{p}{\ns}}.
			\end{equation}
	\end{enumerate}	
\end{lemma}

\subsubsection{Main proof of Theorem \ref{thm: upper bound multitask est error}}
The proof of Theorem \ref{thm: upper bound multitask est error} consists of two cases. In Case 1, we study the scenario $h \gtrsim \sqrt{\frac{p+\log K}{\max_{k \in S}n_k}}$, where we take a fixed contraction radius. In this case, proving a single-task estimation error rate $\sqrt{\frac{K(p+\log K)}{n_S}}$ is sufficient, which is relatively straightforward. In Case 2, we explore the scenario $h \lesssim \sqrt{\frac{p+\log K}{\max_{k \in S}n_k}}$, in which regime multi-task learning can outperform the classical single-task learning. In this case, the classical finite-sample analysis of EM in \citeapp{balakrishnan2017statistical} and \citeapp{cai2019chime}, which uses a fixed contraction radius as we did in Case 1, does not work. This is because the heterogenous $\bmuks{k}_1$ and $\bmuks{k}_2$ lead to an error of $\sqrt{\frac{K(p+\log K)}{n_S}}$ when estimating $\wks{k}$. This term $\sqrt{\frac{K(p+\log K)}{n_S}}$ ultimately affects the estimation error of $\bbetaks{k}$ and $\deltaks{k}$, preventing us from proving the improvement of multi-task learning over single-task learning. To resolve this issue, we creatively use a ``localization" strategy to adaptively shrink the contraction radius in each iteration. This method effectively eliminates the term $\sqrt{\frac{K(p+\log K)}{n_S}}$. By combining the two cases, we complete the proof.

WLOG, in Assumptions \ref{asmp: upper bound multitask est error}.(\rom{3}) and \ref{asmp: upper bound multitask est error}.(\rom{4}), we assume
\begin{itemize}
	\item $\max_{k \in S}\big(\twonorm{\hbeta^{(k)[0]} - \bbetaks{k}} \vee \twonorm{\hmu^{(k)[0]}_1 - \bmuks{k}_1}\vee \twonorm{\hmu^{(k)[0]}_2 - \bmuks{k}_2}\big) \leq C'\min_{k \in S}\Delta^{(k)}$, with a small constant $C' > 0$; 
	\item $\max_{k \in S}\norm{\hw^{(k)[0]}-\wks{k}} \leq c_w/2$.
\end{itemize}

\noindent\underline{(\Rom{1}) Case 1:} Let us consider the case that $h \geq C\sqrt{\frac{p+\log K}{\max_{k \in S}n_k}}$. Consider an event $\mathcal{E}$ defined to be the intersection of the events in Lemmas \ref{lem: concentration w}.(\rom{1}), \ref{lem: concentration delta}.(\rom{1}), \ref{lem: concentration beta}.(\rom{1}), and \ref{lem: concentration beta 2}.(\rom{1}), with $\xi^{(k)} = $ a large constant $C$, which satisfies $\tp(\mathcal{E}) \geq 1-C'(K^{-2}+K^{-2}e^{-C''p})$. Throughout the analysis in Case 1, we condition on $\mathcal{E}$, therefore all the arguments hold with probability at least $1-C'(K^{-2}+K^{-2}e^{-C''p})$.

Consider the case $t = 1$. Lemma \ref{lem: claim B1} tells us that when $\lambda^{[t]} \geq C\max_{k \in S}\{\sqrt{n_k}\twonorm{\hSigma^{(k)[t]}\bbetaks{k} - (\hmu^{(k)[t]}_2 - \hmu^{(k)[t]}_1)}\}$, we have
\begin{equation}\label{eq: beta hat beta 1}
	\twonorm{\hbeta^{(k)[t]} - \bbetaks{k}} \lesssim \twonorma{\sum_{k \in S}\frac{n_k}{\ns}[\hSigma^{(k)[t]}\bbetaks{k} - (\hmu^{(k)[t]}_2 - \hmu^{(k)[t]}_1)]} + h \wedge \frac{\lambda^{[t]}}{\sqrt{n_k}} + \epsilon \frac{\lambda^{[t]}}{\sqrt{\max_{k=1:K}n_k}}.
\end{equation}
And if further $\lambda^{[t]} \geq C\max_{k \in S}\sqrt{n_k}h$, we have \eqref{eq: beta hat beta 1} holds with $\hbeta^{(k)[t]} = \obbeta^{[t]}$ for all $k \in S$. Note that
\begin{equation}\label{eq: lambda related eq 1 stage 1}
	\twonorm{\hSigma^{(k)[t]}\bbetaks{k} - (\hmu^{(k)[t]}_2 - \hmu^{(k)[t]}_1)} \leq \twonorm{(\hSigma^{(k)[t]} - \bSigmaks{k})\bbetaks{k}} + \twonorm{\hmu^{(k)[t]}_2 - \hmu^{(k)[t]}_1 - \bmuks{k}_2 + \bmuks{k}_1}.
\end{equation}
And the first term on the RHS can be controlled as
\begin{align}
	&\twonorm{(\hSigma^{(k)[t]} - \bSigmaks{k})\bbetaks{k}} \\
	&\leq \underbrace{\twonorma{\frac{1}{n_k}\sum_{i=1}^{n_k}\big[\bzk{k}_i(\bzk{k}_i)^\top - \te [\bzk{k}_i(\bzk{k}_i)^\top]\big]\bbetaks{k}}}_{\circled{1}} \\
	&\quad + \underbrace{\twonorma{\big[(1-\hw^{(k)[t]})\hmu_1^{(k)[t]}(\hmu_1^{(k)[t]})^\top - (1-\wks{k})\bmuks{k}_1(\bmuks{k}_1)^\top\big]\bbetaks{k}}}_{\circled{2}} \\
	&\quad + \underbrace{\twonorma{\big[\hw^{(k)[t]}\hmu_2^{(k)[t]}(\hmu_2^{(k)[t]})^\top - \wks{k}\bmuks{k}_2(\bmuks{k}_2)^\top\big]\bbetaks{k}}}_{\circled{3}}. \label{eq: stage 1 Sigma}
\end{align}
Conditioned on $\mathcal{E}$, we have
\begin{equation}
	\circled{1} \lesssim \sqrt{\frac{p+\log K}{n_k}},
\end{equation}
And
\begin{align}
	\circled{2} &\leq \underbrace{\twonorma{(1-\hw^{(k)[t]})(\hmu^{(k)[t]}_1 - \bmuks{k}_1)\cdot (\hmu^{(k)[t]}_1)^\top\bbetaks{k}}}_{\circled{2}.1} \\
	&\quad + \underbrace{\twonorma{\big[(1-\hw^{(k)[t]})\bmuks{k}_1(\hmu^{(k)[t]}_1)^\top - (1-\wks{k})\bmuks{k}_1(\bmuks{k}_1)^\top\big]\bbetaks{k}}}_{\circled{2}.2},
\end{align}
where
\begin{equation}\label{eq: 2.2 eq 1}
	\circled{2}.2 \leq \twonorma{(\hw^{(k)[t]} - \wks{k})\bmuks{k}_1(\hmu^{(k)[t]}_1)^\top\bbetaks{k}} + \twonorma{(1-\hw^{(k)[t]})\bmuks{k}_1(\hmu^{(k)[t]}_1 - \bmuks{k}_1)^\top\bbetaks{k}}.
\end{equation}
Before we discuss how to control the terms on the RHS, let us first try to control $\norm{\hw^{(k)[t]} - \wks{k}}$ as it will be used to bound the existing terms. Note that by Lemma \ref{lem: contraction},
\begin{align}
	\norm{\hw^{(k)[t]} - \wks{k}} &\leq \norm{\wk{k}(\htheta^{(k)[t-1]}) - \wks{k}} + \norm{\hw^{(k)[t]} - \wk{k}(\htheta^{(k)[t-1]})} \\
	&\leq \kappa_0 d(\htheta^{(k)[t-1]}, \bthetaks{k}) + \norma{\frac{1}{n_k}\sum_{i=1}^{n_k}\gamma_{\htheta^{(k)[t-1]}}(\bzk{k}_i) - \te_{\bzk{k}}[\gamma_{\htheta^{(k)[t-1]}}(\bzk{k})]} \\
	&\lesssim \kappa_0 d(\htheta^{(k)[t-1]}, \bthetaks{k}) + \sqrt{\frac{p+\log K}{n_k}} \label{eq: w stage 1 eq 1}\\
	&\leq c,
\end{align}
where $c$ is a small constant. By Lemma \ref{lem: contraction} again,
\begin{align}
	\twonorm{\hmu^{(k)[t]}_1 - \bmuks{k}_1} &= \twonorma{\frac{\frac{1}{n_k}\sum_{i=1}^{n_k}(1-\gamma_{\htheta^{(k)[t-1]}}(\bzk{k}_i))\bzk{k}_i}{1-\hw^{(k)[t]}} - \frac{\te_{\bzk{k}}[(1-\gamma_{\htheta^{(k)[t-1]}}(\bzk{k}))\bzk{k}]}{1-\wk{k}(\htheta^{(k)[t-1]})}}\\
	&\leq \twonorma{\frac{\frac{1}{n_k}\sum_{i=1}^{n_k}(1-\gamma_{\htheta^{(k)[t-1]}}(\bzk{k}_i))\bzk{k}_i - \te_{\bzk{k}}[(1-\gamma_{\htheta^{(k)[t-1]}}(\bzk{k}))\bzk{k}]}{1-\hw^{(k)[t]}}} \\
	&\quad + \twonorma{\frac{\te_{\bzk{k}}[(1-\gamma_{\htheta^{(k)[t-1]}}(\bzk{k}))\bzk{k}]}{(1-\hw^{(k)[t]})(1-\wk{k}(\htheta^{(k)[t-1]}))}(\hw^{(k)[t]}-\wk{k}(\htheta^{(k)[t-1]}))} \\
	&\lesssim \twonorma{\frac{1}{n_k}\sum_{i=1}^{n_k}(1-\gamma_{\htheta^{(k)[t-1]}}(\bzk{k}_i))\bzk{k}_i - \te_{\bzk{k}}[(1-\gamma_{\htheta^{(k)[t-1]}}(\bzk{k}))\bzk{k}]} \\
	&\quad + \norm{\hw^{(k)[t]} - \wks{k}} + \kappa_0 d(\htheta^{(k)[t-1]}, \bthetaks{k}) \\
	&\lesssim \kappa_0 d(\htheta^{(k)[t-1]}, \bthetaks{k}) +  \sqrt{\frac{p+\log K}{n_k}} \\
	&\leq C_b\Delta. 
\end{align}
Therefore, we can bound the RHS of \eqref{eq: 2.2 eq 1} as
\begin{equation}\label{eq: circle 2 eq 1}
	\circled{2}.2 \lesssim \norm{\hw^{(k)[t]} - \wks{k}} + \twonorm{\hmu^{(k)[t]}_1 - \bmuks{k}_1} \lesssim \kappa_0 d(\htheta^{(k)[t-1]}, \bthetaks{k}) +  \sqrt{\frac{p+\log K}{n_k}}.
\end{equation}

Similarly, we have
\begin{equation}\label{eq: circle 2 eq 2}
	\circled{2}.1\lesssim \twonorm{\hmu^{(k)[t]}_1 - \bmuks{k}_1} \lesssim \kappa_0 d(\htheta^{(k)[t-1]}, \bthetaks{k}) +  \sqrt{\frac{p+\log K}{n_k}},
\end{equation}
Combining \eqref{eq: circle 2 eq 1} and \eqref{eq: circle 2 eq 2}, we have
\begin{equation}
	\circled{2} \lesssim \kappa_0 d(\htheta^{(k)[t-1]}, \bthetaks{k}) +  \sqrt{\frac{p+\log K}{n_k}},
\end{equation}
Similarly, we can bound $\circled{3}$ in the same way, and get
\begin{equation}
	\circled{3} \lesssim \kappa_0 d(\htheta^{(k)[t-1]}, \bthetaks{k}) +  \sqrt{\frac{p+\log K}{n_k}},
\end{equation}
Hence
\begin{equation}
	\twonorm{(\hSigma^{(k)[t]} - \bSigmaks{k})\bbetaks{k}} \lesssim \kappa_0 d(\htheta^{(k)[t-1]}, \bthetaks{k}) +  \sqrt{\frac{p+\log K}{n_k}},
\end{equation}
And the second term on the RHS of \eqref{eq: lambda related eq 1 stage 1} satisfies
\begin{equation}\label{eq: mu stage 1}
	\twonorm{\hmu^{(k)[t]}_2 - \hmu^{(k)[t]}_1 - \bmuks{k}_2 + \bmuks{k}_1}\lesssim \twonorm{\hmu^{(k)[t]}_2 -  \bmuks{k}_2} \vee \twonorm{\hmu^{(k)[t]}_1 - \bmuks{k}_1} \lesssim \kappa_0 d(\htheta^{(k)[t-1]}, \bthetaks{k}) +  \sqrt{\frac{p+\log K}{n_k}}.
\end{equation}
All together, we have
\begin{equation}\label{eq: penalty eq 2 stage 1}
	\twonorm{\hSigma^{(k)[t]}\bbetaks{k} - (\hmu^{(k)[t]}_2 - \hmu^{(k)[t]}_1)} \lesssim \kappa_0 d(\htheta^{(k)[t-1]}, \bthetaks{k}) +  \sqrt{\frac{p+\log K}{n_k}}.
\end{equation}
This implies that $\lambda^{[t]} = C_{\lambda}\sqrt{p+\log K} + \kappa\lambda^{[0]} \geq C\max_{k \in S}\{\sqrt{n_k}\twonorm{\hSigma^{(k)[t]}\bbetaks{k} - (\hmu^{(k)[t]}_2 - \hmu^{(k)[t]}_1)}\}$, therefore by \eqref{eq: beta hat beta 1},
\begin{equation}\label{eq: beta stage 1}
	\twonorm{\hbeta^{(k)[t]} - \bbetaks{k}} \lesssim \kappa_0 \sum_{k \in S}\frac{n_k}{\ns}d(\htheta^{(k)[t-1]}, \bthetaks{k}) + \sqrt{\frac{K(p+\log K)}{\ns}} + \epsilon \frac{\lambda^{[t]}}{\sqrt{\max_{k=1:K}n_k}},
\end{equation}
And by \eqref{eq: w stage 1 eq 1},
\begin{equation}\label{eq: w stage 1}
	\sum_{k \in S}\frac{n_k}{\ns}\norm{\hw^{(k)[t]} - \wks{k}} \lesssim \kappa_0 \sum_{k \in S}\frac{n_k}{\ns}d(\htheta^{(k)[t-1]}, \bthetaks{k}) + \sqrt{\frac{K(p+\log K)}{\ns}}.
\end{equation}
Also, 
\begin{align}
	\norm{\hdelta^{(k)[t]} - \deltaks{k}} &= \frac{1}{2}\twonorma{(\hbeta^{(k)[t]})^\top(\hmu^{(k)[t]}_1 + \hmu^{(k)[t]}_2) - (\bbetaks{k})^\top(\bmuks{k}_1 + \bmuks{k}_2)} \\
	&\lesssim \twonorm{\hbeta^{(k)[t]} - \bbetaks{k}} + \twonorm{\hmu^{(k)[t]}_1 - \bmuks{k}_1} + \twonorm{\hmu^{(k)[t]}_2 - \bmuks{k}_2} \\
	&\lesssim  \kappa_0 d(\htheta^{(k)[t-1]}, \bthetaks{k}) + \sqrt{\frac{p+\log K}{n_k}}, \label{eq: delta stage 1 eq 2}
\end{align}
which entails that
\begin{equation}\label{eq: delta stage 1}
	\sum_{k \in S}\frac{n_k}{\ns}\norm{\hdelta^{(k)[t]} - \deltaks{k}} \lesssim \kappa_0 \sum_{k \in S}\frac{n_k}{\ns}d(\htheta^{(k)[t-1]}, \bthetaks{k}) + \sqrt{\frac{K(p+\log K)}{\ns}}.
\end{equation}
Combining \eqref{eq: beta stage 1}, \eqref{eq: w stage 1}, and \eqref{eq: delta stage 1}, we have
\begin{equation}\label{eq: d stage 1}
	\sum_{k \in S}\frac{n_k}{\ns}d(\htheta^{(k)[t]}, \bthetaks{k}) \lesssim \kappa_0 \sum_{k \in S}\frac{n_k}{\ns}d(\htheta^{(k)[t-1]}, \bthetaks{k}) + \sqrt{\frac{K(p+\log K)}{\ns}}+ \epsilon\frac{\lambda^{[t]}}{\sqrt{\max_{k=1:K}n_k}}.
\end{equation}
Also,
\begin{equation}\label{eq: d max stage 1}
	\max_{k \in S}\big\{\sqrt{n_k}d(\htheta^{(k)[t]}, \bthetaks{k})\big\} \lesssim \kappa_0 \max_{k \in S}\big\{\sqrt{n_k}d(\htheta^{(k)[t-1]}, \bthetaks{k})\big\} + \lambda^{[t]}.
\end{equation}

When we assume \eqref{eq: beta stage 1}, \eqref{eq: d stage 1}, \eqref{eq: d max stage 1} hold for all $t = 1:t'$, via the same analysis we will have \eqref{eq: penalty eq 2 stage 1} hold again for $t = t' + 1$. Hence
\begin{equation}
	\max_{k \in S}\big\{\sqrt{n_k}\twonorm{\hSigma^{(k)[t'+1]}\bbetaks{k} - (\hmu^{(k)[t'+1]}_2 - \hmu^{(k)[t'+1]}_1)}\big\} \lesssim \kappa_0 \max_{k \in S}\big\{\sqrt{n_k}d(\htheta^{(k)[t']}, \bthetaks{k})\big\} +  \sqrt{p+\log K}.
\end{equation}
Then by \eqref{eq: d max stage 1} when $t = t'$,
\begin{align}
	\kappa_0 \max_{k \in S}\big\{\sqrt{n_k}d(\htheta^{(k)[t']}, \bthetaks{k})\big\} +  \sqrt{p+\log K} &\lesssim \kappa_0^2 \max_{k \in S}\big\{\sqrt{n_k}d(\htheta^{(k)[t'-1]}, \bthetaks{k})\big\}  +  \sqrt{p+\log K}+ \kappa_0 \lambda^{[t']} \\
	&\leq \kappa_0 \lambda^{[t']}  +  \sqrt{p+\log K} \\
	&\leq  \lambda^{[t'+1]},
\end{align}
where we need $\kappa \geq C\kappa_0$ with a large constant $C > 0$. Recall that $\kappa \in (0, 1)$ is one of the tuning parameters in the update formula of $\lambda^{[t]}$. Therefore we can follow the same arguments as above to obtain \eqref{eq: w stage 1 eq 1}, \eqref{eq: mu stage 1}, \eqref{eq: beta stage 1}, \eqref{eq: delta stage 1 eq 2}, \eqref{eq: d stage 1}, \eqref{eq: d max stage 1} for $t = t'+1$.

So far, we have shown that \eqref{eq: w stage 1 eq 1}, \eqref{eq: mu stage 1}, \eqref{eq: beta stage 1}, \eqref{eq: delta stage 1 eq 2}, \eqref{eq: d stage 1}, \eqref{eq: d max stage 1} hold for any $t$. By the update formula of $\lambda^{[t]}$, when $t \geq 1$, we have
\begin{equation}
	\lambda^{[t]} = \frac{1-\kappa^t}{1-\kappa}C_{\lambda}\sqrt{p+\log K} + \kappa^{t-1}\lambda^{[0]}.
\end{equation} 
Therefore by \eqref{eq: d stage 1},
\begin{align}
	\sum_{k \in S}\frac{n_k}{\ns}d(\htheta^{(k)[t]}, \bthetaks{k}) &\leq C\kappa_0 \sum_{k \in S}\frac{n_k}{\ns}d(\htheta^{(k)[t-1]}, \bthetaks{k}) + C'\sqrt{\frac{K(p+\log K)}{\ns}}+ C'\sqrt{\frac{K}{\ns}}\lambda^{[t]} \\
	&\leq (C\kappa_0)^{t} \sum_{k \in S}\frac{n_k}{\ns}d(\htheta^{(k)[0]}, \bthetaks{k})+ C'\sqrt{\frac{K(p+\log K)}{\ns}}+ C'\sqrt{\frac{K}{\ns}}\sum_{t'=1}^t \lambda^{[t']}\cdot (C\kappa_0)^{t-t'} \\
	&\leq (C\kappa_0)^{t} \sum_{k \in S}\frac{n_k}{\ns}d(\htheta^{(k)[0]}, \bthetaks{k})+ C'\sqrt{\frac{K(p+\log K)}{\ns}}+ C'\sqrt{\frac{K}{\ns}}\sum_{t'=1}^t \lambda^{[t']}\cdot \kappa^{t-t'} \\
	&\leq C''t\kappa^{t} + C''\sqrt{\frac{K(p+\log K)}{\ns}}. \label{eq: nk ns d stage 1 eq 1}
\end{align}

Consider a new event $\mathcal{E}'$ defined to be the intersection of the events in Lemmas \ref{lem: concentration w}.(\rom{1}), \ref{lem: concentration delta}.(\rom{1}), \ref{lem: concentration beta}.(\rom{1}), and \ref{lem: concentration beta 2}.(\rom{1}), with $\xi^{(k)} = C\sqrt{\frac{n_k}{\max_{k \in S}n_k}}$, which satisfies $\tp(\mathcal{E}') \geq 1-C'(K^{-2}+K^{-2}e^{-C''p})$. Throughout the following analysis in Case 1, we condition on $\mathcal{E} \cap \mathcal{E}'$, therefore all the arguments hold with probability at least $1-C'(K^{-2}+K^{-2}e^{-C''p})$. When $h \geq C\sqrt{\frac{p+\log K}{\max_{k \in S}n_k}}$, since $\ns \gtrsim K\max_{k \in S}n_k$, we have $\sqrt{\frac{K(p+\log K)}{\ns}} \lesssim \sqrt{\frac{p+\log K}{\max_{k \in S}n_k}} \lesssim h \wedge \sqrt{\frac{p+\log K}{n_k}}$. Furthermore, when $t \geq C'\log\big(\frac{\max_{k \in S}n_k}{\min_{k \in S}n_k}\big)$ with a large $C' > 0$, we have $\xi^{(k)}\sqrt{\frac{p}{n_k}} \lesssim h \wedge \sqrt{\frac{p+\log K}{n_k}} + \epsilon \sqrt{\frac{p+\log K}{\max_{k=1:K}n_k}} + \sqrt{\frac{\log K}{n_k}}$ and $t\kappa^t + C\sqrt{\frac{K(p+\log K)}{\ns}} + C\epsilon \sqrt{\frac{p+\log K}{\max_{k=1:K}n_k}} + C\sqrt{\frac{\log K}{n_k}}\leq \xi^{(k)}$, where we used the fact $\ns \gtrsim K\max_{k \in S}n_k$ again to get the second inequality.

Plugging \eqref{eq: nk ns d stage 1 eq 1} back into \eqref{eq: beta stage 1}, we have
\begin{align}
	\twonorm{\hbeta^{(k)[t]} - \bbetaks{k}} &\leq C''t\kappa^{t} + C''\sqrt{\frac{K(p+\log K)}{\ns}} + C''\xi^{(k)}\sqrt{\frac{p}{n_k}} + C''\sqrt{\frac{\log K}{n_k}}\\
	&\leq Ct\kappa^{t} + C\sqrt{\frac{K(p+\log K)}{\ns}} + C\cdot h \wedge \sqrt{\frac{p+\log K}{n_k}} + C\epsilon \sqrt{\frac{p+\log K}{n_k}} + C\sqrt{\frac{\log K}{n_k}} \\
	&\leq Ct\kappa^{t} + C\cdot h \wedge \sqrt{\frac{p+\log K}{n_k}} + C\epsilon \sqrt{\frac{p+\log K}{\max_{k=1:K}n_k}} + C\sqrt{\frac{\log K}{n_k}}  \label{eq: beta stage 1 eq x}
\end{align}

Then by \eqref{eq: w stage 1 eq 1},
\begin{align}
	\norm{\hw^{(k)[t]} - \wks{k}} &\lesssim \kappa_0 d(\htheta^{(k)[t-1]}, \bthetaks{k}) + \xi^{(k)}\sqrt{\frac{p}{n_k}} + \sqrt{\frac{\log K}{n_k}} \\
	&\lesssim \kappa_0\twonorm{\hbeta^{(k)[t-1]} - \bbetaks{k}} + \kappa_0\norm{\hw^{(k)[t-1]} - \wks{k}} \vee \norm{\hdelta^{(k)[t-1]} - \deltaks{k}}+ \xi^{(k)}\sqrt{\frac{p}{n_k}} + \sqrt{\frac{\log K}{n_k}} \\
	&\lesssim Ct\kappa^t +\kappa_0\norm{\hw^{(k)[t-1]} - \wks{k}} \vee \norm{\hdelta^{(k)[t-1]} - \deltaks{k}} + h \wedge \sqrt{\frac{p+\log K}{n_k}} \\
	&\quad + \epsilon \sqrt{\frac{p+\log K}{\max_{k=1:K}n_k}} + \sqrt{\frac{\log K}{n_k}}.
\end{align}
Similarly, by \eqref{eq: delta stage 1 eq 2},
\begin{align}
	\norm{\hdelta^{(k)[t-1]} - \deltaks{k}} &\lesssim Ct\kappa^t +\kappa_0\norm{\hw^{(k)[t-1]} - \wks{k}} \vee \norm{\hdelta^{(k)[t-1]} - \deltaks{k}} \\
	&\quad + h \wedge \sqrt{\frac{p+\log K}{n_k}} + \epsilon \sqrt{\frac{p+\log K}{\max_{k=1:K}n_k}} + \sqrt{\frac{\log K}{n_k}}.
\end{align}
Therefore,
\begin{align}
	\norm{\hw^{(k)[t]} - \wks{k}} \vee \norm{\hdelta^{(k)[t]} - \deltaks{k}} &\leq Ct\kappa^t +C\kappa_0\norm{\hw^{(k)[t-1]} - \wks{k}} \vee \norm{\hdelta^{(k)[t-1]} - \deltaks{k}}  \\
	&\quad + h \wedge \sqrt{\frac{p+\log K}{n_k}} + \epsilon \sqrt{\frac{p+\log K}{n_k}} + \sqrt{\frac{\log K}{n_k}} \\
	&\leq C''t^2\kappa^t + C''\cdot h \wedge \sqrt{\frac{p+\log K}{n_k}} + C''\epsilon \sqrt{\frac{p+\log K}{\max_{k=1:K}n_k}} \\
	&\quad + C''\sqrt{\frac{\log K}{n_k}}.
\end{align}
Combine it with \eqref{eq: beta stage 1 eq x}, we obtain that
\begin{equation}
	d(\htheta^{(k)[t]}, \bthetaks{k}) \leq C''t^2\kappa^t  + C''\cdot h \wedge \sqrt{\frac{p+\log K}{n_k}} + C''\epsilon \sqrt{\frac{p+\log K}{\max_{k=1:K}n_k}} + C''\sqrt{\frac{\log K}{n_k}}.
\end{equation}
Plugging this back into \eqref{eq: circle 2 eq 2}, we get
\begin{equation}
	\twonorm{\hmu^{(k)[t]}_1 - \bmuks{k}_1} \leq C''t^2\kappa^t + C''\sqrt{\frac{p+\log K}{n_k}}.
\end{equation}
And the same bound holds for $\twonorm{\hmu^{(k)[t]}_2 - \bmuks{k}_2}$ as well. The same bound for $\twonorm{\hSigma^{(k)[t]} - \bSigmaks{k}}$ can be obtained in the same spirit as in \eqref{eq: stage 1 Sigma}.

\noindent\underline{(\Rom{2}) Case 2:} We now focus on the case that $h \leq C\sqrt{\frac{p+\log K}{\max_{k \in S}n_k}}$. As mentioned at the beginning of this proof, we need to adaptively shrink the radius of the contraction basin to prove the desired convergence rate. The analysis of Case 2 can be divided into two stages. In the first stage, we use the same fixed contraction radius as in Case 1 and follow the same analysis until the iterative error $t^2\kappa^t$ has reduced to the single-task error $\sqrt{\frac{K(p+\log K)}{\ns}}$. In the second stage, we apply the localization argument to shrink the contraction basin until we achieve the desired rate of convergence.

Similar to Case 1, we consider an event $\mathcal{E}$ defined to be the intersection of the events in Lemmas \ref{lem: concentration w}, \ref{lem: concentration delta}, \ref{lem: concentration beta}, and \ref{lem: concentration beta 2}, with $\xi^{(k)} = $ a large constant $C$, which satisfies $\tp(\mathcal{E}) \geq 1-C'(K^{-2}+K^{-2}e^{-C''p})$. Throughout the analysis in Case 2, we condition on $\mathcal{E}$, therefore all the arguments hold with probability at least $1-C'(K^{-2}+K^{-2}e^{-C''p})$. 

Consider $t_0$ as the number of iterations in the first stage which satisfies $t_0^2\kappa^{t_0} \asymp \sqrt{\frac{K(p+\log K)}{\ns}}$. When $t = 1:t_0$, we can go through the same analysis as in Case 1, and show that conditioned on $\mathcal{E}$, 
\begin{equation}\label{eq: d stage 2 t = 1}
	\sum_{k \in S}\frac{n_k}{\ns}d(\htheta^{(k)[t]}, \bthetaks{k}) \lesssim C''t\kappa^{t} + C''\sqrt{\frac{K(p+\log K)}{\ns}},
\end{equation}
and
\begin{align}
	&d(\htheta^{(k)[t]}, \bthetaks{k}) \leq C''t^2\kappa^t  +C''\sqrt{\frac{K(p+\log K)}{\ns}} + C''\cdot h \wedge \sqrt{\frac{p+\log K}{n_k}} + C''\epsilon \sqrt{\frac{p+\log K}{\max_{k=1:K}n_k}} \\
	&\hspace{3cm} + C''\sqrt{\frac{\log K}{n_k}}, \\
	&\twonorm{\hmu^{(k)[t]}_1 - \bmuks{k}_1}\vee \twonorm{\hmu^{(k)[t]}_2 - \bmuks{k}_2} \vee \twonorm{\hSigma^{(k)[t]} - \bSigmaks{k}}\leq C''t^2\kappa^t + C''\sqrt{\frac{p+\log K}{n_k}}.
\end{align}
Since $t_0^2\kappa^{t_0} \asymp \sqrt{\frac{K(p+\log K)}{\ns}}$, the rates above are the desired rates. In the following, we will derive the results for the case $t \geq t_0+1$. 

Define
\begin{align}
	\xi^{(k)}_{t_0} &= C''\sqrt{\frac{K(p+\log K)}{\ns}} + C''\cdot h \wedge \sqrt{\frac{p+\log K}{n_k}} + C''\epsilon \sqrt{\frac{p+\log K}{\max_{k=1:K}n_k}} + C''\sqrt{\frac{\log K}{n_k}},\\
	\overline{\xi}_{t_0} = \sum_{k \in S}\frac{n_k}{\ns}\xi^{(k)}_{t_0} &\leq C''\sqrt{\frac{K(p+\log K)}{\ns}} + C''\cdot h \wedge \sqrt{\frac{K(p+\log K)}{\ns}} + C''\epsilon \sqrt{\frac{K(p+\log K)}{\ns}} \\
	&\quad + C''\sqrt{\frac{K\log K}{\ns}}.
\end{align}
Consider an event $\mathcal{E}_{t_0}$ defined to be the intersection of the events in Lemmas \ref{lem: concentration w}, \ref{lem: concentration delta}, \ref{lem: concentration beta}, and \ref{lem: concentration beta 2}, with $\xi^{(k)} = \xi^{(k)}_{t_0}$, which satisfies $\tp(\mathcal{E}_{t_0}) \geq 1-C'(K^{-2}+K^{-2}e^{-C''p})$. In the following, we condition on $\mathcal{E} \cap \mathcal{E}_{t_0}$, therefore all the arguments hold with probability at least $1-C'K^{-1}$.

Let $t = t_0 + 1$. Since $\lambda^{[t-1]} \geq C\sqrt{p+\log K} \geq C\max_{k \in S}\sqrt{n_k}h$, by Lemma \ref{lem: claim B1}, we also have $\hbeta^{(k)[t-1]} = \obbeta^{[t-1]}$ for all $k \in S$. Similar to \eqref{eq: w stage 1 eq 1},
\begin{align}
	\norm{\hw^{(k)[t]} - \wks{k}} &\leq \norm{\wk{k}(\htheta^{(k)[t-1]}) - \wks{k}} + \norm{\hw^{(k)[t]} - \wk{k}(\htheta^{(k)[t-1]})} \\
	&\leq \kappa_0 d(\htheta^{(k)[t-1]}, \bthetaks{k}) + \norma{\frac{1}{n_k}\sum_{i=1}^{n_k}\gamma_{\htheta^{(k)[t-1]}}(\bzk{k}_i) - \te_{\bzk{k}}[\gamma_{\htheta^{(k)[t-1]}}(\bzk{k})]} \\
	&\leq \kappa_0 d(\htheta^{(k)[t-1]}, \bthetaks{k}) + C''\xi^{(k)}_{t-1}\sqrt{\frac{p}{n_k}} + C''\sqrt{\frac{\log K}{n_k}}\\
	&\leq  \kappa_0 d(\htheta^{(k)[t-1]}, \bthetaks{k}) + \kappa_0\xi^{(k)}_{t-1} + C''\sqrt{\frac{\log K}{n_k}}.
\end{align}
This implies that
\begin{equation}
	\sum_{k\in S} \frac{n_k}{\ns}\norm{\hw^{(k)[t]} - \wks{k}} \leq \kappa_0\sum_{k\in S} \frac{n_k}{\ns} d(\htheta^{(k)[t-1]}, \bthetaks{k}) + \kappa_0\sum_{k\in S} \frac{n_k}{\ns}\xi^{(k)}_{t-1} + C''\sqrt{\frac{K\log K}{\ns}}.
\end{equation}
And by Lemma \ref{lem: claim B1},
\begin{align}
	\twonorm{\hbeta^{(k)[t]} - \bbetaks{k}} &\lesssim \twonorma{\sum_{k \in S}\frac{n_k}{\ns}[\hSigma^{(k)[t]}\bbetaks{k} - (\hmu^{(k)[t]}_2 - \hmu^{(k)[t]}_1)]} + h \wedge \frac{\lambda^{[t]}}{\sqrt{n_k}} + \epsilon \frac{\lambda^{[t]}}{\sqrt{n_k}}\\
	&\lesssim \twonorma{\sum_{k \in S}\frac{n_k}{\ns}[\hSigma^{(k)[t]}\bbetaks{k} - (\hmu^{(k)[t]}_2 - \hmu^{(k)[t]}_1)]} + h \wedge \sqrt{\frac{p+\log K}{n_k}} + \epsilon\sqrt{\frac{p+\log K}{\max_{k=1:K}n_k}}, \label{eq: stage 2 beta eq 1}
\end{align}
where
\begin{align}
	\twonorma{\sum_{k \in S}\frac{n_k}{\ns}[\hSigma^{(k)[t]}\bbetaks{k} - (\hmu^{(k)[t]}_2 - \hmu^{(k)[t]}_1)]} &\leq \twonorma{\sum_{k \in S}\frac{n_k}{\ns}(\hSigma^{(k)[t]} - \bSigmaks{k})\bbetaks{k}} \\
	&\quad+ \twonorma{\sum_{k \in S}\frac{n_k}{\ns}\big(\hmu^{(k)[t]}_2 - \hmu^{(k)[t]}_1 - \bmuks{k}_2 + \bmuks{k}_1\big)}. \label{eq: lambda related eq 1 stage 2}
\end{align}
And the first term on the RHS can be controlled as
\begin{align}
	& \twonorma{\sum_{k \in S}\frac{n_k}{\ns}(\hSigma^{(k)[t]} - \bSigmaks{k})\bbetaks{k}} \\
	&\leq \underbrace{\twonorma{\frac{1}{\ns}\sum_{k \in S}\sum_{i=1}^{n_k}\big[\bzk{k}_i(\bzk{k}_i)^\top - \te [\bzk{k}_i(\bzk{k}_i)^\top]\big]\bbetaks{k}}}_{\circled{4}} \\
	&\quad + \underbrace{\twonorma{\sum_{k \in S}\frac{n_k}{\ns}\big[(1-\hw^{(k)[t]})\hmu_1^{(k)[t]}(\hmu_1^{(k)[t]})^\top - (1-\wks{k})\bmuks{k}_1(\bmuks{k}_1)^\top\big]\bbetaks{k}}}_{\circled{5}} \\
	&\quad + \underbrace{\twonorma{\sum_{k \in S}\frac{n_k}{\ns}\big[\hw^{(k)[t]}\hmu_2^{(k)[t]}(\hmu_2^{(k)[t]})^\top - \wks{k}\bmuks{k}_2(\bmuks{k}_2)^\top\big]\bbetaks{k}}}_{\circled{6}}. \label{eq: stage 2 Sigma}
\end{align}
Conditioned on $\mathcal{E}$, we have
\begin{equation}
	\circled{4} \lesssim \sqrt{\frac{p}{\ns}}.
\end{equation}
And
\begin{align}
	\circled{5} &\leq \underbrace{\twonorma{\sum_{k \in S}\frac{n_k}{\ns}(1-\hw^{(k)[t]})(\hmu^{(k)[t]}_1 - \bmuks{k}_1)\cdot (\hmu^{(k)[t]}_1)^\top\bbetaks{k}}}_{\circled{5}.1} \\
	&\quad + \underbrace{\twonorma{\sum_{k \in S}\frac{n_k}{\ns}\big[(1-\hw^{(k)[t]})\bmuks{k}_1(\hmu^{(k)[t]}_1)^\top - (1-\wks{k})\bmuks{k}_1(\bmuks{k}_1)^\top\big]\bbetaks{k}}}_{\circled{5}.2},
\end{align}
where
\begin{equation}
	\circled{5}.2 \leq \twonorma{\sum_{k \in S}\frac{n_k}{\ns}(\hw^{(k)[t]} - \wks{k})\bmuks{k}_1(\hmu^{(k)[t]}_1)^\top\bbetaks{k}} + \twonorma{\sum_{k \in S}\frac{n_k}{\ns}(1-\hw^{(k)[t]})\bmuks{k}_1(\hmu^{(k)[t]}_1 - \bmuks{k}_1)^\top\bbetaks{k}}.
\end{equation}
For the first term, we have
\begin{align}
	&\twonorma{\sum_{k \in S}\frac{n_k}{\ns}(\hw^{(k)[t]} - \wks{k})\bmuks{k}_1(\hmu^{(k)[t]}_1)^\top\bbetaks{k}} \\
	&\lesssim \sum_{k \in S}\frac{n_k}{\ns}\norm{\wk{k}(\htheta^{(k)[t-1]}) - \wks{k}} \\
	&\quad +\frac{1}{\ns}\twonorma{\sum_{k \in S}\sum_{i=1}^{n_k}\Big\{\gamma_{\htheta^{(k)[t-1]}}(\bzk{k}_i)\bmuks{k}_1(\hmu^{(k)[t]}_1)^\top\bbetaks{k} - \te_{\bzk{k}}\big[\gamma_{\htheta^{(k)[t-1]}}(\bzk{k})\bmuks{k}_1(\hmu^{(k)[t]}_1)^\top\bbetaks{k}\big]\Big\}} \\
	&\lesssim \kappa_0\sum_{k \in S}\frac{n_k}{\ns}d(\htheta^{(k)[t-1]}, \bthetaks{k}) \\
	&\quad + \frac{1}{\ns}\sup_{|\widetilde{w}_k|\leq U}\twonorma{\sum_{k \in S}\sum_{i=1}^{n_k}\widetilde{w}_k\Big\{\gamma_{\htheta^{(k)[t-1]}}(\bzk{k}_i)\bmuks{k}_1 - \te_{\bzk{k}}\big[\gamma_{\htheta^{(k)[t-1]}}(\bzk{k})\bmuks{k}_1\big]\Big\}},
\end{align}
where $U > 0$ is some constant such that $U \geq \twonorm{\hmu^{(k)[t]}_1 - \bmuks{k}_1}\twonorm{\bbetaks{k}} + \twonorm{\bmuks{k}_1}\twonorm{\bbetaks{k}} \geq |(\hmu^{(k)[t]}_1)^\top\bbetaks{k}|$ under event $\mathcal{E}_{t_0}$. Note that the last inequality holds because the expectation $\te_{\bzk{k}}$ is w.r.t. $\bzk{k}$ which is independent of $\hmu^{(k)[t]}$. By Lemma \ref{lem: concentration beta}.(\rom{3}) and the definition of $\mathcal{E}_{t_0}$, the second term can be bounded as
\begin{align}
	\frac{1}{\ns}\sup_{|\widetilde{w}_k|\leq U}\twonorma{\sum_{k \in S}\sum_{i=1}^{n_k}\widetilde{w}_k\Big\{\gamma_{\htheta^{(k)[t-1]}}(\bzk{k}_i)\bmuks{k}_1 - \te_{\bzk{k}}\big[\gamma_{\htheta^{(k)[t-1]}}(\bzk{k})\bmuks{k}_1\big]\Big\}} \lesssim \sqrt{\frac{p+K}{\ns}}.
\end{align}
Therefore,
\begin{equation}
	\twonorma{\sum_{k \in S}\frac{n_k}{\ns}(\hw^{(k)[t]} - \wks{k})\bmuks{k}_1(\hmu^{(k)[t]}_1)^\top\bbetaks{k}} \lesssim \kappa_0\sum_{k \in S}\frac{n_k}{\ns}d(\htheta^{(k)[t-1]}, \bthetaks{k}) + \sqrt{\frac{p+K}{\ns}}.
\end{equation}
On the other hand, by simple calculations, we have
\begin{align}
	&\twonorma{\sum_{k \in S}\frac{n_k}{\ns}(1-\hw^{(k)[t]})\bmuks{k}_1(\hmu^{(k)[t]}_1 - \bmuks{k}_1)^\top\bbetaks{k}} \\
	&\lesssim \kappa_0\sum_{k \in S}\frac{n_k}{\ns}d(\htheta^{(k)[t-1]}, \bthetaks{k}) + \sup_{|\widetilde{w}_k| \leq U} \twonorma{\sum_{k \in S}\frac{n_k}{\ns}\widetilde{w}_k[\wk{k}(\htheta^{(k)[t-1]}) - \hw^{(k)[t]}]\bmuks{k}_1} \\
	&\quad + \frac{1}{\ns}\sup_{|\widetilde{w}_k| \leq U'}\twonorma{\sum_{k \in S}\widetilde{w}_k\sum_{i=1}^{n_k}\big\{\gamma_{\htheta^{(k)[t-1]}}(\bzk{k}_i)(\bzk{k}_i)^\top\bbetaks{k}-\te_{\bzk{k}}[\gamma_{\htheta^{(k)[t-1]}}(\bzk{k})(\bzk{k})^\top\bbetaks{k}]\big\}\bmuks{k}_1} \\
	&\lesssim \kappa_0\sum_{k \in S}\frac{n_k}{\ns}d(\htheta^{(k)[t-1]}, \bthetaks{k}) + \frac{1}{\ns}\sup_{|\widetilde{w}_k| \leq U} \twonorma{\sum_{k \in S}\widetilde{w}_k\sum_{i=1}^{n_k}\big[\gamma_{\htheta^{(k)[t-1]}}(\bzk{k}_i) - \te_{\bzk{k}}[\gamma_{\htheta^{(k)[t-1]}}(\bzk{k})]\big]\bmuks{k}_1} \\
	&\quad + \frac{1}{\ns}\sup_{|\widetilde{w}_k| \leq U'}\sup_{\twonorm{u}\leq 1}\norma{\sum_{k \in S}\widetilde{w}_k\sum_{i=1}^{n_k}\big\{\gamma_{\htheta^{(k)[t-1]}}(\bzk{k}_i)(\bzk{k}_i)^\top\bbetaks{k}-\te_{\bzk{k}}[\gamma_{\htheta^{(k)[t-1]}}(\bzk{k})(\bzk{k})^\top\bbetaks{k}]\big\}(\bmuks{k}_1)^\top\bu} \\
	&\lesssim \kappa_0\sum_{k \in S}\frac{n_k}{\ns}d(\htheta^{(k)[t-1]}, \bthetaks{k}) + \sqrt{\frac{p+K}{\ns}}\\
	&\quad + \frac{1}{\ns}\sup_{|\widetilde{w}_k| \leq U'}\norma{\sum_{k \in S}\widetilde{w}_k\sum_{i=1}^{n_k}\big\{\gamma_{\htheta^{(k)[t-1]}}(\bzk{k}_i)(\bzk{k}_i)^\top\bbetaks{k}-\te_{\bzk{k}}[\gamma_{\htheta^{(k)[t-1]}}(\bzk{k})(\bzk{k})^\top\bbetaks{k}]\big\}} \\
	&\lesssim \kappa_0\sum_{k \in S}\frac{n_k}{\ns}d(\htheta^{(k)[t-1]}, \bthetaks{k}) + \sqrt{\frac{p+K}{\ns}}.
\end{align}
Hence
\begin{equation}
	\circled{5}.2 \lesssim \kappa_0\sum_{k \in S}\frac{n_k}{\ns}d(\htheta^{(k)[t-1]}, \bthetaks{k}) + \sqrt{\frac{p+K}{\ns}}.
\end{equation}
A similar discussion leads to the same bound for $\circled{5}.1$. Therefore,
\begin{equation}
	\circled{5} \lesssim \kappa_0\sum_{k \in S}\frac{n_k}{\ns}d(\htheta^{(k)[t-1]}, \bthetaks{k}) + \sqrt{\frac{p+K}{\ns}}.
\end{equation}
And the same bound holds for $\circled{6}$, which can be shown in the same spirit. Putting all the pieces together,
\begin{equation}
	\twonorma{\sum_{k \in S}\frac{n_k}{\ns}(\hSigma^{(k)[t]} - \bSigmaks{k})\bbetaks{k}} \lesssim \kappa_0\sum_{k \in S}\frac{n_k}{\ns}d(\htheta^{(k)[t-1]}, \bthetaks{k}) + \sqrt{\frac{p+K}{\ns}}.
\end{equation}
Therefore by Lemma \ref{lem: claim B1} and \eqref{eq: stage 2 beta eq 1}, we have $\hbeta^{(k)[t-1]} = \obbeta^{[t-1]}$ for all $k \in S$, and
\begin{align}
	\twonorm{\hbeta^{(k)[t]} - \bbetaks{k}} &\lesssim \kappa_0\sum_{k \in S}\frac{n_k}{\ns}d(\htheta^{(k)[t-1]}, \bthetaks{k}) + \sqrt{\frac{p+K}{\ns}} + h \wedge \sqrt{\frac{p+\log K}{n_k}} + \epsilon\sqrt{\frac{p+\log K}{\max_{k=1:K}n_k}} \\
	&\leq C \kappa_0\cdot \sum_{k \in S}\frac{n_k}{\ns}\xi^{(k)}_{t-1} + \sqrt{\frac{p+K}{\ns}} + h \wedge \sqrt{\frac{p+\log K}{n_k}} + \epsilon\sqrt{\frac{p+\log K}{\max_{k=1:K}n_k}} \\
	&\leq \kappa_0'\overline{\xi}_{t-1}+ C'\sqrt{\frac{p}{\ns}} + C'\cdot h \wedge \sqrt{\frac{p+\log K}{n_k}} + C'\epsilon\sqrt{\frac{p+\log K}{\max_{k=1:K}n_k}}+ C'\sqrt{\frac{K\log K}{\ns}} \\
	&\coloneqq \xi^{(k)}_{t}. \label{eq: xi eq 1}
\end{align}
This entails that
\begin{equation}\label{eq: xi bar eq 1}
	\overline{\xi}_{t} = \sum_{k \in S}\frac{n_k}{\ns}\xi^{(k)}_{t} = \kappa_0'\overline{\xi}_{t-1}+ C'\sqrt{\frac{p}{\ns}} + C'\cdot h \wedge \sqrt{\frac{K(p+\log K)}{\ns}} + C'\epsilon\sqrt{\frac{K(p+\log K)}{\ns}}+ C'\sqrt{\frac{K\log K}{\ns}}
\end{equation}

This implies that
\begin{equation}
	\sum_{k \in S}\frac{n_k}{\ns}\twonorm{\hbeta^{(k)[t]} - \bbetaks{k}} \lesssim \kappa_0\sum_{k \in S}\frac{n_k}{\ns}d(\htheta^{(k)[t-1]}, \bthetaks{k}) + \sqrt{\frac{p+K}{\ns}} + h \wedge \sqrt{\frac{K(p+\log K)}{\ns}} + \epsilon\sqrt{\frac{K(p+\log K)}{\ns}}.
\end{equation}
Also, 
\begin{align}
	\norm{\hdelta^{(k)[t]} - \deltaks{k}} &= \frac{1}{2}\twonorma{(\hbeta^{(k)[t]})^\top(\hmu^{(k)[t]}_1 + \hmu^{(k)[t]}_2) - (\bbetaks{k})^\top(\bmuks{k}_1 + \bmuks{k}_2)} \\
	&\lesssim \twonorm{\hbeta^{(k)[t]} - \bbetaks{k}} + \twonorm{(\bbetaks{k})^\top(\hmu^{(k)[t]}_1 - \bmuks{k}_1)} + \twonorm{(\bbetaks{k})^\top(\hmu^{(k)[t]}_2 - \bmuks{k}_2)} \\
	&\lesssim  \kappa_0\sum_{k \in S}\frac{n_k}{\ns}d(\htheta^{(k)[t-1]}, \bthetaks{k}) + \sqrt{\frac{p+K}{\ns}} + h \wedge \sqrt{\frac{p+\log K}{n_k}} + \epsilon\sqrt{\frac{p+\log K}{n_k}} \\
	&\quad + \twonorma{\frac{1}{n_k}\sum_{i=1}^{n_k}\gamma_{\htheta^{(k)[t-1]}}(\bzk{k}_i)(\bbetaks{k})^\top\bzk{k}_i - \te_{\bzk{k}}\big[\gamma_{\htheta^{(k)[t-1]}}(\bzk{k})(\bbetaks{k})^\top\bzk{k}\big]}	\\
	&\lesssim  \kappa_0\sum_{k \in S}\frac{n_k}{\ns}d(\htheta^{(k)[t-1]}, \bthetaks{k}) + \sqrt{\frac{p+K}{\ns}} + h \wedge \sqrt{\frac{p+\log K}{n_k}} + \epsilon\sqrt{\frac{p+\log K}{n_k}} \\
	&\quad + \xi^{(k)}_{t-1}\sqrt{\frac{p}{n_k}} + \sqrt{\frac{\log K}{n_k}} \\
	&\lesssim  \kappa_0\sum_{k \in S}\frac{n_k}{\ns}d(\htheta^{(k)[t-1]}, \bthetaks{k}) + \sqrt{\frac{p}{\ns}} + h \wedge \sqrt{\frac{p+\log K}{n_k}} + \epsilon\sqrt{\frac{p+\log K}{\max_{k=1:K}n_k}} \\
	&\quad + \kappa_0\xi^{(k)}_{t-1} + \sqrt{\frac{\log K}{n_k}}
\end{align}
Therefore, 
\begin{align}
	\sum_{k \in S}\frac{n_k}{\ns}\norm{\hdelta^{(k)[t]} - \deltaks{k}}&\lesssim  \kappa_0\sum_{k \in S}\frac{n_k}{\ns}d(\htheta^{(k)[t-1]}, \bthetaks{k}) + \sqrt{\frac{p}{\ns}} + h \wedge \sqrt{\frac{K(p+\log K)}{\ns}} \\
	&\quad + \epsilon\sqrt{\frac{K(p+\log K)}{\ns}} + \kappa_0\sum_{k \in S}\frac{n_k}{\ns}\xi^{(k)}_{t-1} + \sqrt{\frac{K\log K}{\ns}}.
\end{align}
Hence
\begin{align}
	\sum_{k \in S}\frac{n_k}{\ns}d(\htheta^{(k)[t]}, \bthetaks{k}) &\leq C\kappa_0\sum_{k \in S}\frac{n_k}{\ns}d(\htheta^{(k)[t-1]}, \bthetaks{k}) + C\sqrt{\frac{p}{\ns}} + C\cdot h \wedge \sqrt{\frac{K(p+\log K)}{\ns}} \\
	&\quad + C\epsilon\sqrt{\frac{K(p+\log K)}{\ns}} + C\kappa_0\overline{\xi}_{t-1} + C\sqrt{\frac{K\log K}{\ns}} \\
	&\leq \kappa_0'\overline{\xi}_{t-1} + C\sqrt{\frac{p}{\ns}} + C\cdot h \wedge \sqrt{\frac{K(p+\log K)}{\ns}} \\
	&\quad + C\epsilon\sqrt{\frac{K(p+\log K)}{\ns}}  + C'\sqrt{\frac{K\log K}{\ns}},
\end{align}
and 
\begin{align}
	d(\htheta^{(k)[t]}, \bthetaks{k}) &\leq C\kappa_0\sum_{k \in S}\frac{n_k}{\ns}d(\htheta^{(k)[t-1]}, \bthetaks{k}) + C\sqrt{\frac{p}{\ns}} + Ch \wedge \sqrt{\frac{p+\log K}{n_k}} + C\epsilon\sqrt{\frac{p+\log K}{\max_{k=1:K}n_k}} \\
	&\quad + C\kappa_0\xi^{(k)}_{t-1} + C\sqrt{\frac{\log K}{n_k}} \\
	&\leq \frac{1}{2}\kappa_0'\sum_{k \in S}\frac{n_k}{\ns}d(\htheta^{(k)[t-1]}, \bthetaks{k}) + C\sqrt{\frac{p}{\ns}} + Ch \wedge \sqrt{\frac{p+\log K}{n_k}} + C\epsilon\sqrt{\frac{p+\log K}{\max_{k=1:K}n_k}} \\
	&\quad + \frac{1}{2}\kappa_0'\xi^{(k)}_{t-1} + C\sqrt{\frac{\log K}{n_k}} \\
	&\leq \kappa_0'\xi^{(k)}_{t-1} + C\sqrt{\frac{p}{\ns}} + Ch \wedge \sqrt{\frac{p+\log K}{n_k}} + C\epsilon\sqrt{\frac{p+\log K}{\max_{k=1:K}n_k}} \\
	&\quad + C\sqrt{\frac{\log K}{n_k}}, \label{eq: d stage x}
\end{align}
where $\kappa_0' = 2C\kappa_0 \in (0, 1)$. 

Therefore, for $t = (t_0+1):(t_0+t_0')$, where $(\kappa_0')^{t_0'} \asymp K^{-1/2}$ hence $t_0' \asymp \log K$, we have updating formulas \eqref{eq: xi eq 1} and \eqref{eq: xi bar eq 1} for $\xi^{(k)}_t$. We can get
\begin{align}
	\overline{\xi}_{t_0+t'}  &= (\kappa_0')^{t'}\overline{\xi}_{t_0}+ C'\sqrt{\frac{p}{\ns}} + C'\cdot h \wedge \sqrt{\frac{K(p+\log K)}{\ns}} + C'\epsilon\sqrt{\frac{K(p+\log K)}{\ns}}+ C'\sqrt{\frac{K\log K}{\ns}} \\
	&\leq C(\kappa_0')^{t'}\sqrt{\frac{K(p+\log K)}{\ns}} + C'\sqrt{\frac{p}{\ns}} + C'\cdot h \wedge \sqrt{\frac{K(p+\log K)}{\ns}} + C'\epsilon\sqrt{\frac{K(p+\log K)}{\ns}} \\
	&\quad + C'\sqrt{\frac{K\log K}{\ns}},
\end{align}
and
\begin{align}
	\xi^{(k)}_{t_0+t'} &= \kappa_0'\overline{\xi}_{t_0+t'-1}+ C'\sqrt{\frac{p}{\ns}} + C'\cdot h \wedge \sqrt{\frac{p+\log K}{n_k}} + C'\epsilon\sqrt{\frac{p+\log K}{n_k}}+ C'\sqrt{\frac{K\log K}{\ns}} \\
	&\leq C(\kappa_0')^{t'}\sqrt{\frac{K(p+\log K)}{\ns}} + C'\sqrt{\frac{p}{\ns}} + C'\cdot h \wedge \sqrt{\frac{p+\log K}{n_k}} + C'\epsilon\sqrt{\frac{p+\log K}{\max_{k=1:K}n_k}} \\
	&\quad + C'\sqrt{\frac{K\log K}{\ns}},
\end{align}
with
\begin{equation}\label{eq: xi stage x+1}
	\xi^{(k)}_{t_0+t_0'} \leq C'\sqrt{\frac{p}{\ns}} + C'\cdot h \wedge \sqrt{\frac{p+\log K}{n_k}} + C'\epsilon\sqrt{\frac{p+\log K}{\max_{k=1:K} n_k}}+ C'\sqrt{\frac{K\log K}{\ns}},
\end{equation}
where the last inequality is due to $(\kappa_0')^{t_0'}\sqrt{\frac{K(p+\log K)}{\ns}} \asymp \sqrt{\frac{p+\log K}{\ns}}$. 

Consider an event series $\{\mathcal{E}_t\}_{t=t_0}^{t_0+t_0'}$ each of which is defined to be the intersection of the events in Lemmas \ref{lem: concentration w}, \ref{lem: concentration delta}, \ref{lem: concentration beta}, and \ref{lem: concentration beta 2}, with $\xi^{(k)} = \xi^{(k)}_t$, which satisfies $\tp(\mathcal{E}_{t}) \geq 1-C'(K^{-2}+K^{-2}e^{-C''p})$ hence $\tp(\bigcap_{t=t_0}^{t_0+t_0'}\mathcal{E}_{t}) \geq 1-C't_0'(K^{-2}+K^{-2}e^{-C''p}) \geq 1-C''(K^{-2}+K^{-2}e^{-C''p})\log K \geq 1-C'K^{-1}$. In the following, we condition on $\mathcal{E} \cap (\cap_{t=t_0}^{t_0+t_0'}\mathcal{E}_t)$, therefore all the arguments hold with probability at least $1-C'K^{-1}$. Therefore, for $t = (t_0+1):(t_0+t_0')$, we have \eqref{eq: d stage x} hold, which leads to
\begin{align}
	d(\htheta^{(k)[t_0+t']}, \bthetaks{k}) &\leq \kappa_0'\xi^{(k)}_{t_0+t'-1} + C\sqrt{\frac{p}{\ns}} + Ch \wedge \sqrt{\frac{p+\log K}{n_k}} + C\epsilon\sqrt{\frac{p+\log K}{\max_{k=1:K} n_k}} + C\sqrt{\frac{\log K}{n_k}} \\
	&\leq C(\kappa_0')^{t'}\sqrt{\frac{K(p+\log K)}{\ns}}+ C\sqrt{\frac{p}{\ns}} + Ch \wedge \sqrt{\frac{p+\log K}{n_k}} + C\epsilon\sqrt{\frac{p+\log K}{\max_{k=1:K} n_k}} \\
	&\quad + C\sqrt{\frac{\log K}{n_k}} \\
	&\leq C'(\kappa_0')^{t'} \cdot t_0^2\kappa^{t_0} + C\sqrt{\frac{p}{\ns}} + Ch \wedge \sqrt{\frac{p+\log K}{n_k}} + C\epsilon\sqrt{\frac{p+\log K}{\max_{k=1:K}n_k}} \\
	&\quad + C\sqrt{\frac{\log K}{n_k}} \\
	&\leq (t_0 + t')^2(\kappa \vee \kappa_0')^{t_0+t'}+ C\sqrt{\frac{p}{\ns}} + Ch \wedge \sqrt{\frac{p+\log K}{n_k}} + C\epsilon\sqrt{\frac{p+\log K}{\max_{k=1:K}n_k}} \\
	&\quad + C\sqrt{\frac{\log K}{n_k}}, 
\end{align} 
where $t' = 1, \ldots, t_0'$, which provides the desired rate for $t = (t_0+1):(t_0+t_0')$. When $t' \geq t_0'$, by \eqref{eq: xi stage x+1}, we have 
\begin{align}
	d(\htheta^{(k)[t_0+t']}, \bthetaks{k}) &\leq (\kappa_0')^{t'-t_0'}\xi^{(k)}_{t_0+t_0'} + C\sqrt{\frac{p}{\ns}} + Ch \wedge \sqrt{\frac{p+\log K}{n_k}} + C\epsilon\sqrt{\frac{p+\log K}{\max_{k=1:K} n_k}} + C\sqrt{\frac{\log K}{n_k}} \\
	&\leq C\sqrt{\frac{p}{\ns}} + Ch \wedge \sqrt{\frac{p+\log K}{n_k}} + C\epsilon\sqrt{\frac{p+\log K}{\max_{k=1:K} n_k}} + C\sqrt{\frac{\log K}{n_k}},
\end{align}
which is the desired rate. We complete the proof for Theorem \ref{thm: upper bound multitask est error}.

\subsubsection{Proofs of lemmas}

\begin{proof}[Proof of Lemma \ref{lem: concentration w}]
We prove part (\rom{1}) first. 

Denote $W = \sup_{\substack{\bthetak{k} \in B_{\text{con}} \\ \twonorm{\bbetak{k} - \bbetaks{k}} \leq \xi^{(k)}}} \norma{\frac{1}{n_k}\sum_{i=1}^{n_k}\gamma_{\bthetak{k}}(\bzk{k}_i) - \te[\gamma_{\bthetak{k}}(\bzk{k})]}$. By bounded difference inequality, 
\begin{equation}\label{eq: W eq 1}
	W \leq \te W + C\sqrt{\frac{\log K}{n_k}},
\end{equation}
with probability at least $1-C'K^{-2}$. By the generalized symmetrization inequality (Proposition 4.11 in \citeapp{wainwright2019high}), with i.i.d. Rademacher variables $\{\epsilonk{k}_i\}_{i=1}^{n_k}$,
\begin{align}
	\te W &\leq \frac{2}{n_k}\te_{\bz}\te_{\bepsilon}\left[\sup_{\substack{\bthetak{k} \in B_{\text{con}} \\ \twonorm{\bbetak{k} - \bbetaks{k}} \leq \xi^{(k)}}} \norma{\sum_{i=1}^{n_k}\gamma_{\bthetak{k}}(\bzk{k}_i)\epsilonk{k}_i}\right] \\
	&\leq \frac{2}{n_k}\te_{\bz}\te_{\bepsilon}\left[\sup_{\substack{\bthetak{k} \in B_{\text{con}} \\ \twonorm{\bbetak{k} - \bbetaks{k}} \leq \xi^{(k)}}} \norma{\sum_{i=1}^{n_k}\frac{\wk{k}\exp\{C_{\bthetak{k}}(\bzk{k}_i)\}}{1-\wk{k} + \wk{k}\exp\{C_{\bthetak{k}}(\bzk{k}_i)\}}\epsilonk{k}_i}\right] \\
	&\leq \frac{2}{n_k}\te_{\bz}\te_{\bepsilon}\left[\sup_{\substack{\bthetak{k} \in B_{\text{con}} \\ \twonorm{\bbetak{k} - \bbetaks{k}} \leq \xi^{(k)}}} \norma{\sum_{i=1}^{n_k}\frac{1}{1 + \exp\{C_{\bthetak{k}}(\bzk{k}_i)-\log((\wk{k})^{-1}-1)\}}\epsilonk{k}_i}\right], \\ \label{eq: W eq 1}
\end{align}
where $C_{\bthetak{k}}(\bzk{k}_i) = (\bbetak{k})^\top\bzk{k}_i - \deltak{k}$. Denote $\bmuks{k} = (1-\wks{k})\bmuks{k}_1 + \wks{k}\bmuk{k}_2 = \te[\bzk{k}_i]$. By the contraction inequality for Rademecher variables (Theorem 11.6 in \citealpapp{boucheron2013concentration}),
\begin{align}
	\text{RHS of \eqref{eq: W eq 1}} &\leq \frac{C}{n_k}\te_{\bz}\te_{\bepsilon}\left[\sup_{\substack{\bthetak{k} \in B_{\text{con}} \\ \twonorm{\bbetak{k} - \bbetaks{k}} \leq \xi^{(k)}}} \norma{\sum_{i=1}^{n_k}\big[C_{\bthetak{k}}(\bzk{k}_i)-\log((\wk{k})^{-1}-1)\big]\epsilonk{k}_i}\right] \\
	&\leq \frac{C}{n_k}\te_{\bz}\te_{\bepsilon}\left[\sup_{\substack{\bthetak{k} \in B_{\text{con}} \\ \twonorm{\bbetak{k} - \bbetaks{k}} \leq \xi^{(k)}}} \norma{\sum_{i=1}^{n_k}(\bbetak{k})^\top(\bzk{k}_i - \bmuks{k})\cdot \epsilonk{k}_i}\right] \\
	&\quad + \frac{C}{n_k}\te_{\bz}\te_{\bepsilon}\left[\sup_{\substack{\bthetak{k} \in B_{\text{con}} \\ \twonorm{\bbetak{k} - \bbetaks{k}} \leq \xi^{(k)}}} \norma{\sum_{i=1}^{n_k}(\bbetak{k})^\top\bmuks{k}\cdot \epsilonk{k}_i}\right] \\
	&\quad + \frac{C}{n_k}\te_{\bepsilon}\left[\sup_{\substack{\bthetak{k} \in B_{\text{con}} \\ \twonorm{\bbetak{k} - \bbetaks{k}} \leq \xi^{(k)}}} \norma{\log((\wk{k})^{-1}-1)}\cdot \norma{\sum_{i=1}^{n_k} \epsilonk{k}_i}\right] \\
	&\leq \frac{C}{n_k}\te_{\bz}\te_{\bepsilon}\left[\sup_{\twonorm{\bbetak{k} - \bbetaks{k}} \leq \xi^{(k)}} \norma{\sum_{i=1}^{n_k}(\bbetak{k}-\bbetaks{k})^\top(\bzk{k}_i - \bmuks{k})\cdot \epsilonk{k}_i}\right] \\
	&\quad + \frac{C}{n_k}\te_{\bz}\te_{\bepsilon}\norma{\sum_{i=1}^{n_k}(\bbetaks{k})^\top(\bzk{k}_i - \bmuks{k})\cdot \epsilonk{k}_i} \\
	&\quad + \frac{C}{n_k}\te_{\bepsilon}\norma{\sum_{i=1}^{n_k} \epsilonk{k}_i} + \frac{C}{n_k}\te_{\bz}\te_{\bepsilon}\left[\sup_{\substack{\bthetak{k} \in B_{\text{con}} \\ \twonorm{\bbetak{k} - \bbetaks{k}} \leq \xi^{(k)}}} \norma{\sum_{i=1}^{n_k}(\bbetak{k})^\top\bmuks{k}\cdot \epsilonk{k}_i}\right].
\end{align}
Since $\{\epsilonk{k}_i\}_{i=1}^{n_k}$ and $\{(\bbetak{k}-\bbetaks{k})^\top(\bzk{k}_i - \bmuks{k})\epsilonk{k}_i\}_{i=1}^{n_k}$ are i.i.d. sub-Gaussian variables, we know that
\begin{equation}
	\frac{C}{n_k}\te_{\bz}\te_{\bepsilon}\norma{\sum_{i=1}^{n_k}(\bbetaks{k})^\top(\bzk{k}_i - \bmuks{k})\cdot \epsilonk{k}_i} +\frac{C}{n_k}\te_{\bepsilon}\norma{\sum_{i=1}^{n_k} \epsilonk{k}_i} \lesssim \sqrt{\frac{1}{n_k}}.
\end{equation}
Suppose $\{\bu_j\}_{j=1}^N$ is a $1/2$-cover of $\mathcal{B}^p \coloneqq \{\bu \in \mathbb{R}^p: \twonorm{\bu} \leq 1\}$ with $N = 5^p$ (see Example 5.8 in \citealpapp{wainwright2019high}). Hence by standard arguments,
\begin{align}
	&\frac{C}{n_k}\te_{\bz}\te_{\bepsilon}\left[\sup_{\twonorm{\bbetak{k} - \bbetaks{k}} \leq \xi^{(k)}} \norma{\sum_{i=1}^{n_k}(\bbetak{k}-\bbetaks{k})^\top(\bzk{k}_i - \bmuks{k})\cdot \epsilonk{k}_i}\right] \\
	&\lesssim \frac{\xi^{(k)}}{n_k}\te_{\bz}\te_{\bepsilon}\left[\sup_{j=1:N}\norma{\sum_{i=1}^{n_k}\bu_j^\top(\bzk{k}_i - \bmuks{k})\cdot \epsilonk{k}_i}\right].
\end{align}
Again, since $\{\bu_j^\top(\bzk{k}_i - \bmuks{k})\epsilonk{k}_i\}_{i=1}^{n_k}$ are i.i.d. sub-Gaussian variables,
\begin{equation}
	\frac{1}{n_k}\te_{\bz}\te_{\bepsilon}\left[\sup_{j=1:N}\norma{\sum_{i=1}^{n_k}\bu_j^\top(\bzk{k}_i - \bmuks{k})\cdot \epsilonk{k}_i}\right] \lesssim \sqrt{\frac{\log N}{n_k}} = \sqrt{\frac{p}{n_k}}.
\end{equation}
Putting all the pieces together,
\begin{equation}\label{eq: EW eq 1}
	\te W \lesssim \xi^{(k)}\sqrt{\frac{p}{n_k}} + \sqrt{\frac{1}{n_k}}.
\end{equation}
Combining \eqref{eq: W eq 1} and \eqref{eq: EW eq 1}, we get the result in (\rom{1}).

Next, we derive part (\rom{2}) using a similar analysis. 

Denote $W' = \sup_{\{\bthetak{k}\}_{k \in S} \in B_{\text{con}}^{J,2}}\sup_{\norm{\widetilde{w}_k} \leq 1} \frac{1}{\ns}\norma{\sum_{k \in S}\widetilde{w}_k\sum_{i=1}^{n_k}\Big[\gamma_{\bthetak{k}}(\bzk{k}_i) - \te[\gamma_{\bthetak{k}}(\bzk{k})]\Big]}$.

By a similar standard symmetrization and contraction arguments we used in part (\rom{1}), with i.i.d. Rademacher variables $\{\epsilonk{k}_i\}_{i=1}^{n_k}$, for any $\lambda \in \mathbb{R}$, we have
\begin{align}
	&\te\exp\{\lambda W'\} \\
	&\leq C\te_{\bz}\te_{\bepsilon} \exp\left\{\frac{2\lambda}{\ns}\sup_{\{\bthetak{k}\}_{k \in S} \in B_{\text{con}}^{J,2}}\sup_{\norm{\widetilde{w}_k} \leq 1}\norma{\sum_{k \in S}\widetilde{w}_k\sum_{i=1}^{n_k}\gamma_{\bthetak{k}}(\bzk{k}_i)\epsilonk{k}_i}\right\} \\
	&\leq C\te_{\bz}\te_{\bepsilon} \exp\left\{\frac{4\lambda}{\ns}\sup_{\{\bthetak{k}\}_{k \in S} \in B_{\text{con}}^{J,2}}\sup_{\widetilde{w}_k = \pm 1/2}\norma{\sum_{k \in S}\widetilde{w}_k\sum_{i=1}^{n_k}\gamma_{\bthetak{k}}(\bzk{k}_i)\epsilonk{k}_i}\right\} \\
	&\leq C\sum_{\widetilde{w}_k = \pm 1/2}\te_{\bz}\te_{\bepsilon}\exp\left\{\frac{4\lambda}{\ns}\sup_{\{\bthetak{k}\}_{k \in S} \in B_{\text{con}}^{J,2}}\norma{\sum_{k \in S}\widetilde{w}_k\sum_{i=1}^{n_k}\gamma_{\bthetak{k}}(\bzk{k}_i)\epsilonk{k}_i}\right\}\\
	&\leq C\sum_{\widetilde{w}_k = \pm 1/2}\te_{\bz}\te_{\bepsilon}\exp\left\{\frac{4\lambda}{\ns}\sup_{\{\bthetak{k}\}_{k \in S} \in B_{\text{con}}^{J,2}}\norma{\sum_{k \in S}\widetilde{w}_k\sum_{i=1}^{n_k}\frac{1}{1 + \exp\{C_{\bthetak{k}}(\bzk{k}_i)-\log((\wk{k})^{-1}-1)\}}\epsilonk{k}_i}\right\} \\
	&\leq C\sum_{\widetilde{w}_k = \pm 1/2}\te_{\bz}\te_{\bepsilon}\exp\left\{\frac{4\lambda}{\ns}\sup_{\{\bthetak{k}\}_{k \in S} \in B_{\text{con}}^{J,2}}\norma{\sum_{k \in S}\widetilde{w}_k\sum_{i=1}^{n_k}\big[C_{\bthetak{k}}(\bzk{k}_i)-\log((\wk{k})^{-1}-1)\big]\epsilonk{k}_i}\right\}, \label{eq: W' eq 1}
\end{align}
where $C_{\bthetak{k}}(\bzk{k}_i) = (\bbetak{k})^\top\bzk{k}_i - \deltak{k}$. Denote $\bmuks{k} = (1-\wks{k})\bmuks{k}_1 + \wks{k}\bmuk{k}_2 = \te[\bzk{k}_i]$. Suppose $\{\bu_j\}_{j=1}^N$ is a $1/2$-cover of $\mathcal{B}^p \coloneqq \{\bu \in \mathbb{R}^p: \twonorm{\bu} \leq 1\}$ with $N = 5^p$. Then by Cauchy-Schwarz inequality and standard arguments,
\begin{align}
	&\te_{\bz, \bepsilon}\exp\left\{\frac{4\lambda}{\ns}\sup_{\{\bthetak{k}\}_{k \in S} \in B_{\text{con}}^{J,2}}\norma{\sum_{k \in S}\widetilde{w}_k\sum_{i=1}^{n_k}\big[C_{\bthetak{k}}(\bzk{k}_i)-\log((\wk{k})^{-1}-1)\big]\epsilonk{k}_i}\right\} \\
	&\lesssim \left[\te_{\bz, \bepsilon}\exp\left\{\frac{C\lambda}{\ns}\sup_{\twonorm{\bbeta}\leq U}\norma{\sum_{k \in S}\sum_{i=1}^{n_k}\widetilde{w}_k\bbeta^\top(\bzk{k}_i - \bmuks{k})\epsilonk{k}_i}\right\}\right]^{1/3} \\
	&\quad + \left[\te_{\bepsilon}\exp\left\{\frac{C\lambda}{\ns}\sup_{\twonorm{\bbeta}\leq U}\norma{\sum_{k \in S}\sum_{i=1}^{n_k}\widetilde{w}_k\bbeta^\top\bmuks{k}\epsilonk{k}_i}\right\}\right]^{1/3} \\
	&\quad + \left[\te_{\bepsilon}\exp\left\{\frac{C\lambda}{\ns}\sup_{c_w/2\leq \wk{k}\leq 1-c_w/2}\norma{\sum_{k \in S}\sum_{i=1}^{n_k}\widetilde{w}_k\log((\wk{k})^{-1}-1)\epsilonk{k}_i}\right\}\right]^{1/3}\\
	&\lesssim \left[\te_{\bz, \bepsilon}\exp\left\{\frac{C\lambda}{\ns}\sup_{j=1:N}\norma{\sum_{k \in S}\sum_{i=1}^{n_k}\widetilde{w}_k\bu_j^\top(\bzk{k}_i - \bmuks{k})\epsilonk{k}_i}\right\}\right]^{1/3} \\
	&\quad + \left[\te_{\bepsilon}\exp\left\{\frac{C\lambda}{\ns}\sup_{j=1:N}\norma{\sum_{k \in S}\sum_{i=1}^{n_k}\widetilde{w}_k\bu_j^\top\bmuks{k}\epsilonk{k}_i}\right\}\right]^{1/3} \\
	&\quad + \left[\te_{\bepsilon}\exp\left\{\frac{C\lambda}{\ns}\sup_{c_w/2\leq \wk{k}\leq 1-c_w/2}\norma{\sum_{k \in S}\sum_{i=1}^{n_k}\widetilde{w}_k\log((\wk{k})^{-1}-1)\epsilonk{k}_i}\right\}\right]^{1/3}\\
	&\lesssim \underbrace{\sum_{j=1}^N\left[\te_{\bz, \bepsilon}\exp\left\{\frac{C\lambda}{\ns}\norma{\sum_{k \in S}\sum_{i=1}^{n_k}\widetilde{w}_k\bu_j^\top(\bzk{k}_i - \bmuks{k})\epsilonk{k}_i}\right\}\right]^{1/3}}_{[1]} \\
	&\quad + \underbrace{\sum_{j=1}^N\left[\te_{\bepsilon}\exp\left\{\frac{C\lambda}{\ns}\norma{\sum_{k \in S}\sum_{i=1}^{n_k}\widetilde{w}_k\bu_j^\top\bmuks{k}\epsilonk{k}_i}\right\}\right]^{1/3}}_{[2]} \\
	&\quad + \underbrace{\left[\te_{\bepsilon}\exp\left\{\frac{C\lambda}{\ns}\sum_{k \in S}\norma{\sum_{i=1}^{n_k}\widetilde{w}_k\epsilonk{k}_i}\right\}\right]^{1/3}}_{[3]}
\end{align}
Since $\{\widetilde{w}_k\bu_j^\top(\bzk{k}_i - \bmuks{k})\epsilonk{k}_i\}_{i,j}$, $\{\widetilde{w}_k(\bbetak{k})^\top\bmuks{k}\epsilonk{k}_i\}_{i,k}$, and $\{\widetilde{w}_k\epsilonk{k}_i\}_{i=1}^{n_k}$ are independent sub-Gaussian variables, we can bound the three terms on the RHS as
\begin{align}
	[1] &\lesssim 5^p \cdot \exp\left\{\frac{C\lambda^2}{\ns}\right\},\\
	[2] &\lesssim 5^p \cdot \exp\left\{\frac{C\lambda^2}{\ns}\right\},\\
	[3] &\leq \left[\prod_{k \in S}\te_{\bepsilon}\exp\left\{\frac{C\lambda}{\ns}\norma{\sum_{i=1}^{n_k}\widetilde{w}_k\epsilonk{k}_i}\right\}\right]^{1/3} \lesssim \left[\prod_{k \in S}\exp\left\{C\frac{\lambda^2}{\ns^2}n_k\right\}\right]^{1/3} \lesssim \exp\left\{\frac{C\lambda^2}{\ns}\right\}.
\end{align}
Putting all pieces together,
\begin{align}
	\te\exp\{\lambda W'\} \leq C'2^K5^p \cdot \exp\left\{\frac{C\lambda^2}{\ns}\right\} = \exp\left\{C\frac{\lambda^2}{\ns} + C''(K + p)\right\}.
\end{align}
Therefore, for any $\delta > 0$
\begin{equation}
	\tp(W' \geq \delta) \leq e^{-\lambda\delta}\te\exp\{\lambda W'\} \leq \exp\left\{C\frac{\lambda^2}{\ns} + C''(K + p) - \lambda\delta\right\}.
\end{equation}
Let $\lambda = \frac{\ns}{2C}\delta$ and $\delta = 4\sqrt{\frac{CC''(K+p)}{\ns}}$, we have 
\begin{equation}
	\tp(W' \geq \delta) \leq \exp\left\{-\frac{\ns}{4C}\delta^2 + C''(K+p)\right\} = \exp\{-3C''(K+p)\} \leq C'K^{-2}\exp\{-3C''p\},
\end{equation}
which completes the proof.
\end{proof}

\begin{proof}[Proof of Lemma \ref{lem: concentration delta}]
	The proof of part (\rom{2}) is the same as the proof of part (\rom{2}) for Lemma \ref{lem: concentration w}, so we omit it. The only difference between the proofs of part (\rom{1}) for two lemmas is that here the bounded difference inequality is not available. Denote
	\begin{equation}
		W = \sup_{\substack{\bthetak{k} \in B_{\text{con}} \\ \twonorm{\bbetak{k} - \bbetaks{k}} \leq \xi^{(k)}}} \norma{\frac{1}{n_k}\sum_{i=1}^{n_k}\big[1-\gamma_{\bthetak{k}}(\bzk{k}_i)\big](\bzk{k}_i)^\top\bbetaks{k} - \te\big[[1-\gamma_{\bthetak{k}}(\bzk{k})](\bzk{k})^\top\bbetaks{k}\big]}.
	\end{equation}
	We need to use Lemma \ref{lem: maurer concentration} to upper bound $W-\te W$. Prior to that, we first verify the conditions required by the lemma. Fix $\bzk{k}_1$, \ldots, $\bzk{k}_{i-1}$, $\bzk{k}_{i+1}$, \ldots, $\bzk{k}_{n_k}$, and define 
	\begin{equation}
		g^{(k)}_i(\bzk{k}_i) = W - \te[W|\bzk{k}_1, \ldots, \bzk{k}_{i-1}, \bzk{k}_{i+1}, \ldots, \bzk{k}_{n_k}].
	\end{equation}
	By triangle inequality,
	\begin{align}
		&\norma{g^{(k)}_i(\bzk{k}_i)} \\
		&= \Bigg|\sup_{\substack{\bthetak{k} \in B_{\text{con}} \\ \twonorm{\bbetak{k} - \bbetaks{k}} \leq \xi^{(k)}}} \bigg|\frac{1}{n_k}\sum_{i=1}^{n_k}\gamma_{\bthetak{k}}(\bzk{k}_i)(\bzk{k}_i)^\top\bbetaks{k} - \te\big[\gamma_{\bthetak{k}}(\bzk{k}_i)(\bzk{k}_i)^\top\bbetaks{k}\big]\bigg|\\
		&\quad - \te \Bigg[\sup_{\substack{\bthetak{k} \in B_{\text{con}} \\ \twonorm{\bbetak{k} - \bbetaks{k}} \leq \xi^{(k)}}} \bigg|\frac{1}{n_k}\sum_{i=1}^{n_k}\gamma_{\bthetak{k}}(\bzk{k}_i)(\bzk{k}_i)^\top\bbetaks{k} - \te\big[\gamma_{\bthetak{k}}(\bzk{k}_i)(\bzk{k}_i)^\top\bbetaks{k}\big]\bigg|\Big|\{\bzk{k}_{i'}\}_{i'\neq i}\Bigg]\Bigg| \\
		&\leq \underbrace{\frac{1}{n_k}\sup_{\substack{\bthetak{k} \in B_{\text{con}} \\ \twonorm{\bbetak{k} - \bbetaks{k}} \leq \xi^{(k)}}} \bigg|\gamma_{\bthetak{k}}(\bzk{k}_i)(\bzk{k}_i)^\top\bbetaks{k}\bigg|}_{W_1} + \underbrace{\frac{2}{n_k}\te\left|\sup_{\substack{\bthetak{k} \in B_{\text{con}} \\ \twonorm{\bbetak{k} - \bbetaks{k}} \leq \xi^{(k)}}}\gamma_{\bthetak{k}}(\bzk{k}_i)(\bzk{k}_i)^\top\bbetaks{k}\right|}_{W_2}.
	\end{align}
	Note that $[\te(W_1+W_2)^d]^{1/d} \leq (\te W_1^d)^{1/d} + (\te W_2^d)^{1/d}$, where
	\begin{align}
		 (\te W_1^d)^{1/d} &\leq \frac{1}{n_k}\left[\te\sup_{\bthetak{k} \in B_{\text{con}}}[\gamma_{\bthetak{k}}(\bzk{k}_i)]^{2d}\right]^{1/2d}\left[\te\big|(\bzk{k}_i)^\top\bbetaks{k}\big|^{2d}\right]^{1/2d} \\
		 &\leq \frac{1}{n_k}\cdot C \cdot \sqrt{d}\\
		 (\te W_2^d)^{1/d} &= \te W_1 \leq (\te W_1^d)^{1/d} \leq \frac{1}{n_k}\cdot C \cdot \sqrt{d}.
	\end{align}
	Therefore $[\te(W_1+W_2)^d]^{1/d} \leq \frac{C}{n_k}\sqrt{d}$. Hence by applying Lemma \ref{lem: maurer concentration}, we have
	\begin{equation}
		W \leq \te W + C\sqrt{\frac{\log K}{n_k}},
	\end{equation} 
	with probability at least $1-C'K^{-2}$.
\end{proof}

\begin{proof}[Proof of Lemma \ref{lem: concentration beta}]
For part (\rom{1}), denote 
\begin{equation}
	W = \sup_{\bthetak{k} \in B_{\text{con}}} \twonorma{\frac{1}{n_k}\sum_{i=1}^{n_k}\gamma_{\bthetak{k}}(\bzk{k}_i)\bzk{k}_i - \te[\gamma_{\bthetak{k}}(\bzk{k})\bzk{k}]}.
\end{equation}
Suppose $\{\bu_j\}_{j=1}^N$ is a $1/2$-cover of $\mathcal{B}^p \coloneqq \{\bu \in \mathbb{R}^p: \twonorm{\bu} \leq 1\}$ with $N = 5^p$. Define $\bmuks{k} = (1-\wks{k})\bmuks{k}_1 + \wks{k}\bmuks{k}_2$. Then by the generalized symmetrization inequality (Proposition 4.11 in \citealpapp{wainwright2019high}), with i.i.d. Rademacher variables $\{\epsilonk{k}_i\}_{i=1}^{n_k}$, for any $\lambda \in \mathbb{R}$,
\begin{align}
	&\te \exp\{\lambda W\} \\
	&\lesssim \te_{\bz,\bepsilon} \exp\left\{\frac{C\lambda}{n_k}\sup_{\bthetak{k} \in B_{\text{con}}}\twonorma{\sum_{i=1}^{n_k}\gamma_{\bthetak{k}}(\bzk{k}_i)\bzk{k}_i\epsilonk{k}_i}\right\} \\
	&\lesssim \te_{\bz,\bepsilon} \exp\left\{\frac{C\lambda}{n_k}\sup_{j=1:N}\sup_{\bthetak{k} \in B_{\text{con}}}\norma{\sum_{i=1}^{n_k}\gamma_{\bthetak{k}}(\bzk{k}_i)(\bzk{k}_i)^\top\bu_j\cdot \epsilonk{k}_i}\right\} \\
	&\lesssim \sum_{j=1}^N\te_{\bz,\bepsilon} \exp\left\{\frac{C\lambda}{n_k}\sup_{\bthetak{k} \in B_{\text{con}}}\norma{\sum_{i=1}^{n_k}\gamma_{\bthetak{k}}(\bzk{k}_i)(\bzk{k}_i)^\top\bu_j\cdot \epsilonk{k}_i}\right\} \\
	&\lesssim \sum_{j=1}^N\te_{\bz,\bepsilon} \exp\left\{\frac{C\lambda}{n_k}\sup_{\bthetak{k} \in B_{\text{con}}}\norma{\sum_{i=1}^{n_k}\big[C_{\bthetak{k}}(\bzk{k}_i) - \log((\wk{k})^{-1}-1)\big](\bzk{k}_i)^\top\bu_j\cdot \epsilonk{k}_i}\right\} \\
	&\lesssim \sum_{j=1}^N\te_{\bz,\bepsilon} \exp\left\{\frac{C\lambda}{n_k}\sup_{\twonorm{\bbetak{k}} \leq U}\norma{\sum_{i=1}^{n_k}(\bbetak{k})^\top(\bzk{k}_i - \bmuks{k})(\bzk{k}_i)^\top\bu_j\cdot \epsilonk{k}_i}\right\} \\
	&\quad + \sum_{j=1}^N\te_{\bz,\bepsilon} \exp\left\{\frac{C\lambda}{n_k}\sup_{\twonorm{\bbetak{k}} \leq U}\norma{\sum_{i=1}^{n_k}\big[(\bbetak{k})^\top\bmuks{k} - \deltak{k}\big](\bzk{k}_i)^\top\bu_j\cdot \epsilonk{k}_i}\right\}\\
	&\quad + \sum_{j=1}^N\te_{\bz,\bepsilon} \exp\left\{\frac{C\lambda}{n_k}\sup_{c_w/2 \leq \wk{k} \leq 1-c_w/2}\norma{\sum_{i=1}^{n_k}\log((\wk{k})^{-1}-1)(\bzk{k}_i)^\top\bu_j\cdot \epsilonk{k}_i}\right\} \\
	&\lesssim \sum_{j=1}^N\sum_{j'=1}^N\te_{\bz,\bepsilon} \exp\left\{\frac{C\lambda}{n_k}\norma{\sum_{i=1}^{n_k}\bu_{j'}^\top(\bzk{k}_i - \bmuks{k})(\bzk{k}_i)^\top\bu_j\cdot \epsilonk{k}_i}\right\} \\
	&\quad + \sum_{j=1}^N\te_{\bz,\bepsilon} \exp\left\{\frac{C\lambda}{n_k}\norma{\sum_{i=1}^{n_k}(\bzk{k}_i)^\top\bu_j\cdot \epsilonk{k}_i}\right\}.
\end{align}
Note that since $\{\bu_{j'}^\top(\bzk{k}_i - \bmuks{k})(\bzk{k}_i)^\top\bu_j\cdot \epsilonk{k}_i\}_{i=1}^{n_k}$ are i.i.d. sub-exponential variables and $\{(\bzk{k}_i)^\top\bu_j\cdot \epsilonk{k}_i\}_{i=1}^{n_k}$ are i.i.d. sub-Gaussian variables, we have
\begin{align}
	\te_{\bz,\bepsilon} \exp\left\{\frac{C\lambda}{n_k}\norma{\sum_{i=1}^{n_k}\bu_{j'}^\top(\bzk{k}_i - \bmuks{k})(\bzk{k}_i)^\top\bu_j\cdot \epsilonk{k}_i}\right\} &\lesssim \exp\left\{C\frac{\lambda^2}{n_k}\right\},\\
	\te_{\bz,\bepsilon} \exp\left\{\frac{C\lambda}{n_k}\norma{\sum_{i=1}^{n_k}(\bzk{k}_i)^\top\bu_j\cdot \epsilonk{k}_i}\right\} &\lesssim \exp\left\{C\frac{\lambda^2}{n_k}\right\},
\end{align}
where the first inequality holds when $\lambda \leq C''n_k$ where $C''$ is small. Therefore,
\begin{equation}
	\te\exp\{\lambda W\} \lesssim \exp\left\{C\frac{\lambda^2}{n_k} + C'p\right\},
\end{equation}
when $\lambda \leq C''n_k$. The desired result follows from Chernoff's bound.

The proofs of parts (\rom{2}) and (\rom{3}) are almost the same as the proofs of part (\rom{2}) of Lemma \ref{lem: concentration w}, so we do not repeat them here.

\end{proof}

\begin{proof}[Proof of Lemma \ref{lem: concentration beta 2}]
Note that
\begin{equation}
	\twonorma{\frac{1}{n_k}\sum_{i=1}^{n_k}\big[\bzk{k}_i(\bzk{k}_i)^\top - \te [\bzk{k}_i(\bzk{k}_i)^\top]\big]\bbetaks{k}} \lesssim \twonorma{\frac{1}{n_k}\sum_{i=1}^{n_k}\big[\bzk{k}_i(\bzk{k}_i)^\top - \te [\bzk{k}_i(\bzk{k}_i)^\top]\big]}.
\end{equation}
The bound of the RHS comes from Theorem 6.5 in \citeapp{wainwright2019high}. And the bound in part (\rom{2}) can be proved in the same way.
\end{proof}


\subsection{Proof of Theorem \ref{thm: lower bound multitask est error}}

\subsubsection{Lemmas}
Recall the parameter space
\begin{equation}
	\overline{\Theta}_{S}(h) 
	= \Big\{\{\overline{\btheta}^{(k)}\}_{k \in S} = \{(\wk{k}, \bmuk{k}_1, \bmuk{k}_2, \bSigmak{k})\}_{k  \in S}: \othetak{k} \in \overline{\Theta}, \inf_{\overline{\bbeta}}\max_{k \in S}\twonorm{\bbetak{k}-\overline{\bbeta}} \leq h\Big\},
\end{equation}
and $\bbetak{k} = (\bSigmak{k})^{-1}(\bmuk{k}_2-\bmuk{k}_1)$ and $\deltak{k} = \frac{1}{2}(\bbetak{k})^\top(\bmuk{k}_1 + \bmuk{k}_2)$. The notation $\overline{\Theta}_{S}(h)$ was first introduced in equation \eqref{eq: parameter space mtl} in Section \ref{subsec: problem mtl} and defines the parameter space used for both the upper and lower bounds of the estimation and mis-clustering errors.

Here, $\overline{\Theta}$ is defined in equation \eqref{eq: parameter space Theta} in Section \ref{subsec: problem mtl} and imposes regularity conditions on $(\wk{k}, \bmuk{k}_1, \bmuk{k}_2, \bSigmak{k})$, as commonly assumed in EM theory (e.g., \citealpapp{balakrishnan2017statistical, yan2017convergence, cai2019chime}). The similarity condition is formalized by requiring that all discriminant coefficients $\bbetak{k}$ for tasks in $S$ lie within Euclidean distance $h$ from some central vector $\obbeta$. We interpret $\obbeta$ as a type of ``global model" or ``average model" in the context of federated multi-task learning \citepapp{kairouz2021advances}, with each individual task-specific model $\bbetak{k}$ allowed to deviate slightly from this global representation. The degree of similarity between the global and individual models determines the potential benefit of multi-task learning. It is also important to note that the global model is defined only over the subset $S$ of non-outlier tasks.

\begin{lemma}[Lemma 8.4 in \citealpapp{cai2019chime}]\label{lem: lem 8.4 with mu}
	For any $\bmu$, $\widetilde{\bmu} \in \mathbb{R}^p$ and $w \in (0, 1)$, denote $\tp_{\bmu} = (1-w)\mathcal{N}(\bmu, \bm{I}_p) + w\mathcal{N}(-\bmu, \bm{I}_p)$ and $\tp_{\widetilde{\bmu}} = (1-w)\mathcal{N}(\widetilde{\bmu}, \bm{I}_p) + w\mathcal{N}(-\widetilde{\bmu}, \bm{I}_p)$. Then
	\begin{equation}
		\textup{KL}(\tp_{\bmu} \|\tp_{\widetilde{\bmu}}) \leq \left(4\twonorm{\bmu}^2+\frac{1}{2}\log\left(\frac{w}{1-w}\right)\right)\cdot 2\twonorm{\bmu - \widetilde{\bmu}}^2.
	\end{equation}
\end{lemma}

\begin{lemma}\label{lem: new lemma 8.4}
	For any $\bmu$, $\bmu'$, $\widetilde{\bmu}$, $\widetilde{\bmu}' \in \mathbb{R}^p$ and $w \in (0, 1)$, denote $\tp_{\bmu, \widetilde{\bmu}} = (1-w)\mathcal{N}(\bmu, \bm{I}_p) + w\mathcal{N}(\widetilde{\bmu}, \bm{I}_p)$ and $\tp_{\bmu', \widetilde{\bmu}'} = (1-w)\mathcal{N}(\bmu', \bm{I}_p) + w\mathcal{N}(\widetilde{\bmu}', \bm{I}_p)$. Then
	\begin{equation}
		\textup{KL}(\tp_{\bmu, \widetilde{\bmu}} \|\tp_{\bmu', \widetilde{\bmu}'}) \leq (1-w)\twonorm{\bmu - \bmu'}^2 + w\twonorm{\widetilde{\bmu} - \widetilde{\bmu}'}^2.
	\end{equation}
\end{lemma}

\begin{lemma}\label{lem: lem 8.4 with w}
	Denote distribution $(1-w)\mathcal{N}(\bmu, \bm{I}_p) + w\mathcal{N}(-\bmu, \bm{I}_p)$ as $\tp_w$ for any $w \in (c_w, 1-c_w)$, where $\bmu \in \mathbb{R}^p$. Then
	\begin{equation}
		\textup{KL}(\tp_w\|\tp_{w'}) \leq \frac{1}{2c_w^2}(w-w')^2.
	\end{equation}
\end{lemma}

\begin{lemma}\label{lem: lem 8.4 with delta}
	Denote distribution $\frac{1}{2}\mathcal{N}((-1/2,\bm{0}_{p-1}^\top)^\top, \bm{I}_p) + \frac{1}{2}\mathcal{N}((1/2+\widetilde{u}, \bm{0}_{p-1}^\top)^\top, \bm{I}_p)$ as $\tp_{\widetilde{u}}$ for any $\widetilde{u} \in [-1,1]$. Then
	\begin{equation}
		\textup{KL}(\tp_{\widetilde{u}}\|\tp_{\widetilde{u}'}) \leq \frac{1}{2}(\widetilde{u}-\widetilde{u}')^2.
	\end{equation}
\end{lemma}

\begin{lemma}\label{lem: multitask d no outlier}
	When there exists an subset $S$ such that $\min_{k \in S} n_k \geq C(p + \log K)$ with some constant $C > 0$, we have
	\begin{align}
		\inf_{\{\htheta^{(k)}\}_{k=1}^K} \sup_{S: |S| \geq s}\sup_{\substack{\{\othetaks{k}\}_{k \in S} \in \overline{\Theta}_S \\ \mathbb{Q}_S}} &\tp\Bigg(\bigcup_{k \in S}\bigg\{d(\htheta^{(k)}, \bthetaks{k}) \geq  C_1\sqrt{\frac{p}{\ns}}+ C_2\sqrt{\frac{\log K}{n_k}}  \\
		&\quad\quad  + C_3 h \wedge \sqrt{\frac{p + \log K}{n_k}}\bigg\}\Bigg) \geq \frac{1}{4}.	
	\end{align}
\end{lemma}

\begin{lemma}\label{lem: multitask d outlier}
	Denote $\widetilde{\epsilon} = \frac{K-s}{s}$. Then
	\begin{align}
		\inf_{\{\htheta^{(k)}\}_{k=1}^K} \sup_{S: |S| \geq s}\sup_{\substack{\{\othetaks{k}\}_{k \in S} \in \overline{\Theta}_S \\ \mathbb{Q}_S}} &\tp\Bigg(\max_{k \in S}d(\htheta^{(k)}, \bthetaks{k}) \geq  C_1\widetilde{\epsilon}\sqrt{\frac{1}{\max_{k=1:K}n_k}}\Bigg) \geq \frac{1}{10}.	
	\end{align}
\end{lemma}

\begin{lemma}[The first variant of Theorem 5.1 in \citealpapp{chen2018robust}]\label{lem: from chen}
	Given a series of distributions $\{\{\tp_{\theta}^{(k)}\}_{k=1}^K: \theta \in \Theta\}$, each of which is indexed by the same parameter $\theta \in \Theta$. Consider $\bxk{k} \sim (1-\widetilde{\epsilon})\tp^{(k)}_{\theta} + \widetilde{\epsilon}\mathbb{Q}^{(k)}$ independently for $k = 1:K$. Denote the joint distribution of $\{\bxk{k}\}_{k=1}^K$ as $\tp_{(\wepsilon, \theta, \{\tq^{(k)}\}_{k=1}^K)}$. Then
	\begin{equation}
		\inf_{\widehat{\theta}} \sup_{\substack{\theta \in \Theta \\ \{\tq^{(k)}\}_{k=1}^K}} \tp_{(\wepsilon, \theta, \{\tq^{(k)}\}_{k=1}^K)}\left(\|\widehat{\theta}-\theta\| \geq C\varpi(\wepsilon, \Theta)\right) \geq \frac{1}{2},
	\end{equation}
	where $\varpi(\wepsilon, \Theta) \coloneqq \sup\big\{\|\theta_1-\theta_2\|: \max_{k = 1:K}d_{\textup{TV}}\big(\tp^{(k)}_{\theta_1}, \tp^{(k)}_{\theta_2}\big) \leq \wepsilon/(1-\wepsilon)\big\}$.
\end{lemma}

\begin{lemma}\label{lem: binomial lower bound}
	Suppose $K-s \geq 1$. Consider two data generating mechanisms:
		\begin{enumerate}[(i)]
			\item $\bxk{k} \sim (1-\wepsilon')\tp_{\theta}^{(k)} + \wepsilon' \mathbb{Q}^{(k)}$ independently for $k = 1:K$, where $\wepsilon' = \frac{K-s}{K}$;
			\item With a preserved set $S \subseteq 1:K$, generate $\{\bxk{k}\}_{k \in S^c} \sim \mathbb{Q}_S$ and $\bxk{k} \sim \tp_{\theta}^{(k)}$ independently for $k \in S$.
		\end{enumerate}
	Denote the joint distributions of $\{\bxk{k}\}_{k=1}^K$ in (\rom{1}) and (\rom{2}) as $\tp_{(\wepsilon, \theta, \{\mathbb{Q}^{(k)}\}_{k=1}^K)}$ and $\tp_{(S, \theta, \mathbb{Q})}$, respectively. We claim that if
	\begin{equation}
		\inf_{\widehat{\theta}} \sup_{\substack{\theta \in \Theta \\ \{\tq^{(k)}\}_{k=1}^K}} \tp_{(\frac{K-s}{50K}, \theta, \{\tq^{(k)}\}_{k=1}^K)}\left(\|\widehat{\theta}-\theta\| \geq C\varpi\left(\frac{K-s}{50K}, \Theta \right)\right) \geq \frac{1}{2},
	\end{equation} 
	then
	\begin{equation}\label{eq: conclusion binomial lemma}
		\inf_{\widehat{\theta}} \sup_{S: |S| \geq s}\sup_{\substack{\theta \in \Theta \\ \tq_S}} \tp_{(S, \theta, \tq_S)}\left(\|\widehat{\theta}-\theta\| \geq C\varpi\left(\frac{K-s}{50K}, \Theta \right)\right) \geq \frac{1}{10},
	\end{equation}
	where $\varpi(\wepsilon, \Theta) \coloneqq \sup\big\{\|\theta_1-\theta_2\|: \max_{k = 1:K}\textup{KL}\big(\tp^{(k)}_{\theta_1}\| \tp^{(k)}_{\theta_2}\big) \leq[ \wepsilon/(1-\wepsilon)]^2\big\}$ for any $\wepsilon \in (0,1)$.
\end{lemma}

\begin{lemma}\label{lem: multitask Sigma no outlier}
	When there exists an subset $S$ such that $\min_{k \in S} n_k \geq C(p \vee \log K)$ with some constant $C > 0$, we have
	\begin{align}
		&\inf_{\{\hSigma^{(k)}\}_{k=1}^K} \sup_{S: |S| \geq s}\sup_{\substack{\{\othetaks{k}\}_{k \in S} \in \overline{\Theta}_S \\ \mathbb{Q}_S}} \tp\Bigg(\bigcup_{k \in S}\bigg\{\min\big\{\twonorm{\hmu^{(k)}_1 - \bmuks{k}_1}\vee \twonorm{\hmu^{(k)}_2 - \bmuks{k}_2}, \\
		&\quad  \twonorm{\hmu^{(k)}_1 - \bmuks{k}_2}\vee \twonorm{\hmu^{(k)}_2 - \bmuks{k}_1}\big\} \vee \twonorm{\hSigma^{(k)} - \bSigmaks{k}} \geq  C\sqrt{\frac{p + \log K}{n_k}}\bigg\}\Bigg) \geq \frac{1}{10}.	
	\end{align}
\end{lemma}
%

\subsubsection{Main proof of Theorem \ref{thm: lower bound multitask est error}}

Combine conclusions of Lemmas \ref{lem: multitask d no outlier} and \ref{lem: multitask d outlier} to get the first lower bound. Lemma \ref{lem: multitask Sigma no outlier} implies the second one.

\subsubsection{Proofs of lemmas}
\begin{proof}[Proof of Lemma \ref{lem: lem 8.4 with w}]
Denote $g(w; \widetilde{z}) = \log \big[(1-w)\widetilde{z} + w\big]$, $g'(w; \widetilde{z}) = \frac{1-\widetilde{z}}{(1-w)\widetilde{z} + w}$, $g''(w; \widetilde{z}) = -\frac{(1-\widetilde{z})^2}{[(1-w)\widetilde{z} + w]^2}$ and $f(w; \bz, \bmu) = \frac{1-w}{(2\pi)^{p/2}}\exp\{-\frac{1}{2}\twonorm{\bz - \bmu}^2\} + \frac{w}{(2\pi)^{p/2}}\exp\{-\frac{1}{2}\twonorm{\bz + \bmu}^2\}$. 

By Taylor expansion,
\begin{equation}
	\log \left[\frac{f(w';\bz, \bmu)}{f(w;\bz, \bmu)}\right] = \frac{\partial \log f(w;\bz, \bmu)}{\partial w}\bigg|_{w} \cdot (w'-w) + \frac{1}{2}\frac{\partial^2 \log f(w;\bz, \bmu)}{\partial w^2}\bigg|_{w_0} \cdot (w'-w)^2,
\end{equation}
where $w_0 = w_0(\bz, \bmu)$ is between $w$ and $w'$. By the property of score function, 
\begin{equation}
	\int \frac{\partial \log f(w;\bz, \bmu)}{\partial w} d\tp_w = 0.
\end{equation}
Besides,
\begin{equation}
	\frac{\partial^2 \log f(w;\bz, \bmu)}{\partial w^2} = \frac{\partial^2 \log \big[f(w;\bz, \bmu)/\big((2\pi)^{-p/2}\exp\{-\frac{1}{2}\twonorm{\bz + \bmu}^2\}\big)\big]}{\partial w^2} = g''(w; \widetilde{z}),
\end{equation}
where $\widetilde{z} = e^{-\bmu^\top \bz}$. Note that
\begin{equation}
	-g''(w; \widetilde{z}) = \frac{1}{(1-w)^2} \cdot \frac{(\widetilde{z}-1)^2}{(\widetilde{z}  + w/(1-w))^2} \leq \frac{1}{c_w^2},
\end{equation}
for any $\widetilde{z} > 0$. Therefore,
\begin{align}
	\textup{KL}(\tp_w\|\tp_{w'}) &= -\int \log \left[\frac{f(w';\bz, \bmu)}{f(w;\bz, \bmu)}\right] d\tp_w \\
	&= -\frac{1}{2}(w'-w)^2 \cdot \int g''(w_0(\bz, \bmu); \widetilde{z}) d\tp_w \\
	&\leq \frac{1}{2c_w^2}(w'-w)^2,
\end{align}
which completes the proof.
\end{proof}

\begin{proof}[Proof of Lemma \ref{lem: lem 8.4 with delta}]
Recall that we denote distribution $\frac{1}{2}\mathcal{N}((-1/2,\bm{0}_{p-1}^\top)^\top, \bm{I}_p) + \frac{1}{2}\mathcal{N}((1/2+\widetilde{u}, \bm{0}_{p-1}^\top)^\top, \bm{I}_p)$ as $\tp_{\widetilde{u}}$ for any $\widetilde{u} \in [-1,1]$. By the bi-convexity of KL divergence, we have
\begin{align}
	\textup{KL}(\tp_{\widetilde{u}}\|\tp_{\widetilde{u}'}) &\leq \frac{1}{2}\textup{KL}(\mathcal{N}((1/2+\widetilde{u}, \bm{0}_{p-1}^\top)^\top, \bm{I}_p)\|\mathcal{N}((1/2+\widetilde{u}', \bm{0}_{p-1}^\top)^\top, \bm{I}_p)) \\
	&= \frac{1}{2}\textup{KL}(\mathcal{N}(1/2+\widetilde{u}, 1)\|\mathcal{N}(1/2+\widetilde{u}', 1)) \\
	&= \frac{1}{2}(\widetilde{u} - \widetilde{u}')^2,
\end{align}
which completes the proof.
\end{proof}

\begin{proof}[Proof of Lemma \ref{lem: multitask d no outlier}]
WLOG, suppose $\Delta \geq 1$. It's easy to see that given any $S$, $\overline{\Theta}_S \supseteq \overline{\Theta}_{S, w} \cup \overline{\Theta}_{S, \bbeta}\cup \overline{\Theta}_{S, \delta} $, where 
\begin{align}
	\overline{\Theta}_{S, w} &= \Big\{\{\overline{\btheta}^{(k)}\}_{k \in S}: \bmuk{k}_1 = \bm{1}_p/\sqrt{p}, \bmuk{k}_2 = - \bmuk{k}_1 = \widetilde{\bmu}, \bSigmak{k} = \bm{I}_p, \wk{k} \in (c_w, 1-c_w)\Big\}, \\
	\overline{\Theta}_{S, \bbeta} &= \Big\{\{\overline{\btheta}^{(k)}\}_{k \in S}: \bSigmak{k} = \bm{I}_p, \wk{k} = \frac{1}{2}, \twonorm{\bmuk{k}_1} \vee \twonorm{\bmuk{k}_2} \leq M, \min_{\bbeta}\max_{k \in S}\twonorm{\bbetak{k}-\bbeta} \leq h\Big\},\\
	\overline{\Theta}_{S, \delta} &= \Big\{\{\overline{\btheta}^{(k)}\}_{k \in S}: \bSigmak{k} = \bm{I}_p, \wk{k} = \frac{1}{2}, \twonorm{\bmuk{k}_1} \vee \twonorm{\bmuk{k}_2} \leq M, \bmuk{k}_1 = -\frac{1}{2}\bmu_0, \bmuk{k}_2 = \frac{1}{2}\bmu_0 + \bm{u}, \\
	&\hspace{3cm} \twonorm{\bu} \leq 1\Big\}.
\end{align}

\noindent(\rom{1}) By fixing an $S$ and a $\mathbb{Q}_S$, we want to show
\begin{equation}
	\inf_{\{\hbeta^{(k)}\}_{k=1}^K} \sup_{\{\othetaks{k}\}_{k \in S} \in \overline{\Theta}_{S, \bbeta}} \tp\Bigg(\bigcup_{k \in S}\bigg\{\twonorm{\hbeta^{(k)} - \bbetaks{k}} \wedge \twonorm{\hbeta^{(k)} + \bbetaks{k}} \geq C\sqrt{\frac{p}{\ns}}\bigg\}\Bigg) \geq \frac{1}{4}
\end{equation}

By Lemma \ref{lem: packing number of sphere quadrant}, $\exists$ a quadrant $\mathcal{Q}_{\bv}$ of $\mathbb{R}^p$ and a $r/8$-packing of $(r\mathcal{S}^p)\cap \mathcal{Q}_{\bv}$ under Euclidean norm: $\{\widetilde{\bmu}_j\}_{j=1}^N$, where $r = (c\sqrt{p/\ns}) \wedge M \leq 1$ with a small constant $c > 0$ and $N \geq (\frac{1}{2})^p 8^{p-1} = \frac{1}{2}\times 4^{p-1} \geq 2^{p-1}$ when $p \geq 2$. For any $\bmu \in \mathbb{R}^p$, denote distribution $\frac{1}{2}\mathcal{N}(\bmu_0+\bmu, \bm{I}_p) + \frac{1}{2}\mathcal{N}(-\bmu_0+\bmu, \bm{I}_p)$ as $\tp_{\bmu}$, where $\bmu_0$ can be any vector in $\mathbb{R}^p$ with $\twonorm{\bmu_0} \geq 1$. Then
\begin{align}
	\text{LHS} &\geq \inf_{\hmu} \sup_{\bmu \in (r\mathcal{S}^p)\cap \mathcal{Q}_{\bv}} \tp\Bigg(\twonorm{\hmu-\bmu} \wedge \twonorm{\hmu+\bmu} \geq C\sqrt{\frac{p}{\ns}}\Bigg) \\
	&\geq \inf_{\hmu} \sup_{\bmu \in (r\mathcal{S}^p)\cap \mathcal{Q}_{\bv}} \tp\Bigg(\twonorm{\hmu-\bmu} \geq C\sqrt{\frac{p}{\ns}}\Bigg), \label{eq: lower bdd eq mu 1}
\end{align}
where the last inequality holds because it suffices to consider estimator $\hmu$ satisfying $\hmu(X)\in (r\mathcal{S}^p)\cap \mathcal{Q}_{\bv}$ almost surely. In addition, for any $\bx$, $\by \in \mathcal{Q}_{\bv}$, $\twonorm{\bx-\by} \leq \twonorm{\bx+\by}$.

By Lemma \ref{lem: lem 8.4 with mu},
\begin{align}
	\text{KL}\left(\prod_{k \in S}\tp_{\widetilde{\bmu}_j}^{\otimes n_k} \cdot \mathbb{Q}_S \bigg\| \prod_{k \in S}\tp_{\widetilde{\bmu}_{j'}}^{\otimes n_k} \cdot \mathbb{Q}_S\right) &= \sum_{k \in S}n_k \text{KL}(\tp_{\widetilde{\bmu}_j} \| \tp_{\widetilde{\bmu}_{j'}})\\
	&\leq \sum_{k \in S}n_k \cdot 8\twonorm{\widetilde{\bmu}_j}^2 \twonorm{\widetilde{\bmu}_j - \widetilde{\bmu}_{j'}}^2 \\
	&\leq 32\ns r^2 \\
	&\leq 32\ns c^2 \cdot \frac{2(p-1)}{\ns}\\
	&\leq \frac{64c^2}{\log 2}\log N.
\end{align}
By Lemma \ref{lem: fano},
\begin{align}
	\text{LHS of \eqref{eq: lower bdd eq mu 1}} \geq 1-\frac{\log 2}{\log N} - \frac{64c^2}{\log 2}\geq 1-\frac{1}{p-1}-\frac{1}{4} \geq \frac{1}{4},
\end{align}
when $C = c/2$, $p \geq 3$ and $c =\sqrt{\log 2}/16$.

\noindent(\rom{2}) By fixing an $S$ and a $\mathbb{Q}_S$, we want to show
\begin{equation}
	\inf_{\{\hbeta^{(k)}\}_{k=1}^K} \sup_{\{\othetaks{k}\}_{k \in S} \in \overline{\Theta}_S} \tp\Bigg(\bigcup_{k \in S}\bigg\{\twonorm{\hbeta^{(k)}-\bbetaks{k}} \wedge \twonorm{\hbeta^{(k)}+\bbetaks{k}} \geq C\bigg[h \wedge \bigg(c\sqrt{\frac{p}{n_k}}\bigg)\bigg]\bigg\}\Bigg) \geq \frac{1}{4}.
\end{equation}
WLOG, suppose $1 \in S$.  We have
\begin{equation}
	\inf_{\hbeta^{(1)}} \sup_{\{\othetaks{k}\}_{k \in S} \in \overline{\Theta}_S} \tp\Bigg(\twonorm{\hbeta^{(1)}-\bbetaks{1}} \wedge \twonorm{\hbeta^{(1)}+\bbetaks{1}} \geq C\bigg[h \wedge \bigg(c\sqrt{\frac{p}{n_1}}\bigg)\bigg]\Bigg) \geq \frac{1}{4},\\ \label{eq: lower bdd eq mu 2}
\end{equation}

By Lemma \ref{lem: packing number of sphere quadrant}, $\exists$ a quadrant $\mathcal{Q}_{\bv}$ of $\mathbb{R}^p$ and a $r/8$-packing of $(r\mathcal{S}^{p-1})\cap \mathcal{Q}_{\bv}$ under Euclidean norm: $\{\widetilde{\bm{\vartheta}}_j\}_{j=1}^N$, where $r = h_{\bbeta} \wedge (c\sqrt{p/n_1}) \wedge M \leq 1$ with a small constant $c > 0$ and $N \geq (\frac{1}{2})^{p-1} 8^{p-2} = \frac{1}{2}\times 4^{p-2} \geq 2^{p-2}$ when $p \geq 3$. WLOG, assume $M \geq 2$. Denote $\widetilde{\bmu}_j = (1, \widetilde{\bm{\vartheta}}_j^\top)^\top \in \mathbb{R}^p$. Let $\bmuks{k}_1 = \widetilde{\bmu} = (1, \bm{0}_{p-1})^\top$ for all $k \in S\backslash \{1\}$. And let $\bmuks{0}_1 = \bmu = (1, \bm{\vartheta})$ with $\bm{\vartheta} \in (r\mathcal{S}^{p-1}) \cap \mathcal{Q}_{\bv}$. For any $\bmu \in \mathbb{R}^p$, denote distribution $\frac{1}{2}\mathcal{N}(\bmu, \bm{I}_p) + \frac{1}{2}\mathcal{N}(-\bmu, \bm{I}_p)$ as $\tp_{\bmu}$. Then similar to the arguments in (\rom{1}),
\begin{align}
	\text{LHS} &\geq \inf_{\hmu} \sup_{\substack{\bm{\vartheta} \in (r\mathcal{S}^{p-1})\cap \mathcal{Q}_{\bv}\\ \bmu = (1, \bm{\vartheta})^\top}} \tp\Bigg(\twonorm{\hmu-\bmu} \wedge \twonorm{\hmu+\bmu} \geq C\bigg[h \wedge \bigg(c\sqrt{\frac{p}{n_1}}\bigg)\bigg]\Bigg) \\
	&\geq \inf_{\hmu} \sup_{\substack{\bm{\vartheta} \in (r\mathcal{S}^{p-1})\cap \mathcal{Q}_{\bv}\\ \bmu = (1, \bm{\vartheta})^\top}}\tp\Bigg(\twonorm{\hmu-\bmu} \geq C\bigg[h \wedge \bigg(c\sqrt{\frac{p}{n_1}}\bigg)\bigg]\Bigg).
\end{align}

Then by Lemma \ref{lem: lem 8.4 with mu},
\begin{align}
	\text{KL}\left(\prod_{k \in S\backslash \{1\}}\tp_{\widetilde{\bmu}}^{\otimes n_k} \cdot \tp_{\widetilde{\bmu}_j}^{\otimes n_1} \cdot \mathbb{Q}_S \bigg\| \prod_{k \in S\backslash \{1\}}\tp_{\widetilde{\bmu}}^{\otimes n_k} \cdot \tp_{\widetilde{\bmu}_{j'}}^{\otimes n_1} \cdot \mathbb{Q}_S \right) &= n_1 \text{KL}(\tp_{\widetilde{\bmu}_j} \| \tp_{\widetilde{\bmu}_{j'}}) \\
	&\leq n_1 \cdot 8\twonorm{\widetilde{\bmu}_j}^2\twonorm{\widetilde{\bmu}_j - \widetilde{\bmu}_{j'}}^2 \\
	&\leq 32n_1 r^2 \\
	&\leq 32n_1 c^2 \cdot \frac{3(p-2)}{n_1} \\
	&\leq \frac{96c^2}{\log 2}\log N,
\end{align}
when $n_1 \geq (c^2 \vee  M^{-2})p$ and $p \geq 3$. By Fano's lemma (See Corollary 2.6 in \citealpapp{tsybakov2009introduction}),
\begin{align}
	\text{LHS of \eqref{eq: lower bdd eq mu 2}} \geq 1-\frac{\log 2}{\log N} - \frac{96c^2}{\log 2} \geq 1-\frac{1}{p-2} - \frac{1}{4}\geq  \frac{1}{4},
\end{align}
when $C = 1/2$, $p \geq 4$ and $c= \sqrt{(\log 2)/384}$.

\noindent(\rom{3}) By fixing an $S$ and a $\mathbb{Q}_S$, we want to show
\begin{equation}
	\inf_{\{\htheta^{(k)}\}_{k=1}^K} \sup_{\{\othetaks{k}\}_{k \in S} \in \overline{\Theta}_S} \tp\Bigg(\bigcup_{k \in S}\bigg\{\twonorm{\hbeta^{(k)}-\bbetaks{k}} \wedge \twonorm{\hbeta^{(k)}+\bbetaks{k}} \geq C\bigg[h \wedge \bigg(c\sqrt{\frac{\log K}{n_k}}\bigg)\bigg]\bigg\}\Bigg) \geq \frac{1}{4}.
\end{equation}
Suppose $\bv = \bm{1}_p$ and denote the associated quadrant $\mathcal{Q}_{\bv} = \mathbb{R}_{+}^p$, $\Upsilon_S = \{\{\bmuk{k}\}_{k \in S}: \bmuk{k} \in \mathbb{R}_{+}^p, \min_{\mu}\max_{k \in S}\twonorm{\bmuk{k}-\bmu} \leq h, \twonorm{\bmuk{k}} \leq M\}$. Let $r_k =  h \wedge (c\sqrt{\log K/n_k})\wedge M$ with a small constant $c > 0$ for $k \in S$. For any $\bm{M} = \{\bmuk{k}\}_{k \in S}$, where $\bmuk{k} \in \mathbb{R}^p$, denote distribution $\prod_{k \in S}\big[\frac{1}{2}\mathcal{N}(\bmuk{k}, \bm{I}_p) + \frac{1}{2}\mathcal{N}(-\bmuk{k}, \bm{I}_p)\big]^{\otimes n_k}$ as $\tp_{\bm{M}}$, and the joint distribution of $\tp_{\bm{M}}$ and $\mathbb{Q}_S$ as $\tp_{\bm{M}} \cdot \mathbb{Q}_S$. And denote distribution $(1-\overline{w})\mathcal{N}(\bmu, \bm{I}_p) + \overline{w}\mathcal{N}(-\bmu, \bm{I}_p)$ as $\tp_{\bmu}$ for any $\bmu \in \mathbb{R}^p$. Similar to the arguments in (\rom{1}), since it suffices to consider the estimators $\{\hmu^{(k)}\}_{k \in S}$ satisfying $\{\hmu^{(k)}\}_{k \in S} \in \Upsilon_S$ almost surely and $\twonorm{\bx-\by} \leq \twonorm{\bx+\by}$ for any $\bx$, $\by \in \mathbb{R}_+^p$, we have
\begin{align}
	\text{LHS} &\geq \inf_{\{\hmu^{(k)}\}_{k \in S}} \sup_{\{\bmuk{k}\}_{k \in S} \in \Upsilon_S} \tp_{\{\bmuk{k}\}_{k \in S}}\cdot \mathbb{Q}_S\Bigg(\bigcup_{k \in S}\bigg\{\twonorm{\hmu^{(k)}-\bmuk{k}} \wedge \twonorm{\hmu^{(k)}+\bmuk{k}}  \\
	&\hspace{8cm} \geq C\bigg[h \wedge \bigg(c\sqrt{\frac{\log K}{n_k}}\bigg)\bigg]\bigg\}\Bigg) \\
	&\geq \inf_{\{\hmu^{(k)}\}_{k \in S}} \sup_{\{\bmuk{k}\}_{k \in S} \in \Upsilon_S} \tp_{\{\bmuk{k}\}_{k \in S}}\cdot \mathbb{Q}_S\Bigg(\bigcup_{k \in S}\bigg\{\twonorm{\hmu^{(k)}-\bmuk{k}} \geq C\bigg[h \wedge \bigg(c\sqrt{\frac{\log K}{n_k}}\bigg)\bigg]\bigg\}\Bigg), \label{eq: lower bdd eq mu 3}
\end{align}
Consider $\bm{M}^{(k)} = \{\bmuk{j}\}_{j \in S}$ where $\bmuk{j} = \frac{r_j}{\sqrt{p-3/4}} \cdot \bm{1}_p + \bmu_0$ for $j \neq k$ and $\bmuk{k} = \frac{r_k}{2\sqrt{p-3/4}} \cdot \bm{1}_p + \bmu_0$, where $\bmu_0 = (1, \bm{0}_{p-1}^\top)^\top$. Define two new ``distances" (which are not rigorously distances because triangle inequalities and the definiteness do not hold) between $\bm{M} = \{\bmuk{k}\}_{k \in S}$ and as $\bm{M}' = \{\bmu'^{(k)}\}_{k \in S}$
\begin{align}
	\widetilde{d}(\bm{M}, \bm{M}') &\coloneqq \sum_{k \in S}\mathds{1}\left(\twonorm{\bmuk{k}-\bmu'^{(k)}} \geq \frac{r_k}{2\sqrt{p-3/4}}\right), \\
	\widetilde{d}'(\bm{M}, \bm{M}') &\coloneqq \sum_{k \in S}\mathds{1}\left(\twonorm{\bmuk{k}-\bmu'^{(k)}} \geq \frac{r_k}{4\sqrt{p-3/4}}\right). 
\end{align} 
Therefore $\widetilde{d}(\bm{M}^{(k)}, \bm{M}^{(k')}) = 2$ when $k \neq k'$. For $\{\hmu^{(k)}\}_{k \in S}$, define $\psi^* = \argmin_{k \in S}\widetilde{d}'(\{\hmu^{(k)}\}_{k \in S},\allowbreak \bm{M}^{(k)})$. Because $\widetilde{d}(\bm{M}_1, \bm{M}_2) \leq \widetilde{d}'(\bm{M}_1, \bm{M}_2) + \widetilde{d}'(\bm{M}_2, \bm{M}_3)$ for any $\bm{M}_1$, $\bm{M}_2$ and $\bm{M}_3$, it's easy to see that
\begin{align}
	&\inf_{\{\hmu^{(k)}\}_{k \in S}} \sup_{\{\bmuk{k}\}_{k \in S} \in \Upsilon_S} \tp_{\{\bmuk{k}\}_{k \in S}}\cdot \mathbb{Q}_S\Bigg(\bigcup_{k \in S}\bigg\{\twonorm{\hmu^{(k)}-\bmuk{k}} \geq \frac{r_k}{4\sqrt{p-3/4}}\bigg\}\Bigg) \\
	&\geq \inf_{\{\hmu^{(k)}\}_{k \in S}} \sup_{k \in S}\tp_{\bm{M}^{(k)}} \left(\widetilde{d}'(\{\hmu^{(k)}_1\}_{k \in S}, \bm{M}^{(k)}) \geq 1\right) \\
	&\geq \inf_{\{\hmu^{(k)}\}_{k \in S}} \sup_{k \in S}\tp_{\bm{M}^{(k)}}\left(\psi^* \neq k\right) \\
	&\geq \inf_{\psi} \sup_{k \in S}\tp_{\bm{M}^{(k)}}\left(\psi \neq k\right). \label{eq: lower bdd eq mu 4}
\end{align}
By Lemma \ref{lem: lem 8.4 with mu},
\begin{align}
	\text{KL}\left(\tp_{\bm{M}^{(k)}} \cdot \mathbb{Q}_S \big\| \tp_{\bm{M}^{(k')}} \cdot \mathbb{Q}_S \right) &= n_k \text{KL}\left(\tp_{\frac{r_k}{\sqrt{p-3/4}}\mathds{1}_p+ \bmu_0}^{\otimes n_k} \| \tp_{\frac{r_k}{2\sqrt{p-3/4}}\mathds{1}_p+ \bmu_0}^{\otimes n_k}\right) \\
	&\quad + n_{k'} \text{KL}\left(\tp_{\frac{r_{k'}}{\sqrt{p-3/4}}\mathds{1}_p+ \bmu_0}^{\otimes n_{k'}} \| \tp_{\frac{r_{k'}}{2\sqrt{p-3/4}}\mathds{1}_p+ \bmu_0}^{\otimes n_{k'}}\right) \\
	&\leq n_k \cdot 8\twonorma{\frac{r_k}{\sqrt{p-3/4}}\mathds{1}_p+ \bmu_0}^2\twonorma{\frac{r_k}{2\sqrt{p-3/4}}\mathds{1}_p}^2 \\
	&\quad + n_{k'} \cdot 8\twonorma{\frac{r_{k'}}{\sqrt{p-3/4}}\mathds{1}_p+ \bmu_0}^2\twonorma{\frac{r_{k'}}{2\sqrt{p-3/4}}\mathds{1}_p}^2 \label{eq: lemma intermediate eq} \\
	&\leq n_k \cdot 8\cdot 2\cdot (2 r_k^2 + 1) \cdot \frac{1}{4}\cdot 2 r_k^2 + n_{k'} \cdot 8\cdot 2\cdot  (2 r_{k'}^2 + 1) \cdot \frac{1}{4}\cdot 2 r_{k'}^2 \\
	&\leq 16c^2\log K,
\end{align}
when $p \geq 3$.
By Fano's lemma (See Corollary 2.6 in \citealpapp{tsybakov2009introduction}),
\begin{align}
	\text{RHS of \eqref{eq: lower bdd eq mu 4}} \geq 1-\frac{\log 2}{\log K} - 16c^2 \geq \frac{1}{4},
\end{align}
when $K \geq 3$, $c = \sqrt{1/160}$, and $\min_{k \in S}n_k \geq (c^2\vee M^{-2})\log K$.

\noindent(\rom{4}) We want to show
\begin{align}
	\inf_{\{\htheta^{(k)}\}_{k=1}^K} \sup_{S: |S| \geq s}\sup_{\substack{\{\othetaks{k}\}_{k \in S} \in \overline{\Theta}_{S, w} \\ \mathbb{Q}_S}} \tp\Bigg(\bigcup_{k \in S}\bigg\{&\norm{\hw^{(k)}-\wks{k}} \wedge \norm{1-\hw^{(k)}-\wks{k}} \\
	&\geq C\sqrt{\frac{\log K}{n_k}}\bigg\}\Bigg) \geq \frac{1}{4}.
\end{align}
The argument is similar to part (\rom{3}). The only two differences here are that the dimension of interested parameter $w$ equals 1, and Lemma \ref{lem: lem 8.4 with mu} is replaced by Lemma \ref{lem: lem 8.4 with w}.

\noindent(\rom{5}) We want to show
\begin{align}
	\inf_{\{\hdelta^{(k)}\}_{k=1}^K} \sup_{S: |S| \geq s}\sup_{\substack{\{\othetaks{k}\}_{k \in S} \in \overline{\Theta}_{S, \delta} \\ \mathbb{Q}_S}} \tp\Bigg(\bigcup_{k \in S}\bigg\{&\norm{\hdelta^{(k)}-\deltaks{k}} \wedge \norm{\hdelta^{(k)}+\deltaks{k}}\\
	&\geq C\sqrt{\frac{\log K}{n_k}}\bigg\}\Bigg) \geq \frac{1}{4}, \label{eq: lower bound eq beta mtl}
\end{align}
The argument is similar to (\rom{3}). The only two differences here are that the dimension of interested parameter $\delta$ equals 1, and Lemma \ref{lem: lem 8.4 with mu} is replaced by Lemma \ref{lem: lem 8.4 with delta}.

Finally, we get the desired conclusion by combining (\rom{1})-(\rom{5}).
\end{proof}

\begin{proof}[Proof of Lemma \ref{lem: multitask d outlier}]
Let $\widetilde{\epsilon} = \frac{K-s}{s}$ and $\wepsilon' = \frac{K-s}{K}$. Since $s/K \geq c > 0$, $\wepsilon \lesssim \wepsilon'$. Denote $\Upsilon_S = \{\{\bmuk{k}\}_{k \in S}: \bmuk{k} \in \mathbb{R}_{+}^p, \min_{\bmu}\max_{k \in S}\twonorm{\bmuk{k}-\bmu} \leq h_{\bbeta}/2, \twonorm{\bmuk{k}} \leq M\}$. For any $\bmu \in \mathbb{R}$, denote distribution $\frac{1}{2}\mathcal{N}(\bmu, \bm{I}_p) + \frac{1}{2}\mathcal{N}(-\bmu, \bm{I}_p)$ as $\tp_{\bmu}$, and denote $\prod_{k \in S}\tp_{\bmuk{k}}^{\otimes n_k}$ as $\tp_{\{\bmuk{k}\}_{k \in S}}$. Note that $\bbetak{k} = 2\bmuk{k}$ for $\tp_{\bmuk{k}}$ with $\{\bmuk{k}\}_{k \in S} \in \Upsilon_S$. Then it suffices to show
\begin{equation}\label{eq: multitask d outlier wts}
	\inf_{\{\hmu^{(k)}\}_{k=1}^K} \sup_{S: |S| \geq s}\sup_{\substack{\{\bmuk{k}\}_{k \in S} \in \Upsilon_S \\ \mathbb{Q}_S}} \tp\Bigg(\max_{k \in S}\twonorm{\hmu^{(k)} - \bmuk{k}} \geq  C_1\widetilde{\epsilon}'\sqrt{\frac{1}{\max_{k=1:K}n_k}}\Bigg) \geq \frac{1}{10},
\end{equation}
where $\tp = \tp_{\{\bmuk{k}\}_{k \in S}} \cdot \mathbb{Q}_S$.
 WLOG, assume $M \geq 1$. For any $\widetilde{\bmu}_1$, $\widetilde{\bmu}_2 \in \mathbb{R}^p$ with $\twonorm{\widetilde{\bmu}_1} = \twonorm{\widetilde{\bmu}_2} = 1$, by Lemma \ref{lem: lem 8.4 with mu},
\begin{equation}
	\max_{k =1:K}\text{KL}\big(\tp_{\widetilde{\bmu}_1}^{\otimes n_k} \| \tp_{\widetilde{\bmu}_2}^{\otimes n_k}\big) \leq \max_{k=1:K}n_k \cdot 8\twonorm{\widetilde{\bmu}_1 - \widetilde{\bmu}_2}^2.
\end{equation}
Let $8\max_{k=1:K}n_k \cdot \twonorm{\widetilde{\bmu}_1 - \widetilde{\bmu}_2}^2 \leq (\frac{\wepsilon'}{1-\wepsilon'})^2$, then $\twonorm{\widetilde{\bmu}_1 - \widetilde{\bmu}_2} \leq C\sqrt{\frac{1}{\max_{k=1:K}n_k}}\wepsilon'$ for some constant $C > 0$. Then \eqref{eq: multitask d outlier wts} follows by Lemma \ref{lem: binomial lower bound}.
\end{proof}

\begin{proof}[Proof of Lemma \ref{lem: from chen}]
The proof is similar to the proof of Theorem 5.1 in \citeapp{chen2018robust}, so we omit it here.
\end{proof}

\begin{proof}[Proof of Lemma \ref{lem: binomial lower bound}]
It's easy to see that
\begin{equation}
	\text{LHS of \eqref{eq: conclusion binomial lemma}} \geq \inf_{\widehat{\theta}}\sup_{\substack{\theta \in \Theta \\ \tq_S}} \te_{S \sim \tp_s}\left[\tp_{(S, \theta, \tq_S)}\left(\|\widehat{\theta}-\theta\| \geq C\varpi\left(\frac{K-s}{50K}, \Theta \right)\right)\right],
\end{equation}
where $\tp_s$ can be any probability measure on all subsets of $1:K$ with $\tp_s(|S|\geq s) = 1$.

Consider a special distribution $\wtp_s$ as $\tp_s$:
\begin{equation}
	\wtp_s(S = S') = \frac{\tp_{|S^c| \sim \text{Bin}(K, \frac{K-s}{50K})}(|S|=|S'|)}{\tp_{|S^c| \sim \text{Bin}(K,\frac{K-s}{50K})}(|S^c|\leq \frac{41(K-s)}{50})}\cdot \frac{1}{\binom{K}{|S'|}},
\end{equation}
for any $S'$ with $|(S')^c| \leq \frac{41(K-s)}{50}$. Given $S$, consider the distribution of $\{\bxk{k}\}_{k=1}^K$ as
\begin{equation}
	\mathbb{P}^S = \prod_{k \in S}\tp^{(k)}_{\theta}\cdot \prod_{k \notin S}\tq^{(k)}
\end{equation}
Then consider the distribution of $\{\bxk{k}\}_{k=1}^K$ as
\begin{equation}
	\tp' = \sum_{S: |S^c| \leq \frac{41(K-s)}{50}}\wtp_s(S) \cdot \mathbb{P}^S.
\end{equation}
It's easy to see that $\tp'$ is the same as $\tp_{(\frac{K-s}{50K}, \theta, \{\tq^{(k)}\}_{k=1}^K)}$ conditioning on the event $\big\{S: |S^c| \leq \frac{41(K-s)}{50}\big\}$. Therefore,
\begin{align}
	&\inf_{\widehat{\theta}}\sup_{\substack{\theta \in \Theta \\ \tq_S}} \te_{S \sim \wtp_s}\left[\tp_{(S, \theta, \tq_S)}\left(\|\widehat{\theta}-\theta\| \geq C\varpi\left(\frac{K-s}{50K}, \Theta \right)\right)\right] \\
	&\geq \inf_{\widehat{\theta}}\sup_{\substack{\theta \in \Theta \\ \{\tq^{(k)}\}_{k=1}^K}} \te_{S \sim \wtp_s}\left[\mathbb{P}^S\left(\|\widehat{\theta}-\theta\| \geq C\varpi\left(\frac{K-s}{50K}, \Theta \right)\right)\right] \\
	&= \inf_{\widehat{\theta}}\sup_{\substack{\theta \in \Theta \\ \{\tq^{(k)}\}_{k=1}^K}} \mathbb{P}'\left(\|\widehat{\theta}-\theta\| \geq C\varpi\left(\frac{K-s}{50K}, \Theta \right)\right) \\
	&\geq \inf_{\widehat{\theta}}\sup_{\substack{\theta \in \Theta \\ \{\tq^{(k)}\}_{k=1}^K}} \tp_{(\frac{K-s}{50K}, \theta, \{\tq^{(k)}\}_{k=1}^K)}\left(\|\widehat{\theta}-\theta\| \geq C\varpi\left(\frac{K-s}{50K}, \Theta \right) \bigg| |S^c| \leq \frac{41(K-s)}{50}\right) \\
	&\geq \inf_{\widehat{\theta}}\sup_{\substack{\theta \in \Theta \\ \{\tq^{(k)}\}_{k=1}^K}} \tp_{(\frac{K-s}{50K}, \theta, \{\tq^{(k)}\}_{k=1}^K)}\left(\|\widehat{\theta}-\theta\| \geq C\varpi\left(\frac{K-s}{50K}, \Theta \right)\right) - \tp_{|S^c| \sim \text{Bin}(K, \frac{K-s}{50K})}\left(|S^c| > \frac{41(K-s)}{50}\right) \\
	&\geq \frac{1}{2} - \exp\left\{-\frac{\frac{1}{2}[\frac{4}{5}(K-s)]^2}{K \cdot \frac{K-s}{50K}\big(1- \frac{K-s}{50K}\big) + \frac{1}{3}\cdot \frac{4}{5}(K-s)}\right\} \\
	&\geq \frac{1}{2}- \exp\left\{-\frac{\frac{1}{2}\cdot (\frac{4}{5})^2}{\frac{1}{50} + \frac{4}{15}}\right\}\\
	&\geq \frac{1}{10},
\end{align} 
where the last third inequality comes from Bernstein's inequality, application of Lemma \ref{lem: lem 8.4 with w} and the fact that $d_{\text{TV}}^2(\tp_{\theta_1}, \tp_{\theta_2}) \leq \text{KL}(\tp_{\theta_1}\| \tp_{\theta_2})$.
\end{proof}

\begin{proof}[Proof of Lemma \ref{lem: multitask Sigma no outlier}]
(\rom{1}) We want to show
\begin{equation}
	\inf_{\{\hSigma^{(k)}\}_{k=1}^K} \sup_{S: |S| \geq s}\sup_{\substack{\{\othetaks{k}\}_{k \in S} \in \overline{\Theta}_{S} \\ \mathbb{Q}_S}} \tp\Bigg(\bigcup_{k \in S}\bigg\{\twonorm{\hSigma^{(k)} - \bSigmaks{k}} \geq  C\sqrt{\frac{p}{n_k}}\bigg\}\Bigg) \geq \frac{1}{10}.	
\end{equation}
Fix $S$ and some $\mathbb{Q}_S$. WLOG, assume $1 \in S$. Then it suffices to show
\begin{equation}
	\inf_{\hSigma^{(k)}} \sup_{\{\othetaks{k}\}_{k \in S} \in \overline{\Theta}_{S}} \tp\Bigg(\twonorm{\hSigma^{(k)} - \bSigmaks{k}} \geq  C\sqrt{\frac{p}{n_k}}\Bigg) \geq \frac{1}{10}.
\end{equation}
Consider a special subset of $\overline{\Theta}_{S}$ as 
\begin{equation}
	\overline{\Theta}_{S, \bSigma} = \{\otheta: w = 1/2, \bmu_1 = \bmu_2 = 0, \bSigma = \bSigma (\bgamma), \bgamma \in \{0, 1\}^p\},
\end{equation}
where 
\begin{equation}
	\bSigma (\bgamma) = \begin{pmatrix}
		\gamma_1 \bm{e}_1^\top \\
		\vdots \\
		\gamma_p \bm{e}_1^\top
	\end{pmatrix}\cdot \tau + \bm{I}_p,
\end{equation}
and $\tau > 0$ is a small constant which we will specify later. For any $\bgamma \in \{0, 1\}^p$, denote $\mathcal{N}(\bm{0}, \bSigma (\bgamma))$ as $\tp_{\bgamma}$. Therefore it suffices to show
\begin{equation}\label{eq: lemma 16 eq 1}
	\inf_{\hSigma^{(1)}} \sup_{\bgamma \in \{0, 1\}^p} \tp\Bigg(\twonorm{\hSigma^{(1)} - \bSigmaks{1}} \geq  C\sqrt{\frac{p}{n_1}}\Bigg) \geq \frac{1}{10}.
\end{equation}
Note that for any $\hSigma^{(1)}$, we can define $\widehat{\bgamma} = \argmin_{\bgamma \in \{0, 1\}^p} \twonorm{\hSigma^{(1)} - \bSigma(\bgamma)}$. Then by triangle inequality and definition of $\widehat{\bgamma}$, $\twonorm{\hSigma^{(1)} - \bSigma(\bgamma)} \geq \twonorm{\bSigma(\widehat{\bgamma}) - \bSigma(\bgamma)}/2$. Therefore
\begin{equation}\label{eq: lemma 16 eq 2}
	\text{LHS of \eqref{eq: lemma 16 eq 1}} \geq \inf_{\widehat{\bgamma} \in \{0,1\}^p} \sup_{\bgamma \in \{0, 1\}^p} \tp\Bigg(\twonorm{\bSigma(\widehat{\bgamma}) - \bSigma(\bgamma)} \geq  2C\sqrt{\frac{p}{n_1}}\Bigg).
\end{equation}
Let $\tau = c\sqrt{1/n_1}$ where $c > 0$ is a small constant. Since $\twonorm{\bSigma(\widehat{\bgamma}) - \bSigma(\bgamma)} \leq \tau$ for any $\widehat{\bgamma}$ and $\bgamma \in \{0, 1\}^p$, by Lemma D.2 in \citeapp{duan2023adaptive},
\begin{equation}\label{eq: lemma 17 eq 2}
	\text{LHS of \eqref{eq: lemma 16 eq 2}} \geq \frac{\inf_{\widehat{\bgamma} \in \{0,1\}^p} \sup_{\bgamma \in \{0, 1\}^p} \te \twonorm{\bSigma(\widehat{\bgamma}) - \bSigma(\bgamma)}^2 - 4C^2\cdot \frac{p}{n_1}}{(c^2-4C^2) \frac{p}{n_1}}.
\end{equation}
Applying Assouad's lemma (Theorem 2.12 in \citealpapp{tsybakov2009introduction} or Lemma 2 in \citealpapp{cai2012optimal}), we get
\begin{align}
	\inf_{\widehat{\bgamma} \in \{0,1\}^p} \sup_{\bgamma \in \{0, 1\}^p} \te \twonorm{\bSigma(\widehat{\bgamma}) - \bSigma(\bgamma)}^2 &\geq \frac{p}{8}\min_{\rho_H(\bgamma, \bgamma') \geq 1}\left[\frac{\twonorm{\bSigma(\bgamma) - \bSigma(\bgamma')}^2}{\rho_H(\bgamma, \bgamma')}\right] \\
	&\quad \cdot \left[1-\max_{\rho_H(\bgamma, \bgamma')=1}\left(\text{KL}(\tp_{\bgamma}^{\otimes n_1}\|\tp_{\bgamma'}^{\otimes n_1})\right)^{1/2}\right], \label{eq: assouad lemma 17}
\end{align}
where $\rho_H$ is the Hamming distance. For the first term on the RHS, it's easy to see that
\begin{equation}
	\twonorm{\bSigma(\bgamma) - \bSigma(\bgamma')}^2 = \tau^2 \rho_H(\bgamma, \bgamma'),
\end{equation}
for any $\bgamma$ and $\bgamma' \in \{0, 1\}^p$.  For the second term, by the density form of Gaussian distribution, we can show that if $\rho_H(\bgamma, \bgamma')=1$, then
\begin{align}
	\text{KL}(\tp_{\bgamma}^{\otimes n_1}\|\tp_{\bgamma'}^{\otimes n_1}) &= n_1\text{KL}(\tp_{\bgamma}\|\tp_{\bgamma'}) \\
	&\leq n_1\cdot \frac{1}{2}\left\{\log(|\bSigma(\bgamma')|/|\bSigma(\bgamma)|) - \text{Tr}\left[(\bSigma(\bgamma)^{-1} - \bSigma(\bgamma')^{-1})\bSigma(\bgamma)\right]\right\} \\
	&\leq n_1\cdot \frac{1}{4}\tau^2 \\ \label{eq: eq 1 lemma 18}
	&\leq \frac{c^2}{4}. 
\end{align}
Plugging this back into \eqref{eq: assouad lemma 17}, combining with \eqref{eq: lemma 17 eq 2}, we have
\begin{equation}
	\text{LHS of \eqref{eq: lemma 16 eq 2}} \geq \frac{c^2\cdot \frac{p}{8n_1}(1-\frac{c}{2}) - 4C^2\cdot \frac{p}{n_1}}{(c^2-4C^2) \frac{p}{n_1}} \geq \frac{1}{10},	
\end{equation}
when $c = 2/9$ and $C \leq c/\sqrt{324}$. 

\noindent(\rom{2}) We want to show
\begin{equation}
	\inf_{\{\hSigma^{(k)}\}_{k=1}^K} \sup_{S: |S| \geq s}\sup_{\substack{\{\othetaks{k}\}_{k \in S} \in \overline{\Theta}_{S} \\ \mathbb{Q}_S}} \tp\Bigg(\bigcup_{k \in S}\bigg\{\twonorm{\hSigma^{(k)} - \bSigmaks{k}} \geq  C\sqrt{\frac{\log K}{n_k}}\bigg\}\Bigg) \geq \frac{1}{10}.	
\end{equation}
The proof idea is similar to part (\rom{3}) of the proof of Lemma \ref{lem: multitask d no outlier}, so we omit the details here. It suffices to consider $\bm{M}^{(k)} = \{\bSigmak{j}\}_{j=1}^K$ where $\bSigmak{j} = \bm{I}_p$ when $j \neq k$ and $\bSigmak{k} = \bm{I}_p + \sqrt{\log K/n_k} \cdot \bm{e}_1\bm{e}_1^\top$.

\end{proof}


\subsection{Proof of Theorem \ref{thm: upper bound multitask classification error}}
We claim that with probability at least $1-CK^{-1}$,
\begin{equation}\label{eq: proof of thm mtl classification error eq 1}
	R_{\othetaks{k}}(\widehat{\mathcal{C}}^{(k)[t]}) - R_{\othetaks{k}}(\mC_{\othetaks{k}}) \lesssim d^2(\htheta^{(k)[t]}, \bthetaks{k}).
\end{equation}
Then the conclusion immediately follows from Theorem \ref{thm: upper bound multitask est error}. Hence it suffices to verify the claim. For convenience, we write $\widehat{\mathcal{C}}^{(k)[t]} = \mC_{\htheta^{(k)[t]}}$ simply as $\mC_{\htheta^{(k)}}$ and $\htheta^{(k)[t]}$ as $\htheta^{(k)}$.

By simple calculations, we have
\begin{align}
	R_{\othetaks{k}}(\mC_{\htheta^{(k)}}) &= (1-\wks{k})\Phi\left(\frac{-\log(\frac{1-\hw^{(k)}}{\hw^{(k)}}) - \hdelta^{(k)}+(\hbeta^{(k)})^\top\bmuks{k}_1}{\sqrt{(\hbeta^{(k)})^\top\bSigmaks{k}\hbeta^{(k)}}}\right) \\
	&\quad + \wks{k}\Phi\left(\frac{\log(\frac{1-\hw^{(k)}}{\hw^{(k)}}) + \hdelta^{(k)} -(\hbeta^{(k)})^\top\bmuks{k}_2}{\sqrt{(\hbeta^{(k)})^\top\bSigmaks{k}\hbeta^{(k)}}}\right),\\
	R_{\othetaks{k}}(\mC_{\bthetaks{k}}) &= (1-\wks{k})\Phi\left(\frac{-\log(\frac{1-\wks{k}}{\wks{k}}) - \deltaks{k}+(\bbetaks{k})^\top\bmuks{k}_1}{\sqrt{(\bbetaks{k})^\top\bSigmaks{k}\bbetaks{k}}}\right) \\
	&\quad + \wks{k}\Phi\left(\frac{\log(\frac{1-\wks{k}}{\wks{k}}) + \deltaks{k} -(\bbetaks{k})^\top\bmuks{k}_2}{\sqrt{(\bbetaks{k})^\top\bSigmaks{k}\bbetaks{k}}}\right).
\end{align}
Then by Taylor expansion,
\begin{align}
	&R_{\othetaks{k}}(\mC_{\htheta^{(k)}}) - R_{\othetaks{k}}(\mC_{\bthetaks{k}}) \leq (1-\wks{k})\Phi'\left(\frac{-\log(\frac{1-\wks{k}}{\wks{k}}) - \deltaks{k}+(\bbetaks{k})^\top\bmuks{k}_1}{\sqrt{(\bbetaks{k})^\top\bSigmaks{k}\bbetaks{k}}}\right) \\
	&\quad \cdot \Bigg[\frac{-\log(\frac{1-\hw^{(k)}}{\hw^{(k)}}) - \hdelta^{(k)}+(\hbeta^{(k)})^\top\bmuks{k}_1}{\sqrt{(\hbeta^{(k)})^\top\bSigmaks{k}\hbeta^{(k)}}} - \frac{-\log(\frac{1-\wks{k}}{\wks{k}}) - \deltaks{k}+(\bbetaks{k})^\top\bmuks{k}_1}{\sqrt{(\bbetaks{k})^\top\bSigmaks{k}\bbetaks{k}}}\Bigg] \\
	&\quad + \wks{k}\Phi'\left(\frac{\log(\frac{1-\wks{k}}{\wks{k}}) + \deltaks{k} -(\bbetaks{k})^\top\bmuks{k}_2}{\sqrt{(\bbetaks{k})^\top\bSigmaks{k}\bbetaks{k}}}\right)\\
	&\cdot \Bigg[\frac{\log(\frac{1-\hw^{(k)}}{\hw^{(k)}}) + \hdelta^{(k)} -(\hbeta^{(k)})^\top\bmuks{k}_2}{\sqrt{(\hbeta^{(k)})^\top\bSigmaks{k}\hbeta^{(k)}}} - \frac{\log(\frac{1-\wks{k}}{\wks{k}}) + \deltaks{k} -(\bbetaks{k})^\top\bmuks{k}_2}{\sqrt{(\bbetaks{k})^\top\bSigmaks{k}\bbetaks{k}}}\Bigg] \\
	&\quad +  C\Bigg[\frac{-\log(\frac{1-\hw^{(k)}}{\hw^{(k)}}) - \hdelta^{(k)}+(\hbeta^{(k)})^\top\bmuks{k}_1}{\sqrt{(\hbeta^{(k)})^\top\bSigmaks{k}\hbeta^{(k)}}} - \frac{-\log(\frac{1-\wks{k}}{\wks{k}}) - \deltaks{k}+(\bbetaks{k})^\top\bmuks{k}_1}{\sqrt{(\bbetaks{k})^\top\bSigmaks{k}\bbetaks{k}}}\Bigg]^2 \\
	&\quad + C\Bigg[\frac{\log(\frac{1-\hw^{(k)}}{\hw^{(k)}}) + \hdelta^{(k)} -(\hbeta^{(k)})^\top\bmuks{k}_2}{\sqrt{(\hbeta^{(k)})^\top\bSigmaks{k}\hbeta^{(k)}}} - \frac{\log(\frac{1-\wks{k}}{\wks{k}}) + \deltaks{k} -(\bbetaks{k})^\top\bmuks{k}_2}{\sqrt{(\bbetaks{k})^\top\bSigmaks{k}\bbetaks{k}}}\Bigg]^2.
\end{align}
Denote $\mathscr{A} = (1-\wks{k})\Phi'\left(\frac{-\log(\frac{1-\wks{k}}{\wks{k}}) - \deltaks{k}+(\bbetaks{k})^\top\bmuks{k}_1}{\sqrt{(\bbetaks{k})^\top\bSigmaks{k}\bbetaks{k}}}\right)\cdot \Bigg[\frac{-\log(\frac{1-\hw^{(k)}}{\hw^{(k)}}) - \hdelta^{(k)}+(\hbeta^{(k)})^\top\bmuks{k}_1}{\sqrt{(\hbeta^{(k)})^\top\bSigmaks{k}\hbeta^{(k)}}} - \frac{-\log(\frac{1-\wks{k}}{\wks{k}}) - \deltaks{k}+(\bbetaks{k})^\top\bmuks{k}_1}{\sqrt{(\bbetaks{k})^\top\bSigmaks{k}\bbetaks{k}}}\Bigg]+ \wks{k}\Phi'\left(\frac{\log(\frac{1-\wks{k}}{\wks{k}}) + \deltaks{k} -(\bbetaks{k})^\top\bmuks{k}_2}{\sqrt{(\bbetaks{k})^\top\bSigmaks{k}\bbetaks{k}}}\right)\cdot \Bigg[\frac{\log(\frac{1-\hw^{(k)}}{\hw^{(k)}}) + \hdelta^{(k)} -(\hbeta^{(k)})^\top\bmuks{k}_2}{\sqrt{(\hbeta^{(k)})^\top\bSigmaks{k}\hbeta^{(k)}}} - \frac{\log(\frac{1-\wks{k}}{\wks{k}}) + \deltaks{k} -(\bbetaks{k})^\top\bmuks{k}_2}{\sqrt{(\bbetaks{k})^\top\bSigmaks{k}\bbetaks{k}}}\Bigg]$ and $\mathscr{B} = C\Bigg[\frac{-\log(\frac{1-\hw^{(k)}}{\hw^{(k)}}) - \hdelta^{(k)}+(\hbeta^{(k)})^\top\bmuks{k}_1}{\sqrt{(\hbeta^{(k)})^\top\bSigmaks{k}\hbeta^{(k)}}} - \frac{-\log(\frac{1-\wks{k}}{\wks{k}}) - \deltaks{k}+(\bbetaks{k})^\top\bmuks{k}_1}{\sqrt{(\bbetaks{k})^\top\bSigmaks{k}\bbetaks{k}}}\Bigg]^2 + C\Bigg[\frac{\log(\frac{1-\hw^{(k)}}{\hw^{(k)}}) + \hdelta^{(k)} -(\hbeta^{(k)})^\top\bmuks{k}_2}{\sqrt{(\hbeta^{(k)})^\top\bSigmaks{k}\hbeta^{(k)}}} - \frac{\log(\frac{1-\wks{k}}{\wks{k}}) + \deltaks{k} -(\bbetaks{k})^\top\bmuks{k}_2}{\sqrt{(\bbetaks{k})^\top\bSigmaks{k}\bbetaks{k}}}\Bigg]^2$. By plugging in the density formula of standard Gaussian distribution, it is easy to see that
\begin{align}
	\mathscr{A} &\lesssim \sqrt{(1-\wks{k})\wks{k}}\cdot \exp\left\{-\frac{\big[\log(\frac{1-\wks{k}}{\wks{k}}) + \frac{1}{2}(\bbetaks{k})^\top\bSigmaks{k}\bbetaks{k}\big]^2}{2(\bbetaks{k})^\top\bSigmaks{k}\bbetaks{k}} + \frac{1}{2}\log\bigg(\frac{1-\wks{k}}{\wks{k}}\bigg)\right\} \\
	&\quad \cdot \Bigg[\frac{-\log(\frac{1-\hw^{(k)}}{\hw^{(k)}}) - \hdelta^{(k)}+(\hbeta^{(k)})^\top\bmuks{k}_1}{\sqrt{(\hbeta^{(k)})^\top\bSigmaks{k}\hbeta^{(k)}}} - \frac{-\log(\frac{1-\wks{k}}{\wks{k}}) - \deltaks{k}+(\bbetaks{k})^\top\bmuks{k}_1}{\sqrt{(\bbetaks{k})^\top\bSigmaks{k}\bbetaks{k}}}\Bigg]\\
	&\quad + \sqrt{(1-\wks{k})\wks{k}}\cdot \exp\left\{-\frac{\big[\log(\frac{1-\wks{k}}{\wks{k}}) - \frac{1}{2}(\bbetaks{k})^\top\bSigmaks{k}\bbetaks{k}\big]^2}{2(\bbetaks{k})^\top\bSigmaks{k}\bbetaks{k}} + \frac{1}{2}\log\bigg(\frac{\wks{k}}{1-\wks{k}}\bigg)\right\} \\
	&\quad\cdot \Bigg[\frac{\log(\frac{1-\hw^{(k)}}{\hw^{(k)}}) + \hdelta^{(k)} -(\hbeta^{(k)})^\top\bmuks{k}_2}{\sqrt{(\hbeta^{(k)})^\top\bSigmaks{k}\hbeta^{(k)}}} - \frac{\log(\frac{1-\wks{k}}{\wks{k}}) + \deltaks{k} -(\bbetaks{k})^\top\bmuks{k}_2}{\sqrt{(\bbetaks{k})^\top\bSigmaks{k}\bbetaks{k}}}\Bigg] \\
	&= \sqrt{(1-\wks{k})\wks{k}}\cdot \exp\left\{-\frac{1}{8}[(\bbetaks{k})^\top\bSigmaks{k}\bbetaks{k}]^2 - \frac{1}{2}\cdot \frac{\log^2(\frac{1-\wks{k}}{\wks{k}})}{(\bbetaks{k})^\top\bSigmaks{k}\bbetaks{k}}\right\} \\
	&\quad \cdot \norma{\frac{(\hbeta^{(k)})^\top(\bmuks{k}_1-\bmuks{k}_2)}{\sqrt{(\hbeta^{(k)})^\top\bSigmaks{k}\hbeta^{(k)}}} - \frac{(\bbetaks{k})^\top(\bmuks{k}_1-\bmuks{k}_2)}{\sqrt{(\bbetaks{k})^\top\bSigmaks{k}\bbetaks{k}}}} \\
	&\lesssim \norma{\frac{(\widehat{\bm{\xi}}^{(k)})^\top\bm{\xi}^{(k)*}}{\twonorm{\widehat{\bm{\xi}}^{(k)}}} - \twonorm{\bm{\xi}^{(k)*}}} \\
	&\lesssim \twonorm{\widehat{\bm{\xi}}^{(k)} - \bm{\xi}^{(k)*}}^2, \label{eq: proof of thm mtl classification error eq 2}
\end{align}
with probability at least $1-C'K^{-1}$, where $\widehat{\bm{\xi}}^{(k)} = (\bSigmaks{k})^{1/2}\hbeta^{(k)}$ and $\bm{\xi}^{(k)*} = (\bSigmaks{k})^{1/2}\bbetaks{k}$, so $\twonorm{\widehat{\bm{\xi}}^{(k)} - \bm{\xi}^{(k)*}}^2 \lesssim \twonorm{\hbeta^{(k)}-\bbetaks{k}}^2$. Only the last inequality in \eqref{eq: proof of thm mtl classification error eq 2} holds with high probability and the others are deterministic. It comes from the fact that $\twonorm{\widehat{\bm{\xi}}^{(k)} - \bm{\xi}^{(k)*}} \leq c \leq \twonorm{\bm{\xi}^{(k)*}}$ for some $c > 0$ with probability at least $1-C'K^{-1}$ and a direct application of Lemma 8.1 in \citeapp{cai2019chime}. On the other hand, it is easy to see that $\mathscr{B} \lesssim d^2(\htheta^{(k)}, \bthetaks{k})$. Combining these two facts leads to \eqref{eq: proof of thm mtl classification error eq 1}.

\subsection{Proof of Theorem \ref{thm: lower bound multitask classification error}}
\subsubsection{Lemmas}

Recall that for GMM associated with parameter set $\otheta = (w, \bmu_1, \bmu_2, \bSigma)$, we define the mis-clustering error rate of any classifier $\mC$ as $R_{\otheta}(\mC) = \min_{\pi: \{1,2\} \rightarrow \{1, 2\}}\tp_{\otheta}(\mC(Z^{\textup{new}}) \neq \pi(Y^{\textup{new}}))$, where $\tp_{\otheta}$ represents the distribution of $(Z^{\textup{new}}, Y^{\textup{new}})$, i.e. $(1-w)\mathcal{N}(\bmu_1,\bSigma) + w\mathcal{N}(\bmu_2,\bSigma)$. Denote $\mC_{\otheta}$ as the Bayes classifier corresponding to $\otheta$. Define a surrogate loss $L_{\otheta}(\mC) =  \min_{\pi: \{1,2\} \rightarrow \{1, 2\}}\tp_{\otheta}(\mC(Z^{\textup{new}}) \neq \pi(\mC_{\otheta}(Y^{\textup{new}})))$.

\begin{lemma}\label{lem: multitask misclassification error no outlier}
	Assume there exists an subset $S$ such that $\min_{k \in S} n_k \geq C(p \vee \log K)$ and $\min_{k \in S}\Delta^{(k)} \geq \sigma^2 > 0$ with some constants $C > 0$. We have
	\begin{align}
		\inf_{\{\widehat{\mC}^{(k)}\}_{k=1}^K} \sup_{S: |S| \geq s}\sup_{\substack{\{\othetaks{k}\}_{k \in S} \in \overline{\Theta}_{S} \\ \mathbb{Q}_S}} &\tp\Bigg(\bigcup_{k \in S}\bigg\{R_{\othetaks{k}}(\widehat{\mC}^{(k)}) - R_{\othetaks{k}}(\mC_{\othetaks{k}}) \geq  C_1\frac{p}{\ns} + C_2\frac{\log K}{n_k} \\
		&\quad\quad  + C_3 h^2 \wedge \frac{p+\log K}{n_k} + C_4\frac{\epsilon^2}{\max_{k \in S}n_k}\bigg\}\Bigg) \geq \frac{1}{10}.	
	\end{align}
\end{lemma}

\begin{lemma}\label{lem: surrogate loss and classification error}
	Suppose $\otheta = (w, \bmu_1, \bmu_2, \bbeta, \bSigma)$ satisfies $\Delta^2 \coloneqq (\bmu_1 - \bmu_2)^\top\bSigma^{-1}(\bmu_1 - \bmu_2) \geq \sigma^2 > 0$ with some constant $\sigma^2 > 0$ and $w, w' \in (c_w, 1-c_w)$. Then $\exists c > 0$ such that
	\begin{equation}
		cL_{\otheta}^2(\mC) \leq R_{\otheta}(\mC) - R_{\otheta}(\mC_{\otheta}),
	\end{equation}
	for any classifier $\mC$, where $R_{\otheta}(\mC) \coloneqq \min_{\pi: 1:2 \rightarrow 1:2}\tp_{\otheta}(\mC(\bz) \neq \pi(y))$, $L_{\otheta}(\mC) \coloneqq \min_{\pi: \{0,1\} \rightarrow \{0, 1\}} \allowbreak\tp_{\otheta}(\mC(\bz) \neq \pi(\mC_{\otheta}(\bz)))$, and $\mC_{\otheta}$ is the corresponding Bayes classifier.
\end{lemma}

\begin{lemma}\label{lem: surrogate loss mismatch bound}
	Consider $\otheta = (w, \bmu_1, \bmu_2, \bbeta, \bSigma)$ and $\otheta' = (w', \bmu_1, \bmu_2, \bbeta, \bSigma)$ satisfies $\Delta^2 \coloneqq (\bmu_1 - \bmu_2)^\top\bSigma^{-1}(\bmu_1 - \bmu_2) \geq \sigma^2 > 0$ with some constant $\sigma^2 > 0$.. We have
	\begin{equation}
		c\norm{w-w'} \leq L_{\otheta}(\mC_{\otheta'}) \leq c' \norm{w - w'},
	\end{equation}
	for some constants $c$, $c' > 0$.
\end{lemma}

\begin{lemma}\label{lem: surrogate loss mismatch bound delta}
	Consider $\otheta = (w, \bmu_1, \bmu_2, \bSigma)$ and $\otheta' = (w, \bmu_1', \bmu_2',  \bSigma)$ satisfies $w = 1/2$, $\bmu_1 = -\bmu_0/2 + \bu$, $\bmu_2 = \bmu_0/2 + \bu$, $\bmu_1' = -\bmu_0/2 + \bu'$, $\bmu_2' = \bmu_0/2 + \bu'$, $\bSigma = \bm{I}_p$, $\bmu_0 = (1, \bm{0}_{p-1}^\top)^\top$, $\bu = (\widetilde{u}, \bm{0}_{p-1}^\top)^\top$, and $\bu' = (\widetilde{u}', \bm{0}_{p-1}^\top)^\top$. We have
	\begin{equation}
		c\norm{\widetilde{u}-\widetilde{u}'} \leq L_{\otheta}(\mC_{\otheta'}) \leq c' \norm{\widetilde{u}-\widetilde{u}'},
	\end{equation}
	for some constants $c$, $c' > 0$.
\end{lemma}

\begin{lemma}\label{lem: multitask misclassification error outlier}
	Denote $\widetilde{\epsilon} = \frac{K-s}{s}$. We have
	\begin{equation}
		\inf_{\{\widehat{\mC}^{(k)}\}_{k=1}^K} \sup_{S: |S| \geq s}\sup_{\substack{\{\othetaks{k}\}_{k \in S} \in \overline{\Theta}_{S} \\ \mathbb{Q}_S}} \tp\Bigg(\max_{k \in S}\left[R_{\othetaks{k}}(\widehat{\mC}^{(k)}) - R_{\othetaks{k}}(\mC_{\othetaks{k}})\right] \geq  C_1\frac{\widetilde{\epsilon}'^2}{\max_{k=1:K}n_k}\Bigg) \geq \frac{1}{10}.
	\end{equation}
\end{lemma}

\subsubsection{Main proof of Theorem \ref{thm: lower bound multitask classification error}}
Combine conclusions of Lemmas \ref{lem: multitask misclassification error no outlier} and \ref{lem: multitask misclassification error outlier} to get the lower bound.

\subsubsection{Proof of lemmas}

\begin{proof}[Proof of Lemma \ref{lem: multitask misclassification error no outlier}]
Recall the definitions and proof idea of Lemma \ref{lem: multitask d no outlier}.
We have $\overline{\Theta}_{S} \supseteq \overline{\Theta}_{|S|, w} \cup \overline{\Theta}_{|S|, \delta} \cup \overline{\Theta}_{|S|, \bbeta}$, where 
\begin{align}
	\overline{\Theta}_{S, w} &= \Big\{\{\overline{\btheta}^{(k)}\}_{k \in S}: \bmuk{k}_1 = \bm{1}_p/\sqrt{p}, \bmuk{k}_2 = - \bmuk{k}_1 = \widetilde{\bmu}, \bSigmak{k} = \bm{I}_p, \wk{k} \in (c_w, 1-c_w)\Big\}, \\
	\overline{\Theta}_{S, \bbeta} &= \Big\{\{\overline{\btheta}^{(k)}\}_{k \in S}: \bSigmak{k} = \bm{I}_p, \wk{k} = \frac{1}{2}, \twonorm{\bmuk{k}_1} \vee \twonorm{\bmuk{k}_2} \leq M, \bmuk{k}_2 = - \bmuk{k}_1, \\
	&\hspace{3cm} \min_{\bbeta}\max_{k \in S}\twonorm{\bbetak{k}-\bbeta} \leq h\Big\},\\
	\overline{\Theta}_{S, \delta} &= \Big\{\{\overline{\btheta}^{(k)}\}_{k \in S}: \bSigmak{k} = \bm{I}_p, \wk{k} = \frac{1}{2}, \twonorm{\bmuk{k}_1} \vee \twonorm{\bmuk{k}_2} \leq M, \bmuk{k}_1 = -\frac{1}{2}\bmu_0, \bmuk{k}_2 = \frac{1}{2}\bmu_0 + \bm{u}, \\
	&\hspace{3cm} \twonorm{\bu} \leq 1\Big\}.
\end{align}
Recall the mis-clustering error for GMM associated with parameter set $\otheta$ of any classifier $\mC$ is $R_{\otheta}(\mC) = \min_{\pi: \{1,2\} \rightarrow \{1,2\}}\tp_{\otheta}(\mC(Z) \neq \pi(Y))$. To help the analysis, following \citeapp{azizyan2013minimax} and \citeapp{cai2019chime}, we define a surrogate loss $L_{\otheta}(\mC) = \min_{\pi: \{1,2\} \rightarrow \{1,2\}}\tp_{\otheta}(\mC(Z) \neq \pi(\mC_{\otheta}(Z)))$, where $\mC_{\otheta}$ is the Bayes classifier. Suppose $\sigma = \sqrt{0.005}$.

\noindent(\rom{1}) We want to show
\begin{equation}\label{eq: lemma 19 eq 1}
	\inf_{\{\widehat{\mC}^{(k)}\}_{k=1}^K} \sup_{S: |S| \geq s}\sup_{\substack{\{\othetaks{k}\}_{k \in S} \in \overline{\Theta}_{S} \\ \mathbb{Q}_S}} \tp\Bigg(\bigcup_{k \in S}\bigg\{R_{\othetaks{k}}(\widehat{\mC}^{(k)}) - R_{\othetaks{k}}(\mC_{\othetaks{k}}) \geq  C\sqrt{\frac{p}{\ns}}\bigg\} \Bigg) \geq \frac{1}{4}.
\end{equation}
Consider $S = 1:K$ and space $\overline{\Theta}_0 = \{\{\overline{\btheta}^{(k)}\}_{k=1}^K: \bSigmak{k} = \bm{I}_p, \wk{k} = 1/2, \bmuk{k}_1 = \bmu_1, \bmuk{k}_2 = \bmu_2, \twonorm{\bmu_1} \vee \twonorm{\bmu_2} \leq M\}$. And 
\begin{equation}
	\text{LHS of \eqref{eq: lemma 19 eq 1}} \geq \inf_{\widehat{\mC}^{(1)}}\sup_{\{\othetaks{k}\}_{k=1}^K \in \overline{\Theta}_0} \tp\Bigg(R_{\othetaks{1}}(\widehat{\mC}^{(1)}) - R_{\othetaks{1}}(\mC_{\othetaks{1}}) \geq  C\sqrt{\frac{p}{\ns}} \Bigg).
\end{equation}
Let $r = c\sqrt{p/\ns} \leq 0.001$ with some small constant $c > 0$. For any $\bmu \in \mathbb{R}^p$, denote distribution $\frac{1}{2}\mathcal{N}(\bmu, \bm{I}_p) + \frac{1}{2}\mathcal{N}(-\bmu, \bm{I}_p)$ as $\tp_{\bmu}$. Consider a $r/4$-packing of $r\mathcal{S}^{p-1}$: $\{\widetilde{\bm{v}}_j\}_{j=1}^N$. By Lemma \ref{lem: packing number of sphere}, $N \geq 4^{p-1}$. Denote $\widetilde{\bmu}_j = (\sigma, \widetilde{\bm{v}}_j^\top)^\top \in \mathbb{R}^p$, where $\sigma = \sqrt{0.005}$. Then by definition of KL divergence and Lemma 8.4 in \citeapp{cai2019chime},
\begin{align}
	\text{KL}\left(\prod_{k \in S}\tp_{\widetilde{\bmu}_j}^{\otimes n_k} \cdot \mathbb{Q}_S \bigg\| \prod_{k \in S}\tp_{\widetilde{\bmu}_{j'}}^{\otimes n_k} \cdot \mathbb{Q}_S\right) &= \sum_{k \in S}n_k \text{KL}(\tp_{\widetilde{\bmu}_j} \| \tp_{\widetilde{\bmu}_{j'}}) \\
	&\leq n_S \cdot 8(1+\sigma^2)\twonorm{\widetilde{\bmu}_j - \widetilde{\bmu}_{j'}}^2 \\
	&\leq 32(1+\sigma^2)\ns r^2 \\
	&\leq 32(1+\sigma^2)\ns \cdot c^2\frac{2(p-1)}{\ns} \\
	&\leq \frac{32(1+\sigma^2)c^2}{\log 2}\log N.
\end{align}
For simplicity, we write $L_{\otheta}$ with $\otheta \in \overline{\Theta}_0$ and $\bmu_1 = -\bmu_2 = \bmu$ as $L_{\bmu}$. By Lemma 8.5 in \citeapp{cai2019chime},
\begin{equation}
	L_{\widetilde{\bmu}_i}(\mC_{\widetilde{\bmu}_j}) \geq \frac{1}{\sqrt{2}}g\left(\frac{\sqrt{\sigma^2+r^2}}{2}\right)\frac{\twonorm{\widetilde{\bmu}_i - \widetilde{\bmu}_j}}{\twonorm{\widetilde{\bmu}_i}} \geq \frac{1}{\sqrt{2}}\cdot 0.15\cdot \frac{r/4}{\sqrt{\sigma^2+r^2}} \geq 2r,
\end{equation}
where $g(x) = \phi(x)[\phi(x)-x\Phi(x)]$. The last inequality holds because $\sqrt{\sigma^2+r^2} \geq \sqrt{2}\sigma$ and $g(\sqrt{\sigma^2+r^2}/2) \geq 0.15$ when $r^2 \leq \sigma^2 =0.001$. Then by Lemma 3.5 in \citeapp{cai2019chime} (Proposition 2 in \citealpapp{azizyan2013minimax}), for any classifier $\mC$, and $i \neq j$,
\begin{equation}\label{eq: lemma 19 eq 2}
	L_{\widetilde{\bmu}_i}(\mC) + L_{\widetilde{\bmu}_j}(\mC) \geq L_{\widetilde{\bmu}_i}(\mC_{\widetilde{\bmu}_j}) - \sqrt{\text{KL}(\tp_{\widetilde{\bmu}_i}\| \tp_{\widetilde{\bmu}_j})/2} \geq 2r - r = c\sqrt{\frac{p}{\ns}}.
\end{equation}
For any $\widehat{\mC}^{(1)}$, consider a test $\psi^* = \argmin_{j =1:N} L_{\widetilde{\bmu}_j}(\widehat{\mC}^{(1)})$. Therefore if there exists $j_0$ such that $L_{\widetilde{\bmu}_{j_0}}(\widehat{\mC}^{(1)}) < \frac{c}{2}\sqrt{\frac{p}{\ns}}$, then by \eqref{eq: lemma 19 eq 2}, we must have $\psi^* = j_0$. Let $C_1 \leq c/2$, then by Fano's lemma (Corollary 6 in \citealpapp{tsybakov2009introduction}) 
\begin{align}
	\inf_{\widehat{\mC}^{(1)}}\sup_{\{\othetaks{k}\}_{k=1}^K \in \overline{\Theta}_0} \tp\Bigg(L_{\othetaks{1}}(\widehat{\mC}^{(1)}) \geq  C_1\sqrt{\frac{p}{\ns}} \Bigg) 
	&\geq \inf_{\widehat{\mC}^{(1)}}\sup_{j=1:N} \tp\Bigg(L_{\widetilde{\bmu}^{(j)}}(\widehat{\mC}^{(1)}) \geq  C_1\sqrt{\frac{p}{\ns}} \Bigg) \\
	&\geq \inf_{\widehat{\mC}^{(1)}}\sup_{j=1:N} \tp\Bigg(\psi^* \neq j \Bigg) \\
	&\geq \inf_{\psi}\sup_{j=1:N} \tp\Bigg(\psi \neq j \Bigg) \\
	&\geq 1-\frac{\log 2}{\log N} - \frac{32(1+\sigma^2)c^2}{\log 2}\\
	&\geq \frac{1}{4},
\end{align}
when $p \geq 2$ and  $c = \sqrt{\frac{\log 2}{128(1+\sigma^2)}}$. Then apply Lemma \ref{lem: surrogate loss and classification error} to get the \eqref{eq: lemma 19 eq 1}.

\noindent(\rom{2}) We want to show
\begin{equation}\label{eq: lemma 19 eq 3}
	\inf_{\{\widehat{\mC}^{(k)}\}_{k=1}^K} \sup_{S: |S| \geq s}\sup_{\substack{\{\othetaks{k}\}_{k \in S} \in \overline{\Theta}_{S, \bbeta} \\ \mathbb{Q}_S}} \tp\Bigg(\bigcup_{k \in S}\bigg\{R_{\othetaks{k}}(\widehat{\mC}^{(k)}) - R_{\othetaks{k}}(\mC_{\othetaks{k}}) \geq  C h\wedge \sqrt{\frac{p}{n_k}}\bigg\} \Bigg) \geq \frac{1}{4}.
\end{equation}
Fixing an $S$ and a $\mathbb{Q}_S$. Suppose $1 \in S$. We have
\begin{equation}\label{eq: lemma 19 eq 4}
	\text{LHS of \eqref{eq: lemma 19 eq 3}} \geq \inf_{\widehat{\mC}^{(1)}} \sup_{\substack{\{\othetaks{k}\}_{k \in S} \in \overline{\Theta}_{S, \bbeta} \\ \mathbb{Q}_S}} \tp\bigg(R_{\othetaks{1}}(\widehat{\mC}^{(1)}) - R_{\othetaks{1}}(\mC_{\othetaks{1}}) \geq Ch\wedge \sqrt{\frac{p}{n_k}}\Bigg).
\end{equation}
Let $r = h \wedge (c\sqrt{p/n_1}) \wedge M$ with a small constant $c > 0$. For any $\bmu \in \mathbb{R}^p$, denote distribution $\frac{1}{2}\mathcal{N}(\bmu, \bm{I}_p) + \frac{1}{2}\mathcal{N}(-\bmu, \bm{I}_p)$ as $\tp_{\bmu}$. Consider a $r/4$-packing of $r\mathcal{S}^{p-1}$. By Lemma \ref{lem: packing number of sphere}, $N \geq 4^{p-1}$. Denote $\widetilde{\bmu}_j = (\sigma, \widetilde{\bm{v}}_j^\top)^\top \in \mathbb{R}^p$. WLOG, assume $M \geq 2$. Let $\bmuks{k}_1 = \widetilde{\bmu} = (\sigma, \bm{0}_{p-1})^\top$ for all $k \in S\backslash \{1\}$. Then by following the same arguments in (\rom{1}) and part (\rom{2}) of the proof of Lemma \ref{lem: multitask d no outlier}, we can show that the RHS of \eqref{eq: lemma 19 eq 4} is larger than or equal to $1/4$ when $p \geq 3$.

\noindent(\rom{3}) We want to show
\begin{equation}
	\inf_{\{\widehat{\mC}^{(k)}\}_{k=1}^K} \sup_{S: |S| \geq s}\sup_{\substack{\{\othetaks{k}\}_{k \in S} \in \overline{\Theta}_{S, \bbeta} \\ \mathbb{Q}_S}} \tp\Bigg(\bigcup_{k \in S}\bigg\{R_{\othetaks{k}}(\widehat{\mC}^{(k)}) - R_{\othetaks{k}}(\mC_{\othetaks{k}}) \geq  C h\wedge \sqrt{\frac{\log K}{n_k}}\bigg\} \Bigg) \geq \frac{1}{4}.
\end{equation}
This can be proved by following similar ideas used in step (\rom{3}) of the proof of Lemma \ref{lem: multitask d no outlier}, so we omit the proof here.

\noindent(\rom{4}) We want to show
\begin{equation}
	\inf_{\{\widehat{\mC}^{(k)}\}_{k=1}^K} \sup_{S: |S| \geq s}\sup_{\substack{\{\othetaks{k}\}_{k \in S} \in \overline{\Theta}_{S, w} \\ \mathbb{Q}_S}} \tp\Bigg(\bigcup_{k \in S}\bigg\{R_{\othetaks{k}}(\widehat{\mC}^{(k)}) - R_{\othetaks{k}}(\mC_{\othetaks{k}}) \geq  C \sqrt{\frac{\log K}{n_k}}\bigg\} \Bigg) \geq \frac{1}{4}.
\end{equation}
This can be similarly proved by following the arguments in part (\rom{1}) with Lemmas \ref{lem: surrogate loss and classification error} and \ref{lem: surrogate loss mismatch bound}.

\noindent(\rom{5}) We want to show
\begin{equation}\label{eq: lower bound eq beta mtl 2}
	\inf_{\{\widehat{\mC}^{(k)}\}_{k=1}^K} \sup_{S: |S| \geq s}\sup_{\substack{\{\othetaks{k}\}_{k \in S} \in \overline{\Theta}_{S, \delta} \\ \mathbb{Q}_S}} \tp\Bigg(\bigcup_{k \in S}\bigg\{R_{\othetaks{k}}(\widehat{\mC}^{(k)}) - R_{\othetaks{k}}(\mC_{\othetaks{k}}) \geq  C \sqrt{\frac{\log K}{n_k}}\bigg\} \Bigg) \geq \frac{1}{4}.
\end{equation}
This can be similarly proved by following the arguments in part (\rom{1}) with Lemmas \ref{lem: surrogate loss and classification error} and \ref{lem: surrogate loss mismatch bound delta}.

Finally, we get the desired conclusion by combining (\rom{1})-(\rom{5}).
\end{proof}

\begin{proof}[Proof of Lemma \ref{lem: surrogate loss and classification error}]
We follow a similar proof idea used in the proof of Lemma 3.4 in \citeapp{cai2019chime}. Let $\phi_1$ and $\phi_2$ be the density of $\mathcal{N}(\bmu, \bSigma)$ and $\mathcal{N}(-\bmu, \bSigma)$, respectively. Denote $\eta_{\otheta}(\bz) = \frac{(1-w)\phi_1(\bz)}{(1-w)\phi_1(\bz) + w \phi_2(\bz)}$ and $S_{\mC} = \{\bz \in \mathbb{R}^p: \mC(\bz) = 1\}$ for any classifier $\mC$. Note that $S_{\mC_{\otheta}} = \{\bz \in \mathbb{R}^p: (1-w)\phi_1(\bz) \geq w\phi_2(\bz)\}$. The permutation actually doesn't matter in the proof. WLOG, we drop the permutations in the definition of misclassification error and surrogate loss by assuming $\pi$ to be the identity function. If $\pi$ is not identity in the definition of $R_{\otheta}(\mC)$, for example, we can define $S_{\mC} = \{\bz \in \mathbb{R}^p: \mC(\bz) = 2\}$ instead and all the following steps still follow.

By definition,
\begin{align}
	\tp_{\otheta}(\mC(\bz) \neq y) &= (1-w)\int_{S_{\mC}^c}\phi_1 d\bz  + w\int_{S_{\mC}}\phi_2 d\bz, \\
	\tp_{\otheta}(\mC_{\otheta}(\bz) \neq y) &= (1-w)\int_{S_{\mC_{\otheta}}^c}\phi_1 d\bz  + w\int_{S_{\mC_{\otheta}}}\phi_2 d\bz,
\end{align}
which leads to
\begin{align}
	\tp_{\otheta}(\mC(\bz) \neq y) - \tp_{\otheta}(\mC_{\otheta}(\bz) \neq y) &= \int_{S_{\mC_{\otheta}} \backslash S_{\mC}} [(1-w)\phi_1 - w\phi_2] d\bz + \int_{S_{\mC_{\otheta}}^c \backslash S_{\mC}^c} [w\phi_2 - (1-w)\phi_1] d\bz \\
	&= \int_{S_{\mC_{\otheta}} \triangle S_{\mC}} \norm{(1-w)\phi_1 - w\phi_2} d\bz \\
	&= \te_{\bz \sim (1-w)\phi_1 + w\phi_2}\big[\norma{2\eta_{\otheta}(\bz)-1}\mathds{1}(S_{\mC_{\otheta}} \triangle S_{\mC})\big]\\
	&\geq 2t\cdot \tp_{\otheta}\left(S_{\mC_{\otheta}} \triangle S_{\mC}, \norma{2\eta_{\otheta}(\bz)-1} > 2t \right) \\
	&= 2t\left[\tp_{\otheta}(S_{\mC_{\otheta}} \triangle S_{\mC}) - \tp_{\otheta}(\norma{2\eta_{\otheta}(\bz)-1} \leq 2t)\right] \\
	&\geq 2t\left[\tp_{\otheta}(S_{\mC_{\otheta}} \triangle S_{\mC}) - ct\right] \\
	&\geq\frac{1}{2c}\tp_{\otheta}^2(S_{\mC_{\otheta}} \triangle S_{\mC}),
\end{align}
where we let $t = \frac{1}{2c}\tp_{\otheta}(S_{\mC_{\otheta}} \triangle S_{\mC})$ with $c = 1+\frac{8}{\sqrt{2\pi}\sigma}$. This completes the proof. The last second inequality depends on the fact that
\begin{equation}
	\tp_{\otheta}(\norma{\eta_{\otheta}(\bz)-1/2} \leq t) \leq ct,
\end{equation}
holds for all $t \leq 1/(2c)$. This is because
\begin{align}
	&\tp_{\otheta}(\norma{\eta_{\otheta}(\bz)-1/2} \leq t) \\
	&= \tp_{\otheta}\left(\log\left(\frac{w}{1-w}\right) + \log\left(\frac{1-2t}{1+2t}\right) \leq \log\left(\frac{\phi_1}{\phi_2}(\bz)\right) \leq \log\left(\frac{w}{1-w}\right) + \log\left(\frac{1+2t}{1-2t}\right)\right) \\
	&= \tp_{\otheta}\bigg(\log\left(\frac{w}{1-w}\right) + \log\left(\frac{1-2t}{1+2t}\right) \leq (\bmu_1-\bmu_2)^\top\bSigma^{-1}\left(\bz - \frac{\bmu_1+\bmu_2}{2}\right) \\
	&\hspace{6.6cm} \leq \log\left(\frac{w}{1-w}\right) + \log\left(\frac{1+2t}{1-2t}\right)\bigg) \\
	&= \frac{1}{2}\tp_{\otheta}\bigg(\log\left(\frac{w}{1-w}\right) + \log\left(\frac{1-2t}{1+2t}\right) \leq \mathcal{N}(\Delta^2/2,\Delta^2) \leq \log\left(\frac{w}{1-w}\right) + \log\left(\frac{1+2t}{1-2t}\right)\bigg) \\
	&\quad + \frac{1}{2}\tp_{\otheta}\bigg(\log\left(\frac{w}{1-w}\right) + \log\left(\frac{1-2t}{1+2t}\right) \leq \mathcal{N}(-\Delta^2/2,\Delta^2) \leq \log\left(\frac{w}{1-w}\right) + \log\left(\frac{1+2t}{1-2t}\right)\bigg)\label{eq: lem 20 eq x} \\
	&\leq \frac{1}{\sqrt{2\pi}\sigma} \cdot \left[\log\left(\frac{1+2t}{1-2t}\right) - \log\left(\frac{1-2t}{1+2t}\right)\right] \\
	&\leq  \frac{1}{\sqrt{2\pi}\sigma}\cdot \frac{8t}{1-2t} \\
	&\leq ct, \label{eq: GMM margin = 1}
\end{align}
when $t \leq 1/(2c)$. Note that \eqref{eq: GMM margin = 1} implies that a binary GMM under the separation assumption $\Delta \gtrsim 1$ has Tsybakov's margin with margin parameter $1$. For the notion of Tsybakov's margin, see \citeapp{audibert2007fast}. We will prove a more general result showing that a multi-cluster GMM under the separation assumption also has Tsybakov's margin with margin parameter $1$. This turns out to be useful in proving the upper and lower bounds of misclassification error.
\end{proof}

\begin{proof}[Proof of Lemma \ref{lem: surrogate loss mismatch bound}]
WLOG, suppose $w \geq w'$. Similar to \eqref{eq: lem 20 eq x}, it's easy to see that
\begin{align}
	L_{\otheta}(\mC_{\otheta'}) &= (1-w)\tp_{\otheta}\left(\mC_{\otheta}(\bz) \neq \mC_{\otheta'}(\bz)|\bz = 1\right) + w\tp_{\otheta}\left(\mC_{\otheta}(\bz) \neq \mC_{\otheta'}(\bz)|\bz = 2\right) \\
	&= (1-w)\tp\left(\log\left(\frac{w'}{1-w'}\right) \leq  \mathcal{N}(\Delta^2/2,\Delta^2) \leq \log\left(\frac{w}{1-w}\right)\right) \\
	&\quad + w\tp\left(\log\left(\frac{w'}{1-w'}\right) \leq  \mathcal{N}(-\Delta^2/2,\Delta^2) \leq \log\left(\frac{w}{1-w}\right)\right) \label{eq: lemma 26 eq} \\
	&\leq \frac{1}{\sqrt{2\pi}\sigma}\left[\log\left(\frac{w}{1-w}\right) - \log\left(\frac{w'}{1-w'}\right)\right] \\
	&= \frac{1}{\sqrt{2\pi}c_w(1-c_w)\sigma}\cdot \norm{w - w'}.
\end{align}
On the other hand,
\begin{align}
	\eqref{eq: lemma 26 eq} &\geq \frac{1}{\sqrt{2\pi}Mc_{\bSigma}}\cdot \exp\left\{-\frac{1}{2\sigma^2}\left[\log\left(\frac{1-c_w}{c_w}\right)+\frac{1}{2}M^2c_{\bSigma}\right]^2\right\}\left[\log\left(\frac{w}{1-w}\right) - \log\left(\frac{w'}{1-w'}\right)\right] \\
	&\geq \frac{1}{\sqrt{2\pi}Mc_{\bSigma}c_w(1-c_w)}\cdot \exp\left\{-\frac{1}{2\sigma^2}\left[\log\left(\frac{1-c_w}{c_w}\right)+\frac{1}{2}M^2c_{\bSigma}\right]^2\right\}\norm{w-w'},
\end{align}
which completes the proof.
\end{proof}

\begin{proof}[Proof of Lemma \ref{lem: multitask misclassification error outlier}]
By Lemma \ref{lem: surrogate loss and classification error}, it suffices to prove
\begin{equation}
	\inf_{\{\widehat{\mC}^{(k)}\}_{k=1}^K} \sup_{S: |S| \geq s}\sup_{\substack{\{\othetaks{k}\}_{k \in S} \in \overline{\Theta}_S \\ \mathbb{Q}_S}} \tp\Bigg(\max_{k \in S} L_{\othetaks{k}}(\widehat{\mC}^{(k)})  \geq  C_1\frac{\widetilde{\epsilon}'^2}{\max_{k=1:K}n_k}\Bigg) \geq \frac{1}{10}.
\end{equation}
For any $\bmu \in \mathbb{R}^p$, denote distribution $\frac{1}{2}\mathcal{N}(\bmu, \bm{I}_p) + \frac{1}{2}\mathcal{N}(-\bmu, \bm{I}_p)$ as $\tp_{\bmu}$. For simplicity, we write $L_{\otheta}$ with $\otheta$ satisfying $\bmu_1 = -\bmu_2 = \bmu$, $w = 1/2$ and $\bSigma = \bm{I}_p$ as $L_{\bmu}$. Consider $L_{\bmu}(\mC_{\bmu'})$ as a loss function between $\bmu$ and $\bmu'$ in Lemmas \ref{lem: from chen} and \ref{lem: binomial lower bound}. Considering $\twonorm{\bmu} = \twonorm{\bmu'} = 1$, by Lemma \ref{lem: lem 8.4 with mu}, note that
\begin{equation}
	\max_{k = 1:K}\text{KL}(\tp_{\bmu}^{\otimes n_k}\| \tp_{\bmu'}^{\otimes n_{k}}) \leq 8\max_{k = 1:K}n_k \cdot \twonorm{\bmu - \bmu'}^2.
\end{equation}
By Lemma 8.5 in \citeapp{cai2019chime}, this implies for some constants $c, C > 0$
\begin{align}
	&\sup\left\{L_{\bmu}(\mC_{\bmu'}): \max_{k = 1:K}\text{KL}(\tp_{\bmu}^{\otimes n_k}\| \tp_{\bmu'}^{\otimes n_{k}}) \leq (\wepsilon'/(1-\wepsilon))^2\right\} \\
	&\geq \sup\left\{c\twonorm{\bmu - \bmu'}:  \max_{k = 1:K}\text{KL}(\tp_{\bmu}^{\otimes n_k}\| \tp_{\bmu'}^{\otimes n_{k}}) \leq (\wepsilon'/(1-\wepsilon))^2\right\} \\
	&\geq \sup\left\{c\twonorm{\bmu - \bmu'}:  8\max_{k = 1:K}n_k \cdot\twonorm{\bmu - \bmu'}^2  \leq (\wepsilon'/(1-\wepsilon))^2\right\} \\
	&= C\cdot \frac{\wepsilon'}{\sqrt{\max_{k = 1:K}n_k}}.
\end{align}
Then apply Lemmas \ref{lem: from chen} and \ref{lem: binomial lower bound} to get the desired bound.
\end{proof}

\subsection{Proof of Theorem \ref{thm: brute force alignment}}
Denote $\xi = \max_{k \in S} \min_{r_k = \pm 1}\twonorm{r_k\hbeta^{(k)[0]} - \bbetaks{k}} = \max_{k \in S}(\twonorm{\hbeta^{(k)[0]} - \bbetaks{k}} \wedge \twonorm{\hbeta^{(k)[0]} + \bbetaks{k}})$. WLOG, assume $S = \{1, \ldots, s\}$ and $r^*_k = 1$ for all $k \in S$. Hence $\xi = \max_{k \in S}\twonorm{\hbeta^{(k)[0]} - \bbetaks{k}}$. For any $k' = 1, \ldots, s$, define 
\begin{align}
	\bm{r} &= (\underbrace{r_1, \ldots, r_{k'}}_{=-1}, \underbrace{r_{k'+1}, \ldots, r_s}_{=1}, \underbrace{r_{s+1}, \ldots, r_K}_{\text{outlier tasks}}), \\
    \bm{r}' &= (1, 1, \ldots, 1, 1, 1, \ldots, 1, 1, \underbrace{r_{s+1}, \ldots, r_K}_{\text{outlier tasks}}),\\
    \bm{r}'' &= (-1, \ldots, -1, -1, \ldots, -1, \underbrace{r_{s+1}, \ldots, r_K}_{\text{outlier tasks}}).
\end{align}
WLOG, it suffices to prove that
\begin{align}
	\text{score}(\bm{r}) - \text{score}(\bm{r}') > 0 \quad \text{ when } k' \leq \lfloor s/2\rfloor, \label{eq: brute force alignment} \\
	\text{score}(\bm{r}) - \text{score}(\bm{r}'') > 0 \quad \text{ when } k' > \lfloor s/2\rfloor. \label{eq: brute force alignment 2}
\end{align}
In fact, if this holds, then we must have
\begin{equation}
	\widehat{r}_k = 1  \text{ for all } k \in S \text{\quad or \quad} \widehat{r}_k = -1  \text{ for all } k \in S.
\end{equation}
Otherwise, according to \eqref{eq: brute force alignment}, if $\#\{k\in S: \widehat{r}_k = -1\} \leq \lfloor s/2\rfloor$, by replacing the first $s$ entries of $\widehat{\bm{r}}$ with $1$, we get a different alignment whose score is smaller than the score of $\widehat{\bm{r}}$, which is contradicted with the definition of $\widehat{\bm{r}}$.  If $\#\{k\in S: \widehat{r}_k = -1\} > \lfloor s/2\rfloor$, based on \eqref{eq: brute force alignment 2}, by replacing the first $s$ entries of $\widehat{\bm{r}}$ with $-1$, we get a different alignment whose score is smaller than the score of $\widehat{\bm{r}}$, which is again contradicted with the definition of $\widehat{\bm{r}}$.

In the following, we prove \eqref{eq: brute force alignment}. The proof of \eqref{eq: brute force alignment 2} is almost the same, so we do not repeat it. Under the conditions we assume, it can be shown that
\begin{align}
	\text{score}(\bm{r}) - \text{score}(\bm{r}') &= \sum_{k_1=1}^{k'}\sum_{k_2=1}^{k'}\twonorm{\hbeta^{(k_1)[0]} - \hbeta^{(k_2)[0]}} + 2\sum_{k_1=1}^{k'}\sum_{k_2=k'+1}^{s}\twonorm{\hbeta^{(k_1)[0]} + \hbeta^{(k_2)[0]}} \\
	&\quad + 2\sum_{k_1=1}^{k'}\sum_{k_2=s+1}^{K}\twonorm{\hbeta^{(k_1)[0]} + r_{k_2}\hbeta^{(k_2)[0]}} \\
	&\quad - \sum_{k_1=1}^{k'}\sum_{k_2=1}^{k'}\twonorm{\hbeta^{(k_1)[0]} - \hbeta^{(k_2)[0]}} - 2\sum_{k_1=1}^{k'}\sum_{k_2=k'+1}^{s}\twonorm{\hbeta^{(k_1)[0]} - \hbeta^{(k_2)[0]}} \\
	&\quad - 2\sum_{k_1=1}^{k'}\sum_{k_2=s+1}^{K}\twonorm{-\hbeta^{(k_1)[0]} + r_{k_2}\hbeta^{(k_2)[0]}} \\
	&= 2\underbrace{\sum_{k_1=1}^{k'}\sum_{k_2=k'+1}^{s}\twonorm{\hbeta^{(k_1)[0]} + \hbeta^{(k_2)[0]}}}_{(1)} + 2\underbrace{\sum_{k_1=1}^{k'}\sum_{k_2=s+1}^{K}\twonorm{\hbeta^{(k_1)[0]} + r_{k_2}\hbeta^{(k_2)[0]}}}_{(2)} \\
	&\quad - 2\underbrace{\sum_{k_1=1}^{k'}\sum_{k_2=k'+1}^{s}\twonorm{\hbeta^{(k_1)[0]} - \hbeta^{(k_2)[0]}}}_{(1)'} - 2\underbrace{\sum_{k_1=1}^{k'}\sum_{k_2=s+1}^{K}\twonorm{-\hbeta^{(k_1)[0]} + r_{k_2}\hbeta^{(k_2)[0]}}}_{(2)'}. \label{eq: decomp brute force}
\end{align}
And
\begin{align}
	(1) - (1)' &= \sum_{k_1=1}^{k'}\sum_{k_2=k'+1}^{s}(\twonorm{\hbeta^{(k_1)[0]} + \hbeta^{(k_2)[0]}} - \twonorm{\hbeta^{(k_1)[0]} - \hbeta^{(k_2)[0]}}) \\
	&\geq \sum_{k_1=1}^{k'}\sum_{k_2=k'+1}^{s}(\twonorm{\bbetaks{k_1} + \bbetaks{k_2}} - \twonorm{\bbetaks{k_1} - \bbetaks{k_2}} - 4\xi) \\
	&\geq \sum_{k_1=1}^{k'}\sum_{k_2=k'+1}^{s}(2\twonorm{\bbetaks{k_1}} - 2\twonorm{\bbetaks{k_1} - \bbetaks{k_2}} - 4\xi) \\
	&\geq 2(s-k')\sum_{k_1=1}^{k'}\twonorm{\bbetaks{k_1}} - 4k'(s-k')h_{\bbeta} - 4k'(s-k')\xi,
\end{align}
\begin{equation}
	(2) - (2)' \geq - \sum_{k_1=1}^{k'}\sum_{k_2=s+1}^{K}2\twonorm{\hbeta^{(k_1)[0]}} \geq -2(K-s)\sum_{k_1=1}^{k'}\twonorm{\bbetaks{k_1}} - 2k'(K-s)\xi.
\end{equation}
Combining all these pieces,
\begin{align}
	&\text{score}(\bm{r}) - \text{score}(\bm{r}') \\
	&\geq 2(2s-k'-K)\sum_{k_1=1}^{k'}\twonorm{\bbetaks{k_1}} - 4k'(s-k')h_{\mu} -2k'(K-s)\xi -4k'(s-k')\xi \\
	&\geq  2k'\left[(2s-k'-K)\min_{k\in S}\twonorm{\bbetaks{k}}- 2(s-k')h_{\mu} -(K-s)\xi -2(s-k')\xi\right]\\
	&> 2k'\left[\bigg(\frac{3}{2}s-K\bigg)\min_{k\in S}\twonorm{\bbetaks{k}}- 2sh_{\mu} -(K-s)\xi -2s\xi\right] \label{eq: brute force 2}\\
	&\geq 0, \label{eq: brute force 3}
\end{align}
where \eqref{eq: brute force 2} holds because $1 \leq k' \leq \lfloor s/2 \rfloor$ and \eqref{eq: brute force 3} is due to the condition (\rom{2}).

\subsection{Proof of Theorem \ref{thm: greedy alignment}}
Denote $\xi = \max_{k \in S} \min_{r_k = \pm 1}\twonorm{r_k\hbeta^{(k)[0]} - \bbetaks{k}} = \max_{k \in S}(\twonorm{\hbeta^{(k)[0]} - \bbetaks{k}} \wedge \twonorm{\hbeta^{(k)[0]} + \bbetaks{k}})$. WLOG, assume $S = \{1, \ldots, s\}$ and $r^*_k = 1$ for all $k \in S$. Hence $\xi = \max_{k \in S}\twonorm{\hbeta^{(k)[0]} - \bbetaks{k}}$. For any $k' = 1, \ldots, sp_a$, define 
\begin{align}
	\bm{r} &= (\underbrace{r_1, \ldots, \notes{r_{k'}}}_{=-1}, \underbrace{r_{k'+1}, \ldots, r_s}_{=1}, \underbrace{r_{s+1}, \ldots, r_K}_{\text{outlier tasks}}), \\
	\bm{r}' &= (\underbrace{r_1, \ldots, r_{k'-1}}_{=-1}, \underbrace{\notes{r_{k'}'}, r_{k'+1},\ldots, r_s}_{=1}, \underbrace{r_{s+1}, \ldots, r_K}_{\text{outlier tasks}}).
\end{align}
By the definition of $p_a$, we must have $\#\{k \in S: \widehat{r}_k = -1\} = sp_a$ or $\#\{k \in S: \widehat{r}_k = 1\} = sp_a$. If $\#\{k \in S: \widehat{r}_k = -1\} = sp_a$ and we have
\begin{equation}\label{eq: L1 L2 greedy}
	\text{score}(\bm{r}) - \text{score}(\bm{r}') > 0,
\end{equation}
then for each $k \in S$ in the for loop of Algorithm \ref{algo: greedy alignment}, the algorithm will flip the sign of $\widehat{r}_{k'}$ to decrease the mis-alignment proportion $p_a$. Then after the for loop, the mis-alignment proportion $p_a$ will become zero, which means the correct alignment is achieved. The case that $\#\{k \in S: \widehat{r}_k = 1\} = sp_a$ can be similarly discussed.

Now we derive \eqref{eq: L1 L2 greedy}. Similar to the decomposition in \eqref{eq: decomp brute force}, we have
\begin{align}
	\text{score}(\bm{r}) - \text{score}(\bm{r}') &= 2\underbrace{\sum_{k=k'+1}^{s}\twonorm{\hbeta^{(k')[0]} + \hbeta^{(k)[0]}}}_{(1)} + 2\underbrace{\sum_{k=1}^{k'-1}\twonorm{\hbeta^{(k')[0]} - \hbeta^{(k)[0]}}}_{(2)}\\
	&\quad + 2\underbrace{\sum_{k=s+1}^{K}\twonorm{-\hbeta^{(k')[0]} - r_k\hbeta^{(k)[0]}}}_{(3)} \\
	&\quad - 2\underbrace{\sum_{k=k'+1}^{s}\twonorm{\hbeta^{(k')[0]} - \hbeta^{(k)[0]}}}_{(1)'} - 2\underbrace{\sum_{k=1}^{k'-1}\twonorm{\hbeta^{(k')[0]} + \hbeta^{(k)[0]}}}_{(2)'}\\
	&\quad - 2\underbrace{\sum_{k=s+1}^{K}\twonorm{\hbeta^{(k')[0]} - r_k\hbeta^{(k)[0]}}}_{(3)'}.
\end{align}
Note that
\begin{align}
	(1) - (1)' &= \sum_{k=k'+1}^{s}(\twonorm{\hbeta^{(k')[0]} + \hbeta^{(k)[0]}} - \twonorm{\hbeta^{(k')[0]} - \hbeta^{(k)[0]}}) \\
	&\geq \sum_{k=k'+1}^{s}(\twonorm{\bbetaks{k'} + \bbetaks{k}} - \twonorm{\bbetaks{k'} - \bbetaks{k}} - 4\xi) \\
	&\geq \sum_{k=k'+1}^{s}(2\twonorm{\bbetaks{k'}} - 2\twonorm{\bbetaks{k'} - \bbetaks{k}} - 4\xi) \\
	&\geq (s-k')(2\twonorm{\bbetaks{k'}} - 4h_{\bbeta} - 4\xi),
\end{align}
\begin{equation}
	(2) - (2)' \geq -\sum_{k=1}^{k'-1}2\twonorm{\hbeta^{(k')[0]}} \geq -2(k'-1)\twonorm{\bbetaks{k'}} - 2(k'-1)\xi,
\end{equation}
\begin{equation}
	(3) - (3)' \geq -\sum_{k=s+1}^{K}2\twonorm{\hbeta^{(k')[0]}} \geq -2(K-s)\twonorm{\bbetaks{k'}} - 2(K-s)\xi.
\end{equation}
Putting all pieces together,
\begin{align}
	&\text{score}(\bm{r}) - \text{score}(\bm{r}') \\
	&\geq 2\left[(2s-2k'-K+1)\twonorm{\bbetaks{k'}} - 2(s-k')h_{\bbeta} -(s-k'+K-1)\xi\right] \\
	&> 2\left[(2s-2sp_a-K)\twonorm{\bbetaks{k'}} - 2sh_{\bbeta} -2(s+K)\xi\right] \label{eq: greedy 2}\\
	&\geq 0. \label{eq: greedy 3}
\end{align}
where \eqref{eq: greedy 2} holds because $1 \leq k' \leq sp_a$ and \eqref{eq: greedy 3} is due to the condition (\rom{3}).

\subsection{Proof of Theorem \ref{thm: upper bound transfer est error}}

\subsubsection{Lemmas}
Define the contraction basin of one GMM as 
\begin{equation}
	B_{\text{con}}(\bthetaks{k}) = \{\btheta = \{w, \bbeta, \delta\}: w_r\in[c_w/2, 1-c_w/2], \twonorm{\bbeta - \bbetaks{k}} \leq C_b\Delta, \norm{\delta - \deltaks{k}} \leq C_b\Delta\},
\end{equation}
for which we may shorthand as $B_{\text{con}}$ in the following.

For GMM $\bz \sim  (1-w^*)\mathcal{N}(\bmu_1^*, \bSigma^*) + w^*\mathcal{N}(\bmu_2^*, \bSigma^*)$ and any $\btheta = (w, \bbeta, \delta)$, define
\begin{align}
	\gamma_{\btheta}(\bz) = \frac{w\exp\{\bbeta^\top\bz - \delta\}}{1-w+w\exp\{\bbeta^\top\bz - \delta_r\}},& \quad w(\btheta) = \te [\gamma_{\btheta}(\bz)], \\
	\bmu_1(\btheta) = \frac{\te [(1-\gamma_{\btheta}(\bz))\bz]}{\te [1-\gamma_{\btheta}(\bz)]},& \quad  \bmu_2(\btheta) = \frac{\te [\gamma_{\btheta}(\bz)\bz]}{\te [\gamma_{\btheta}(\bz)]}.
\end{align}

\begin{lemma}\label{lem: concentration transfer}
	Suppose Assumption \ref{asmp: upper bound transfer est error} holds.
	\begin{enumerate}[(i)]
		\item With probability at least $1-\tau$,
			\begin{equation}
				\sup_{\substack{\bthetak{0} \in B_{\text{con}} \\ \twonorm{\bbetak{0} - \bbetaks{0}} \leq \xi^{(0)}}} \norma{\frac{1}{n_0}\sum_{i=1}^{n_0}\gamma_{\bthetak{0}}(\bzk{0}_i) - \te[\gamma_{\bthetak{0}}(\bzk{0})]} \lesssim \xi^{(0)}\sqrt{\frac{p}{n_0}} + \sqrt{\frac{\log (1/\tau)}{n_0}}.
			\end{equation}
		\item With probability at least $1-\tau$,
			\begin{equation}
				\sup_{\substack{\bthetak{0} \in B_{\text{con}} \\ \twonorm{\bbetak{0} - \bbetaks{0}} \leq \xi^{(0)}}} \norma{\frac{1}{n_0}\sum_{i=1}^{n_0}\gamma_{\bthetak{k}}(\bzk{0}_i)(\bzk{0}_i)^\top\bbetaks{0} - \te[\gamma_{\bthetak{0}}(\bzk{0})(\bzk{0}_i)^\top\bbetaks{0}]} \lesssim \xi^{(0)}\sqrt{\frac{p}{n_0}} + \sqrt{\frac{\log (1/\tau)}{n_0}}.
			\end{equation}
		\item With probability at least $1-\tau$,
			\begin{equation}
				\sup_{\substack{\bthetak{0} \in B_{\text{con}} \\ \twonorm{\bbetak{0} - \bbetaks{0}} \leq \xi^{(0)}}} \twonorma{\frac{1}{n_0}\sum_{i=1}^{n_0}\gamma_{\bthetak{k}}(\bzk{0}_i)\bzk{0}_i - \te[\gamma_{\bthetak{0}}(\bzk{0})\bzk{0}_i]} \lesssim \xi^{(0)}\sqrt{\frac{p}{n_0}} + \sqrt{\frac{\log (1/\tau)}{n_0}}.
			\end{equation}
	\end{enumerate}
\end{lemma}

\subsubsection{Main proof of Theorem \ref{thm: upper bound transfer est error}}
WLOG, in Assumptions \ref{asmp: upper bound transfer est error}.(\rom{3}) and \ref{asmp: upper bound transfer est error}.(\rom{4}), we assume
\begin{itemize}
	\item $\twonorm{\hbeta^{(0)[0]} - \bbetaks{0}} \vee \norm{\hdelta^{(0)[0]} - \deltaks{0}}\leq C_4\Delta^{(0)}$, with a sufficiently small constant $C_4$;
	\item $\norm{\hw^{(0)[0]}-\wks{0}} \leq c_w/2$.
\end{itemize}

\noindent\underline{(\Rom{1}) Case 1:} We first consider the case that $h \geq C\sqrt{\frac{p}{n_0}}$. 
Consider an event $\mathcal{E}$ defined to be the intersection of the events in Lemma \ref{lem: concentration transfer}, with $\xi^{(k)} = $ a large constant $C$, which satisfies $\tp(\mathcal{E}) \geq 1-\tau$. Throughout the analysis in Case 1, we condition on $\mathcal{E}$, therefore all the arguments hold with probability at least $1-\tau$. 

Similar to our analysis in the proof of Theorem \ref{thm: upper bound multitask est error}, conditioned on $\mathcal{E}$, we have
\begin{align}
	\norm{\hw^{(0)[t]} - \wks{0}} &\lesssim \kappa_0''d(\htheta^{(0)[t-1]}, \bthetaks{0}) + \sqrt{\frac{p}{n_0}}, \\
	\max_{r=1:2}\twonorm{\hmu^{(0)[t]}_r - \bmuks{0}_r} &\lesssim \kappa_0''d(\htheta^{(0)[t-1]}, \bthetaks{0}) + \sqrt{\frac{p}{n_0}}, \label{eq: mu transfer}\\
	\twonorm{(\hSigma^{(0)[t]} - \bSigmaks{0})\bbetaks{0}} &\lesssim \kappa_0''d(\htheta^{(0)[t-1]}, \bthetaks{0}) + \sqrt{\frac{p}{n_0}}.\label{eq: Sigma transfer}
\end{align}
Hence 
\begin{align}
	\twonorm{(\hSigma^{(0)[t]})\bbetaks{0} - (\hmu^{(0)[t-1]}_2 - \hmu^{(0)[t-1]}_1)} &\lesssim \twonorm{(\hSigma^{(0)[t]} - \bSigmaks{0})\bbetaks{0}} + \max_{r=1:2}\twonorm{\hmu^{(0)[t]}_r - \bmuks{0}_r} \\
	&\lesssim \kappa_0''d(\htheta^{(0)[t-1]}, \bthetaks{0}) + \sqrt{\frac{p}{n_0}}.
\end{align}
By Lemma \ref{lem: transfer lemma}, we have
\begin{align}
	\twonorm{\hbeta^{(0)[t]} - \bbetaks{0}} &\lesssim \twonorm{(\hSigma^{(0)[t]})\bbetaks{0} - (\hmu^{(0)[t-1]}_2 - \hmu^{(0)[t-1]}_1)} + \frac{\lambda_0^{[t]}}{\sqrt{n_0}} \\
	&\lesssim  \kappa_0''d(\htheta^{(0)[t-1]}, \bthetaks{0}) + \sqrt{\frac{p}{n_0}}+ \frac{\lambda_0^{[t]}}{\sqrt{n_0}}.
\end{align}
Combining these results, we have
\begin{equation}
	d(\htheta^{(0)[t]}, \bthetaks{0}) \leq C\kappa_0''d(\htheta^{(0)[t-1]}, \bthetaks{0}) + C'\sqrt{\frac{p}{n_0}}+ C'\frac{\lambda_0^{[t]}}{\sqrt{n_0}}.
\end{equation}
By the construction of $\lambda_0^{[t]}$, we know that
\begin{equation}
	\lambda_0^{[t]} = \frac{1-\kappa_0^{t}}{1-\kappa_0}C_{\lambda_0}\sqrt{p} + \kappa_0^t \lambda_0^{[0]},
\end{equation}
implies that
\begin{align}
	d(\htheta^{(0)[t]}, \bthetaks{0}) &\leq (C\kappa_0'')^{t}d(\htheta^{(0)[0]}, \bthetaks{0}) + C''\sqrt{\frac{p}{n_0}}+ C'\sum_{t'=1}^{t}\frac{\lambda_0^{[t']}}{\sqrt{n_0}}(C\kappa_0'')^{t-t'} \\
	&\leq (\kappa_0')^t d(\htheta^{(0)[0]}, \bthetaks{0}) + C'''\sqrt{\frac{p}{n_0}} + C'''t(\kappa_0')^t \\
	&\leq Ct(\kappa_0')^t+ C'''\sqrt{\frac{p}{n_0}},
\end{align}
which is the desired rate. The bound of $\max_{r=1:2}\twonorm{\hmu^{(0)[t]}_r - \bmuks{0}_r}$ and $\twonorm{\hSigma^{(0)[t]} - \bSigmaks{0}}$ can be derived similar to \eqref{eq: mu transfer} and \eqref{eq: Sigma transfer}.

\noindent\underline{(\Rom{2}) Case 2:} Next, we consider the case that $h \leq C\sqrt{\frac{p}{n_0}}$. According to Assumption \ref{asmp: upper bound transfer est error}, we have $\sqrt{\frac{p}{n_0}} \lesssim \sqrt{\frac{p+\log K}{\max_{k \in S}n_k}}$. It is easy to see that the analysis in part (\Rom{1}) does not depend on the condition $h \geq C\sqrt{\frac{p}{n_0}}$. Hence we have proved the desired bounds of $\max_{r=1:2}\twonorm{\hmu^{(0)[t]}_r - \bmuks{0}_r}$ and $\twonorm{\hSigma^{(0)[t]} - \bSigmaks{0}}$. Denote $t_0$ as an integer such that $t_0\kappa_0^{t_0} \asymp \sqrt{\frac{p}{n_0}}$. When $1 \leq t \leq t_0$, the bound in part (\Rom{1}) is the desired bound since the term $t\kappa_0^t$ dominates the other terms. Let us consider the case $t = t_0 + 1$.

Consider an event $\mathcal{E}'$ defined to be the event of 
\begin{equation}
	\twonorm{\obbeta^{[T]} - \bbetaks{k'}} \lesssim h + \sqrt{\frac{\log K}{n_{k'}}} + \sqrt{\frac{p}{\ns}} + \epsilon\sqrt{\frac{p+\log K}{\max_{k=1:K}n_k}}, \quad k' \in \argmin_{k \in S}n_k.
\end{equation}
Note that since $h \leq C\sqrt{\frac{p+\log K}{\max_{k \in S}n_k}}$, by part (\Rom{2}) of the proof of Theorem \ref{thm: upper bound multitask est error}, we know that $\tp(\mathcal{E}') \geq 1-C(K^{-1}+\exp\{-C'p\})$. And $\mathcal{E}'$ implies that
\begin{equation}
	\twonorm{\obbeta^{[T]} - \bbetaks{0}} \lesssim h + \sqrt{\frac{\log K}{\max_{k \in S}n_k}} + \sqrt{\frac{p}{\ns}} + \epsilon\sqrt{\frac{p+\log K}{\max_{k=1:K}n_k}}\lesssim \sqrt{\frac{p}{n_0}},
\end{equation}
where the second inequality comes from Assumption \ref{asmp: upper bound transfer est error}.

Also consider another event $\mathcal{E}''$ defined to be the intersection of the events in Lemma \ref{lem: concentration transfer}, with $\xi = C\Big[h + \sqrt{\frac{\log K}{\max_{k \in S}n_k}} + \sqrt{\frac{p}{\ns}} + \epsilon\sqrt{\frac{p+\log K}{\max_{k=1:K}n_k}}\Big]$, which satisfies $\tp(\mathcal{E}'') \geq 1-\tau$. Throughout the analysis in Case 1, we condition on $\mathcal{E}\cap \mathcal{E}' \cap \mathcal{E}''$, therefore all the arguments hold with probability at least $1-\tau-C(K^{-1}+\exp\{-C'p\})$. 

Note that $\lambda_0^{[t]} \geq C\sqrt{p} \geq C'\twonorm{\obbeta^{[T]} - \bbetaks{0}}$ and $\lambda_0^{[t]} \geq C\sqrt{p} \geq C'\sqrt{n_0}\twonorm{(\hSigma^{(0)[t]})\bbetaks{0} - (\hmu^{(0)[t-1]}_2 - \hmu^{(0)[t-1]}_1)}$. Hence by Lemma \ref{lem: transfer lemma}, we have $\hbeta^{(0)[t]} = \obbeta^{[T]}$ thus
\begin{equation}
	\twonorm{\hbeta^{(0)[t]} - \bbetaks{0}} \lesssim h + \sqrt{\frac{\log K}{\max_{k \in S}n_k}} + \sqrt{\frac{p}{\ns}} + \epsilon\sqrt{\frac{p+\log K}{\max_{k=1:K}n_k}}.
\end{equation}
Similar to the analysis in part (\Rom{2}) in the proof of Theorem \ref{thm: upper bound multitask est error}, we have
\begin{align}
	\norm{\hw^{(0)[t]} - \wks{0}} &\lesssim \kappa_0''d(\htheta^{(0)[t-1]}, \bthetaks{0}) + \xi\sqrt{\frac{p}{n_0}} + \sqrt{\frac{1}{n_0}} \\
	&\lesssim \kappa_0''d(\htheta^{(0)[t-1]}, \bthetaks{0}) + \xi + \sqrt{\frac{1}{n_0}}, \\
	\norm{\hdelta^{(0)[t]} - \deltaks{0}} &\lesssim \kappa_0''d(\htheta^{(0)[t-1]}, \bthetaks{0}) +  \xi + \xi\sqrt{\frac{p}{n_0}} + \sqrt{\frac{1}{n_0}},\\
	&\lesssim \kappa_0''d(\htheta^{(0)[t-1]}, \bthetaks{0}) +  \xi + \sqrt{\frac{1}{n_0}}.
\end{align}
Putting all pieces together,
\begin{equation}
	d(\htheta^{(0)[t]}, \bthetaks{0}) \leq C\kappa_0''d(\htheta^{(0)[t-1]}, \bthetaks{0})  + h + \sqrt{\frac{\log K}{\max_{k \in S}n_k}} + \sqrt{\frac{p}{\ns}} + \epsilon\sqrt{\frac{p+\log K}{\max_{k=1:K}n_k}}+ \sqrt{\frac{1}{n_0}}.
\end{equation}
We can continue this analysis from $t = t_0+1$ to $t_0+2$, and so on. Hence for any $t' \geq 1$, we have
\begin{align}
	d(\htheta^{(0)[t_0+t']}, \bthetaks{0}) &\leq (C\kappa_0'')^{t'}d(\htheta^{(0)[t_0]}, \bthetaks{0})  + C'h + C'\sqrt{\frac{\log K}{\max_{k \in S}n_k}} + C'\sqrt{\frac{p}{\ns}} \\
	&\quad + C'\epsilon\sqrt{\frac{p+\log K}{\max_{k=1:K}n_k}}+ C'\sqrt{\frac{1}{n_0}} \\
	&\leq (\kappa_0')^{t'}d(\htheta^{(0)[t_0]}, \bthetaks{0})  + C'h + C'\sqrt{\frac{\log K}{\max_{k \in S}n_k}} + C'\sqrt{\frac{p}{\ns}} \\
	&\quad + C'\epsilon\sqrt{\frac{p+\log K}{\max_{k=1:K}n_k}}+ C'\sqrt{\frac{1}{n_0}} \\
	&\leq (t'+t_0)(\kappa_0')^{t'+t_0} + C'h + C'\sqrt{\frac{\log K}{\max_{k \in S}n_k}} + C'\sqrt{\frac{p}{\ns}} \\
	&\quad + C'\epsilon\sqrt{\frac{p+\log K}{\max_{k=1:K}n_k}}+ C'\sqrt{\frac{1}{n_0}},
\end{align}
where the last inequality holds because $t_0$ is chosen to be the integer satisfying $t_0(\kappa_0)^{t_0} \asymp \sqrt{\frac{p}{n_0}} \gtrsim d(\htheta^{(0)[t_0]}, \bthetaks{0})$.

\subsubsection{Proof of lemmas}

\begin{proof}[Proof of Lemma \ref{lem: concentration transfer}]
	The proof is almost the same as the proofs of lemmas in Theorem \ref{thm: upper bound multitask est error}, so we do not repeat it here.
\end{proof}

\subsection{Proof of Theorem \ref{thm: lower bound transfer est error}}

\subsubsection{Lemmas}
Recall
\begin{align}
	\overline{\Theta}_{S}'(h) = \Big\{\{\overline{\btheta}^{(k)}\}_{k \in \{0\} \cup S} = \{(\wk{k}, \bmuk{k}_1, \bmuk{k}_2, \bSigmak{k})\}_{k \in \{0\} \cup S}: \bthetak{k} \in \overline{\Theta}, \max_{k \in S}\twonorm{\bbetak{k}-\bbetak{0}} \leq h\Big\}.
\end{align}

\begin{lemma}\label{lem: transfer d no outlier}
	When $n_0 \geq Cp$ with some constant $C > 0$, we have
	\begin{align}
		\inf_{\htheta^{(0)}} \sup_{S: |S| \geq s}\sup_{\substack{\{\othetaks{k}\}_{k \in \{0\}\cup S} \in \overline{\Theta}_{S}' \\ \mathbb{Q}_S}} &\tp\Bigg(d(\htheta^{(0)}, \bthetaks{0}) \geq  C_1\sqrt{\frac{p}{\ns + n_0}}+ C_1h\wedge \sqrt{\frac{p}{n_0}} + C_1\sqrt{\frac{1}{n_0}}\Bigg) \geq \frac{1}{10}.
	\end{align}
\end{lemma}

\begin{lemma}\label{lem: transfer d outlier}
	Denote $\widetilde{\epsilon} = \frac{K-s}{s}$. Then
	\begin{align}
		\inf_{\htheta^{(0)}} \sup_{S: |S| \geq s}\sup_{\substack{\{\othetaks{k}\}_{k \in  \{0\} \cup S} \in \overline{\Theta}_{S}' \\ \mathbb{Q}_S}} &\tp\Bigg(d(\htheta^{(0)}, \bthetaks{0}) \geq  \bigg(C_1\frac{\widetilde{\epsilon}}{\sqrt{\max_{k=1:K}n_k}}\bigg) \wedge \bigg(C_2\sqrt{\frac{1}{n_0}}\bigg)\Bigg) \geq \frac{1}{10}.	
	\end{align}
\end{lemma}

\begin{lemma}[The second variant of Theorem 5.1 in \citealpapp{chen2018robust}]\label{lem: from chen second}
	Given a series of distributions $\{\{\tp_{\theta}^{(k)}\}_{k=0}^K: \theta \in \Theta\}$, each of which is indexed by the same parameter $\theta \in \Theta$. Consider $\bxk{k} \sim (1-\widetilde{\epsilon})\tp^{(k)}_{\theta} + \widetilde{\epsilon}\mathbb{Q}^{(k)}$ independently for $k = 1:K$ and $\bxk{0} \sim \tp^{(0)}_{\theta}$. Denote the joint distribution of $\{\bxk{k}\}_{k=0}^K$ as $\tp_{(\wepsilon, \theta, \{\tq^{(k)}\}_{k=1}^K)}$. Then
	\begin{equation}
		\inf_{\widehat{\theta}} \sup_{\substack{\theta \in \Theta \\ \{\tq^{(k)}\}_{k=1}^K}} \tp_{(\wepsilon, \theta, \{\tq^{(k)}\}_{k=1}^K)}\left(\|\widehat{\theta}-\theta\| \geq C\varpi'(\wepsilon, \Theta)\right) \geq \frac{9}{20},
	\end{equation}
	where $\varpi'(\wepsilon, \Theta) \coloneqq \sup\big\{\|\theta_1-\theta_2\|: \max_{k = 1:K}d_{\textup{TV}}\big(\tp^{(k)}_{\theta_1}, \tp^{(k)}_{\theta_2}\big) \leq \wepsilon/(1-\wepsilon), d_{\textup{TV}}\big(\tp^{(0)}_{\theta_1}, \tp^{(0)}_{\theta_2}\big) \leq 1/20\big\}$.
\end{lemma}

\begin{lemma}\label{lem: transfer binomial lower bound}
	Consider two data generating mechanisms:
		\begin{enumerate}[(i)]
			\item $\bxk{k} \sim (1-\wepsilon')\tp_{\theta}^{(k)} + \wepsilon' \mathbb{Q}^{(k)}$ independently for $k = 1:K$ and $\bxk{0} \sim \tp^{(0)}_{\theta}$, where $\wepsilon' = \frac{K-s}{K}$;
			\item With a preserved set $S \subseteq 1:K$, generate $\{\bxk{k}\}_{k \in S^c} \sim \mathbb{Q}_S$ and $\bxk{k} \sim \tp_{\theta}^{(k)}$ independently for $k \in S$. And $\bxk{0} \sim \tp^{(0)}_{\theta}$.
		\end{enumerate}
	Denote the joint distributions of $\{\bxk{k}\}_{k=0}^K$ in (\rom{1}) and (\rom{2}) as $\tp_{(\wepsilon, \theta, \{\mathbb{Q}^{(k)}\}_{k=1}^K)}$ and $\tp_{(S, \theta, \mathbb{Q})}$, respectively. We claim that if
	\begin{equation}
		\inf_{\widehat{\theta}} \sup_{\substack{\theta \in \Theta \\ \{\tq^{(k)}\}_{k=1}^K}} \tp_{(\frac{K-s}{50K}, \theta, \{\tq^{(k)}\}_{k=1}^K)}\left(\|\widehat{\theta}-\theta\| \geq C\varpi'\left(\frac{K-s}{50K}, \Theta \right)\right) \geq \frac{9}{20},
	\end{equation} 
	then
	\begin{equation}
		\inf_{\widehat{\theta}} \sup_{S: |S| \geq s}\sup_{\substack{\theta \in \Theta \\ \tq_S}} \tp_{(S, \theta, \tq_S)}\left(\|\widehat{\theta}-\theta\| \geq C\varpi'\left(\frac{K-s}{50K}, \Theta \right)\right) \geq \frac{1}{10},
	\end{equation}
	where $\varpi'(\wepsilon, \Theta) \coloneqq \sup\big\{\|\theta_1-\theta_2\|: \max_{k = 1:K}\textup{KL}\big(\tp^{(k)}_{\theta_1}\| \tp^{(k)}_{\theta_2}\big) \leq[ \wepsilon/(1-\wepsilon)]^2, \textup{KL}\big(\tp^{(0)}_{\theta_1}\| \tp^{(0)}_{\theta_2}\big) \leq 1/400\big\}$ for any $\wepsilon \in (0,1)$.
\end{lemma}

\begin{lemma}\label{lem: transfer Sigma no outlier}
	When $n_0 \geq Cp$ with some constant $C > 0$, we have
	\begin{equation}
		\inf_{\hSigma^{(0)}} \sup_{S: |S| \geq s}\sup_{\substack{\{\othetaks{k}\}_{k \in \{0\}\cup S} \in \overline{\Theta}_{S}' \\ \mathbb{Q}_S}} \tp\Bigg(\twonorm{\hSigma^{(0)} - \bSigmaks{0}} \geq  C\sqrt{\frac{p}{n_0}}\Bigg) \geq \frac{1}{10}.	
	\end{equation}
\end{lemma}

\subsubsection{Main proof of Theorem \ref{thm: lower bound transfer est error}}

\subsubsection{Proof of lemmas}

\begin{proof}[Proof of Lemma \ref{lem: transfer d no outlier}]
	It's easy to see that $\overline{\Theta}_{S}' \supseteq \overline{\Theta}_{S, w}'  \cup \overline{\Theta}_{S, \bbeta}'\cup \overline{\Theta}_{S, \delta}'$, where 
\begin{align}
	\overline{\Theta}_{S, w}' &= \Big\{\{\overline{\btheta}^{(k)}\}_{k \in \{0\}\cup S}: \bmuk{k}_1 = \bm{1}_p/\sqrt{p}, \bmuk{k}_2 = - \bmuk{k}_1, \bSigmak{k} = \bm{I}_p, \wk{k} \in (c_w, 1-c_w)\Big\}, \\
	\overline{\Theta}_{S, \bbeta}' &= \Big\{\{\overline{\btheta}^{(k)}\}_{k \in \{0\}\cup S}: \bSigmak{k} = \bm{I}_p, \wk{k} = \frac{1}{2}, \twonorm{\bmuk{k}_1} \vee \twonorm{\bmuk{k}_2} \leq M, \max_{k \in S}\twonorm{\bbetak{k}-\bbetak{0}} \leq h\Big\}, \\
	\overline{\Theta}_{S, \delta}' &= \Big\{\{\overline{\btheta}^{(k)}\}_{k \in \{0\}\cup S}: \bSigmak{k} = \bm{I}_p, \wk{k} = \frac{1}{2}, \twonorm{\bmuk{k}_1} \vee \twonorm{\bmuk{k}_2} \leq M, \bmuk{k}_1 = -\frac{1}{2}\bmu_0, \\
	&\hspace{3cm} \bmuk{k}_2 = \frac{1}{2}\bmu_0 + \bm{u}, \twonorm{\bu} \leq 1\Big\}.
\end{align}

\noindent(\rom{1}) By fixing an $S$ and a $\mathbb{Q}_S$, we want to show
\begin{align}
	\inf_{\htheta^{(0)}} \sup_{\{\othetaks{k}\}_{k \in \{0\}\cup S} \in \overline{\Theta}_{S, \bbeta}'} \tp\Bigg(\twonorm{\hbeta^{(0)}-\bbetaks{0}} \vee \twonorm{\hbeta^{(0)} +\bbetaks{0}} \geq C\sqrt{\frac{p}{\ns+n_0}}\Bigg) \geq \frac{1}{4}.
\end{align}
By Lemma \ref{lem: packing number of sphere quadrant}, $\exists$ a quadrant $\mathcal{Q}_{\bv}$ of $\mathbb{R}^p$ and a $r/8$-packing of $(r\mathcal{S}^p)\cap \mathcal{Q}_{\bv}$ under Euclidean norm: $\{\widetilde{\bmu}_j\}_{j=1}^N$, where $r = (c\sqrt{p/(\ns+n_0)}) \wedge M \leq 1$ with a small constant $c > 0$ and $N \geq (\frac{1}{2})^p 8^{p-1} = \frac{1}{2}\times 4^{p-1} \geq 2^{p-1}$ when $p \geq 2$. For any $\bmu \in \mathbb{R}^p$, denote distribution $\frac{1}{2}\mathcal{N}(\bmu, \bm{I}_p) + \frac{1}{2}\mathcal{N}(-\bmu, \bm{I}_p)$ as $\tp_{\bmu}$. Then
\begin{align}
	\text{LHS} &\geq \inf_{\hmu} \sup_{\bmu \in (r\mathcal{S}^p)\cap \mathcal{Q}_{\bv}} \tp\Bigg(\twonorm{\hmu-\bmu} \wedge \twonorm{\hmu+\bmu} \geq C\sqrt{\frac{p}{\ns}}\Bigg) \\
	&\geq \inf_{\hmu} \sup_{\bmu \in (r\mathcal{S}^p)\cap \mathcal{Q}_{\bv}} \tp\Bigg(\twonorm{\hmu-\bmu} \geq C\sqrt{\frac{p}{\ns}}\Bigg), \label{eq: lower bdd eq mu 1 transfer}
\end{align}
where the last inequality holds because it suffices to consider estimator $\hmu$ satisfying $\hmu(X)\in (r\mathcal{S}^p)\cap \mathcal{Q}_{\bv}$ almost surely. In addition, for any $\bx$, $\by \in \mathcal{Q}_{\bv}$, $\twonorm{\bx-\by} \leq \twonorm{\bx+\by}$.

By Lemma \ref{lem: lem 8.4 with mu},
\begin{align}
	\text{KL}\left(\prod_{k \in \{0\}\cup S}\tp_{\widetilde{\bmu}_j}^{\otimes n_k} \cdot \mathbb{Q}_S \bigg\| \prod_{k \in \{0\}\cup S}\tp_{\widetilde{\bmu}_{j'}}^{\otimes n_k} \cdot \mathbb{Q}_S\right) &= \sum_{k \in \{0\}\cup S}n_k \text{KL}(\tp_{\widetilde{\bmu}_j} \| \tp_{\widetilde{\bmu}_{j'}})\\
	&\leq \sum_{k \in \{0\}\cup S}n_k \cdot 8\twonorm{\widetilde{\bmu}_j}^2 \twonorm{\widetilde{\bmu}_j - \widetilde{\bmu}_{j'}}^2 \\
	&\leq 32(\ns+n_0) r^2 \\
	&\leq 32\ns c^2 \cdot \frac{2(p-1)}{\ns+n_0}\\
	&\leq \frac{64c^2}{\log 2}\log N.
\end{align}
By Lemma \ref{lem: fano},
\begin{align}
	\text{LHS of \eqref{eq: lower bdd eq mu 1 transfer}} \geq 1-\frac{\log 2}{\log N} - \frac{64c^2}{\log 2}\geq 1-\frac{1}{p-1}-\frac{1}{4} \geq \frac{1}{4},
\end{align}
when $C = c/2$, $p \geq 3$ and $c =\sqrt{\log 2}/16$.

\noindent(\rom{2}) By fixing an $S$ and a $\mathbb{Q}_S$, we want to show
\begin{align}
	\inf_{\htheta^{(0)}} \sup_{\{\othetaks{k}\}_{k \in \{0\} \cup S} \in \overline{\Theta}_S'} \tp\Bigg(\twonorm{\hbeta^{(0)}-\bbetaks{0}} \vee \twonorm{\hbeta^{(0)}+\bbetaks{0}} \geq C\bigg[h \wedge \bigg(c\sqrt{\frac{p}{n_0}}\bigg)\bigg]\bigg\}\Bigg) \geq \frac{1}{4}. \\ \label{eq: lower bdd eq mu 2 transfer}
\end{align}
By Lemma \ref{lem: packing number of sphere quadrant}, $\exists$ a quadrant $\mathcal{Q}_{\bv}$ of $\mathbb{R}^p$ and a $r/8$-packing of $(r\mathcal{S}^{p-1})\cap \mathcal{Q}_{\bv}$ under Euclidean norm: $\{\widetilde{\bm{\vartheta}}_j\}_{j=1}^N$, where $r = h \wedge (c\sqrt{p/n_0}) \wedge M \leq 1$ with a small constant $c > 0$ and $N \geq (\frac{1}{2})^{p-1} 8^{p-2} = \frac{1}{2}\times 4^{p-2} \geq 2^{p-2}$ when $p \geq 3$. WLOG, assume $M \geq 2$. Denote $\widetilde{\bmu}_j = (1, \widetilde{\bm{\vartheta}}_j^\top)^\top \in \mathbb{R}^p$. Let $\bmuks{k}_1 = \widetilde{\bmu} = (1, \bm{0}_{p-1})^\top$ for all $k \in S$ and $\bmuks{0}_1 = \bmu = (1, \bm{\vartheta})$ with $\bm{\vartheta} \in (r\mathcal{S}^{p-1}) \cap \mathcal{Q}_{\bv}$. For any $\bmu \in \mathbb{R}^p$, denote distribution $\frac{1}{2}\mathcal{N}(\bmu, \bm{I}_p) + \frac{1}{2}\mathcal{N}(-\bmu, \bm{I}_p)$ as $\tp_{\bmu}$. Then similar to the arguments in (\rom{1}),
\begin{align}
	\text{LHS} &\geq \inf_{\hmu} \sup_{\substack{\bm{\vartheta} \in (r\mathcal{S}^{p-1})\cap \mathcal{Q}_{\bv}\\ \bmu = (1, \bm{\vartheta})^\top}} \tp\Bigg(\twonorm{\hmu-\bmu} \wedge \twonorm{\hmu+\bmu} \geq C\bigg[h \wedge \bigg(c\sqrt{\frac{p}{n_0}}\bigg)\bigg]\Bigg) \\
	&\geq \inf_{\hmu} \sup_{\substack{\bm{\vartheta} \in (r\mathcal{S}^{p-1})\cap \mathcal{Q}_{\bv}\\ \bmu = (1, \bm{\vartheta})^\top}}\tp\Bigg(\twonorm{\hmu-\bmu} \geq C\bigg[h \wedge \bigg(c\sqrt{\frac{p}{n_0}}\bigg)\bigg]\Bigg).
\end{align}

Then by Lemma \ref{lem: lem 8.4 with mu},
\begin{align}
	\text{KL}\left(\tp_{\widetilde{\bmu}_j}^{\otimes n_0} \cdot\prod_{k \in S}\tp_{\widetilde{\bmu}}^{\otimes n_k} \cdot  \mathbb{Q}_S \bigg\| \tp_{\widetilde{\bmu}_{j'}}^{\otimes n_0} \cdot\prod_{k \in S}\tp_{\widetilde{\bmu}}^{\otimes n_k} \cdot  \mathbb{Q}_S \right) &= n_0 \text{KL}(\tp_{\widetilde{\bmu}_j} \| \tp_{\widetilde{\bmu}_{j'}}) \\
	&\leq n_0 \cdot 8\twonorm{\widetilde{\bmu}_j}^2\twonorm{\widetilde{\bmu}_j - \widetilde{\bmu}_{j'}}^2 \\
	&\leq 32n_0 r^2 \\
	&\leq 32n_0 c^2 \cdot \frac{3(p-2)}{n_0} \\
	&\leq \frac{96c^2}{\log 2}\log N,
\end{align}
when $n_0 \geq (c^2 \vee M^{-2})p$ and $p \geq 3$. By Fano's lemma (See Corollary 2.6 in \citealpapp{tsybakov2009introduction}),
\begin{align}
	\text{LHS of \eqref{eq: lower bdd eq mu 2 transfer}} \geq 1-\frac{\log 2}{\log N} - \frac{96c^2}{\log 2} \geq 1-\frac{1}{p-2} - \frac{1}{4}\geq  \frac{1}{4},
\end{align}
when $C = 1/2$, $p \geq 4$ and $c= \sqrt{(\log 2)/384}$.

\noindent(\rom{3}) We want to show
\begin{equation}
	\inf_{\htheta^{(0)}} \sup_{S: |S| \geq s}\sup_{\substack{\{\othetaks{k}\}_{k \in \{0\} \cup S} \in \overline{\Theta}_{S, w}' \\ \mathbb{Q}_S}} \tp\bigg(\norm{\hw^{(0)}-\wks{0}} \geq C\sqrt{\frac{1}{n_0}}\Bigg) \geq \frac{1}{4}.
\end{equation}
The argument is similar to (\rom{2}). The only two differences here are that the dimension of interested parameter $w$ equals 1, and Lemma \ref{lem: lem 8.4 with mu} is replaced by Lemma \ref{lem: lem 8.4 with w}.

\noindent(\rom{4}) We want to show
\begin{equation}\label{eq: lower bound eq delta tl}
	\inf_{\htheta^{(0)}} \sup_{S: |S| \geq s}\sup_{\substack{\{\othetaks{k}\}_{k \in \{0\} \cup S} \in \overline{\Theta}_{S, \delta}' \\ \mathbb{Q}_S}} \tp\bigg(\norm{\hdelta^{(0)}-\deltaks{0}} \geq C\sqrt{\frac{1}{n_0}}\Bigg) \geq \frac{1}{4}
\end{equation}
The argument is similar to (\rom{2}).

Finally, we get the desired conclusion by combining (\rom{1})-(\rom{4}).

\end{proof}

\begin{proof}[Proof of Lemma \ref{lem: transfer d outlier}]
	Let $\widetilde{\epsilon} = \frac{K-s}{s}$ and $\wepsilon' = \frac{K-s}{K}$. Since $s/K \geq c > 0$, $\wepsilon \lesssim \wepsilon'$. Denote $\Upsilon_S = \{\{\bmuk{k}\}_{k \in \{0\}\cup S}: \bmuk{k} \in \mathbb{R}_{+}^p, \max_{k \in S}\twonorm{\bmuk{k}-\bmuk{0}} \leq h/2, \twonorm{\bmuk{k}} \leq M\}$. For any $\bmu \in \mathbb{R}$, denote distribution $\frac{1}{2}\mathcal{N}(\bmu, \bm{I}_p) + \frac{1}{2}\mathcal{N}(-\bmu, \bm{I}_p)$ as $\tp_{\bmu}$, and denote $\prod_{k \in S}\tp_{\bmuk{k}}^{\otimes n_k}$ as $\tp_{\{\bmuk{k}\}_{k \in S}}$. It suffices to show
\begin{equation}\label{eq: transfer d outlier ts}
	\inf_{\htheta^{(0)}} \sup_{S: |S| \geq s}\sup_{\substack{\{\othetaks{k}\}_{k \in \{0\} \cup S} \in \overline{\Theta}_{S} \\ \mathbb{Q}_S}} \tp\Bigg(\twonorm{\hmu^{(0)} - \bmuks{0}} \geq  \bigg(C_1\widetilde{\epsilon}'\sqrt{\frac{1}{\max_{k=1:K}n_k}} \bigg) \wedge \bigg(C_2\sqrt{\frac{1}{n_0}}\bigg)\Bigg) \geq \frac{1}{10}.	
\end{equation}
where $\tp = \tp_{\bmuk{0}}^{\otimes n_0}\cdot \tp_{\{\bmuk{k}\}_{k \in S}} \cdot \mathbb{Q}_S$.

For any $\bmu \in \mathbb{R}$, denote distribution $\frac{1}{2}\mathcal{N}(\bmu, \bm{I}_p) + \frac{1}{2}\mathcal{N}(-\bmu, \bm{I}_p)$ as $\tp_{\bmu}$. WLOG, assume $M \geq 1$. For any $\widetilde{\bmu}_1$, $\widetilde{\bmu}_2 \in \mathbb{R}^p$ with $\twonorm{\widetilde{\bmu}_1} = \twonorm{\widetilde{\bmu}_2} = 1$, by Lemma \ref{lem: lem 8.4 with mu},
\begin{equation}
	\max_{k =1:K}\text{KL}\big(\tp_{\widetilde{\bmu}_1}^{\otimes n_k} \| \tp_{\widetilde{\bmu}_2}^{\otimes n_k}\big) \leq \max_{k=1:K}n_k \cdot 8\twonorm{\widetilde{\bmu}_1 - \widetilde{\bmu}_2}^2.
\end{equation}
for any $k = 1:K$. Let $8\max_{k=1:K}n_k \cdot \twonorm{\widetilde{\bmu}_1 - \widetilde{\bmu}_2}^2 \leq (\frac{\wepsilon'}{1-\wepsilon'})^2$, then $\twonorm{\widetilde{\bmu}_1 - \widetilde{\bmu}_2} \leq C\sqrt{\frac{1}{\max_{k=1:K}n_k}}\wepsilon'$ for some constant $C > 0$. On the other hand, let $\text{KL}\big(\tp_{\widetilde{\bmu}_1}^{\otimes n_0} \| \tp_{\widetilde{\bmu}_2}^{\otimes n_0}\big) = 8n_0 \cdot \twonorm{\widetilde{\bmu}_1 - \widetilde{\bmu}_2}^2 \leq 1/100$, then $\twonorm{\widetilde{\bmu}_1 - \widetilde{\bmu}_2} \leq \sqrt{\frac{1}{800}}\sqrt{\frac{1}{n_0}}$ for some constant $C > 0$. Then \eqref{eq: transfer d outlier ts} follows by Lemma \ref{lem: transfer binomial lower bound}.
\end{proof}

\begin{proof}[Proof of Lemma \ref{lem: from chen second}]
	The proof is similar to the proof of Theorem 5.1 in \citeapp{chen2018robust}, so we omit it here.
\end{proof}

\begin{proof}[Proof of Lemma \ref{lem: transfer Sigma no outlier}]
	This can be similarly shown by Assouad's Lemma as in the proof of Lemma \ref{lem: multitask Sigma no outlier}. We omit the proof here.
\end{proof}

\subsection{Proof of Theorem \ref{thm: upper bound transfer classification error}}
The result follows from \eqref{eq: proof of thm mtl classification error eq 1} and Theorem \ref{thm: upper bound transfer est error}.

\subsection{Proof of Theorem \ref{thm: lower bound transfer classification error}}

\subsubsection{Lemmas}
\begin{lemma}\label{lem: transfer misclassification error no outlier}
	Assume $n_0 \geq Cp$ and $\Delta^{(0)} \geq C' > 0$ with some constants $C$, $C' > 0$. We have
	\begin{align}
		\inf_{\widehat{\mC}^{(0)}} \sup_{S: |S| \geq s}\sup_{\substack{\{\othetaks{k}\}_{k \in \{0\} \cup S} \in \overline{\Theta}_{S}' \\ \mathbb{Q}_S}} &\tp\Bigg(R_{\othetaks{0}}(\widehat{\mC}^{(0)}) - R_{\othetaks{0}}(\mC_{\othetaks{0}}) \geq  C_1\frac{p}{\ns + n_0} + C_2h^2\wedge \frac{p}{n_0} + \frac{1}{n_0}\Bigg) \geq \frac{1}{10}.	
	\end{align}
\end{lemma}

\begin{lemma}\label{lem: transfer misclassification error outlier}
	Denote $\widetilde{\epsilon} = \frac{K-s}{s}$. We have
	\begin{equation}
		\inf_{\widehat{\mC}^{(0)}} \sup_{S: |S| \geq s}\sup_{\substack{\{\othetaks{k}\}_{k \in \{0\} \cup S} \in \overline{\Theta}_{S}^{(T)} \\ \mathbb{Q}_S}} \tp\Bigg(R_{\othetaks{0}}(\widehat{\mC}^{(0)}) - R_{\othetaks{0}}(\mC_{\othetaks{0}}) \geq  \bigg(C_1\frac{\widetilde{\epsilon}'^2}{\max_{k=1:K}n_k}\bigg) \wedge \bigg(C_2\sqrt{\frac{1}{n_0}}\bigg)\Bigg) \geq \frac{1}{10}.
	\end{equation}
\end{lemma}

\subsubsection{Main proof of Theorem \ref{thm: lower bound transfer classification error}}
Combine Lemmas \ref{lem: transfer misclassification error no outlier} and \ref{lem: transfer misclassification error outlier} to finish the proof.

\subsubsection{Proof of lemmas}
\begin{proof}[Proof of Lemma \ref{lem: transfer misclassification error no outlier}]
	We proceed with similar proof ideas used in the proof of Lemma \ref{lem: transfer misclassification error no outlier}. Recall the definitions and proof idea of Lemma \ref{lem: transfer d no outlier}.
We have $\overline{\Theta}_{S}' \supseteq \overline{\Theta}_{S, w}'  \cup \overline{\Theta}_{S, \bbeta}'\cup \overline{\Theta}_{S, \delta}'$, where 
\begin{align}
	\overline{\Theta}_{S, w}' &= \Big\{\{\overline{\btheta}^{(k)}\}_{k \in \{0\}\cup S}: \bmuk{k}_1 = \bm{1}_p/\sqrt{p}, \bmuk{k}_2 = - \bmuk{k}_1, \bSigmak{k} = \bm{I}_p, \wk{k} \in (c_w, 1-c_w)\Big\}, \\
	\overline{\Theta}_{S, \bbeta}' &= \Big\{\{\overline{\btheta}^{(k)}\}_{k \in \{0\}\cup S}: \bSigmak{k} = \bm{I}_p, \wk{k} = \frac{1}{2}, \twonorm{\bmuk{k}_1} \vee \twonorm{\bmuk{k}_2} \leq M, \max_{k \in S}\twonorm{\bbetak{k}-\bbetak{0}} \leq h\Big\}, \\
	\overline{\Theta}_{S, \delta}' &= \Big\{\{\overline{\btheta}^{(k)}\}_{k \in \{0\}\cup S}: \bSigmak{k} = \bm{I}_p, \wk{k} = \frac{1}{2}, \twonorm{\bmuk{k}_1} \vee \twonorm{\bmuk{k}_2} \leq M, \bmuk{k}_1 = -\frac{1}{2}\bmu_0, \\
	&\hspace{3cm} \bmuk{k}_2 = \frac{1}{2}\bmu_0 + \bm{u}, \twonorm{\bu} \leq 1\Big\}.
\end{align}

Recall the mis-clustering error for GMM associated with parameter set $\otheta$ of any classifier $\mC$ is $R_{\otheta}(\mC) = \min_{\pi: \{1,2\} \rightarrow \{1,2\}}\tp_{\otheta}(\mC(Z) \neq \pi(Y))$. To help the analysis, following \citeapp{azizyan2013minimax} and \citeapp{cai2019chime}, we define a surrogate loss $L_{\otheta}(\mC) = \min_{\pi: \{1,2\} \rightarrow \{1,2\}}\tp_{\otheta}(\mC(Z) \neq \pi(\mC_{\otheta}(Z)))$, where $\mC_{\otheta}$ is the Bayes classifier. Suppose $\sigma = \sqrt{0.005}$.

\noindent(\rom{1}) We want to show
\begin{equation}\label{eq: lemma 36 eq 1}
	\inf_{\widehat{\mC}^{(0)}} \sup_{S: |S| \geq s}\sup_{\substack{\{\othetaks{k}\}_{k \in \{0\}\cup S} \in \overline{\Theta}_{S}' \\ \mathbb{Q}_S}} \tp\Bigg(R_{\othetaks{0}}(\widehat{\mC}^{(0)}) - R_{\othetaks{0}}(\mC_{\othetaks{0}}) \geq  C\sqrt{\frac{p}{\ns+n_0}}\Bigg) \geq \frac{1}{4}.
\end{equation}
Consider $S = 1:K$ and space $\overline{\Theta}_0' = \{\{\overline{\btheta}^{(k)}\}_{k=0}^K: \bSigmak{k} = \bm{I}_p, \wk{k} = 1/2, \bmuk{k}_1 = \bmu_1, \bmuk{k}_2 = \bmu_2, \twonorm{\bmu_1} \vee \twonorm{\bmu_2} \leq M\}$. And 
\begin{equation}
	\text{LHS of \eqref{eq: lemma 19 eq 1}} \geq \inf_{\widehat{\mC}^{(0)}}\sup_{\{\othetaks{k}\}_{k=0}^K \in \overline{\Theta}_0'} \tp\Bigg(R_{\othetaks{0}}(\widehat{\mC}^{(0)}) - R_{\othetaks{0}}(\mC_{\othetaks{0}}) \geq  C\sqrt{\frac{p}{\ns+n_0}} \Bigg).
\end{equation}
Let $r = c\sqrt{p/(\ns+n_0)} \leq 0.001$ with some small constant $c > 0$. For any $\bmu \in \mathbb{R}^p$, denote distribution $\frac{1}{2}\mathcal{N}(\bmu, \bm{I}_p) + \frac{1}{2}\mathcal{N}(-\bmu, \bm{I}_p)$ as $\tp_{\bmu}$. Consider a $r/4$-packing of $r\mathcal{S}^{p-1}$: $\{\widetilde{\bm{v}}_j\}_{j=1}^N$. By Lemma \ref{lem: packing number of sphere}, $N \geq 4^{p-1}$. Denote $\widetilde{\bmu}_j = (\sigma, \widetilde{\bm{v}}_j^\top)^\top \in \mathbb{R}^p$, where $\sigma = \sqrt{0.005}$. Then by definition of KL divergence and Lemma 8.4 in \citeapp{cai2019chime},
\begin{align}
	\text{KL}\left(\prod_{k \in \{0\}\cup S}\tp_{\widetilde{\bmu}_j}^{\otimes n_k} \cdot \mathbb{Q}_S \bigg\| \prod_{k \in \{0\}\cup S}\tp_{\widetilde{\bmu}_{j'}}^{\otimes n_k} \cdot \mathbb{Q}_S\right) &= \sum_{k \in \{0\}\cup S}n_k \text{KL}(\tp_{\widetilde{\bmu}_j} \| \tp_{\widetilde{\bmu}_{j'}}) \\
	&\leq (n_S+n_0) \cdot 8(1+\sigma^2)\twonorm{\widetilde{\bmu}_j - \widetilde{\bmu}_{j'}}^2 \\
	&\leq 32(1+\sigma^2)(n_S+n_0) r^2 \\
	&\leq 32(1+\sigma^2)(n_S+n_0) \cdot c^2\frac{2(p-1)}{n_S+n_0} \\
	&\leq \frac{32(1+\sigma^2)c^2}{\log 2}\log N.
\end{align}
For simplicity, we write $L_{\otheta}$ with $\otheta \in \overline{\Theta}_0$ and $\bmu_1 = -\bmu_2 = \bmu$ as $L_{\bmu}$. By Lemma 8.5 in \citeapp{cai2019chime},
\begin{equation}
	L_{\widetilde{\bmu}_i}(\mC_{\widetilde{\bmu}_j}) \geq \frac{1}{\sqrt{2}}g\left(\frac{\sqrt{\sigma^2+r^2}}{2}\right)\frac{\twonorm{\widetilde{\bmu}_i - \widetilde{\bmu}_j}}{\twonorm{\widetilde{\bmu}_i}} \geq \frac{1}{\sqrt{2}}\cdot 0.15\cdot \frac{r/4}{\sqrt{\sigma^2+r^2}} \geq 2r,
\end{equation}
where $g(x) = \phi(x)[\phi(x)-x\Phi(x)]$. The last inequality holds because $\sqrt{\sigma^2+r^2} \geq \sqrt{2}\sigma$ and $g(\sqrt{\sigma^2+r^2}/2) \geq 0.15$ when $r^2 \leq \sigma^2 =0.001$. Then by Lemma 3.5 in \citealpapp{cai2019chime} (Proposition 2 in \citeapp{azizyan2013minimax}), for any classifier $\mC$, and $i \neq j$,
\begin{equation}\label{eq: lemma 36 eq 2}
	L_{\widetilde{\bmu}_i}(\mC) + L_{\widetilde{\bmu}_j}(\mC) \geq L_{\widetilde{\bmu}_i}(\mC_{\widetilde{\bmu}_j}) - \sqrt{\text{KL}(\tp_{\widetilde{\bmu}_i}\| \tp_{\widetilde{\bmu}_j})/2} \geq 2r - r = c\sqrt{\frac{p}{n_S+n_0}}.
\end{equation}
For any $\widehat{\mC}^{(0)}$, consider a test $\psi^* = \argmin_{j =1:N} L_{\widetilde{\bmu}_j}(\widehat{\mC}^{(0)})$. Therefore if there exists $j_0$ such that $L_{\widetilde{\bmu}_{j_0}}(\widehat{\mC}^{(0)}) < \frac{c}{2}\sqrt{\frac{p}{n_S+n_0}}$, then by \eqref{eq: lemma 36 eq 2}, we must have $\psi^* = j_0$. Let $C_1 \leq c/2$, then by Fano's lemma (Corollary 6 in \citealpapp{tsybakov2009introduction}) 
\begin{align}
	\inf_{\widehat{\mC}^{(0)}}\sup_{\{\othetaks{k}\}_{k=0}^K \in \overline{\Theta}_0'} \tp\Bigg(L_{\othetaks{0}}(\widehat{\mC}^{(0)}) \geq  C_1\sqrt{\frac{p}{n_S+n_0}} \Bigg) 
	&\geq \inf_{\widehat{\mC}^{(0)}}\sup_{j=1:N} \tp\Bigg(L_{\widetilde{\bmu}^{(j)}}(\widehat{\mC}^{(0)}) \geq  C_1\sqrt{\frac{p}{n_S+n_0}} \Bigg) \\
	&\geq \inf_{\widehat{\mC}^{(0)}}\sup_{j=1:N} \tp\Bigg(\psi^* \neq j \Bigg) \\
	&\geq \inf_{\psi}\sup_{j=1:N} \tp\Bigg(\psi \neq j \Bigg) \\
	&\geq 1-\frac{\log 2}{\log N} - \frac{32(1+\sigma^2)c^2}{\log 2}\\
	&\geq \frac{1}{4},
\end{align}
when $p \geq 2$ and  $c = \sqrt{\frac{\log 2}{128(1+\sigma^2)}}$. Then apply Lemma \ref{lem: surrogate loss and classification error} to get the \eqref{eq: lemma 36 eq 1}.

\noindent(\rom{2}) We want to show
\begin{equation}\label{eq: lemma 36 eq 3}
	\inf_{\widehat{\mC}^{(0)}}\sup_{S: |S| \geq s}\sup_{\substack{\{\othetaks{k}\}_{k \in \{0\}\cup S} \in \overline{\Theta}_{S, \bbeta}' \\ \mathbb{Q}_S}} \tp\Bigg(R_{\othetaks{0}}(\widehat{\mC}^{(0)}) - R_{\othetaks{0}}(\mC_{\othetaks{0}}) \geq  C \left(h\wedge \sqrt{\frac{p}{n_0}}\right)\Bigg) \geq \frac{1}{4}.
\end{equation}
Fixing an $S$ and a $\mathbb{Q}_S$. Suppose $1 \in S$. We have
\begin{equation}\label{eq: lemma 36 eq 4}
	\text{LHS of \eqref{eq: lemma 36 eq 3}} \geq \inf_{\widehat{\mC}^{(0)}} \sup_{\substack{\{\othetaks{k}\}_{k \in \{0\} \cup S} \in \overline{\Theta}_{S, \bbeta}' \\ \mathbb{Q}_S}} \tp\bigg(R_{\othetaks{0}}(\widehat{\mC}^{(0)}) - R_{\othetaks{0}}(\mC_{\othetaks{0}}) \geq C \left(h\wedge \sqrt{\frac{p}{n_0}}\right)\Bigg).
\end{equation}
Let $r = h \wedge (c\sqrt{p/n_0}) \wedge M$ with a small constant $c > 0$. For any $\bmu \in \mathbb{R}^p$, denote distribution $\frac{1}{2}\mathcal{N}(\bmu, \bm{I}_p) + \frac{1}{2}\mathcal{N}(-\bmu, \bm{I}_p)$ as $\tp_{\bmu}$. Consider a $r/4$-packing of $r\mathcal{S}^{p-1}$. By Lemma \ref{lem: packing number of sphere}, $N \geq 4^{p-1}$. Denote $\widetilde{\bmu}_j = (\sigma, \widetilde{\bm{v}}_j^\top)^\top \in \mathbb{R}^p$. WLOG, assume $M \geq 2$. Let $\bmuks{k}_1 = \widetilde{\bmu} = (\sigma, \bm{0}_{p-1})^\top$ for all $k \in S$ and $\bmuks{0}_1 = \bmu = (1, \bm{\vartheta})$ with $\bm{\vartheta} \in (r\mathcal{S}^{p-1}) \cap \mathcal{Q}_{\bv}$. Then by following the same arguments in part (\rom{2}) of the proof of Lemma \ref{lem: transfer d no outlier}, we can show that the RHS of \eqref{eq: lemma 36 eq 4} is larger than or equal to $1/4$ when $p \geq 3$.

\noindent(\rom{3}) We want to show
\begin{equation}
	\inf_{\widehat{\mC}^{(0)}} \sup_{S: |S| \geq s}\sup_{\substack{\{\othetaks{k}\}_{k \in \{0\} \cup S} \in \overline{\Theta}_{S}' \\ \mathbb{Q}_S}} \tp\Bigg(R_{\othetaks{0}}(\widehat{\mC}^{(0)}) - R_{\othetaks{0}}(\mC_{\othetaks{0}}) \geq  C h_w^2\wedge \frac{1}{n_0} \Bigg) \geq \frac{1}{4}.
\end{equation}
This can be similarly proved by following the arguments in part (\rom{1}) with Lemmas \ref{lem: surrogate loss and classification error} and \ref{lem: surrogate loss mismatch bound}.

\noindent(\rom{4}) We want to show
\begin{equation}
	\inf_{\widehat{\mC}^{(0)}} \sup_{S: |S| \geq s}\sup_{\substack{\{\othetaks{k}\}_{k \in \{0\} \cup S} \in \overline{\Theta}_{S}' \\ \mathbb{Q}_S}} \tp\Bigg(R_{\othetaks{0}}(\widehat{\mC}^{(0)}) - R_{\othetaks{k}}(\mC_{\othetaks{0}}) \geq  C h_{\bbeta}^2\wedge \frac{p}{n_0} \Bigg) \geq \frac{1}{4}.
\end{equation}

The conclusion can be obtained immediately from (\rom{2}), by noticing that $\overline{\Theta}_{S, \bbeta}' \supseteq \overline{\Theta}_{S, \bmu}'$.

Finally, we get the desired conclusion by combining (\rom{1})-(\rom{4}).
\end{proof}

\begin{proof}[Proof of Lemma \ref{lem: transfer misclassification error outlier}]
	By Lemma \ref{lem: surrogate loss and classification error}, it suffices to prove
\begin{equation}
	\inf_{\widehat{\mC}^{(0)}} \sup_{S: |S| \geq s}\sup_{\substack{\{\othetaks{k}\}_{k \in \{0\} \cup S} \in \overline{\Theta}_{S}' \\ \mathbb{Q}_S}} \tp\Bigg(L_{\othetaks{0}}(\widehat{\mC}^{(0)})  \geq  C_1\frac{\widetilde{\epsilon}'^2}{\max_{k=1:K}n_k}\wedge \frac{1}{n_0}\Bigg) \geq \frac{1}{10}.
\end{equation}
For any $\bmu \in \mathbb{R}^p$, denote distribution $\frac{1}{2}\mathcal{N}(\bmu, \bm{I}_p) + \frac{1}{2}\mathcal{N}(-\bmu, \bm{I}_p)$ as $\tp_{\bmu}$. For simplicity, we write $L_{\otheta}$ with $\otheta$ satisfying $\bmu_1 = -\bmu_2 = \bmu$, $w = 1/2$ and $\bSigma = \bm{I}_p$ as $L_{\bmu}$. Consider $L_{\bmu}(\mC_{\bmu'})$ as a loss function between $\bmu$ and $\bmu'$ in Lemmas \ref{lem: from chen second} and \ref{lem: transfer binomial lower bound}. Considering $\twonorm{\bmu} = \twonorm{\bmu'} = 1$, by Lemma \ref{lem: lem 8.4 with mu}, note that
\begin{align}
	\max_{k = 1:K}\text{KL}(\tp_{\bmu}^{\otimes n_k}\| \tp_{\bmu'}^{\otimes n_{k}}) &\leq 8\max_{k = 1:K}n_k \cdot \twonorm{\bmu - \bmu'}^2,\\
	\text{KL}(\tp_{\bmu}^{\otimes n_0}\| \tp_{\bmu'}^{\otimes n_0}) &\leq 8n_0 \cdot \twonorm{\bmu - \bmu'}^2.
\end{align}
By Lemma 8.5 in \citeapp{cai2019chime}, this implies for some constants $c, C > 0$
\begin{align}
	&\sup\left\{L_{\bmu}(\mC_{\bmu'}): \max_{k = 1:K}\text{KL}(\tp_{\bmu}^{\otimes n_k}\| \tp_{\bmu'}^{\otimes n_{k}}) \leq (\wepsilon'/(1-\wepsilon))^2, \text{KL}(\tp_{\bmu}^{\otimes n_0}\| \tp_{\bmu'}^{\otimes n_0}) \leq 1/100\right\} \\
	&\geq \sup\left\{c\twonorm{\bmu - \bmu'}:  \max_{k = 1:K}\text{KL}(\tp_{\bmu}^{\otimes n_k}\| \tp_{\bmu'}^{\otimes n_{k}}) \leq (\wepsilon'/(1-\wepsilon))^2, \text{KL}(\tp_{\bmu}^{\otimes n_0}\| \tp_{\bmu'}^{\otimes n_0}) \leq 1/100\right\} \\
	&\geq \sup\left\{c\twonorm{\bmu - \bmu'}:  8\max_{k = 1:K}n_k \cdot\twonorm{\bmu - \bmu'}^2  \leq (\wepsilon'/(1-\wepsilon))^2, 8n_0 \cdot \twonorm{\bmu - \bmu'}^2 \leq 1/800\right\} \\
	&= C\cdot \frac{\wepsilon'}{\sqrt{\max_{k = 1:K}n_k}} \wedge \sqrt{\frac{1}{n_0}}.
\end{align}
Then apply Lemmas \ref{lem: from chen second} and \ref{lem: transfer binomial lower bound} to get the desired bound.
\end{proof}

\subsection{Proof of Theorem \ref{thm: tl alignment}}
Denote $\xi = \max_{k \in \{0\}\cup S} \min_{r_k = \pm 1}\twonorm{r_k\hbeta^{(k)[0]} - \bbetaks{k}} = \max_{k \in \{0\}\cup S}(\twonorm{\hbeta^{(k)[0]} - \bbetaks{k}} \wedge \twonorm{\hbeta^{(k)[0]} + \bbetaks{k}})$. WLOG, assume $S = \{1, \ldots, s\}$ and $r^*_k = 1$ for all $k \in \{0\}\cup S$. Hence $\xi = \max_{k \in \{0\}\cup S}\twonorm{\hbeta^{(k)[0]} - \bbetaks{k}}$. WLOG, consider $\widehat{r}_k = 1$ for all $k \in S$ (i.e., the tasks in $S$ are already well-aligned). Consider
\begin{align}
    (1, \widehat{\bm{r}}) &= (\underbrace{1}_{\text{target}}, \underbrace{1, \ldots, 1, 1}_{S}, \underbrace{r_{s+1}, \ldots, r_K}_{\text{outlier tasks}}),\\
    (-1, \widehat{\bm{r}}) &= (\underbrace{-1}_{\text{target}}, \underbrace{1, \ldots, 1, 1}_{S}, \underbrace{r_{s+1}, \ldots, r_K}_{\text{outlier tasks}}).
\end{align}
It suffices to prove that
\begin{align}
	 \text{score}((-1, \widehat{\bm{r}})) - \text{score}((1, \widehat{\bm{r}})) > 0.
\end{align}
In fact,
\begin{align}
	\text{score}((-1, \widehat{\bm{r}})) - \text{score}((1, \widehat{\bm{r}})) &= 2\underbrace{\sum_{k=1}^s \twonorm{\hbeta^{(0)[0]} + \hbeta^{(k)[0]}}}_{[1]} + 2\underbrace{\sum_{k=s+1}^K \twonorm{\hbeta^{(0)[0]} + r_k\hbeta^{(k)[0]}}}_{[2]} \\
	&\quad - 2\underbrace{\sum_{k=1}^s \twonorm{\hbeta^{(0)[0]} - \hbeta^{(k)[0]}}}_{[1]'} - 2\underbrace{\sum_{k=s+1}^K \twonorm{-\hbeta^{(0)[0]} + r_k\hbeta^{(k)[0]}}}_{[2]'}, 
\end{align}
where 
\begin{align}
	[1] - [1]' &\geq \sum_{k=1}^s (\twonorm{\bbetaks{0} + \bbetaks{k}} - \twonorm{\bbetaks{0} - \bbetaks{k}} - 4\xi) \\
	&\geq \sum_{k=1}^s (2\twonorm{\bbetaks{0}} - 2\twonorm{\bbetaks{0} - \bbetaks{k}} - 4\xi) \\
	&\geq s(2\twonorm{\bbetaks{0}} - 4h - 4\xi),
\end{align}
and
\begin{equation}
	[2] - [2]' \geq  -4\sum_{k=s+1}^K \twonorm{\hbeta^{(0)[0]}} \geq -4(K-s)(\twonorm{\bbetaks{0}} + \xi).
\end{equation}
Hence
\begin{align}
	\text{score}((-1, \widehat{\bm{r}})) - \text{score}((1, \widehat{\bm{r}})) & = 2([1] - [1]') + 2([2] - [2]') \\
	&\geq 4[(2s-K)\twonorm{\bbetaks{0}} - 2sh - (K+s)\xi] \\
	&> 0, 
\end{align}
when $\twonorm{\bbetaks{0}} > \frac{2(1-\epsilon)}{1-2\epsilon}h + \frac{2-\epsilon}{1-2\epsilon}\xi$, which completes our proof.

\subsection{Proof of Theorem \ref{thm: upper bound multitask est error multi-cluster}}

Define the contraction basin of one GMM as 
\begin{align}
	B_{\text{con}}(\bthetaks{k}) &= \{\btheta = \{\{w_r\}_{r=2}^R, \{\bbeta_r\}_{r=2}^R, \{\delta_r\}_{r=2}^R\}: w_r^*\in(c_w/2, 1-c_w/2), \\
	&\quad \twonorm{\bbeta_r - \bbeta_r^*} \leq C_b\Delta, \norm{\delta_r - \delta_r^*} \leq C_b\Delta\}.
\end{align}
And the joint contraction basin is defined as $B_{\text{con}}(\{\bthetaks{k}\}_{k \in S}) = \bigcap_{r=1}^R B_{\text{con}}(\bthetaks{k})$.

For $\btheta = (\{w_r\}_{r=2}^R, \{\bbeta_r\}_{r=2}^R, \{\delta_r\}_{r=2}^R)$ and $\btheta' = (\{w_r'\}_{r=2}^R, \{\bbeta_r'\}_{r=2}^R, \{\delta_r'\}_{r=2}^R)$, define
\begin{equation}
	d(\btheta, \btheta') = \max_{r=2:R} \{\norm{w_r-w_r'}\vee \twonorm{\bbeta_r - \bbeta_r'} \vee \norm{\delta_r - \delta_r'}\}.
\end{equation}

\subsubsection{Lemmas}
For GMM $\bz \sim \sum_{r=1}^R w_r^*\mathcal{N}(\bmu_r^*, \bSigma^*)$ and any $\btheta$, define
\begin{equation}
	\gammak{r}_{\btheta}(\bz) = \frac{w_r\exp\{\bbeta_r^\top\bz - \delta_r\}}{w_1+\sum_{r=2}^R w_r\exp\{\bbeta_r^\top\bz - \delta_r\}}, r = 2:R, \quad \gammak{1}_{\btheta}(\bz) = \frac{w_1}{w_1+\sum_{r=2}^R w_r\exp\{\bbeta_r^\top\bz - \delta_r\}}.
\end{equation}
Denote $w_r(\btheta) = \te [\gammak{r}_{\btheta}(\bz)]$ and $\bmu_r(\btheta) = \frac{\te [\gammak{r}_{\btheta}(\bz)\bz]}{\te [\gammak{r}_{\btheta}(\bz)]}$.

\begin{lemma}[Contraction of multi-cluster GMM]\label{lem: contraction lemma multi-cluster}
	When $C_b \leq cc_{\bSigma}^{-1/2}$ with a small constant $c > 0$ and $\Delta \geq C\log(c_{\bSigma}Mc_w^{-1})$ with a large constant $C > 0$, there exist positive constants $C '> 0$ and $C'' > 0$, for any $\btheta \in B_{\textup{con}}(\bthetaks{k})$, 
	\begin{equation}
		\norm{w_r(\btheta) - w_r^*} \leq C'\exp\{-C''\Delta^2\}\cdot d(\btheta, \btheta^*), \quad \twonorm{\bmu_r(\btheta) - \bmu_r^*} \leq C'\exp\{-C''\Delta^2\}\cdot d(\btheta, \btheta^*),
	\end{equation}
	where $C'\exp\{-C''\Delta^2\} \leq \kappa_0 < 1$ with a constant $\kappa_0$.
\end{lemma}


\begin{lemma}[Vectorized contraction of Rademacher complexity, Corollary 1 in \citealpapp{maurer2016vector}]\label{lem: vec contraction}
	Suppose $\{\epsilon_{ir}\}_{i\in [n], r \in [R]}$ and $\{\epsilon_i\}_{i=1}^n$ are independent Rademacher variables. Let $\mathcal{F}$ be a class of functions $f: \mathbb{R}^d \rightarrow  \mathcal{S}\subseteq \mathbb{R}^R$ and $h: \mathcal{S} \rightarrow\mathbb{R}$ is $L$-Lipschitz under $\ell_2$-norm, i.e., $\norm{h(\by) - h(\by')} \leq L\twonorm{\by - \by'}$, where $\by = (y_1,\ldots,y_R)^\top$, $\by' = (y_1',\ldots,y_R')^\top \in \mathcal{S}$. Then
	\begin{equation}
		\te \sup_{f \in \mathcal{F}} \sum_{i=1}^n \epsilon_i h(f(x_i)) \leq \sqrt{2}L\te \sup_{f \in \mathcal{F}} \sum_{i=1}^n\sum_{r=1}^R \epsilon_{ir}f_r(x_i), 
	\end{equation}
	where $f_r(x_i)$ is the $r$-th component of $f(x_i) \in \mathcal{S}\subseteq \mathbb{R}^R$.
\end{lemma}

\subsubsection{Main proof of Theorem \ref{thm: upper bound multitask est error multi-cluster}}
The proof idea is almost the same as the idea used in the proof of Theorem \ref{thm: upper bound multitask est error}. We still need to show similar results presented in the lemmas associated with Theorem \ref{thm: upper bound multitask est error}, then go through the same arguments in the proof of Theorem \ref{thm: upper bound multitask est error multi-cluster}. We only sketch the key steps and the differences here. 

The biggest difference appears in the proofs of the lemmas associated with Theorem \ref{thm: upper bound multitask est error} under the context of multi-cluster GMM. The original arguments in the proofs of Lemmas \ref{lem: concentration w}-\ref{lem: concentration beta 2} involve the contraction inequality for Rademacher variables and univariate Lipschitz functions, which is not available anymore. We replace this part with an argument through a vectorized Rademacher contraction inequality \citepapp{maurer2016vector}.

First, we will show that
\begin{equation}\label{eq: eq 1 proof thm 12}
	\sup_{\substack{\bthetak{k} \in B_{\text{con}} \\ \twonorm{\bbetak{k}_r - \bbetaks{k}_r} \leq \xi^{(k)}}} \norma{\frac{1}{n_k}\sum_{i=1}^{n_k}\gammak{r}_{\bthetak{k}}(\bzk{k}_i) - \te[\gammak{r}_{\bthetak{k}}(\bzk{k})]} \lesssim \xi^{(k)}\sqrt{\frac{p}{n_k}} + \sqrt{\frac{\log K}{n_k}},
\end{equation}
for all $k \in S$ and $r \in 1:R$, with probability at least $1-CK^{-2}$. Denote the LHS as $W$. By changing one observation $\bzk{k}_i$, denote the new $W$ as $W'$. Since $\gammak{r}_{\btheta}(\bz)$ is bounded for all $\bz \in \mathbb{R}^p$, we know that $|W-W'| \leq 1/n_k$. Then by bounded difference inequality, we have
\begin{equation}
	W \leq \te W + C\sqrt{\frac{\log K}{n_k}},
\end{equation}
with probability at least $1-C'K^{-2}$. On the other hand, by symmetrization,
\begin{equation}
	\te W \leq \frac{2}{n_k}\te_{\bz}\te_{\bm{\epsilon}}\sup_{\substack{\bthetak{k} \in B_{\text{con}} \\ \twonorm{\bbetak{k}_r - \bbetaks{k}_r} \leq \xi^{(k)}}}\norma{\sum_{i=1}^{n_k}\epsilon^{(k)}_i\gammak{r}_{\bthetak{k}}(\bzk{k}_i)}.
\end{equation}
Note that $\gammak{r}_{\bthetak{k}}(\bz) = \frac{\wk{k}_r\cdot \exp\{(\bbetak{k}_r)^\top\bz - \deltak{k}_r\}}{\wk{k}_1 + \sum_{r=2}^R \wk{k}_r\exp\{(\bbetak{k}_r)^\top\bz - \deltak{k}_r\}} = \frac{\exp\{(\bbetak{k}_r)^\top\bz - \deltak{k}_r + \log \wk{k}_r - \log \wk{k}_1\}}{1 + \sum_{r=2}^R \exp\{(\bbetak{k}_r)^\top\bz - \deltak{k}_r + \log \wk{k}_r - \log \wk{k}_1\}} = \varphi(\{(\bbetak{k}_r)^\top\bz - \deltak{k}_r + \log \wk{k}_r - \log \wk{k}_1\}_{r=2}^R)$, where $\varphi(\bx) = \frac{\exp\{x_r\}}{1+\sum_{r=2}^R\exp\{x_r\}}$ is a 1-Lipschitz function (w.r.t. $\ell_2$-norm). By Lemma \ref{lem: vec contraction}, 
\begin{align}
	\frac{2}{n_k}\te_{\bz}\te_{\bm{\epsilon}}\sup_{\substack{\bthetak{k} \in B_{\text{con}} \\ \twonorm{\bbetak{k}_r - \bbetaks{k}_r} \leq \xi^{(k)}}}\norma{\sum_{i=1}^{n_k}\epsilon^{(k)}_i\gammak{r}_{\bthetak{k}}(\bzk{k}_i)} &\lesssim \frac{1}{n_k}\te_{\bz}\te_{\bepsilon}\sup_{\substack{\bthetak{k} \in B_{\text{con}} \\ \twonorm{\bbetak{k}_r - \bbetaks{k}_r} \leq \xi^{(k)}}} \norma{\sum_{i=1}^{n_k}\sum_{r=2}^R \epsilonk{k}_{ir}g^{(k)}_{ir}} \\
	&\lesssim \frac{1}{n_k}\sum_{r=2}^R\te_{\bz}\te_{\bepsilon}\sup_{\substack{\bthetak{k} \in B_{\text{con}} \\ \twonorm{\bbetak{k}_r - \bbetaks{k}_r} \leq \xi^{(k)}}}\norma{\sum_{i=1}^{n_k}\epsilonk{k}_{ir}g^{(k)}_{ir}},
\end{align}
where $g^{(k)}_{ir} \coloneqq (\bbetak{k}_r)^\top\bzk{k}_i - \deltak{k}_r + \log \wk{k}_r - \log \wk{k}_1$. It follows that
\begin{align}
	&\frac{1}{n_k}\te_{\bz}\te_{\bepsilon}\sup_{\substack{\bthetak{k} \in B_{\text{con}} \\ \twonorm{\bbetak{k}_r - \bbetaks{k}_r} \leq \xi^{(k)}}}\norma{\sum_{i=1}^{n_k}\epsilonk{k}_{ir}g^{(k)}_{ir}}\\
	&\leq \frac{1}{n_k}\te_{\bz,\bepsilon}\sup_{\substack{\bthetak{k} \in B_{\text{con}} \\ \twonorm{\bbetak{k}_r - \bbetaks{k}_r} \leq \xi^{(k)}}}\norma{\sum_{i=1}^{n_k}\epsilonk{k}_{ir}(\bbetak{k}_r)^\top\bzk{k}_i} \\
	&\quad+ \frac{1}{n_k}\te_{\bepsilon}\sup_{\substack{\bthetak{k} \in B_{\text{con}} \\ \twonorm{\bbetak{k}_r - \bbetaks{k}_r} \leq \xi^{(k)}}}\norma{\sum_{i=1}^{n_k}\epsilonk{k}_{ir}(\deltak{k}_r - \log \wk{k}_r + \log \wk{k}_1)} \\
	&\leq \frac{1}{n_k}\te_{\bz,\bepsilon}\sup_{\twonorm{\bbetak{k}_r - \bbetaks{k}_r} \leq \xi^{(k)}}\norma{\sum_{i=1}^{n_k}\epsilonk{k}_{ir}(\bbetak{k}_r)^\top(\bzk{k}_i-\bmuks{k})} \\
	&\quad  + \frac{1}{n_k}\te_{\bz,\bepsilon}\sup_{\twonorm{\bbetak{k}_r - \bbetaks{k}_r} \leq \xi^{(k)}}\norma{\sum_{i=1}^{n_k}\epsilonk{k}_{ir}(\bbetak{k}_r)^\top\bmuks{k}} \\
	&\quad + \frac{1}{n_k}\te_{\bepsilon}\sup_{\substack{|\deltak{k}_r|\leq U \\ c_w/2\leq \wk{k}_r \leq 1-c_w/2}}\norma{\sum_{i=1}^{n_k}\epsilonk{k}_{ir}(\deltak{k}_r - \log \wk{k}_r + \log \wk{k}_1)}\\
	&\leq \frac{\xi^{(k)}}{n_k}\te_{\bz,\bepsilon}\sup_{j=1:N}\norma{\sum_{i=1}^{n_k}\epsilonk{k}_{ir}\bu_j^\top(\bzk{k}_i-\bmuks{k})} + \frac{1}{n_k}\te_{\bz,\bepsilon}\norma{\sum_{i=1}^{n_k}\epsilonk{k}_{ir}(\bbetaks{k}_r)^\top(\bzk{k}_i-\bmuks{k})}\\
	&\quad + \frac{C}{n_k}\te_{\bepsilon}\norma{\sum_{i=1}^{n_k}\epsilonk{k}_{ir}} \label{eq: proof of thm multi-cluster eq 1}
\end{align}
where $\bmuks{k} \coloneqq \sum_{r=1}^R \wks{k}_r\bmuks{k}_r$, $\{\bu_j\}_{j=1}^N$ is a $1/2$-cover of $\mathcal{S}^{d-1}$ with $N = 5^p$, and $\{\epsilonk{k}_{ir}\bu_j^\top(\bzk{k}_i-\bmuks{k})\}_{i=1}^{n_k}$, $\{\epsilonk{k}_{ir}(\bbetaks{k}_r)^\top(\bzk{k}_i-\bmuks{k})\}_{i=1}^{n_k}$, and $\{\epsilonk{k}_{ir}\}_{i=1}^{n_k}$ are all sub-Gaussian processes. Then by the property of sub-Gaussian variables,
\begin{equation}
	\textup{RHS of }\eqref{eq: proof of thm multi-cluster eq 1} \lesssim \xi^{(k)}\sqrt{\frac{p}{n_k}} + \sqrt{\frac{\log K}{n_k}}.
\end{equation}
Putting all the pieces together, we obtain $W \lesssim \xi^{(k)}\sqrt{\frac{p}{n_k}} + \sqrt{\frac{\log K}{n_k}}$ with probability at least $1-CK^{-2}$.

The second bound we want to show is 
\begin{equation}\label{eq: eq 2 proof thm 12}
	\sup_{\{\bthetak{k}\}_{k \in S} \in B_{\text{con}}^{J,2}}\sup_{\norm{\widetilde{w}_k} \leq 1} \frac{1}{\ns}\norma{\sum_{k \in S}\widetilde{w}_k\sum_{i=1}^{n_k}\Big[\gammak{r}_{\bthetak{k}}(\bzk{k}_i) - \te[\gammak{r}_{\bthetak{k}}(\bzk{k})]\Big]} \lesssim \sqrt{\frac{p+K}{\ns}}.
\end{equation}
Denote the LHS as $W'$. Again by bounded difference inequality,
\begin{equation}
	W' \leq \te W' + C\sqrt{\frac{p}{\ns}},
\end{equation}
with probability at least $1-C'\exp\{-C''p\}$. It remains to control $\te W'$. By symmetrization,
\begin{equation}
	\te W' \leq  \frac{2}{\ns}\te \sup_{\{\bthetak{k}\}_{k \in S} \in B_{\text{con}}^{J,2}}\sup_{\norm{\widetilde{w}_k} \leq 1}\norma{\sum_{k \in S}\sum_{i=1}^{n_k}\widetilde{w}_k\gammak{r}_{\bthetak{k}}(\bzk{k}_i)}.
\end{equation}
Denote 
\begin{align}
	&\varphi(\widetilde{w}, \{(\bbetak{k}_r)^\top\bz - \deltak{k}_r + \log \wk{k}_r - \log \wk{k}_1\}_{r=2}^R) \\
	&= \widetilde{w}_k\gammak{r}_{\bthetak{k}}(\bzk{k}_i) \\
	&= \widetilde{w}_k\cdot \frac{\exp\{(\bbetak{k}_r)^\top\bz - \deltak{k}_r + \log \wk{k}_r - \log \wk{k}_1\}}{1 + \sum_{r=2}^R \exp\{(\bbetak{k}_r)^\top\bz - \deltak{k}_r + \log \wk{k}_r - \log \wk{k}_1\}},
\end{align}
which is C-Lipschitz w.r.t. $(\widetilde{w}, \{(\bbetak{k}_r)^\top\bz - \deltak{k}_r + \log \wk{k}_r - \log \wk{k}_1\}_{r=2}^R)$ as a $R$-dimensional vector with a constant $C$. Denote $g^{(k)}_{ir} = (\bbetak{k}_r)^\top\bzk{k}_i - \deltak{k}_r + \log \wk{k}_r - \log \wk{k}_1$. A direct application of Lemma \ref{lem: vec contraction} implies that
\begin{align}
	&\frac{1}{\ns}\te \sup_{\{\bthetak{k}\}_{k \in S} \in B_{\text{con}}^{J,2}}\sup_{\norm{\widetilde{w}_k} \leq 1}\norma{\sum_{k \in S}\sum_{i=1}^{n_k}\widetilde{w}_k\gammak{r}_{\bthetak{k}}(\bzk{k}_i)} \\
	&\lesssim \frac{1}{\ns}\te \sup_{\norm{\widetilde{w}_k} \leq 1} \norma{\sum_{k \in S}\sum_{i=1}^{n_k}\widetilde{w}_k\epsilonk{k}_{i1}} + \frac{1}{\ns}\sum_{r=2}^R\te \sup_{\{\bthetak{k}\}_{k \in S} \in B_{\text{con}}^{J,2}} \norma{\sum_{k \in S}\sum_{i=1}^{n_k}g^{(k)}_{ir}\epsilonk{k}_{ir}}.
\end{align}
By a similar argument involving covering number as before, we can show that
\begin{equation}
	\frac{1}{\ns}\te \sup_{\norm{\widetilde{w}_k} \leq 1} \norma{\sum_{k \in S}\sum_{i=1}^{n_k}\widetilde{w}_k\epsilonk{k}_{i1}} + \frac{1}{\ns}\sum_{r=2}^R\te \sup_{\{\bthetak{k}\}_{k \in S} \in B_{\text{con}}^{J,2}} \norma{\sum_{k \in S}\sum_{i=1}^{n_k}g^{(k)}_{ir}\epsilonk{k}_{ir}} \lesssim \sqrt{\frac{p+ K}{\ns}}.
\end{equation}
Therefore $W \lesssim \sqrt{\frac{p+ K}{\ns}}$ with probability at least $1-C'\exp\{-C''p\}$..

The third bound we want to show is 
\begin{align}
	&\sup_{\substack{\bthetak{k} \in B_{\text{con}} \\ \twonorm{\bbetak{k} - \bbetaks{k}} \leq \xi^{(k)}}} \norma{\frac{1}{n_k}\sum_{i=1}^{n_k}\big[1-\gammak{r}_{\bthetak{k}}(\bzk{k}_i)\big](\bzk{k}_i)^\top\bbetaks{k} - \te\big[[1-\gammak{r}_{\bthetak{k}}(\bzk{k})](\bzk{k})^\top\bbetaks{k}\big]} \\
	&\quad\lesssim \xi^{(k)}\sqrt{\frac{p}{n_k}} + \sqrt{\frac{\log K}{n_k}}, \label{eq: eq 3 proof thm 12}
\end{align}
for all $k \in S$ and $r = 1:R$, with probability at least $1-C'(K^{-2}+K^{-1}e^{-C''p})$. Denote the LHS as $W''$. Similar to the previous two proofs, we derive an upper bound for $W''$ by controlling $W'' - \te W''$ and $\te W''$, separately. The first part involving $W'' - \te W''$ is similar to the proof of part (\rom{1}) in Lemma \ref{lem: concentration delta} and the second part involving $\te W''$ is similar to the proof of \eqref{eq: eq 1 proof thm 12}, so we omit the details.

The arguments to derive these three bounds can be used to derive other results similar to the lemmas used in the proof of Theorem \ref{thm: upper bound multitask est error}. With these lemmas in hand, the remaining proof is almost the same as the proof of Theorem \ref{thm: upper bound multitask est error}. 

\subsubsection{Proof of lemmas}

\begin{proof}[Proof of Lemma \ref{lem: contraction lemma multi-cluster}]
We will prove the contraction of $w_r$ first, and only sketch the different part for the proof of contraction of $\bmu_r$ because the proofs are quite similar. 

\noindent\underline{Part 1: Contraction of $\norm{w_r(\btheta) - w_r^*}$}:

First, note that $w_r(\btheta^*) = w^*_r$ and $\bmu_r(\btheta^*) = \bmu_r^*$. Therefore,
\begin{align}
	\norm{w_r(\btheta) - w_r^*} &= \norma{\te[\gammak{r}_{\btheta}(\bz) - \gammak{r}_{\btheta^*}(\bz)]} \\
	&\leq \sum_{\widetilde{r}=1}^R \wks{k}_r\norma{\te[\gammak{r}_{\btheta}(\bz) - \gammak{r}_{\btheta^*}(\bz)| y = \widetilde{r}]}\\
	&\leq \sum_{\widetilde{r}=1}^R \wks{k}_r \sum_{r'=2}^R \norma{\te \left[\frac{\partial \gammak{r}_{\btheta}(\bz)}{\partial w_{r'}}\bigg|_{\btheta = \widetilde{\btheta}_t}\bigg| y = \widetilde{r}\right]}\cdot \norm{w_{r'}-w_r^*} \\
	&\quad + \sum_{\widetilde{r}=1}^R \wks{k}_r \sum_{r'=2}^R \norma{\te \left[\frac{\partial \gammak{r}_{\btheta}(\bz)}{\partial \delta_{r'}}\bigg|_{\btheta = \widetilde{\btheta}_t}\bigg| y = \widetilde{r}\right]}\cdot \norm{\delta_{r'}-\delta_r^*}  \\
	&\quad + \sum_{\widetilde{r}=1}^R \wks{k}_r \sum_{r'=2}^R \norma{\te \left[\frac{\partial \gammak{r}_{\btheta}(\bz)}{\partial \bbeta_{r'}}\bigg|_{\btheta = \widetilde{\btheta}_t}\bigg| y = \widetilde{r}\right]^\top (\bbeta_{r'}-\bbeta_r^*)},
\end{align}
We only show how to bound $\norma{\te[\gammak{r}_{\btheta}(\bz) - \gammak{r}_{\btheta^*}(\bz)| y = 1]}$, i.e. the case when $\widetilde{r} = 1$. For the other $\widetilde{r} = 2:R$, the proof is the same by changing the reference level from $y = 1$ to $y = \widetilde{r}$.
Note that
\begin{align}
	\norma{\te[\gammak{r}_{\btheta}(\bz) - \gammak{r}_{\btheta^*}(\bz)| y = 1]} &\leq \sum_{r'=2}^R \norma{\te \left[\frac{\partial \gammak{r}_{\btheta}(\bz)}{\partial w_{r'}}\bigg|_{\btheta = \widetilde{\btheta}_t}\bigg| y = 1\right]}\cdot \norm{w_{r'}-w_r^*} \\
	&\quad + \sum_{r'=2}^R \norma{\te \left[\frac{\partial \gammak{r}_{\btheta}(\bz)}{\partial \delta_{r'}}\bigg|_{\btheta = \widetilde{\btheta}_t}\bigg| y = 1\right]}\cdot \norm{\delta_{r'}-\delta_r^*}  \\
	&\quad + \sum_{r'=2}^R \norma{\te \left[\frac{\partial \gammak{r}_{\btheta}(\bz)}{\partial \bbeta_{r'}}\bigg|_{\btheta = \widetilde{\btheta}_t}\bigg| y = 1\right]^\top (\bbeta_{r'}-\bbeta_r^*)}.
\end{align}

where $\widetilde{\btheta}_t = (\{\widetilde{w}_r\}_{r=2}^R, \{\widetilde{\bbeta}_r\}_{r=2}^R, \{\widetilde{\delta}_r\}_{r=2}^R)$ with $\widetilde{w}_r = tw_r + (1-t)w_r^*$, $\widetilde{\bbeta}_r = t\bbeta_r + (1-t)\bbeta_r^*$, $\widetilde{\delta}_r = t\delta_r + (1-t)\delta_r^*$, and $\delta_r = \frac{1}{2}\bbeta_r^\top(\bmu_r + \bmu_1)$. We will bound the three terms on the RHS separately. Note that when $\btheta \in B_{\textup{con}}(\btheta^*)$, we have $w_r \in [c_w/2, 1-c_w]$, $\twonorm{\bbeta_r - \bbeta_r^*} \leq C_b\Delta$, and $\max_{r=1:R}\twonorm{\bmu_r - \bmu_r^*} \leq C_b\Delta$, hence $\widetilde{w}_r \in [c_w/2, 1-c_w]$, $\twonorm{\widetilde{\bbeta}_r - \bbeta^*_r} \leq tC_b\Delta$..

\noindent (\rom{1}) Bounding $\norm{\te [\frac{\partial \gammak{r}_{\btheta}(\bz)}{\partial w_{r'}}|_{\btheta = \widetilde{\btheta}_t}|y=1]}$: Note that
\begin{align}
	\frac{\partial \gammak{r}_{\btheta}(\bz)}{\partial w_{r'}} &= \frac{\exp\{\widetilde{\bbeta}_r^\top\bz - \delta_r\}}{\widetilde{w}_1+\sum_{r=2}^R \widetilde{w}_r\exp\{\widetilde{\bbeta}_r^\top\bz - \delta_r\}} - \frac{\widetilde{w}_r\exp\{\widetilde{\bbeta}_r^\top\bz - \delta_r\}(\exp\{\widetilde{\bbeta}_r^\top\bz - \delta_r\} - 1)}{(\widetilde{w}_1+\sum_{r=2}^R \widetilde{w}_r\exp\{\widetilde{\bbeta}_r^\top\bz - \delta_r\})^2} \\
	&= \begin{cases}
		\frac{\exp\{\widetilde{\bbeta}_r^\top\bz - \delta_r\}(\widetilde{w}_1 + \widetilde{w}_r + \sum_{r'\neq r} \widetilde{w}_{r'}\exp\{\widetilde{\bbeta}_{r'}^\top\bz - \delta_{r'}\})}{(\widetilde{w}_1+\sum_{r=2}^R \widetilde{w}_r\exp\{\widetilde{\bbeta}_r^\top\bz - \delta_r\})^2}, \quad &r' = r, \\
		-\frac{\widetilde{w}_r\cdot \exp\{\widetilde{\bbeta}_r^\top\bz - \delta_r\}(\exp\{\widetilde{\bbeta}_{r'}^\top\bz - \delta_{r'}\} - 1)}{(\widetilde{w}_1+\sum_{r=2}^R \widetilde{w}_r\exp\{\widetilde{\bbeta}_r^\top\bz - \delta_r\})^2}, \quad &r' \neq r.
	\end{cases}
\end{align}
Hence
\begin{equation}
	\te\left[\frac{\partial \gammak{r}_{\btheta}(\bz)}{\partial w_r}\bigg|y = 1\right] = \underbrace{\te\left[\frac{\exp\{\widetilde{\bbeta}_r^\top\bzk{1}-\delta_r\}(\widetilde{w}_1 + \widetilde{w}_r + \sum_{r'\neq r} \widetilde{w}_{r'}\exp\{\widetilde{\bbeta}_{r'}^\top\bzk{1} - \delta_{r'}\})}{(\widetilde{w}_1+\sum_{r=2}^R \widetilde{w}_r\exp\{\widetilde{\bbeta}_r^\top\bzk{1} - \delta_r\})^2}\right]}_{(*)}.
\end{equation}
Let $\widetilde{z}_{r'} = \widetilde{\bbeta}^\top_{r'}(\bzk{1}-\bmu_1^*)\ \sim \mathcal{N}(0, \widetilde{\bbeta}_{r'}^\top\bSigma^*\widetilde{\bbeta}_{r'})$. And notice that
\begin{equation}
	\widetilde{\bbeta}^\top_{r'}\bmu_1^* - \widetilde{\delta}_{r'} = t(\bbeta_{r'}^\top\bmu_1^* - \delta_{r'}) + (1-t)[(\bbeta_{r'}^*)^\top\bmu_1^* - \delta_{r'}^*],
\end{equation}
where 
\begin{align}
	\bbeta_{r'}^\top\bmu_1^* - \delta_{r'} &= [\bbeta_{r'}^* + (\bbeta_{r'} - \bbeta^*_{r'})]^\top\Big[\frac{1}{2}(\bmu_1^* - \bmu_{r'}^*) + \frac{1}{2}(\bmu_{r'}^* - \bmu_{r'}) + \frac{1}{2}(\bmu_1^* - \bmu_1)\Big] \\
	&= -\frac{1}{2}\underbrace{(\bmu_{r'}^* - \bmu_1^*)^\top(\bSigma^*)^{-1}(\bmu_{r'}^* - \bmu_1^*)}_{A_{r'}^2} + \frac{1}{2}(\bbeta_{r'}^*)^\top(\bSigma^*)^{1/2}(\bSigma^*)^{-1/2}[(\bmu_{r'}^* - \bmu_{r'}) + (\bmu_1^* - \bmu_1)]\\
	&\quad + \frac{1}{2}(\bbeta_{r'} - \bbeta^*_{r'})^\top(\bSigma^*)^{1/2}(\bSigma^*)^{-1/2}(\bmu_1^* - \bmu_{r'}^*)  \\
	&\quad + \frac{1}{2}(\bbeta_{r'} - \bbeta^*_{r'})^\top[(\bmu_{r'}^* - \bmu_{r'}) + (\bmu_1^* - \bmu_1)],\\
	(\bbeta_{r'}^*)^\top\bmu_1^* - \delta_{r'}^* &= -\frac{1}{2}A_{r'},
\end{align}
and $A_{r'} = \sqrt{(\bmu_{r'}^* - \bmu_1^*)^\top(\bSigma^*)^{-1}(\bmu_{r'}^* - \bmu_1^*)} = \sqrt{(\bbeta_r^*)^\top\bSigma^*\bbeta_r^*}$. By the fact that $\max_{r=1:R}\twonorm{\bmu_r - \bmu_r^*} \leq C_b\Delta$ and $\max_{r=1:R}\twonorm{\bbeta_r - \bbeta_r^*} \leq C_b\Delta$, we have
\begin{equation}
	-\frac{1}{2}A_{r'}^2 - 2c_{\bSigma}^{1/2}C_b\Delta A_{r'} - C_b^2\Delta^2 \leq \bbeta_{r'}^\top\bmu_1^* - \delta_{r'} \leq -\frac{1}{2}A_{r'}^2 + 2c_{\bSigma}^{1/2}C_b\Delta A_{r'}+ C_b^2\Delta^2,
\end{equation}
implying that
\begin{equation}
	-\frac{1}{2}A_{r'}^2 - 2c_{\bSigma}^{1/2}C_b\Delta A_{r'} - C_b^2\Delta^2 \leq \widetilde{\bbeta}^\top_{r'}\bmu_1^* - \widetilde{\delta}_{r'} \leq -\frac{1}{2}A_{r'}^2 + 2c_{\bSigma}^{1/2}C_b\Delta A_{r'}+ C_b^2\Delta^2.
\end{equation}

By Gaussian tail, we have
\begin{equation}
	\tp\left(\bigcap_{r'=2}^R\Big\{|\widetilde{z}_{r'}| \leq \frac{1}{4}\widetilde{\bbeta}_{r'}^\top\bSigma^*\widetilde{\bbeta}_{r'}\Big\}\right) \geq 1-CR\exp\Big\{-\frac{1}{32}\widetilde{\bbeta}_{r'}^\top\bSigma^*\widetilde{\bbeta}_{r'}\Big\}.
\end{equation}
Denote event $\mathcal{E} = \bigcap_{r'=1}^R\big\{|\widetilde{z}_{r'}| \leq \frac{1}{4}c_{\bSigma}C_b^2 \Delta^2 + \frac{1}{2}C_bc_{\bSigma}^{1/2}\Delta A_{r'} + \frac{1}{4}A_{r'}^2\big\}$. Since 
\begin{align}
	\frac{1}{4}\widetilde{\bbeta}_{r'}^\top\bSigma^*\widetilde{\bbeta}_{r'} &= \frac{1}{4}(\widetilde{\bbeta}_{r'} - \bbeta_{r'}^*)^\top\bSigma^*(\widetilde{\bbeta}_{r'} - \bbeta_{r'}^*) + \frac{1}{2}(\widetilde{\bbeta}_{r'} - \bbeta_{r'}^*)^\top(\bSigma^*)^{1/2}(\bSigma^*)^{1/2}\bbeta_{r'}^* + \frac{1}{4}(\bbeta_{r'}^*)^\top\bSigma^*\bbeta_{r'}^* \\
	&\leq \frac{1}{4}c_{\bSigma}C_b^2 \Delta^2 + \frac{1}{2}C_bc_{\bSigma}^{1/2}\Delta A_{r'} + \frac{1}{4}A_{r'}^2,
\end{align}
and
\begin{equation}
	\widetilde{\bbeta}_{r'}^\top\bSigma^*\widetilde{\bbeta}_{r'} \geq A_{r'}^2 - c_{\bSigma}C_b^2 \Delta^2 - 2C_bc_{\bSigma}^{1/2}\Delta A_{r'} \geq (1-c_{\bSigma}C_b^2 - 2C_bc_{\bSigma}^{1/2})\Delta^2 \geq \frac{1}{2}\Delta^2,
\end{equation}
we have
\begin{equation}
	\tp(\mathcal{E}) \geq 1-CR\exp\Big\{-\frac{1}{64}\Delta^2\Big\}.
\end{equation}
Then since $\min_{r' =1:R}A_{r'} \geq \Delta \geq 5c_{\bSigma}^{1/2}C_b\Delta$ and $C_b \leq \frac{c_{\bSigma}^{-1/2}}{40}\wedge (2c_{\bSigma}+8)^{-1/2}$, we have
\begin{align}
	&(*) \\
	&\leq \te\Bigg[\frac{\exp\{-\frac{1}{4}A_r^2 + \frac{5}{2}c_{\bSigma}^{1/2}C_b\Delta A_r+ (\frac{1}{4}c_{\bSigma}+1)C_b^2\Delta^2\}}{\widetilde{w}^2_1}\\
	&\quad\quad \quad \cdot \bigg(\widetilde{w}_1+ \widetilde{w}_r + \sum_{r'\neq r}\widetilde{w}_{r'}\exp\Big\{-\frac{1}{4}A_r^2 + \frac{5}{2}c_{\bSigma}^{1/2}C_b\Delta A_r+ \Big(\frac{1}{4}c_{\bSigma}+1\Big)C_b^2\Delta^2\Big\}\bigg)\bigg|\mathcal{E}\Bigg] + \tp(\mathcal{E}^c) \\
	&\lesssim c_w^{-2}\exp\{-C\Delta^2\}.
\end{align}
Hence,
\begin{equation}
	\norma{\te \left[\frac{\partial \gammak{r}_{\btheta}(\bz)}{\partial w_r}\Big|_{\btheta = \widetilde{\btheta}_t}\bigg|y = 1\right]} \lesssim c_w^{-2}\exp\{-C\Delta^2\}.
\end{equation}
Similarly, it can be shown that 
\begin{equation}
	\norma{\te \left[\frac{\partial \gammak{r}_{\btheta}(\bz)}{\partial w_{r'}}\Big|_{\btheta = \widetilde{\btheta}_t}\bigg|y = 1\right]} \lesssim c_w^{-2}\exp\{-C\Delta^2\}.
\end{equation}
for any $r' = 2:R$.

\noindent (\rom{2}) Bounding $\norm{\te [\frac{\partial \gammak{r}_{\btheta}(\bz)}{\partial \delta_{r'}}|_{\btheta = \widetilde{\btheta}_t}|y=1]}$: Note that
\begin{equation}
	\frac{\partial \gammak{r}_{\btheta}(\bz)}{\partial \delta_{r'}} 
	= \begin{cases}
		\frac{-w_r\cdot \exp\{\bbeta_r^\top\bz - \delta_r\}\cdot \sum_{r'\neq r} w_{r'}\exp\{\bbeta_{r'}^\top\bz - \delta_{r'}\}}{(w_1+\sum_{r=2}^R w_r\exp\{\bbeta_r^\top\bz - \delta_r\})^2}, \quad &r' = r, \\
		-\frac{w_r\cdot \exp\{\bbeta_r^\top\bz - \delta_r\}\cdot w_{r'}\cdot \exp\{\bbeta_{r'}^\top\bz - \delta_{r'}\}}{(w_1+\sum_{r=2}^R w_r\exp\{\bbeta_r^\top\bz - \delta_r\})^2}, \quad &r' \neq r.
	\end{cases}
\end{equation}
The analysis is almost the same as in (\rom{1}), which leads to
\begin{equation}
	\norma{\te \left[\frac{\partial \gammak{r}_{\btheta}(\bz)}{\partial \delta_{r'}}\Big|_{\btheta = \btheta}\right]} \lesssim c_w^{-2}\exp\{-C\Delta^2\},
\end{equation}
for any $r' = 2:R$. We omit the proof here.

\noindent (\rom{3}) Bounding $\norm{\te [\frac{\partial \gammak{r}_{\btheta}(\bz)}{\partial \bbeta_{r'}}|_{\btheta = \widetilde{\btheta}_t}|y=1]^\top(\bbeta_{r'}-\bbeta_r^*)}$: Note that
\begin{equation}
	\frac{\partial \gammak{r}_{\btheta}(\bz)}{\partial \bbeta_{r'}} 
	= \begin{cases}
		\frac{-w_r\cdot \exp\{\bbeta_r^\top\bz - \delta_r\}\cdot \sum_{r'\neq r} w_{r'}\exp\{\bbeta_{r'}^\top\bz - \delta_{r'}\}\bz}{(w_1+\sum_{r=2}^R w_r\exp\{\bbeta_r^\top\bz - \delta_r\})^2}, \quad &r' = r, \\
		-\frac{w_r\cdot \exp\{\bbeta_r^\top\bz - \delta_r\}\cdot w_{r'}\cdot \exp\{\bbeta_{r'}^\top\bz - \delta_{r'}\}\bz}{(w_1+\sum_{r=2}^R w_r\exp\{\bbeta_r^\top\bz - \delta_r\})^2}, \quad &r' \neq r.
		\end{cases}
\end{equation}

\noindent $\bullet$ When $r' = r$:
\begin{align}
	&\te\left[\left(\frac{\partial \gammak{r}_{\btheta}(\bz)}{\partial \bbeta_{r'}}\bigg|_{\btheta = \widetilde{\btheta}_t}\right)^\top(\bbeta_r - \bbeta_r^*)\bigg| y = 1\right] \\
	&= \te\left[\frac{\widetilde{w}_r\exp\{\widetilde{\bbeta}_r^\top\bz^{(1)} - \widetilde{\delta}_r\}\sum_{r'\neq r} \widetilde{w}_{r'}\cdot \exp\{\widetilde{\bbeta}_{r'}^\top\bz^{(1)} - \widetilde{\delta}_{r'}\}\cdot (\bzk{1})^\top(\bbeta_r - \bbeta_r^*)}{(\widetilde{w}_1 + \sum_{r=2}^R \widetilde{w}_r \exp\{\widetilde{\bbeta}_r^\top\bzk{1} - \widetilde{\delta}_r\})^2}\right]\\
	&\leq \underbrace{\sqrt{\te\left[\frac{\widetilde{w}_r\exp\{\widetilde{\bbeta}_r^\top\bz^{(1)} - \widetilde{\delta}_r\}\sum_{r'\neq r} \widetilde{w}_{r'}\cdot \exp\{\widetilde{\bbeta}_{r'}^\top\bz^{(1)} - \widetilde{\delta}_{r'}\}}{(\widetilde{w}_1 + \sum_{r=2}^R \widetilde{w}_r \exp\{\widetilde{\bbeta}_r^\top\bzk{1} - \widetilde{\delta}_r\})^2}\right]^2}}_{(1)}\cdot \underbrace{\sqrt{\te [(\bzk{1})^\top(\bbeta_r - \bbeta_r^*)]^2}}_{(2)}.
\end{align}
Similar to the previous argument in (\rom{1}), let $\widetilde{z}_{r'} = \widetilde{\bbeta}^\top_{r'}(\bzk{\widetilde{r}}-\bmu_1^*) \sim \mathcal{N}(0, \widetilde{\bbeta}^\top_{r'}\bSigma^*\widetilde{\bbeta}_{r'})$ and event $\mathcal{E} = \bigcap_{r'=1}^R\big\{|\widetilde{z}_{r'}| \leq \frac{1}{4}c_{\bSigma}C_b^2 \Delta^2 + \frac{1}{2}C_bc_{\bSigma}^{1/2}\Delta A_{r'} + \frac{1}{4}A_{r'}^2\big\}$, then
\begin{equation}
	\tp(\mathcal{E}) \geq 1-CR\exp\Big\{-\frac{1}{64}\Delta^2\Big\}.
\end{equation}
Similar to (\rom{1}), we have
\begin{equation}
	(1) \lesssim \sqrt{\te\left[\left(\frac{\widetilde{w}_r\exp\{\widetilde{\bbeta}_r^\top\bz^{(1)} - \widetilde{\delta}_r\}\sum_{r'\neq r} \widetilde{w}_{r'}\cdot \exp\{\widetilde{\bbeta}_{r'}^\top\bz^{(1)} - \widetilde{\delta}_{r'}\}}{(\widetilde{w}_1 + \sum_{r=2}^R \widetilde{w}_r \exp\{\widetilde{\bbeta}_r^\top\bzk{1} - \widetilde{\delta}_r\})^2}\right)^2\bigg|\mathcal{E}\right] + \tp(\mathcal{E}^c)} \lesssim \exp\{-C\Delta^2\}.
\end{equation}
Moreover, $(\bzk{1})^\top(\bbeta_r - \bbeta_r^*) = (\bzk{1} - \bmu_1^*)^\top(\bbeta_r - \bbeta_r^*) + (\bmu_1^*)^\top(\bbeta_r - \bbeta_r^*)$, where $(\bzk{1} - \bmu_1^*)^\top(\bbeta_r - \bbeta_r^*) \sim \mathcal{N}(0, (\bbeta_r - \bbeta_r^*)^\top\bSigma^*(\bbeta_r - \bbeta_r^*))$ and $\norm{(\bmu_1^*)^\top(\bbeta_r - \bbeta_r^*)} \leq M\twonorm{\bbeta_r - \bbeta_r^*}$, hence
\begin{equation}
	(2) \lesssim \sqrt{(\bbeta_r - \bbeta_r^*)^\top\bSigma^*(\bbeta_r - \bbeta_r^*)} + M\twonorm{\bbeta_r - \bbeta_r^*} \lesssim (c_{\bSigma}^{1/2} + M)\twonorm{\bbeta_r - \bbeta_r^*} \lesssim (c_{\bSigma}^{1/2} + M)C_b\Delta.
\end{equation}
Therefore, since $\Delta \leq 2Mc_{\bSigma}^{1/2}$ and $\Delta \gtrsim \log^{1/2}(c_{\bSigma}Mc_w^{-1})$,
\begin{equation}
	\te\left[\left(\frac{\partial \gammak{r}_{\btheta}(\bz)}{\partial \bbeta_{r'}}\right)^\top(\bbeta_r - \bbeta_r^*)\bigg| y = 1\right] \lesssim c_w^{-2}\exp\{-C'\Delta^2\}\cdot (c_{\bSigma}^{1/2} + M)Mc_{\bSigma}^{1/2}\lesssim \exp\{-C'\Delta^2\}.
\end{equation}

\noindent $\bullet$ When $r' \neq r$: we can obtain 
\begin{equation}
	\te\left[\left(\frac{\partial \gammak{r}_{\btheta}(\bz)}{\partial \bbeta_{r'}}\right)^\top(\bbeta_r - \bbeta_r^*)\bigg| y = 1\right] \lesssim \exp\{-C'\Delta^2\}.
\end{equation}
similarly. 

Combining (\rom{1})-(\rom{3}), we have
\begin{equation}
	\norm{w_r(\btheta) - w_r^*} \lesssim \exp\{-C''\Delta^2\}\cdot \sum_{r=2}^R(\norm{w_r-w_r^*} + \norm{\delta_r - \delta_r^*} + \twonorm{\bbeta_r - \bbeta_r^*}).
\end{equation}

\noindent\underline{Part 2: Contraction of $\twonorm{\bmu_r(\btheta) - \bmu_r^*}$}:

By definition,
\begin{equation}\label{eq: contraction multi-cluster part 2}
	\twonorm{\bmu_r(\btheta) - \bmu_r^*} \leq \frac{\twonorm{\te[\gammak{r}_{\btheta}(\bz)\bz]}}{w_r(\btheta)w_r^*}\cdot \norm{w_r^* - w_r(\btheta)} + \frac{\twonorm{\te[(\gammak{r}_{\btheta}(\bz) - \gammak{r}_{\btheta^*}(\bz))\bz]}}{w_r^*},
\end{equation}
implying that
\begin{align}
	\twonorm{\te[(\gammak{r}_{\btheta}(\bz) - \gammak{r}_{\btheta^*}(\bz))\bz]} &\leq \sum_{r'=2}^R \twonorma{\te\bigg[\frac{\partial \gammak{r}_{\btheta}(\bz)}{\partial w_{r'}}\bigg|_{\btheta = \widetilde{\btheta}_t}\bz\bigg]}\cdot \norm{w_{r'}-w_{r'}^*} \\
	&\quad + \sum_{r=2}^R \twonorma{\te \bigg[\frac{\partial \gammak{r}_{\btheta}(\bz)}{\partial \delta_{r'}}\bigg|_{\btheta = \widetilde{\btheta}_t}\bz\bigg]}\cdot \norm{\delta_{r'} - \delta_r^*} \\
	&\quad + \sum_{r=2}^R\twonorma{\te \bigg[\frac{\partial \gammak{r}_{\btheta}(\bz)}{\partial \bbeta_{r'}}\bigg|_{\btheta = \widetilde{\btheta}_t}\bz\bigg]}\cdot \twonorm{\bbeta_{r'} - \bbeta_r^*},
\end{align}
where $\widetilde{\btheta}_t = (\{\widetilde{w}_r\}_{r=2}^R, \{\widetilde{\bbeta}_r\}_{r=2}^R, \{\widetilde{\delta}_r\}_{r=2}^R)$ with $\widetilde{w}_r = tw_r + (1-t)w_r^*$, $\widetilde{\bbeta}_r = t\bbeta_r + (1-t)\bbeta_r^*$, and $\widetilde{\delta}_r = t\delta_r + (1-t)\delta_r^*$. We will bound the three terms on the RHS separately. Note that when $\btheta \in B_{\textup{con}}(\btheta^*)$, we have $\widetilde{w}_r \in (c_w/2, 1-c_w)$, $\twonorm{\widetilde{\bbeta}_r - \bbeta^*_r} \leq C_b\Delta$, and $\norm{\widetilde{\delta}_r - \delta_r} \leq C_b\Delta$.

For any $\bu \in \mathbb{R}^p$ with $\twonorm{\bu} \leq 1$ and any $\widetilde{r} \in 1:R$, similar to our previous arguments, we have
\begin{equation}
	\norma{\te\bigg[\frac{\partial \gammak{r}_{\btheta}(\bz)}{\partial w_{r'}}\bigg|_{\btheta = \widetilde{\btheta}_t}\bz^\top\bu\bigg|y = \widetilde{r}\bigg]} \leq \sqrt{\te \left[\frac{\partial \gammak{r}_{\btheta}(\bz)}{\partial w_{r'}}\right]^2}\cdot \sqrt{\te[(\bzk{\widetilde{r}})^\top\bu]^2} \lesssim \exp\{-C''\Delta^2\},
\end{equation}
which leads to 
\begin{equation}
	\twonorma{\te\bigg[\frac{\partial \gammak{r}_{\btheta}(\bz)}{\partial w_{r'}}\bz\bigg]} \lesssim \exp\{-C''\Delta^2\},
\end{equation}
for any $r' \in 2:R$. Similarly, we have
\begin{equation}
	\twonorma{\te\bigg[\frac{\partial \gammak{r}_{\btheta}(\bz)}{\partial \delta_{r'}}\bz\bigg]}, \twonorma{\te\bigg[\frac{\partial \gammak{r}_{\btheta}(\bz)}{\partial \bbeta_{r'}}\bz\bigg]} \lesssim \exp\{-C''\Delta^2\}.
\end{equation}
for any $r' \in 2:R$. Therefore, $\twonorm{\te[(\gammak{r}_{\btheta}(\bz) - \gammak{r}_{\btheta^*}(\bz))\bz]}\lesssim \exp\{-C''\Delta^2\}$. By part 1, we have $\frac{\twonorm{\te[\gammak{r}_{\btheta}(\bz)\bz]}}{w_r(\btheta)w_r^*}\cdot \norm{w_r^* - w_r(\btheta)} \lesssim \exp\{-C''\Delta^2\}\cdot d(\btheta, \btheta^*)$. Hence by \eqref{eq: contraction multi-cluster part 2}, we have $\twonorm{\bmu_r(\btheta) - \bmu_r^*} \lesssim \exp\{-C''\Delta^2\}\cdot d(\btheta, \btheta^*)$.

Combining part 1 and part 2, we complete the proof.

\end{proof}

\subsection{Proof of Theorem \ref{thm: upper bound multitask classification error multi-cluster}}
Note that the excess risk
\begin{align}
	&R_{\othetaks{k}}(\mC_{\htheta^{(k)}}) - R_{\othetaks{k}}(\mC_{\bthetaks{k}})\\
	&= \tp(\yk{k} \neq \mC_{\htheta^{(k)}}(\bzk{k})) - \tp(\yk{k} \neq \mC_{\bthetaks{k}}(\bzk{k})) \\
	&= \int_{\mC_{\htheta^{(k)}} \neq \mC_{\bthetaks{k}}}\left[\tp(\yk{k} = \mC_{\bthetaks{k}}(\bz)|\bzk{k} = \bz) - \tp(\yk{k} = \mC_{\htheta^{(k)}}(\bz)|\bzk{k} = \bz) \right] d\tp_{\bthetaks{k}}(\bz) \\
	&= \int_{\mC_{\htheta^{(k)}} \neq \mC_{\bthetaks{k}}}\left[\max_{r=1:R}\tp(\yk{k} = r|\bzk{k} = \bz) - \tp(\yk{k} = \mC_{\htheta^{(k)}}(\bz)|\bzk{k} = \bz) \right] d\tp_{\bthetaks{k}}(\bz). \label{eq: proof mtl classification multi-cluster eq 1}
\end{align}
Let event $\mathcal{E} = \big\{\bz: \max_{r}\tp(\yk{k} = r|\bzk{k} = \bz) - \max_{j}\{\tp(\yk{k} = j|\bzk{k} = \bz): \tp(\yk{k} = j|\bzk{k} = \bz) < \max_{r}\tp(\yk{k} = r|\bzk{k} = \bz)\} \leq t\big\}$. We claim that the margin condition $\tp(\mathcal{E}) \lesssim t$ holds for any $t \leq $ a small constant $c$ (to be verified). If this is the case, then denote $\widetilde{\mathcal{E}} = \big\{\max_{r}\norm{\eta^{(r)}_{\htheta^{(k)}}(\bzk{k}) - \eta^{(r)}_{\bthetaks{k}}(\bzk{k})} \leq t/2\big\}$.
\begin{align}
	r^* &= \argmax_r \eta^{(r)}_{\bthetaks{k}}(\bz),\\
	\widehat{r} &= \argmax_r \eta^{(r)}_{\htheta^{(k)}}(\bz).
\end{align}
We have 
\begin{align}
	&\textup{RHS of } \eqref{eq: proof mtl classification multi-cluster eq 1} \\
	&\leq \int_{\substack{r^* \neq \widehat{r} \\ \mathcal{E}, \widetilde{\mathcal{E}}}} \big[\eta^{(r^*)}_{\bthetaks{k}}(\bz) - \eta^{(\widehat{r})}_{\bthetaks{k}}(\bz)\big] d\tp_{\bthetaks{k}}(\bz) +  \int_{\substack{r^* \neq \widehat{r} \\ \mathcal{E}^c, \widetilde{\mathcal{E}}}} \big[\eta^{(r^*)}_{\bthetaks{k}}(\bz) - \eta^{(\widehat{r})}_{\htheta^{(k)}}(\bz)\big] d\tp_{\bthetaks{k}}(\bz) + \tp(\widetilde{\mathcal{E}}^c) \\
	&\leq t\tp(\mathcal{E})  + \tp(\widetilde{\mathcal{E}}^c), \label{eq: proof mtl classification multi-cluster eq 2}
\end{align} 
where the last inequality comes from the fact that when $r^* \neq \widehat{r}$, $\eta^{(r^*)}_{\bthetaks{k}}(\bz) - \eta^{(\widehat{r})}_{\bthetaks{k}}(\bz) \leq t$ on $\mathcal{E}$. And notice that on $\mathcal{E}^c \cap \widetilde{\mathcal{E}}$, we must have $\widehat{r} = r^*$ because if $\widehat{r} \neq r^*$, then
\begin{equation}
	\eta^{(\widehat{r})}_{\htheta^{(k)}}(\bz)-\eta^{(r^*)}_{\htheta^{(k)}}(\bz) \leq \eta^{(\widehat{r})}_{\bthetaks{k}}(\bz) -\eta^{(r^*)}_{\bthetaks{k}}(\bz)+ \frac{t}{2} + \frac{t}{2} \leq -t + t < 0,  
\end{equation}
which is a contradiction with the definition of $\widehat{r}$. Hence $\{r^* \neq \widehat{r}\} \cap \mathcal{E} \cap \widetilde{\mathcal{E}}$ is empty. Therefore $\int_{\substack{r^* \neq \widehat{r} \\ \mathcal{E}^c, \widetilde{\mathcal{E}}}} \big[\eta^{(r^*)}_{\bthetaks{k}}(\bz) - \eta^{(\widehat{r})}_{\htheta^{(k)}}(\bz)\big] d\tp_{\bthetaks{k}}(\bz) = 0$.
Finally, by Lipschitzness,
\begin{align}
	\tp(\widetilde{\mathcal{E}}^c)&= \tp\left(\max_{r}\norm{\eta^{(r)}_{\htheta^{(k)}}(\bzk{k}) - \eta^{(r)}_{\bthetaks{k}}(\bzk{k})} > t/2\right) \\
	&\leq \sum_{r=1}^R \tp\left(\norm{\eta^{(r)}_{\htheta^{(k)}}(\bzk{k}) - \eta^{(r)}_{\bthetaks{k}}(\bzk{k})} > t/2\right) \\
	&\lesssim \tp(\norm{(\hbeta^{(k)} - \bbetaks{k})^\top\bzk{k} - \hdelta^{(k)} + \deltaks{k} - \log \hw^{(k)}_r + \log \hw^{(k)}_1 + \log \wks{k}_r - \log \wks{k}_1} > Ct)  \\
	&\lesssim \tp(\norm{(\hbeta^{(k)} - \bbetaks{k})^\top(\bzk{k}-\bmuk{k})} > C' t)\\
	&\lesssim \exp\left\{-\frac{Ct^2}{\twonorm{\hbeta^{(k)}-\bbetaks{k}}^2}\right\},
\end{align}
if $t \gtrsim d(\htheta^{(k)}, \bthetaks{k}) \gtrsim |\hdelta^{(k)}-\deltaks{k}| + |\hw^{(k)}_r-\wks{k}_r| + |\hw^{(k)}_1-\wks{k}_1| \gtrsim  |\hdelta^{(k)}-\deltaks{k}| + |\log (\hw^{(k)}_r/\wks{k}_r)|  + |\log (\hw^{(k)}_1/\wks{k}_1)|$ and $d(\htheta^{(k)}, \bthetaks{k}) \lesssim 1$.
Plugging back into \eqref{eq: proof mtl classification multi-cluster eq 2}, we have
\begin{equation}
	R_{\othetaks{k}}(\mC_{\htheta^{(k)}}) - R_{\othetaks{k}}(\mC_{\bthetaks{k}}) \lesssim t^2 + \exp\left\{-\frac{Ct^2}{\twonorm{\hbeta^{(k)}-\bbetaks{k}}^2}\right\}. 
\end{equation}
Let $t \asymp d(\htheta^{(k)}, \bthetaks{k})\sqrt{\log d^{-1}(\htheta^{(k)}, \bthetaks{k})}$:
\begin{equation}
	R_{\othetaks{k}}(\mC_{\htheta^{(k)}}) - R_{\othetaks{k}}(\mC_{\bthetaks{k}}) \lesssim d^2(\htheta^{(k)}, \bthetaks{k})\log d^{-1}(\htheta^{(k)}, \bthetaks{k}) \lesssim d^2(\htheta^{(k)}, \bthetaks{k})\log\left(\frac{\ns}{p+\log \ns}\right).
\end{equation}
Then plugging in the upper bound of $d(\htheta^{(k)}, \bthetaks{k})$ in Theorem \ref{thm: upper bound multitask est error multi-cluster} completes the proof.

It remains to verify the margin condition $\tp(\mathcal{E}) \lesssim t$ for any $t \leq $ a small constant $c$. In fact,
\begin{align}
	\tp(\mathcal{E}) &= \sum_{r=1}^R \sum_{j \neq r}\tp\bigg(\argmax_{r'} \tp(y=r'|\bzk{k}) = r, \argmax_{r' \neq r} \tp(y=r'|\bzk{k}) = j,\\
	&\hspace{2.8cm} \tp(y=r|\bzk{k}) - \tp(y=j|\bzk{k}) \leq t\bigg) \\
	&\leq \sum_{r=1}^R\sum_{j \neq r} \tp\bigg(\argmax_{r'} \tp(y=r'|\bzk{k}) = r, \argmax_{r' \neq r} \tp(y=r'|\bzk{k}) = j, \\
	&\hspace{2.8cm} 1-\frac{\tp(y=j|\bzk{k})}{\tp(y=r|\bzk{k})}\leq \frac{t}{\tp(y=r|\bzk{k})} \leq Rt\bigg)\\
	&\leq \sum_{r=1}^R\sum_{j \neq r}\tp\bigg(1-Rt \leq \frac{\tp(y=j|\bzk{k})}{\tp(y=r|\bzk{k})} \\
	&\hspace{2.8cm}= \exp\{(\bbetaks{k}_j - \bbetaks{k}_r)^\top\bzk{k} - \delta_j^{(k)*} + \delta_r^{(k)*} + \log \wks{k}_j-\log \wks{k}_r\} \bigg)\\
	&\lesssim \sum_{r=1}^R\sum_{j \neq r}\sum_{r'=1}^R \tp\bigg(\log(1-Rt) \leq \mathcal{N}\big((\bbetaks{k}_j - \bbetaks{k}_r)^\top\bmuks{k}_r - \delta_j^{(k)*} + \delta_r^{(k)*} + \log \wks{k}_j-\log \wks{k}_r, \\
	&\hspace{3.2cm} (\bbetaks{k}_j - \bbetaks{k}_r)^\top\bSigmaks{k}(\bbetaks{k}_j - \bbetaks{k}_r)\big) \leq 0\bigg) \\
	&\lesssim -\log(1-Rt) \\
	&\lesssim t,
\end{align}
when $t > 0$ is less than some constant $c > 0$. Note that we used the fact that $(\bbetaks{k}_j - \bbetaks{k}_r)^\top\bSigmaks{k}(\bbetaks{k}_j - \bbetaks{k}_r) \geq \Delta^2 \geq $ some constant $C$, which implies that the Gaussian density is upper bounded by a constant. Hence the marginal condition is true.

We want to point out that this multi-class extension of margin condition in binary case has been widely used in literature of multi-class classification. For example, see \citeapp{chen2006consistency} and \citeapp{vigogna2022multiclass}.

\subsection{Proof of Theorem \ref{thm: lower bound multitask est error multi-cluster}}
The proof is almost the same as the proof of Theorem \ref{thm: lower bound multitask est error}, by noticing that we can make the GMM parameters the same across $r$-th task with $r\geq 3$ to reduce the problem to the case $R = 2$, so we do not repeat it here.

\subsection{Proof of Theorem \ref{thm: lower bound multitask classification error multi-cluster}}

\subsubsection{Lemmas}
\begin{lemma}\label{lem: misclassification error multi-cluster w}
	Consider $\otheta = \{\{w_r\}_{r=2}^R, \{\bbeta_r\}_{r=2}^R, \{\delta_r\}_{r=2}^R, \{\bmu_r\}_{r=1}^R, \bSigma\}$ and $\otheta' = \{\{w_r'\}_{r=2}^R, \{\bbeta_r'\}_{r=2}^R,\allowbreak \{\delta_r'\}_{r=2}^R, \{\bmu_r'\}_{r=1}^R, \bSigma'\}$ with $w_r = w_r'$ for $r \geq 3$, $\bmu_r = \bmu_r' = \bm{e}_r$ for $r \geq 2$, $\bmu_1 = \bmu_1' = \bm{0}$, and $\bSigma = \bSigma' = \bm{I}_p$. Then $\bbeta_r = \bbeta_r' = \bm{e}_r$, $\delta_r =\delta_r' = \bbeta_r^\top(\frac{\bmu_1+\bmu_2}{2}) = \frac{1}{2}$ for $r \geq 2$, and
	\begin{equation}
		\tp_{\otheta}(\mC_{\otheta} \neq \mC_{\otheta'}) \gtrsim \norm{w_2 - w_2'}.
	\end{equation}
\end{lemma}

\begin{lemma}\label{lem: misclassification error multi-cluster delta}
	Consider $\otheta = \{\{w_r\}_{r=2}^R, \{\bbeta_r\}_{r=2}^R, \{\delta_r\}_{r=2}^R, \{\bmu_r\}_{r=1}^R, \bSigma\}$ and $\otheta' = \{\{w_r'\}_{r=2}^R, \{\bbeta_r'\}_{r=2}^R,\allowbreak \{\delta_r'\}_{r=2}^R, \{\bmu_r'\}_{r=1}^R, \bSigma'\}$ with $w_r = w_r' = \frac{1}{R}$ for $r = 1:R$, $\bmu_r = (u+1)\bm{e}_r$,  $\bmu_r' = \bm{e}_r$ for $r \geq 2$, $\bmu_1 = u\bm{e}_r$, $\bmu_1' = \bm{0}$, and $\bSigma = \bSigma' = \bm{I}_p$, where $0<u\leq C$. Then $\bbeta_r = \bbeta_r' = \bm{e}_r$, $\delta_r  = \bm{e}_r^\top(\frac{\bmu_1+\bmu_2}{2}) = u+\frac{1}{2}$, $\delta_r' = \bm{e}_r^\top(\frac{\bmu_1'+\bmu_2'}{2}) = \frac{1}{2}$ for $r \geq 2$, and
	\begin{equation}
		\tp_{\otheta}(\mC_{\otheta} \neq \mC_{\otheta'}) \gtrsim u.
	\end{equation}
\end{lemma}

\begin{lemma}\label{lem: misclassification error multi-cluster mu}
	Consider $\otheta = \{\{w_r\}_{r=2}^R, \{\bbeta_r\}_{r=2}^R, \{\delta_r\}_{r=2}^R, \{\bmu_r\}_{r=1}^R, \bSigma\}$ and $\otheta' = \{\{w_r'\}_{r=2}^R, \{\bbeta_r'\}_{r=2}^R,\allowbreak \{\delta_r'\}_{r=2}^R, \{\bmu_r'\}_{r=1}^R, \bSigma'\}$ with $w_r = w_r' = \frac{1}{R}$ for $r = 1:R$, $\bmu_r  = \bmu_r' = \bm{e}_r$ for $r \geq 3$, $\bmu_1 = \bmu_1' = \bm{0}$, and $\bSigma = \bSigma' = \bm{I}_p$.  Suppose $\bmu_2$ and $\bmu_2'$ satisfy $(\bmu_2)_{3:R} = (\bmu_2)_{3:R} = \bm{0}$. Then $\bbeta_r = \bbeta_r' = \bm{e}_r$ for $r \geq 3$, $\delta_r  = \delta_r'$ for $r \geq 3$, $\bbeta_2 = \bmu_2$, $\bbeta_2' = \bmu_2'$, $\delta_2 = \frac{1}{2}\twonorm{\bmu_2}^2$, $\delta_2' = \frac{1}{2}\twonorm{\bmu_2'}^2$ where $\twonorm{\bmu_2} = \twonorm{\bmu_2'} = 1$ with $\bmu_2^\top\bmu_2' > \frac{\sqrt{2}}{2}$, and
	\begin{equation}
		\tp_{\otheta}(\mC_{\otheta} \neq \mC_{\otheta'}) \gtrsim \twonorm{\bmu_2 - \bmu_2'}.
	\end{equation}

\end{lemma}

\subsubsection{Main proof of Theorem \ref{thm: lower bound multitask classification error multi-cluster}}
Given the three lemmas we presented, the proof is almost the same as the proof of Theorem \ref{thm: lower bound multitask classification error}. We do not repeat it here.

\subsubsection{Proof of lemmas}

\begin{proof}[Proof of Lemma \ref{lem: misclassification error multi-cluster w}]
Note that $z_j$'s are independent given $y = 3$. We have
	\begin{align}
		\tp_{\otheta}(\mC_{\otheta} \neq \mC_{\otheta'}) &\geq \tp_{\otheta}(\mC_{\otheta}(\bz) = 2, \mC_{\otheta'}(\bz)\neq 2) \\
		&\geq \tp_{\otheta}\Big(z_2 - \frac{1}{2} - \log w_1 + \log w_2 \geq 0, z_2 - \frac{1}{2} - \log w_1' + \log w_2' \leq 0 \\
		&\quad\quad\quad z_r - \frac{1}{2} - \log w_1 + \log w_r \leq 0 \text{ for all } r \geq 3\Big) \\
		&\geq w_3\tp_{\otheta}\Big(z_2 - \frac{1}{2} - \log w_1 + \log w_2 \geq 0, z_2 - \frac{1}{2} - \log w_1' + \log w_2' \leq 0 \\
		&\quad\quad\quad z_r - \frac{1}{2} - \log w_1 + \log w_r \leq 0 \text{ for all } r \geq 3\Big| y = 3 \Big) \\
		&\gtrsim \tp_{\otheta}\bigg(\frac{1}{2}+\log w_1 - \log w_2 \leq z_2 \leq \frac{1}{2}+\log w_1' - \log w_2'\Big| y = 3\bigg)\\
		&\quad \cdot \prod_{r=3}^R \tp\Big(z_r - \frac{1}{2}- \log w_1 + \log w_r \leq 0\Big) \\
		&\gtrsim \norma{\log w_1 - \log w_2 - \log w_1' + \log w_2'} \\
		&= \norma{\log (1-w_2) - \log w_2 - \log (1-w_2') + \log w_2'}\\
		&\gtrsim \norm{w_2 - w_2'}.
	\end{align}
\end{proof}

\begin{proof}[Proof of Lemma \ref{lem: misclassification error multi-cluster delta}]
Note that $z_j$'s are independent given $y = 3$. We have
	\begin{align}
		\tp_{\otheta}(\mC_{\otheta} \neq \mC_{\otheta'}) &\geq \tp_{\otheta}(\mC_{\otheta}(\bz) = 2, \mC_{\otheta'}(\bz)\neq 2) \\
		&\geq \tp_{\otheta}\Big(z_2 - \frac{1}{2} - u \geq 0, z_2 - \frac{1}{2} > 0, z_r - \frac{1}{2} - u \leq 0, z_r - \frac{1}{2} \leq 0 \text{ for all } r \geq 3\Big) \\
		&\geq \tp_{\otheta}\Big(\frac{1}{2} \leq z_2 \leq \frac{1}{2} + u, z_r \leq \frac{1}{2} \text{ for all } r \geq 3\Big)\\
		&\gtrsim w_3\tp_{\otheta}\bigg(\frac{1}{2} \leq z_2 \leq \frac{1}{2} + u, z_r \leq \frac{1}{2} \text{ for all } r \geq 3\Big| y = 3\bigg)\\
		&\gtrsim w_3\tp_{\otheta}\bigg(\frac{1}{2} \leq z_2 \leq \frac{1}{2} + u\Big| y = 3\bigg) \cdot \prod_{r=3}^R \tp\Big(z_r \leq  \frac{1}{2}\Big) \\
		&\gtrsim u,
	\end{align}
	where we used the fact that $\tp\big(z_r \leq  \frac{1}{2}|y=3\big) \geq $ some constant $C$.
\end{proof}

\begin{proof}[Proof of Lemma \ref{lem: misclassification error multi-cluster delta}]
Note that $z_j$'s are independent given $y = 3$. We have
	\begin{align}
		\tp_{\otheta}(\mC_{\otheta} \neq \mC_{\otheta'}) &\geq \tp_{\otheta}(\mC_{\otheta}(\bz) = 2, \mC_{\otheta'}(\bz)\neq 2) \\
		&\geq \tp_{\otheta}\Big(\bmu_2^\top\bz - \frac{1}{2}\twonorm{\bmu_2}^2 \geq 0, (\bmu_2')^\top \bz - \frac{1}{2}\twonorm{\bmu_2'}^2 \leq 0, z_r \leq \frac{1}{2} \text{ for all } r \geq 3\Big) \\
		&\geq w_3\tp_{\otheta}\Big(\bmu_2^\top\bz - \frac{1}{2} \geq 0, (\bmu_2')^\top \bz - \frac{1}{2} \leq 0, z_r \leq \frac{1}{2} \text{ for all } r \geq 3\Big| y = 3\Big) \\
		&\gtrsim w_3\tp_{\otheta}\Big(\bmu_2^\top\bz \geq \frac{1}{2}, (\bmu_2')^\top\bz \leq \frac{1}{2}\Big| y = 3\Big)\cdot \prod_{r=3}^R\tp\Big(z_r \leq \frac{1}{2}\Big|y=3\Big)\\
		&\gtrsim \sqrt{1-\frac{|\bmu_2^\top\bmu_2'|^2}{\twonorm{\bmu_2}^4}}\cdot \frac{|\bmu_2^\top\bmu_2'|}{\twonorm{\bmu_2}^2} \\
		&\gtrsim \twonorm{\bmu_2 - \bmu_2'}
	\end{align}
	The second last inequality is due to the fact that $\tp\big(z_r \leq  \frac{1}{2}|y=3\big) \geq $ some constant $C$ and Proposition 23 in \citeapp{azizyan2013minimax}. The last inequality comes from Lemma 8.1 in \citeapp{cai2019chime}.
\end{proof}

\subsection{Proof of Theorem \ref{thm: brute force alignment multi component}}
WLOG, suppose that $\pi_k^*$ satisfies $\pi_k^*(r) = $ the ``majority class" $\widetilde{r}$, if $\#\{k \in S: \pi_k(r) = \widetilde{r}\} \geq \frac{1}{2}|S|$. Note that we can make this assumption because it suffices to recover $\{\iota(\pi^*_k)\}_{k \in S}$ \textit{with a permutation $\iota$}. And WLOG, suppose $\pi^*_k(r) = r$ for all $k \in S$. Let us consider any $\pi = \{\pi_k\}_{k=1}^K$ with $\pi_k(r) = \pi_k^*(r)$ for all $k \in S^c$ and $\pi \neq \pi^*$. It suffices to prove that $\text{score}(\pi) > \text{score}(\pi^*)$.

Recall that $\xi = \max_{k \in S}\min_{\pi}\max_{r \in [R]}\twonorm{(\hSigma^{(k)})^{-1}(\hmu^{(k)[0]}_{\pi_k(r)} - \hmu^{(k)[0]}_{\pi_k(1)}) - \bbetaks{k}_r}$. Note that
\begin{align}
	\text{score}(\pi) - \text{score}(\pi^*) &= \underbrace{\sum_{\substack{k \neq k' \in S \\ \pi_k(1) = \pi_{k'}(1)}}\sum_{r = 2}^R\twonorm{(\hSigma^{(k)[0]})^{-1}(\hmu^{(k)[0]}_{\pi_k(r)} - \hmu^{(k)[0]}_{\pi_k(1)}) - (\hSigma^{(k')[0]})^{-1}(\hmu^{(k')[0]}_{\pi_{k'}(r)} - \hmu^{(k')[0]}_{\pi_{k'}(1)})}}_{[1]} \\
	&\quad+  \underbrace{\sum_{\substack{k \neq k' \in S \\ \pi_k(1) \neq \pi_{k'}(1)}}\sum_{r = 2}^R\twonorm{(\hSigma^{(k)[0]})^{-1}(\hmu^{(k)[0]}_{\pi_k(r)} - \hmu^{(k)[0]}_{\pi_k(1)}) - (\hSigma^{(k')[0]})^{-1}(\hmu^{(k')[0]}_{\pi_{k'}(r)} - \hmu^{(k')[0]}_{\pi_{k'}(1)})}}_{[2]} \\
	&\quad+  \underbrace{2\sum_{\substack{k \in S \\ k' \in S^c}}\sum_{r = 2}^R\twonorm{(\hSigma^{(k)[0]})^{-1}(\hmu^{(k)[0]}_{\pi_k(r)} - \hmu^{(k)[0]}_{\pi_k(1)}) - (\hSigma^{(k')[0]})^{-1}(\hmu^{(k')[0]}_{\pi_{k'}(r)} - \hmu^{(k')[0]}_{\pi_{k'}(1)})}}_{[3]} \\
	&- \underbrace{\sum_{\substack{k \neq k' \in S \\ \pi_k(1) = \pi_{k'}(1)}}\sum_{r = 2}^R\twonorm{(\hSigma^{(k)[0]})^{-1}(\hmu^{(k)[0]}_r - \hmu^{(k)[0]}_1) - (\hSigma^{(k')[0]})^{-1}(\hmu^{(k')[0]}_r - \hmu^{(k')[0]}_1)}}_{[1]'}\\ 
	&- \underbrace{\sum_{\substack{k \neq k' \in S \\ \pi_k(1) \neq \pi_{k'}(1)}}\sum_{r = 2}^R\twonorm{(\hSigma^{(k)[0]})^{-1}(\hmu^{(k)[0]}_r - \hmu^{(k)[0]}_1) - (\hSigma^{(k')[0]})^{-1}(\hmu^{(k')[0]}_r - \hmu^{(k')[0]}_1)}}_{[2]'}\\ 
	&- \underbrace{2\sum_{\substack{k \in S \\ k' \in S^c}}\sum_{r = 2}^R\twonorm{(\hSigma^{(k)[0]})^{-1}(\hmu^{(k)[0]}_r - \hmu^{(k)[0]}_1) - (\hSigma^{(k')[0]})^{-1}(\hmu^{(k')[0]}_{\pi_{k'}(r)} - \hmu^{(k')[0]}_{\pi_{k'}(1)})}}_{[3]'}.
\end{align}
We have
\begin{align}
	[2] &\geq \sum_{\substack{k \neq k' \in S \\ \pi_k(1) \neq \pi_{k'}(1)}}\sum_{r = 2}^R\left(\twonorm{(\hSigma^{(k)[0]})^{-1}(\hmu^{(k)[0]}_{\pi_k(r)} - \hmu^{(k)[0]}_{\pi_k(1)}) - (\hSigma^{(k)[0]})^{-1}(\hmu^{(k)[0]}_{\pi_{k'}(r)} - \hmu^{(k)[0]}_{\pi_{k'}(1)})}-2h-\xi\right) \\
	&\geq \sum_{\substack{k \neq k' \in S \\ \pi_k(1) \neq \pi_{k'}(1)}}R\left(\twonorm{(\hSigma^{(k)[0]})^{-1}(\hmu^{(k)[0]}_{\pi_k(1)} - \hmu^{(k)[0]}_{\pi_{k'}(1)})}-2h-\xi\right) \\
	&\geq \sum_{\substack{k \neq k' \in S \\ \pi_k(1) \neq \pi_{k'}(1)}}R(c_{\bSigma}^{-1/2}\Delta - 2h -2\xi).
\end{align}
Hence
\begin{equation}
	[2] - [2]' \geq \sum_{\substack{k \neq k' \in S \\ \pi_k(1) \neq \pi_{k'}(1)}}[R(c_{\bSigma}^{-1/2}\Delta - 2h -2\xi) - R(2\xi + 2h)] = \sum_{\substack{k \neq k' \in S \\ \pi_k(1) \neq \pi_{k'}(1)}}R(c_{\bSigma}^{-1/2}\Delta - 4h - 4\xi).
\end{equation}
Therefore,
\begin{align}
	[1] &= \underbrace{\sum_{\substack{k \neq k' \in S \\ \pi_k(1) =\pi_{k'}(1)}}\sum_{r = 2}^R \sum_{\pi_k(r) = \pi_{k'}(r)}\twonorm{(\hSigma^{(k)[0]})^{-1}(\hmu^{(k)[0]}_{\pi_k(r)} - \hmu^{(k)[0]}_{\pi_k(1)}) - (\hSigma^{(k')[0]})^{-1}(\hmu^{(k')[0]}_{\pi_{k'}(r)} - \hmu^{(k')[0]}_{\pi_{k'}(1)})}}_{[1]_1} \\
	&\quad + \underbrace{\sum_{\substack{k \neq k' \in S \\ \pi_k(1) =\pi_{k'}(1)}}\sum_{r = 2}^R \sum_{\pi_k(r) \neq \pi_{k'}(r)}\twonorm{(\hSigma^{(k)[0]})^{-1}(\hmu^{(k)[0]}_{\pi_k(r)} - \hmu^{(k)[0]}_{\pi_k(1)}) - (\hSigma^{(k')[0]})^{-1}(\hmu^{(k')[0]}_{\pi_{k'}(r)} - \hmu^{(k')[0]}_{\pi_{k'}(1)})}}_{[1]_2}.
\end{align}
Correspondingly, we can decompose $[1]'$ in the same way as $[1]' = [1]'_1 + [1]'_2$ with
\begin{align}
	[1]_1' &= \sum_{\substack{k \neq k' \in S \\ \pi_k(1) =\pi_{k'}(1)}}\sum_{r = 2}^R \sum_{\pi_k(r) = \pi_{k'}(r)}\twonorm{(\hSigma^{(k)[0]})^{-1}(\hmu^{(k)[0]}_r - \hmu^{(k)[0]}_1) - (\hSigma^{(k')[0]})^{-1}(\hmu^{(k')[0]}_r - \hmu^{(k')[0]}_1)},\\
	[1]_2' &= \sum_{\substack{k \neq k' \in S \\ \pi_k(1) =\pi_{k'}(1)}}\sum_{r = 2}^R \sum_{\pi_k(r) = \pi_{k'}(r)}\twonorm{(\hSigma^{(k)[0]})^{-1}(\hmu^{(k)[0]}_r - \hmu^{(k)[0]}_1) - (\hSigma^{(k')[0]})^{-1}(\hmu^{(k')[0]}_r - \hmu^{(k')[0]}_1)}.
\end{align}
Note that
\begin{align}
	[1]_2 &= \sum_{\substack{k \neq k' \in S \\ \pi_k(1) =\pi_{k'}(1)}}\sum_{r = 2}^R \sum_{\pi_k(r) \neq \pi_{k'}(r)}\left(\twonorm{(\hSigma^{(k)[0]})^{-1}(\hmu^{(k)[0]}_{\pi_k(r)} - \hmu^{(k)[0]}_{\pi_k(1)}) - (\hSigma^{(k)[0]})^{-1}(\hmu^{(k)[0]}_{\pi_{k'}(r)} - \hmu^{(k)[0]}_{\pi_{k'}(1)})} - 2h - \xi\right) \\
	&= \sum_{\substack{k \neq k' \in S \\ \pi_k(1) =\pi_{k'}(1)}}\sum_{r = 2}^R \sum_{\pi_k(r) \neq \pi_{k'}(r)}\left(\twonorm{(\hSigma^{(k)[0]})^{-1}(\hmu^{(k)[0]}_{\pi_k(r)} - \hmu^{(k)[0]}_{\pi_{k'}(r)})} - 2h - \xi\right) \\
	&\geq \sum_{\substack{k \neq k' \in S \\ \pi_k(1) =\pi_{k'}(1)}}\sum_{r = 2}^R \sum_{\pi_k(r) \neq \pi_{k'}(r)}\left(c_{\bSigma}^{-1/2}\Delta - 2h - 2\xi\right),
\end{align}
hence
\begin{equation}
	[1]_2 - [1]'_2 \geq \sum_{\substack{k \neq k' \in S \\ \pi_k(1) =\pi_{k'}(1)}}\sum_{r = 2}^R \sum_{\pi_k(r) \neq \pi_{k'}(r)}\left(c_{\bSigma}^{-1/2}\Delta - 4h - 4\xi\right).
\end{equation}
And 
\begin{align}
	[1]_1 &= \sum_{\substack{k \neq k' \in S \\ \pi_k(1) =\pi_{k'}(1)}}\sum_{r = 2}^R \sum_{\pi_k(r) = \pi_{k'}(r)}\twonorm{(\hSigma^{(k)[0]})^{-1}(\hmu^{(k)[0]}_{\pi_k(r)} - \hmu^{(k)[0]}_{\pi_k(1)}) - (\hSigma^{(k')[0]})^{-1}(\hmu^{(k')[0]}_{\pi_{k'}(r)} - \hmu^{(k')[0]}_{\pi_{k'}(1)})} \\
	&= \underbrace{\sum_{\substack{k \neq k' \in S \\ \pi_k(1) =\pi_{k'}(1) = 1}}\sum_{r = 2}^R \sum_{\pi_k(r) = \pi_{k'}(r)}\twonorm{(\hSigma^{(k)[0]})^{-1}(\hmu^{(k)[0]}_{\pi_k(r)} - \hmu^{(k)[0]}_{\pi_k(1)}) - (\hSigma^{(k')[0]})^{-1}(\hmu^{(k')[0]}_{\pi_{k'}(r)} - \hmu^{(k')[0]}_{\pi_{k'}(1)})}}_{[1]_{11}} \\
	&\quad + \underbrace{\sum_{\substack{k \neq k' \in S \\ \pi_k(1) =\pi_{k'}(1) \neq 1}}\sum_{r = 2}^R \sum_{\pi_k(r) = \pi_{k'}(r)}\twonorm{(\hSigma^{(k)[0]})^{-1}(\hmu^{(k)[0]}_{\pi_k(r)} - \hmu^{(k)[0]}_{\pi_k(1)}) - (\hSigma^{(k')[0]})^{-1}(\hmu^{(k')[0]}_{\pi_{k'}(r)} - \hmu^{(k')[0]}_{\pi_{k'}(1)})}}_{[1]_{12}}.
\end{align}
\begin{align}
	[1]_{12} &= \sum_{\substack{k \neq k' \in S \\ \pi_k(1) =\pi_{k'}(1) \neq 1}}\sum_{r = 2}^R \sum_{\pi_k(r) = \pi_{k'}(r)}(\twonorm{(\hSigma^{(k)[0]})^{-1}(\hmu^{(k)[0]}_{\pi_k(r)} - \hmu^{(k)[0]}_{\pi_k(1)}) - (\hSigma^{(k)[0]})^{-1}(\hmu^{(k)[0]}_{\pi_{k'}(r)} - \hmu^{(k)[0]}_{\pi_{k'}(1)})}-2h-\xi) \\
	&\geq -\sum_{\substack{k \neq k' \in S \\ \pi_k(1) =\pi_{k'}(1) \neq 1}}\sum_{r = 2}^R \sum_{\pi_k(r) = \pi_{k'}(r)}(2h + \xi).
\end{align}
An important observation is that $[1]_{11} = [1]_{11}'$. Therefore,
\begin{equation}
	[1]_1 - [1]'_1 = [1]_{12} - [1]_{12}' \geq -\sum_{\substack{k \neq k' \in S \\ \pi_k(1) =\pi_{k'}(1) \neq 1}}\sum_{r = 2}^R \sum_{\pi_k(r) = \pi_{k'}(r)}(4h + 3\xi).
\end{equation}
And
\begin{align}
	[1] - [1]' &= [1]_2 - [1]'_2 + [1]_1 - [1]_1' \\
	&\geq \sum_{\substack{k \neq k' \in S \\ \pi_k(1) =\pi_{k'}(1)}}\sum_{r = 2}^R \sum_{\pi_k(r) \neq \pi_{k'}(r)}(c_{\bSigma}^{-1/2}\Delta - 4h - 4\xi) -\sum_{\substack{k \neq k' \in S \\ \pi_k(1) =\pi_{k'}(1) \neq 1}}\sum_{r = 2}^R \sum_{\pi_k(r) = \pi_{k'}(r)}(4h + 3\xi).
\end{align}
Furthermore, by triangle inequality,
\begin{align}
	[3] - [3]' &= 2\sum_{\substack{k \in S \\ k' \in S^c}}\sum_{r = 2}^R\twonorm{(\hSigma^{(k)[0]})^{-1}(\hmu^{(k)[0]}_{\pi_k(r)} - \hmu^{(k)[0]}_{\pi_k(1)}) - (\hSigma^{(k')[0]})^{-1}(\hmu^{(k')[0]}_{\pi_{k'}(r)} - \hmu^{(k')[0]}_{\pi_{k'}(1)})} \\
	&\quad - 2\sum_{\substack{k \in S \\ k' \in S^c}}\sum_{r = 2}^R\twonorm{(\hSigma^{(k)[0]})^{-1}(\hmu^{(k)[0]}_r - \hmu^{(k)[0]}_1) - (\hSigma^{(k')[0]})^{-1}(\hmu^{(k')[0]}_{\pi_{k'}(r)} - \hmu^{(k')[0]}_{\pi_{k'}(1)})} \\
	&\geq - 2\sum_{\substack{k \in S \\ k' \in S^c}}\sum_{r = 2}^R\twonorm{(\hSigma^{(k)[0]})^{-1}(\hmu^{(k)[0]}_{\pi_k(r)} - \hmu^{(k)[0]}_{\pi_k(1)} - \hmu^{(k')[0]}_r + \hmu^{(k')[0]}_1)}.
\end{align}
Putting all pieces together,
\begin{align}
	\text{score}(\pi) - \text{score}(\pi^*) &\geq \sum_{\substack{k \neq k' \in S \\ \pi_k(1) \neq\pi_{k'}(1)}}R(c_{\bSigma}^{-1/2}\Delta - 4h - 4\xi) + \sum_{\substack{k \neq k' \in S \\ \pi_k(1) =\pi_{k'}(1)}}\sum_{r=2}^R \sum_{\pi_k(r) \neq \pi_{k'}(r)}(c_{\bSigma}^{-1/2}\Delta - 4h - 4\xi) \\
	&\quad - \sum_{\substack{k \neq k' \in S \\ \pi_k(1) =\pi_{k'}(1) \neq 1}}\sum_{r=2}^R \sum_{\pi_k(r) = \pi_{k'}(r)}(4h + 3\xi) \\
	&\quad - 2\sum_{k \in S}\sum_{r=2}^R \twonorm{(\hSigma^{(k)[0]})^{-1}(\hmu^{(k)[0]}_{\pi_k(r)} - \hmu^{(k)[0]}_{\pi_k(1)} - \hmu^{(k')[0]}_r + \hmu^{(k')[0]}_1)}\cdot |S^c|.
\end{align}
Denote $m_r = $ the majority class among $\{\pi_k(r)\}_{k \in S}$ and $S^{(r)}_{\widetilde{r}} = \{k\in S: \pi_k(r) = \widetilde{r}\}$.

\noindent (\rom{1})\underline{\textbf{Case 1:}} $|S^{(1)}_1| \leq \frac{2}{3}|S|$. 

Since $\pi^*_k(1) = 1$, we have $|S^{(r)}_1| \leq \frac{2}{3}|S|$ for all $r \in [R]$, otherwise by our assumption, $m_1 = r_0$ since $r_0$ satisfies $|S^{(r_0)}_1| > \frac{2}{3}|S| \geq |S^{(1)}_1|$, which is a contradition. Therefore,
\begin{align}
	\sum_{\substack{k \neq k' \in S \\ \pi_k(1) \neq \pi_{k'}(1)}}1 &= 2\sum_{r \neq r'}|S^{(r)}_1|\cdot |S^{(r')}_1| \\
	&= \Big(\sum_{r=1}^R |S_1^{(r)}|\Big)^2 - \sum_{r=1}^R|S_1^{(r)}|^2 \\
	&= |S|^2 - \sum_{r=1}^R|S_1^{(r)}|^2  \\
	&\geq |S|^2 - \Big(\frac{4}{9}|S|^2 + \frac{1}{9}|S|^2\Big) \\
	&= \frac{4}{9}|S|^2,
\end{align}
and
\begin{align}
	\sum_{\substack{k \neq k' \in S \\ \pi_k(1) = \pi_{k'}(1) \neq 1}} \sum_{r=2}^R \sum_{\pi_k(r) = \pi_{k'}(r)}1 &\leq R\cdot 2\binom{|S| - |S^{(1)}_1|}{2} \\
	&\leq R(|S| - |S^{(1)}_1|)^2 \\
	&\leq R|S|^2.
\end{align}
Also,
\begin{equation}
	\sum_{k \in S}\sum_{r=2}^R 1 \leq 2R|S|.
\end{equation}
Hence, since $\frac{4}{9}|S| - 4D|S^c| = \frac{4}{9}(1-\epsilon)K - 4D\epsilon K > 0$,
\begin{align}
	\text{score}(\pi) - \text{score}(\pi^*) &\geq \frac{4}{9}|S|^2 R (c_{\bSigma}^{-1/2}\Delta - 4h - 4\xi) - |S|^2R(4h + 4\xi) - 2R|S||S^c|(2Dc_{\bSigma}^{1/2}\Delta + 2\xi) \\
	&= |S|R\Big[\Big(\frac{4}{9}c_{\bSigma}^{-1/2}|S| - 4Dc_{\bSigma}^{1/2}|S^c|\Big)\Delta - \frac{52}{9}|S|h - \Big(\frac{52}{9}|S| + 4|S^c|\Big)\xi\Big] \\
	&> 0.
\end{align}

\noindent (\rom{2})\underline{\textbf{Case 2:}} $|S^{(1)}_1| > \frac{2}{3}|S|$. 

In this case, by our assumption, we must have $m_1 = 1$. And
\begin{align}
	\sum_{\substack{k \neq k' \in S \\ \pi_k(1) \neq \pi_{k'}(1)}}1 &\geq 2|S^{(1)}_1|(|S| - |S^{(1)}_1|), \\
	\sum_{\substack{k \neq k' \in S \\ \pi_k(1) = \pi_{k'}(1) \neq 1}} \sum_{r=2}^R \sum_{\pi_k(r) = \pi_{k'}(r)}1 &\leq R\sum_{k, k'}\mathds{1}(k \neq k' \in S, \pi_k(1) = \pi_{k'}(1) \neq 1) \\
	&\leq R\cdot 2\binom{|S|-|S^{(1)}_1|}{2} \\
	&\leq R(|S|-|S^{(1)}_1|)^2 \\
	&\leq R\frac{1}{2}|S^{(1)}_1|(|S|-|S^{(1)}_1|).
\end{align}
Moreover,
\begin{align}
	&\sum_{k \in S}\sum_{r=2}^R \twonorm{(\hSigma^{(k)[0]})^{-1}(\hmu^{(k)[0]}_{\pi_k(r)} - \hmu^{(k)[0]}_{\pi_k(1)} - \hmu^{(k)[0]}_r + \hmu^{(k)[0]}_1)}\cdot |S^c| \\
	&\leq \sum_{\substack{k \in S \\ \pi_k(1) \neq 1}}\sum_{r=2}^R \twonorm{(\hSigma^{(k)[0]})^{-1}(\hmu^{(k)[0]}_{\pi_k(r)} - \hmu^{(k)[0]}_{\pi_k(1)} - \hmu^{(k)[0]}_r + \hmu^{(k)[0]}_1)}\cdot |S^c| \\
	&\quad + \sum_{k \in S}\sum_{\substack{\pi_k(1) = 1\\ \pi_k(r) \neq r}}\twonorm{(\hSigma^{(k)[0]})^{-1}(\hmu^{(k)[0]}_{\pi_k(r)} - \hmu^{(k)[0]}_r)}\cdot |S^c| \\
	&\leq (|S| - |S^{(1)}_1|)R\cdot (2c_{\bSigma}^{1/2}\Delta + 2\xi)|S^c| + \sum_{k \in S}\sum_{\substack{\pi_k(1) = 1\\ \pi_k(r) \neq r}}(Dc_{\bSigma}^{1/2}\Delta + \xi)|S^c|.
\end{align}
\noindent For $r$ satisfying $|S^{(r)}_r| \leq \frac{1}{2}|S|$, we have
\begin{align}
	\sum_{k \in S}\sum_{\substack{\pi_k(1) = 1 \\ \pi_k(r) \neq r}} &\leq 2|S^{(1)}_1|, \\
	\sum_{\substack{k \neq k' \in S \\ \pi_k(1) = \pi_{k'}(1)}}\sum_{\pi_k(r) \neq \pi_{k'}(r)}1 &\geq 2 \cdot \frac{1}{2}|S|\Big(\frac{2}{3}|S| - \frac{1}{2}|S|\Big) =\frac{1}{6}|S|^2.
\end{align}
\noindent For $r$ satisfying $|S^{(r)}_r| > \frac{1}{2}|S|$, we have $m_r = r$. Define $\widetilde{S}^{(r)}_r = \{k \in S: \pi_k(1) = 1, \pi_k(r) = r\}$. Note that $|\widetilde{S}^{(r)}_r| \geq \frac{2}{3}|S| - \frac{1}{2}|S| = \frac{1}{6}|S|$. Furthermore,
\begin{align}
	\sum_{k \in S}\sum_{\substack{\pi_k(1) = 1 \\ \pi_k(r) \neq r}} &\leq |S^{(1)}_1| - |\widetilde{S}^{(r)}_r|, \\
	\sum_{\substack{k \neq k' \in S \\ \pi_k(1) = \pi_{k'}(1)}}\sum_{\pi_k(r) \neq \pi_{k'}(r)}1 &\geq 2 |\widetilde{S}^{(r)}_r|(|S^{(1)}_1|-|\widetilde{S}^{(r)}_r|) \geq \frac{1}{3}|S|(|S^{(1)}_1|-|\widetilde{S}^{(r)}_r|).
\end{align}
This implies that
\begin{align}
	&\text{score}(\pi) - \text{score}(\pi^*) \\
	&\geq 2|S_1^{(1)}|(|S| - |S_1^{(1)}|)R(c_{\bSigma}^{-1/2}\Delta - 4h - 4\xi) - R\frac{1}{2}|S^{(1)}_1|(|S| - |S^{(1)}_1|)\cdot (4h+3\xi) \\
	&\quad + \sum_{r: |S^{(r)}_r| \leq |S|/2}\Big[\frac{1}{6}|S|^2(c_{\bSigma}^{-1/2}\Delta - 4h-4\xi) - 2|S_1^{(1)}|\cdot 2|S^c|(c_{\bSigma}^{1/2}D\Delta+\xi)\Big] \\
	&\quad + \sum_{r: |S^{(r)}_r| > |S|/2}\Big[\frac{1}{3}|S|(|S_1^{(1)}| - \widetilde{S}^{(r)}_r)(c_{\bSigma}^{-1/2}\Delta - 4h-4\xi) - (|S^{(1)}_1| - |\widetilde{S}^{(r)}_r|)\cdot |S^c|\cdot 2(c_{\bSigma}^{1/2}D\Delta + \xi)\Big] \\
	&\geq |S^{(1)}_1|(|S| - |S^{(1)}_1|)R\Big(2c_{\bSigma}^{-1/2}\Delta - 10h - \frac{19}{2}\xi\Big) \\
	&\quad + \sum_{r: |S^{(r)}_r| \leq |S|/2}|S|\bigg[\Big(\frac{1}{6}c_{\bSigma}^{-1/2}|S|-4Dc_{\bSigma}^{1/2}|S^c|\Big)\Delta - \frac{2}{3}|S|h - \Big(\frac{2}{3}|S| + 4|S^c|\Big)\xi\bigg] \\
	&\quad + \sum_{r: |S^{(r)}_r| > |S|/2}\bigg[\Big(\frac{1}{3}c_{\bSigma}^{-1/2}|S| - 2Dc_{\bSigma}^{1/2}|S^c|\Big)\Delta - \frac{4}{3}|S|h - \Big(\frac{4}{3}|S| + 2|S^c|\Big)\xi\bigg](|S^{(1)}_1|-|\widetilde{S}^{(r)}_r|)\\
	&> 0.
\end{align}

\subsection{Proof of Theorem \ref{thm: greedy alignment multi component}}
WLOG, consider the step $\widetilde{K} \in [K]$ in the for loop and the case that $\iota = $ indentify mapping from $[K]$ to $[K]$, and $\widetilde{K} \in S$. Denote $\widetilde{S} = S \cap [\widetilde{K}]$ and $\widetilde{S}^c = S^c \cap [\widetilde{K}]$, hence $[\widetilde{K}] = \widetilde{S} \cup \widetilde{S}^c$. WLOG, consider $\pi_1 = \pi_2 = \cdots \pi_{\widetilde{K}-1} = $ identify from $[R]$ to $[R]$. Denote $\bm{\pi} = \{\pi_k\}_{k=1}^{\widetilde{K}}$ and $\widetilde{\bm{\pi}} = \{\pi_k\}_{k=1}^{\widetilde{K}-1} \cup \{\widetilde{\pi}_{\widetilde{K}}\}$ with $\widetilde{\pi}_{\widetilde{K}} = $ identify from $[R]$ to $[R]$. It suffices to show that
\begin{equation}\label{eq: proof greedy align}
	\text{score}(\bm{\pi}) > \text{score}(\widetilde{\bm{\pi}}),
\end{equation}
for any $\bm{\pi}$ with $\pi_{\widetilde{K}} \neq \widetilde{\pi}_{\widetilde{K}} = $ identify from $[R]$ to $[R]$. If this is the case, then $\widehat{\bm{\pi}} = \{\widehat{\pi}_k\}_{k=1}^K$ satisfies $\widehat{\pi}_k = $ identity for all $k \in S$, which completes the proof. 

We focus on the derivation of \eqref{eq: proof greedy align} in the remaining part of the proof. 

\noindent\underline{(\rom{1}) Case 1:} $\pi_{\widetilde{K}}(1) = 1$.
\begin{align}
	\text{score}(\pi) - \text{score}(\pi^*) &= \underbrace{\sum_{r = 2}^R\sum_{k \in \widetilde{S}}\twonorm{(\hSigma^{(k)[0]})^{-1}(\hmu^{(k)[0]}_r - \hmu^{(k)[0]}_1) - (\hSigma^{(\widetilde{K})[0]})^{-1}(\hmu^{(\widetilde{K})[0]}_{\pi_{\widetilde{K}}(r)} - \hmu^{(\widetilde{K})[0]}_1)}}_{[1]} \\
	&\quad+  \underbrace{\sum_{r = 2}^R\sum_{k \in \widetilde{S}^c}\twonorm{(\hSigma^{(k)[0]})^{-1}(\hmu^{(k)[0]}_{\pi_k(r)} - \hmu^{(k)[0]}_{\pi_k(1)}) - (\hSigma^{(\widetilde{K})[0]})^{-1}(\hmu^{(\widetilde{K})[0]}_{\pi_{\widetilde{K}}(r)} - \hmu^{(\widetilde{K})[0]}_1)}}_{[2]} \\
	&- \underbrace{\sum_{r = 2}^R\sum_{k \in \widetilde{S}}\twonorm{(\hSigma^{(k)[0]})^{-1}(\hmu^{(k)[0]}_r - \hmu^{(k)[0]}_1) - (\hSigma^{(\widetilde{K})[0]})^{-1}(\hmu^{(\widetilde{K})[0]}_r - \hmu^{(\widetilde{K})[0]}_1)}}_{[1]'}\\ 
	&- \underbrace{\sum_{r = 2}^R\sum_{k \in \widetilde{S}^c}\twonorm{(\hSigma^{(k)[0]})^{-1}(\hmu^{(k)[0]}_{\pi_k(r)} - \hmu^{(k)[0]}_{\pi_k(1)}) - (\hSigma^{(\widetilde{K})[0]})^{-1}(\hmu^{(\widetilde{K})[0]}_r - \hmu^{(\widetilde{K})[0]}_1)}}_{[2]'}.
\end{align}
Note that
\begin{align}
	[1] - [1]' &= \#\{r \in 2:R: \pi_{\widetilde{K}}(r) \neq r\}\cdot \\
	&\quad \sum_{k \in \widetilde{S}}\Big[\twonorm{(\hSigma^{(k)[0]})^{-1}(\hmu^{(k)[0]}_r - \hmu^{(k)[0]}_1) - (\hSigma^{(\widetilde{K})[0]})^{-1}(\hmu^{(\widetilde{K})[0]}_{\pi_{\widetilde{K}}(r)} - \hmu^{(\widetilde{K})[0]}_1)} \\
	&\quad\quad\quad - \twonorm{(\hSigma^{(k)[0]})^{-1}(\hmu^{(k)[0]}_r - \hmu^{(k)[0]}_1) - (\hSigma^{(\widetilde{K})[0]})^{-1}(\hmu^{(\widetilde{K})[0]}_r - \hmu^{(\widetilde{K})[0]}_1)}\Big] \\
	&\geq \#\{r \in 2:R: \pi_{\widetilde{K}}(r) \neq r\}\cdot \Big[\twonorm{(\bSigmaks{k})^{-1}(\bmuks{k}_r - \bmuks{k}_{\pi_{\widetilde{K}}(r)})}-2\xi - 2h - 2\xi - 2h\Big] \\
	&\geq \#\{r \in 2:R: \pi_{\widetilde{K}}(r) \neq r\}\cdot |\widetilde{S}|\cdot \big(\Delta c_{\bSigma}^{-1/2} - 4\xi - 4h\big).
\end{align}
\begin{align}
	[2] - [2]' &= \#\{r \in 2:R: \pi_{\widetilde{K}}(r) \neq r\}\cdot \\
	&\quad \sum_{k \in \widetilde{S}^c}\Big[\twonorm{(\hSigma^{(k)[0]})^{-1}(\hmu^{(k)[0]}_{\pi_k(r)} - \hmu^{(k)[0]}_{\pi_k(1)}) - (\hSigma^{(\widetilde{K})[0]})^{-1}(\hmu^{(\widetilde{K})[0]}_{\pi_{\widetilde{K}}(r)} - \hmu^{(\widetilde{K})[0]}_1)}\\
	&\quad\quad\quad -\twonorm{(\hSigma^{(k)[0]})^{-1}(\hmu^{(k)[0]}_{\pi_k(r)} - \hmu^{(k)[0]}_{\pi_k(1)}) - (\hSigma^{(\widetilde{K})[0]})^{-1}(\hmu^{(\widetilde{K})[0]}_r - \hmu^{(\widetilde{K})[0]}_1)}\Big] \\
	&\geq -\#\{r \in 2:R: \pi_{\widetilde{K}}(r) \neq r\}\cdot \sum_{k \in \widetilde{S}^c}\twonorm{(\hSigma^{(\widetilde{K})[0]})^{-1}(\hmu^{(\widetilde{K})[0]}_{\pi_{\widetilde{K}}(r)} - \hmu^{(\widetilde{K})[0]}_r)}\\
	&\geq -\#\{r \in 2:R: \pi_{\widetilde{K}}(r) \neq r\}\cdot |\widetilde{S}^c|\cdot \big[\twonorm{(\bSigmaks{\widetilde{K}})^{-1}(\bmuks{\widetilde{K}}_{\pi_{\widetilde{K}}(r)} - \bmuks{\widetilde{K}}_r)} + \xi\big] \\
	&\geq -\#\{r \in 2:R: \pi_{\widetilde{K}}(r) \neq r\}\cdot |\widetilde{S}^c|\cdot(Dc_{\bSigma}^{1/2}\Delta + \xi).
\end{align}
These imply that
\begin{align}
	\text{score}(\bm{\pi}) - \text{score}(\widetilde{\bm{\pi}}) &= [1] - [1]' + [2] - [2]' \\
	&\geq \#\{r \in 2:R: \pi_{\widetilde{K}}(r) \neq r\}\cdot \big[(|\widetilde{S}|c_{\bSigma}^{-1/2} - |\widetilde{S}^c|Dc_{\bSigma}^{1/2})\Delta - (4|\widetilde{S}| + |\widetilde{S}^c|)\xi - 4|\widetilde{S}|h\big]\\
	&> 0,
\end{align}
where we used the fact that $|\widetilde{S}|/|\widetilde{S}^c| \leq \frac{K_0}{K\epsilon}$.

\noindent\underline{(\rom{2}) Case 2:} $\pi_{\widetilde{K}}(1) \neq 1$.
\begin{align}
	\text{score}(\pi) - \text{score}(\pi^*) &= \underbrace{\sum_{r = 2}^R\sum_{k \in \widetilde{S}}\twonorm{(\hSigma^{(k)[0]})^{-1}(\hmu^{(k)[0]}_r - \hmu^{(k)[0]}_1) - (\hSigma^{(\widetilde{K})[0]})^{-1}(\hmu^{(\widetilde{K})[0]}_{\pi_{\widetilde{K}}(r)} - \hmu^{(\widetilde{K})[0]}_{\pi_{\widetilde{K}}(1)})}}_{[1]} \\
	&\quad+  \underbrace{\sum_{r = 2}^R\sum_{k \in \widetilde{S}^c}\twonorm{(\hSigma^{(k)[0]})^{-1}(\hmu^{(k)[0]}_{\pi_k(r)} - \hmu^{(k)[0]}_{\pi_k(1)}) - (\hSigma^{(\widetilde{K})[0]})^{-1}(\hmu^{(\widetilde{K})[0]}_{\pi_{\widetilde{K}}(r)} - \hmu^{(\widetilde{K})[0]}_{\pi_{\widetilde{K}}(1)})}}_{[2]} \\
	&- \underbrace{\sum_{r = 2}^R\sum_{k \in \widetilde{S}}\twonorm{(\hSigma^{(k)[0]})^{-1}(\hmu^{(k)[0]}_r - \hmu^{(k)[0]}_1) - (\hSigma^{(\widetilde{K})[0]})^{-1}(\hmu^{(\widetilde{K})[0]}_r - \hmu^{(\widetilde{K})[0]}_1)}}_{[1]'}\\ 
	&- \underbrace{\sum_{r = 2}^R\sum_{k \in \widetilde{S}^c}\twonorm{(\hSigma^{(k)[0]})^{-1}(\hmu^{(k)[0]}_{\pi_k(r)} - \hmu^{(k)[0]}_{\pi_k(1)}) - (\hSigma^{(\widetilde{K})[0]})^{-1}(\hmu^{(\widetilde{K})[0]}_r - \hmu^{(\widetilde{K})[0]}_1)}}_{[2]'}.
\end{align}
By previous results,
\begin{align}
	[1] &= \sum_{k \in \widetilde{S}}\sum_{r = 2}^R\twonorm{(\hSigma^{(k)[0]})^{-1}(\hmu^{(k)[0]}_r - \hmu^{(k)[0]}_1) - (\hSigma^{(\widetilde{K})[0]})^{-1}(\hmu^{(\widetilde{K})[0]}_{\pi_{\widetilde{K}}(r)} - \hmu^{(\widetilde{K})[0]}_{\pi_{\widetilde{K}}(1)})} \\
	&\geq \sum_{k \in \widetilde{S}}\sum_{r = 2}^R\left(\twonorm{(\hSigma^{(k)[0]})^{-1}(\hmu^{(k)[0]}_r - \hmu^{(k)[0]}_1) - (\hSigma^{(\widetilde{K})[0]})^{-1}(\hmu^{(k)[0]}_{\pi_{\widetilde{K}}(r)} - \hmu^{(k)[0]}_{\pi_{\widetilde{K}}(1)})} - 2h - \xi\right) \\
	&\geq |\widetilde{S}|R\cdot \left(\twonorm{(\hSigma^{(k)[0]})^{-1}(\hmu^{(k)[0]}_{\pi_{\widetilde{K}}(1)} - \hmu^{(k)[0]}_1)}- 2h - \xi\right) \\
	&\geq |\widetilde{S}|R\cdot\left(\twonorm{(\bSigmaks{k})^{-1}(\bmuks{k}_{\pi_{\widetilde{K}}(1)} - \bmuks{k}_1)} - \xi - 2h - \xi\right) \\
	&\geq |\widetilde{S}|R\cdot (c_{\bSigma}^{-1/2}\Delta - 2\xi - 2h),
\end{align}
and
\begin{equation}
	-[1]' \geq -|\widetilde{S}|R\cdot (2\xi + 2h).
\end{equation}
Similar to case 1,
\begin{align}
	[2] - [2]' &\geq -\sum_{r = 2}^R\sum_{k \in \widetilde{S}}\twonorm{(\hSigma^{(\widetilde{K})})^{-1}(\hmu^{(\widetilde{K})[0]}_{\pi_{\widetilde{K}(r)}} - \hmu^{(\widetilde{K})[0]}_{\pi_{\widetilde{K}(1)}}) - (\hSigma^{(\widetilde{K})})^{-1}(\hmu^{(\widetilde{K})[0]}_r - \hmu^{(\widetilde{K})[0]}_1)} \\
	&\geq -R|\widetilde{S}^c|\cdot (2Dc_{\bSigma}^{1/2}\Delta + 2\xi).
\end{align}
Therefore,
\begin{align}
	\text{score}(\bm{\pi}) - \text{score}(\widetilde{\bm{\pi}}) &= [1] - [1]' + [2] - [2]' \\
	&\geq R\big[(|\widetilde{S}|c_{\bSigma}^{-1/2} - |\widetilde{S}^c|\cdot 2Dc_{\bSigma}^{1/2})\Delta - (2|\widetilde{S}| + 2|\widetilde{S}^c|)\xi - 2|\widetilde{S}|h\big]\\
	&>0,
\end{align}
where we used the fact that $|\widetilde{S}|/|\widetilde{S}^c| \leq \frac{K_0}{K\epsilon}$.

\subsection{General lemmas}
Denote the unit ball $\mathcal{B}^p = \{\bu \in \mathbb{R}^p: \twonorm{\bu}\leq 1\}$ and the unit sphere $\mathcal{S}^{p-1} = \{\bu \in \mathbb{R}^p: \twonorm{\bu}=1\}$.
\begin{lemma}[Covering number of the unit ball under Euclidean norm, Example 5.8 in \citealpapp{wainwright2019high}]\label{lem: covering number of ball}
	Denote the $\epsilon$-covering number of a unit ball $\mathcal{B}^p$ in $\mathbb{R}^p$ under Euclidean norm as $N(\epsilon, \mathcal{B}^p, \twonorm{\cdot})$, where the centers of covering balls are required to be on the sphere. We have $(1/\epsilon)^p \leq N(\epsilon, \mathcal{B}^{p}, \twonorm{\cdot}) \leq (1+2/\epsilon)^p$.
\end{lemma}

\begin{lemma}[Packing number of the unit sphere under Euclidean norm]\label{lem: packing number of sphere}
	Denote the $\epsilon$-packing number of the unit sphere $\mathcal{S}^{p-1}$ in $\mathbb{R}^p$ under Euclidean norm as $M(\epsilon, \mathcal{S}^{p-1}, \|\cdot\|_2)$. When $p \geq 2$, we have $M(\epsilon, \mathcal{S}^{p-1}, \|\cdot\|_2) \geq N(\epsilon, \mathcal{B}^{p-1}, \twonorm{\cdot}) \geq (1/\epsilon)^{p-1}$.
\end{lemma}


\begin{lemma}[Fano's lemma, see Chapter 2 of \citealpapp{tsybakov2009introduction}, Chapter 15 of \citealpapp{wainwright2019high}]\label{lem: fano}
	Suppose $(\Theta, d)$ is a metric space and each $\theta$ in this space is associated with a probability measure $\tp_{\theta}$. If $\{\theta_j\}_{j = 1}^N$ is an $s$-separated set (i.e. $d(\theta_j, \theta_k) \geq s$ for any $j \neq k$), and $\textup{KL}(\tp_{\theta_j}, \tp_{\theta_k}) \leq \alpha \log N$, then
	\begin{equation}
		\inf_{\widehat{\theta}}\sup_{\theta \in \Theta}\tp_{\theta}(d(\widehat{\theta}, \theta) \geq s/2) \geq \inf_{\psi}\sup_{j=1:N}\tp_{\theta_j}(\psi \neq j) \geq 1-\alpha -\frac{\log 2}{\log N},
	\end{equation}
	where $\psi: X \mapsto \psi(X) \in \{1,\ldots, N\}$.
\end{lemma}

\begin{lemma}[Packing number of the unit sphere in a quadrant under Euclidean norm]\label{lem: packing number of sphere quadrant}
	In $\mathbb{R}^p$, we can use a vector $\bv \in \{\pm 1\}^{\otimes p}$ to indicate each quadrant $\mathcal{Q}_{\bv} = \{[0, +\infty)\cdot \mathds{1}(v_j =+1) + (-\infty, 0)\cdot \mathds{1}(v_j =-1)\}^{\otimes p}$. Then when $p \geq 2$, there exists a quadrant $\mathcal{Q}_{\bv_0}$ such that $M(\epsilon, \mathcal{S}^{p-1} \cap \mathcal{Q}_{\bv_0}, \twonorm{\cdot}) \geq (\frac{1}{2})^p (\frac{1}{\epsilon})^{p-1}$.
\end{lemma}

\begin{lemma}\label{lem: GMM subgaussian}
For one-dimensional Gaussian mixture variable $Z \sim (1-w)\mathcal{N}(\mu_1, \sigma^2) + w\mathcal{N}(\mu_2, \sigma^2)$ with $(1-w)\mu_1 + w\mu_2 = 0$, it is a $\sqrt{\sigma^2 + \frac{1}{4}\norm{\mu_1-\mu_2}^2}$-subGaussian variable. That means,
\begin{equation}
	\te e^{\lambda Z} \leq \exp\left\{\frac{1}{2}\lambda^2\left(\sigma^2 + \frac{1}{4}\norm{\mu_1-\mu_2}^2\right)\right\}.
\end{equation}
\end{lemma}

\begin{lemma}[\citealpapp{duan2023adaptive}]\label{lem: claim B1}
	Let
	\begin{equation}
		(\{\htheta_j\}_{k=1}^K, \hbeta) = \argmin_{\btheta_k, \bbeta \in \mathbb{R}^p}\left\{\sum_{k=1}^K \omega_k f^{(k)}(\btheta_k) + \sqrt{\omega_k}\lambda\twonorm{\bbeta-\btheta_k})\right\}.
	\end{equation}
	Suppose there exists $S \subseteq 1:K$ such that the following conditions are satisfied:
	\begin{enumerate}[(i)]
		\item For any $k \in S$, $f_k$ is $(\btheta^*_k, M, \rho, L)$-regular, that is
			\begin{itemize}
				\item $f_k$ is convex and twice differentiable;
				\item $\rho \bm{I} \preceq \nabla^2 f_k(\btheta) \preceq L\bm{I}$ for all $\btheta \in \mathcal{B}(\btheta^*_k, M)$;
				\item $\twonorm{\nabla f_k(\btheta^*_k)} \leq \rho M/2$. 
			\end{itemize}
		\item $\min_{\btheta \in \mathbb{R}^d}\max_{k \in S}\{\twonorm{\btheta_k^*-\btheta}\} \leq h$, $\sum_{k \in S^c}\sqrt{\omega_k} \leq \epsilon' \sum_{k \in S}\omega_k/\max_{k \in S}\sqrt{\omega_k}$, with $\epsilon' = \frac{|S^c|}{|S|}$.
	\end{enumerate}
	Then we have the following conclusions:
	\begin{enumerate}[(i)]
		\item $\twonorm{\htheta_k - \btheta_k^*} \leq \frac{1}{\rho}(\twonorm{\nabla f^{(k)}(\btheta^*_k)} + \lambda/\sqrt{w_k})$ for all $k \in S$.
		\item If
			\begin{equation}
				 \frac{5\varrho\kappa_w\max_{k \in S}\{\sqrt{w_k}\twonorm{\nabla f^{(k)}(\btheta^*_k)}\}}{1-\varrho \epsilon'} < \lambda < \frac{\rho M}{2}\min_{k \in S}\sqrt{\omega_k},
			\end{equation}
			where $\varrho = L/\rho$, $\varrho \epsilon' < 1$, and $\max_{k \in S}\sqrt{n_k}\cdot \frac{\sum_{k \in S}\sqrt{n_k}}{\ns} \leq \sqrt{\frac{|S|\max_{k \in S}n_k}{\ns}}\coloneqq \kappa_w$, then
			\begin{equation}
				\twonorm{\htheta_k - \btheta^*_k} \leq \frac{\twonorm{\sum_{k \in S}\omega_k\nabla f^{(k)}(\btheta_k^*)}}{\rho\sum_{k \in S}\omega_k} + \frac{6}{1-\varrho\epsilon'}\min\left\{3\varrho^2\kappa_w h,\frac{2\lambda}{5\rho\sqrt{w_k}}\right\} + \frac{\lambda \epsilon'}{\rho \max_{k \in S}\sqrt{\omega_k}}.
			\end{equation}
			Furthermore, if we \textbf{also} have
			\begin{equation}
				\lambda \geq \frac{15\varrho \kappa_w L\max_{k \in S}\sqrt{\omega_k}h}{1-\varrho\epsilon'},
			\end{equation}
			then $\htheta_k = \hbeta$ for all $k \in S$, and
			\begin{equation}
				\sup_{k \in S}\twonorm{\htheta_k - \btheta^*_k} \leq \frac{\twonorm{\sum_{k \in S}\omega_k\nabla f^{(k)}(\btheta_k^*)}}{\rho\sum_{k \in S}\omega_k} + 2\varrho \kappa_w h + \frac{\lambda \epsilon'}{\rho \max_{k \in S}\sqrt{\omega_k}}.
			\end{equation}
	\end{enumerate}
	\end{lemma}

\begin{lemma}\label{lem: transfer lemma}
	Suppose
	\begin{equation}
		\htheta = \argmin_{\btheta}\left\{f^{(0)}(\btheta) + \frac{\lambda}{\sqrt{n_0}}\twonorm{\htheta - \otheta}\right\}
	\end{equation}
	with some $\otheta \in \mathbb{R}^p$. Assume $f^{(0)}$ is convex and twice differentiable, and $\rho \bm{I}_p \leq \nabla^2 f^{(0)}(\btheta) \leq L\bm{I}_p$ for any $\btheta \in \mathbb{R}^p$. Then
	\begin{enumerate}[(i)]
		\item $\twonorm{\htheta - \btheta^*} \leq \frac{\nabla f^{(0)}(\btheta)}{\rho} + \frac{\lambda}{\rho\sqrt{n_0}}$, for any $\btheta^* \in \mathbb{R}^p$ and $\lambda \geq 0$;
		\item $\htheta = \otheta$, if $\lambda \geq 2\twonorm{\nabla f^{(0)}(\btheta^*)}\sqrt{n_0}$ and $\twonorm{\otheta - \btheta^*} \leq (\lambda/\sqrt{n_0} - \twonorm{\nabla f^{(0)}(\btheta^*)})/L$.
	\end{enumerate}
\end{lemma}

\subsection{Proofs of general lemmas}


%

\subsubsection{Proof of Lemma \ref{lem: packing number of sphere}}
The second half inequality is due to Lemma \ref{lem: covering number of ball}. It suffices to show the first half inequality. For any $\bx = (x_1, \ldots, x_{p-1})^\top \in \mathcal{B}^{p-1}$, define $x_p = \sqrt{1-\sum_{j=1}^{p-1}x_j^2}$. Then we can define a mapping
\begin{equation}
	\bx \in \mathcal{B}^{p-1} \mapsto \widetilde{\bx} = (\widetilde{x}_1, \ldots, \widetilde{x}_{p-1}, \widetilde{x}_p) \in \mathcal{S}^p,
\end{equation}
with $\widetilde{x}_j = x_j$ for $j \leq p-1$ and $\widetilde{x}_p = \pm x_p$. And it's easy to see that for any $\bx$, $\by \in \mathcal{B}^{p-1}$, we have $\twonorm{\bx-\by} \leq \twonorm{\widetilde{\bx} - \widetilde{\by}}$. Therefore, if $\{\widetilde{\bx}_j\}_{j=1}^N$ is an $\epsilon$-cover of $\mathcal{S}^{p-1}$ under Euclidean norm, then it $\{\bx_j\}_{j=1}^N$ must be an $\epsilon$-cover of $\mathcal{B}^{p-1}$ under Euclidean norm. Then 
\begin{equation}
	N(\epsilon, \mathcal{B}^{p-1}, \twonorm{\cdot}) \leq N(\epsilon, \mathcal{S}^{p-1}, \twonorm{\cdot}) \leq M(\epsilon, \mathcal{S}^{p-1}, \twonorm{\cdot}).
\end{equation}

\subsubsection{Proof of Lemma \ref{lem: packing number of sphere quadrant}}
If $\{\bu_j\}_{j=1}^N$ is an $\epsilon$-packing of $\mathcal{S}^{p-1}$ under Euclidean norm, then $\{\bu_j\}_{j=1}^N \cap \mathcal{Q}_{\bv}$ must be an $\epsilon$-packing of $\mathcal{S}^p \cap \mathcal{Q}_{\bv}$ under Euclidean norm.  Then by Lemma \ref{lem: packing number of sphere},
\begin{align}
	2^p \max_{\bv \in \{\pm 1\}^{\otimes p}} M(\epsilon, \mathcal{S}^{p-1} \cap \mathcal{Q}_{\bv}, \twonorm{\cdot}) &\geq \sum_{\bv \in \{\pm 1\}^{\otimes p}} M(\epsilon, \mathcal{S}^{p-1} \cap \mathcal{Q}_{\bv}, \twonorm{\cdot}) \\
	&\geq M(\epsilon, \mathcal{S}^{p-1}, \twonorm{\cdot}) \\
	&\geq \left(\frac{1}{\epsilon}\right)^{p-1},
\end{align}
implying
\begin{equation}
	\max_{\bv \in \{\pm 1\}^{\otimes p}} M(\epsilon, \mathcal{S}^p \cap \mathcal{Q}_{\bv}, \twonorm{\cdot}) \geq \left(\frac{1}{2}\right)^p \left(\frac{1}{\epsilon}\right)^{p-1}.
\end{equation}

\subsubsection{Proof of Lemma \ref{lem: GMM subgaussian}}
Suppose $Z_1 \sim \mathcal{N}(\mu_1, \sigma^2) \ind Z_2 \sim \mathcal{N}(\mu_2, \sigma^2)$, then we can write $Z = (1-I)Z_1 + IZ_2 = (1-I)(Z_1-\mu_1) + I(Z_2-\mu_2) + [\mu_1(1-I) + \mu_2I]$, where $I \sim \text{Bernoulli}(w)$ that is independent with $Z_1$ and $Z_2$. Then
\begin{align}
	\te e^{\lambda Z} &\leq \te e^{\lambda(1-I)(Z_1-\mu_1) +\lambda I(Z_2-\mu_2)} \cdot \te e^{\lambda[\mu_1(1-I) + \mu_2I]} \\
	&\leq \te_I\big[(1-I)\te e^{\lambda Z_1} + I \te e^{\lambda Z_2}\big]\cdot \te e^{\lambda[\mu_1(1-I) + \mu_2I]} \\
	&\leq \exp\left\{\frac{1}{2}\sigma^2\lambda^2 + \frac{1}{8}(\mu_2-\mu_1)^2\lambda^2\right\},
\end{align}
where the last second inequality comes from Jensen's inequality and the independence between $Z_1$, $Z_2$, and $I$. This completes the proof.

\subsubsection{Proof of Lemma \ref{lem: claim B1}}
The result follows from Theorem A.2, Lemma B.1, and Claim B.1 in \citeapp{duan2023adaptive}.

\spacingset{0.9}
{\small\bibliographyapp{reference.bib}}
\bibliographystyleapp{apalike}
\end{document}